\documentclass[10pt,journal,compsoc]{IEEEtran}
\ifCLASSOPTIONcompsoc
  \usepackage[nocompress]{cite}
\else
  \usepackage{cite}
\fi
\ifCLASSINFOpdf
  \usepackage[pdftex]{graphicx}
\else
\fi
\usepackage{amsmath}
\usepackage{algorithmic}

\def\swseven{0.135\linewidth}
\def\swedm{0.23\linewidth}

\def\swceleba{0.123\linewidth}
\def\swfigfirst{0.105\linewidth}
\def\swlimitation{0.23\linewidth}
\def\swgray{0.30\linewidth}
\def\swedm1{0.30\linewidth}

\usepackage{bm}
\usepackage{amssymb}
\usepackage{booktabs}
\usepackage{color}

\usepackage{microtype}

\usepackage[pagewise]{lineno}

\begin{document}
%
\title{RestoreFormer++: Towards Real-World \\ Blind Face Restoration from Undegraded Key-Value Pairs}
%
%
%
%


\author{Zhouxia~Wang, Jiawei~Zhang, Tianshui~Chen, 
        Wenping~Wang,~\IEEEmembership{Fellow,~IEEE,} \\
        and~Ping~Luo,~\IEEEmembership{Member,~IEEE}
\IEEEcompsocitemizethanks{\IEEEcompsocthanksitem Zhouxia Wang and Wenping Wang are with The University of Hong Kong, Hong Kong SAR, China.\protect\\
E-mail: \{wzhoux@.connect, wenping@cs\}.hku.hk
\IEEEcompsocthanksitem Jiawei Zhang is with SenseTime Research, China.\protect\\
E-mail: zhjw1988@gmail.com
\IEEEcompsocthanksitem Tianshui Chen is with The Guangdong University of Technology, Guangzhou, China.\protect\\
E-mail: tianshuichen@gmail.com
\IEEEcompsocthanksitem Ping Luo is with The University of Hong Kong and Shanghai AI Laboratory, China.\protect\\
E-mail: pluo@cs.hku.hk
}
}

%
%

\markboth{IEEE TRANSACTIONS ON PATTERN ANALYSIS AND MACHINE INTELLIGENCE}
{Shell \MakeLowercase{\textit{et~al.}}: Bare Demo of IEEEtran.cls for Computer Society Journals}
%



\IEEEtitleabstractindextext{%
\begin{abstract}
Blind face restoration aims at recovering high-quality face images from those with unknown degradations. Current algorithms mainly introduce priors to complement high-quality details and achieve impressive progress. 
However, most of these algorithms ignore abundant contextual information in the face and its interplay with the priors, leading to sub-optimal performance. 
Moreover, they pay less attention to the gap between the synthetic and real-world scenarios, limiting the robustness and generalization to real-world applications. 
In this work, we propose RestoreFormer++, which on the one hand introduces fully-spatial attention mechanisms to model the contextual information and the interplay with the priors, and on the other hand, explores an extending degrading model to help generate more realistic degraded face images to alleviate the synthetic-to-real-world gap. 
Compared with current algorithms, RestoreFormer++ has several crucial benefits. 
First, instead of using a multi-head self-attention mechanism like the traditional visual transformer, we introduce multi-head cross-attention over multi-scale features to fully explore spatial interactions between corrupted information and high-quality \textcolor{black}{priors}. In this way, it can facilitate RestoreFormer++ to restore face images with higher realness and fidelity. 
Second, in contrast to the recognition-oriented dictionary, we learn a reconstruction-oriented dictionary as priors, which contains more diverse high-quality facial details and better accords with the restoration target. 
Third, we introduce an extending degrading model that contains more realistic degraded scenarios for training data synthesizing, and thus helps to enhance the robustness and generalization of our RestoreFormer++ model. 
Extensive experiments show that RestoreFormer++ outperforms state-of-the-art algorithms on both synthetic and real-world datasets. 
Code will be available at https://github.com/wzhouxiff/RestoreFormerPlusPlus.

\end{abstract}

}

\maketitle

\IEEEdisplaynontitleabstractindextext

%
\IEEEpeerreviewmaketitle

\IEEEraisesectionheading{\section{Introduction}\label{sec:introduction}}

%
%
%
%

 

\begin{figure*}[t]
\setlength\tabcolsep{1pt}
\begin{center}
\begin{tabular}{ccccccccc}
    \includegraphics[width=\swfigfirst]{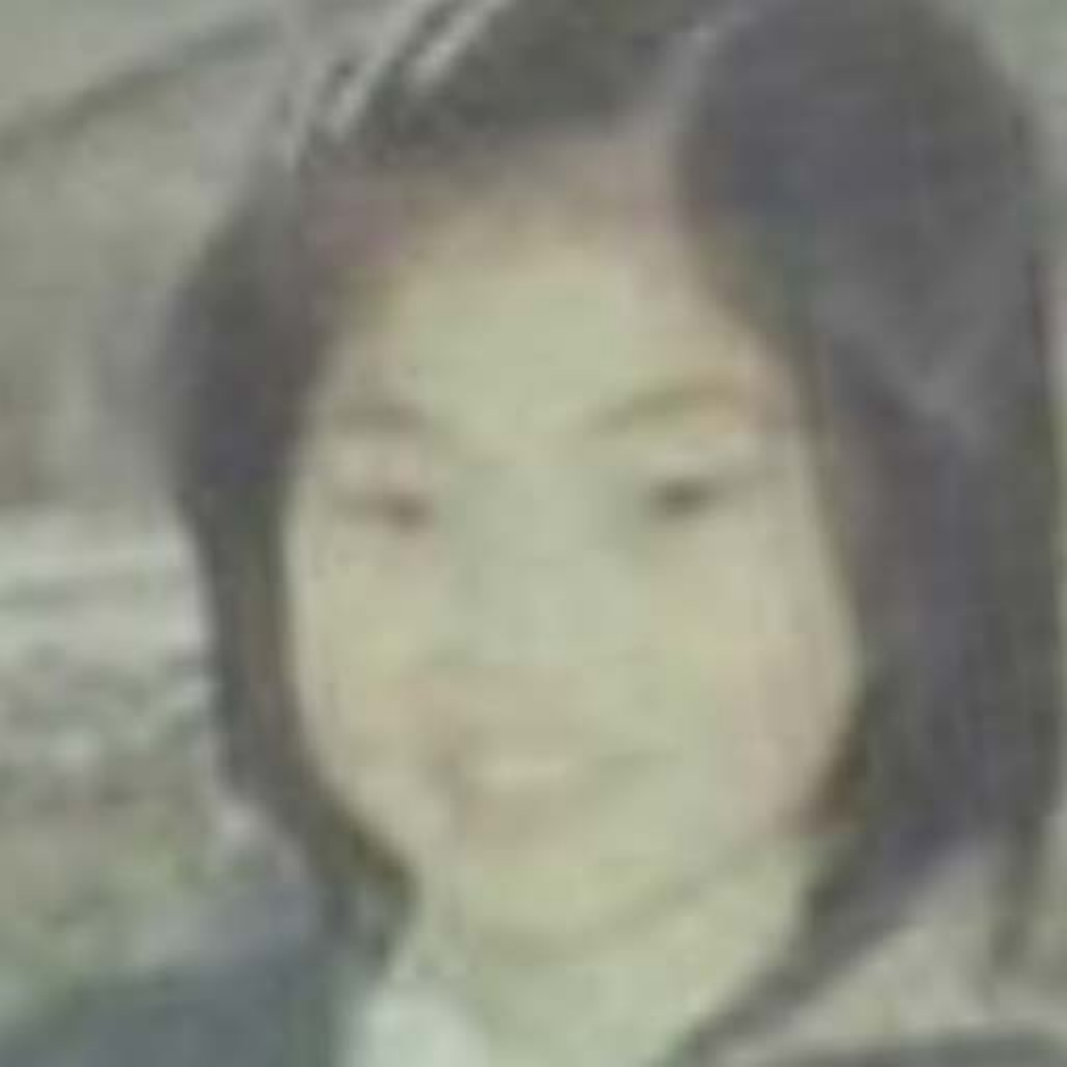}&
    \includegraphics[width=\swfigfirst]{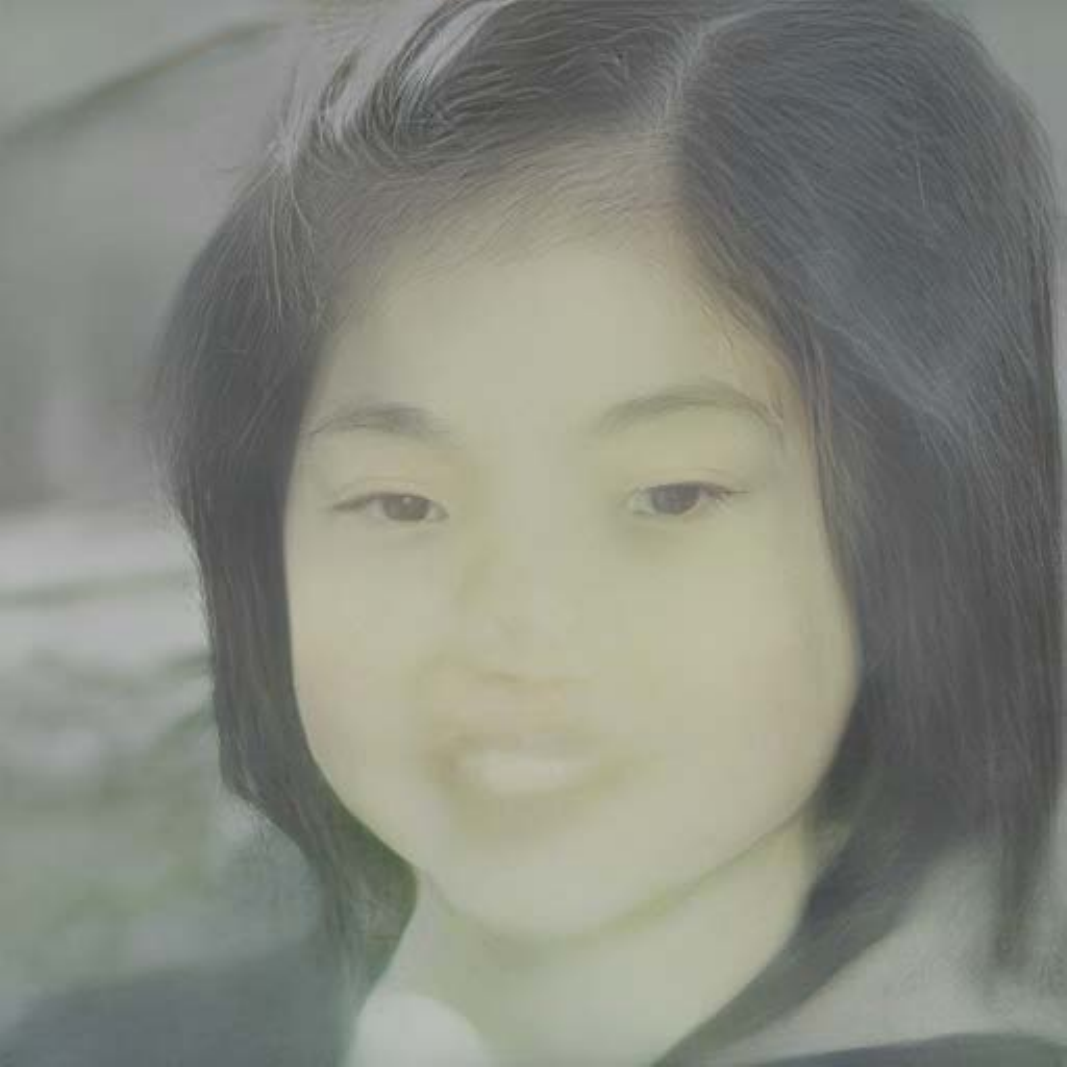}&
    \includegraphics[width=\swfigfirst]{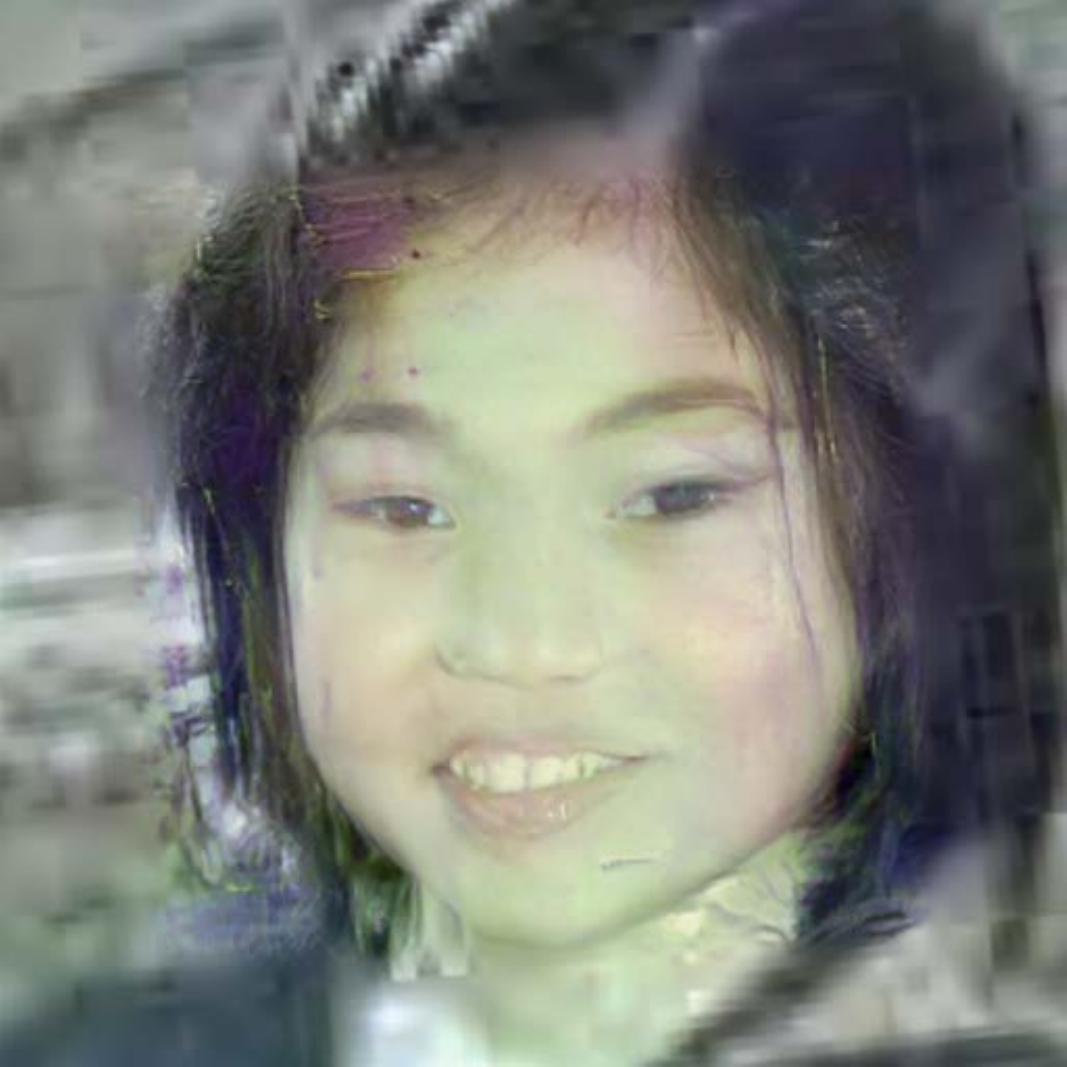}&
    \includegraphics[width=\swfigfirst]{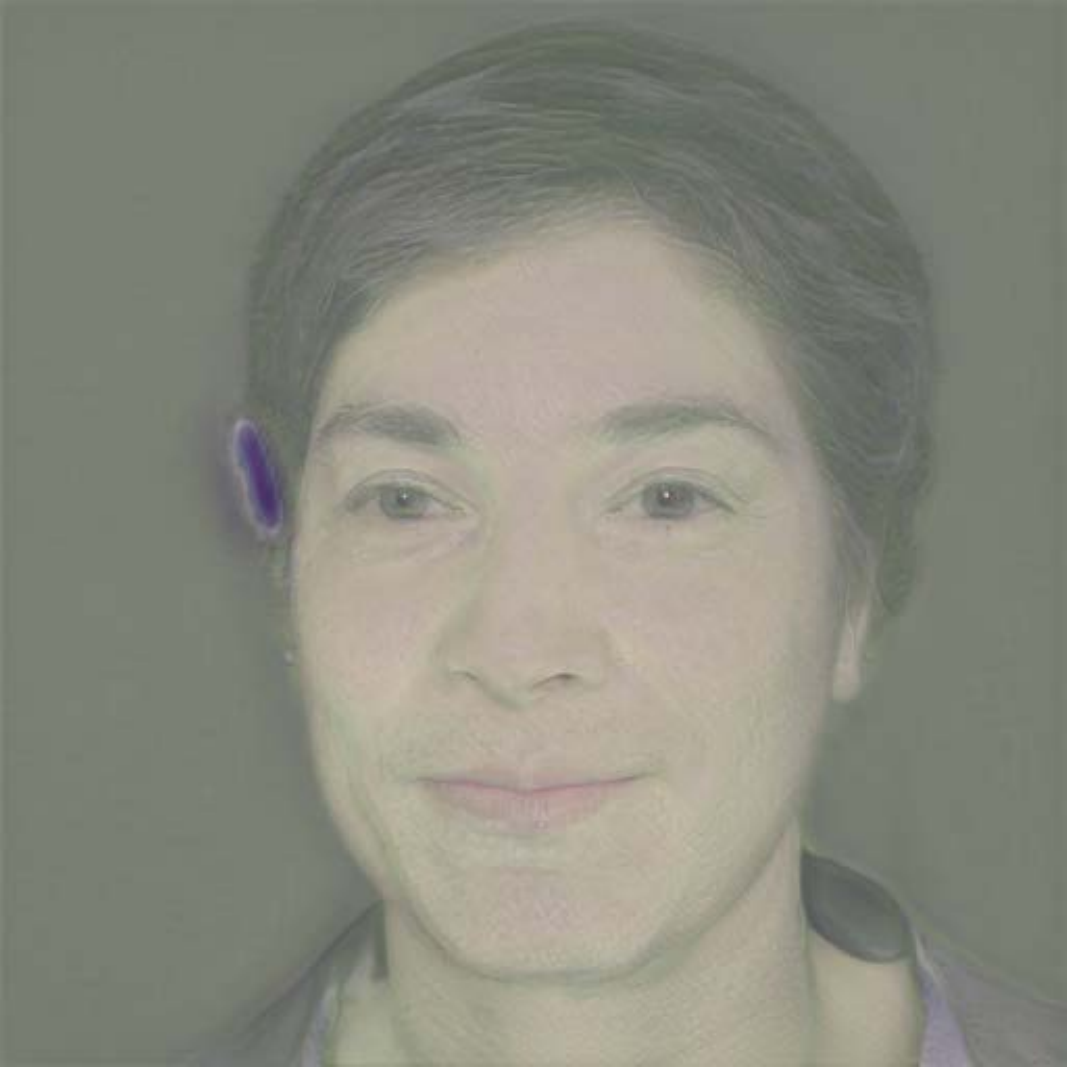}&
    \includegraphics[width=\swfigfirst]{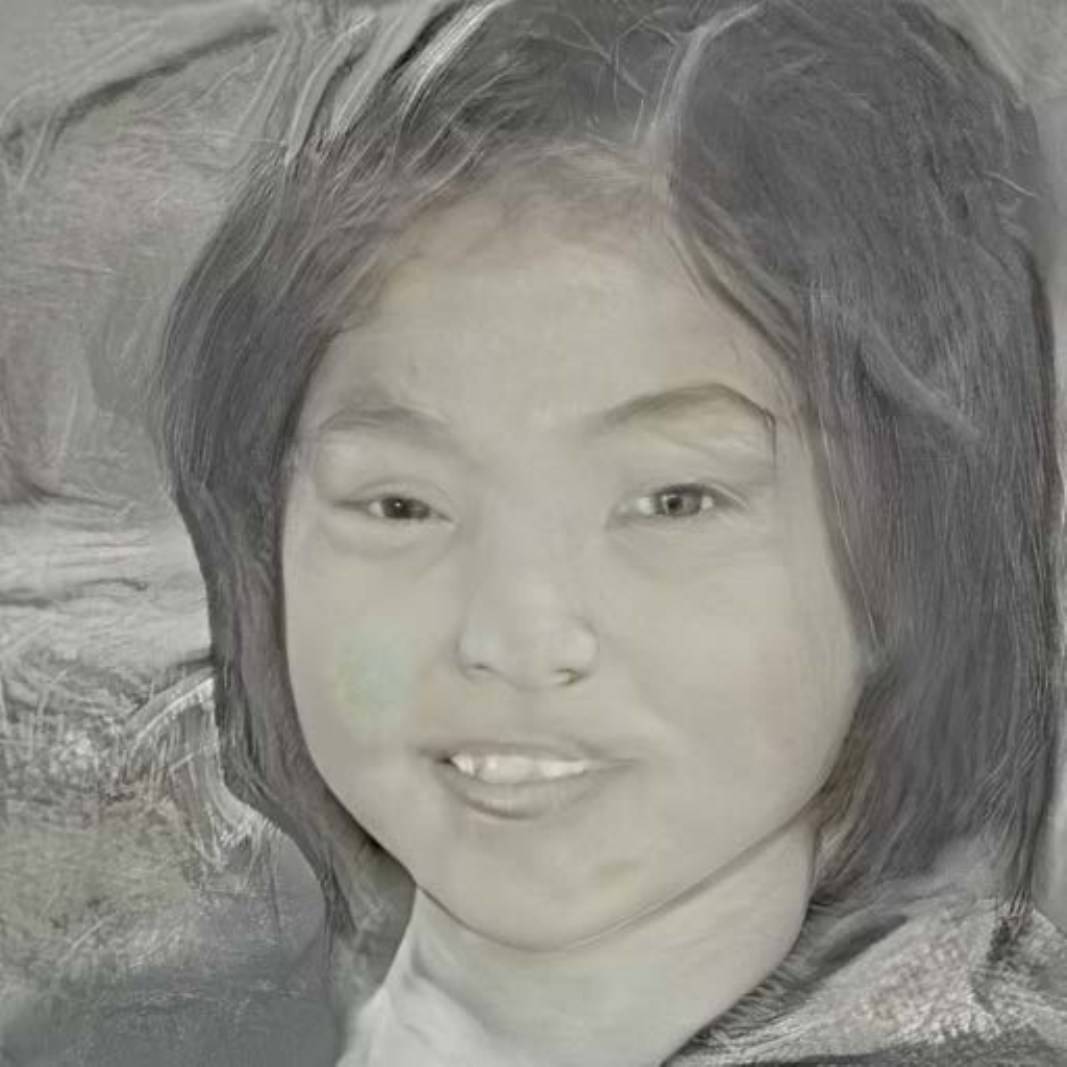}&
    \includegraphics[width=\swfigfirst]{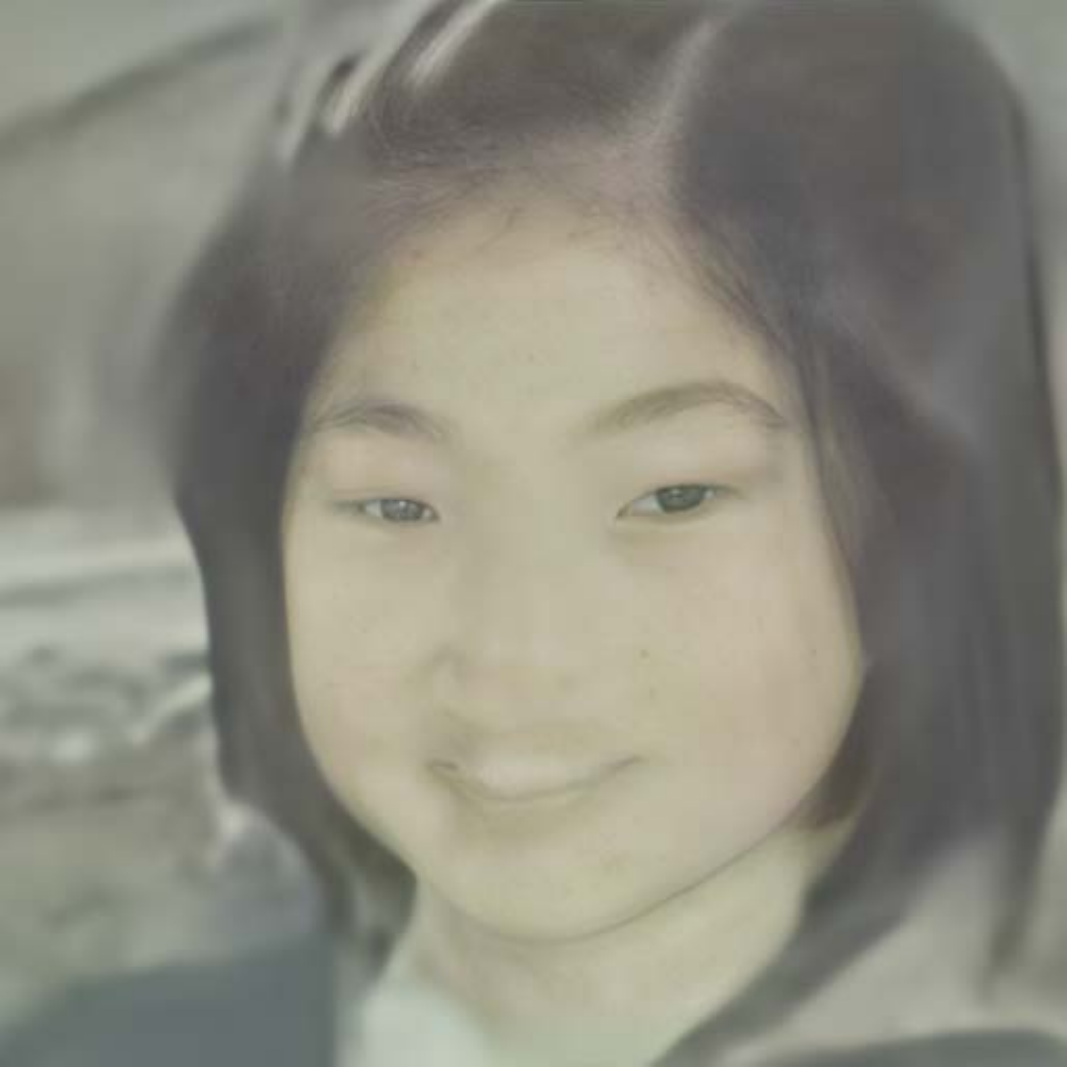}&
    \includegraphics[width=\swfigfirst]{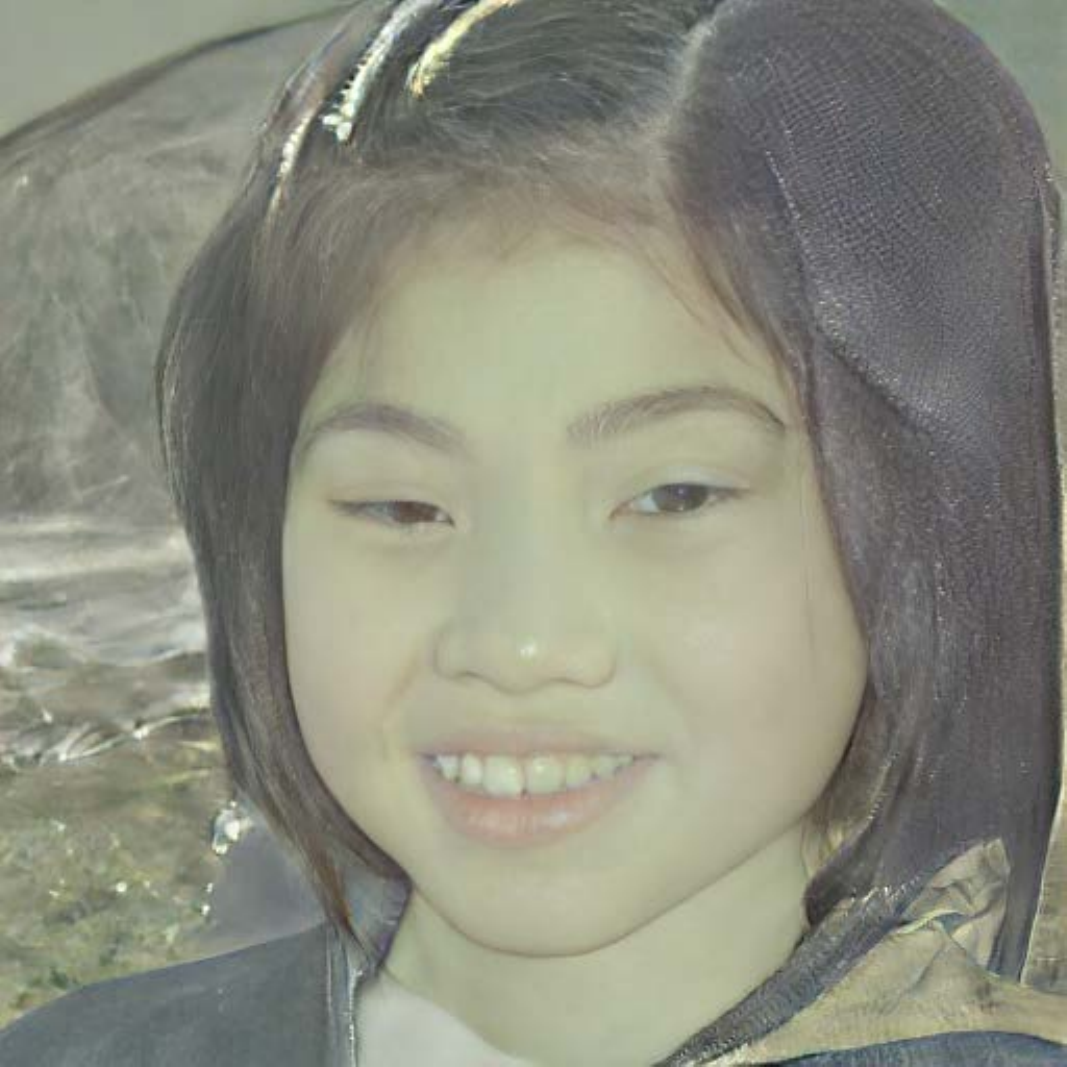}&
    \includegraphics[width=\swfigfirst]{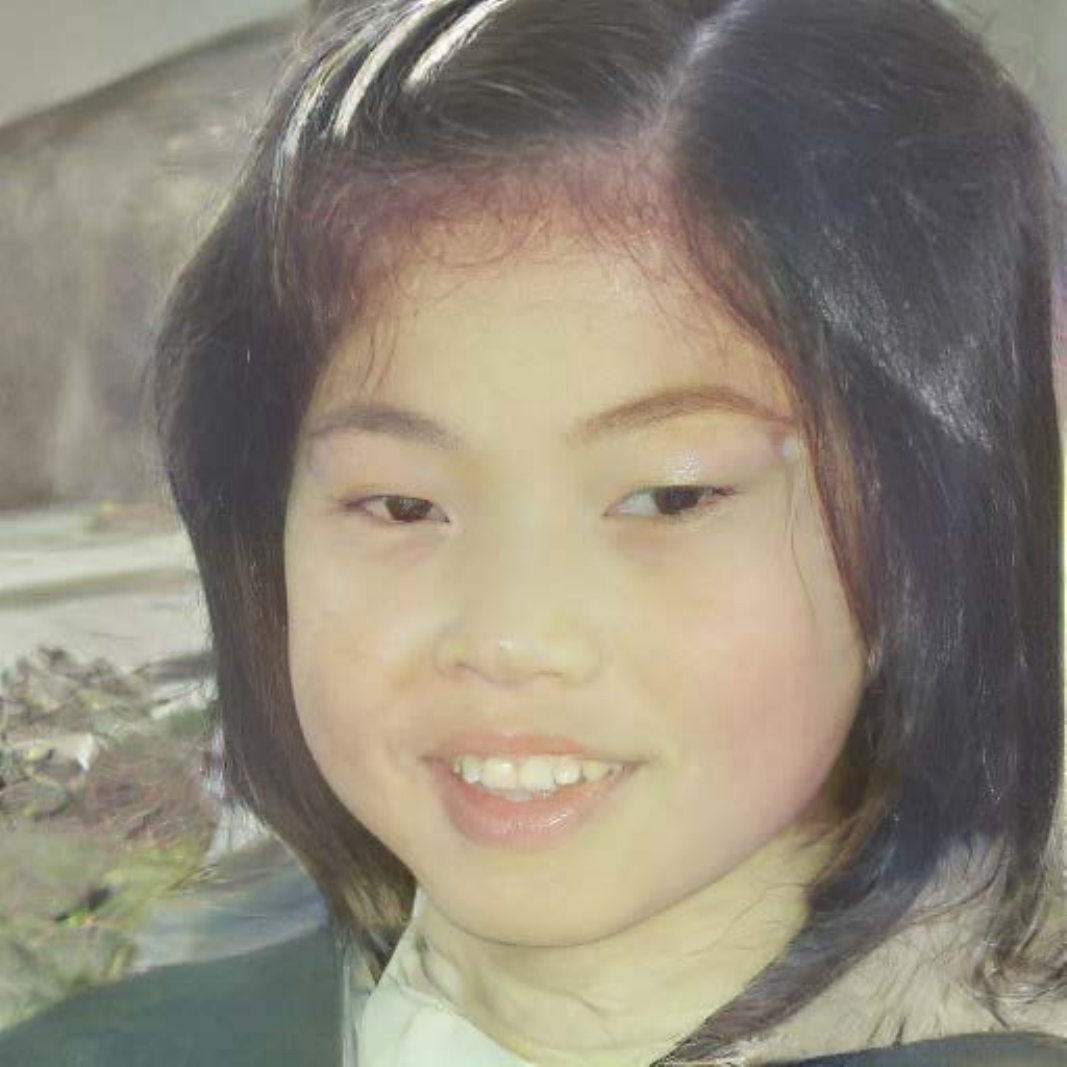}&
    \includegraphics[width=\swfigfirst]{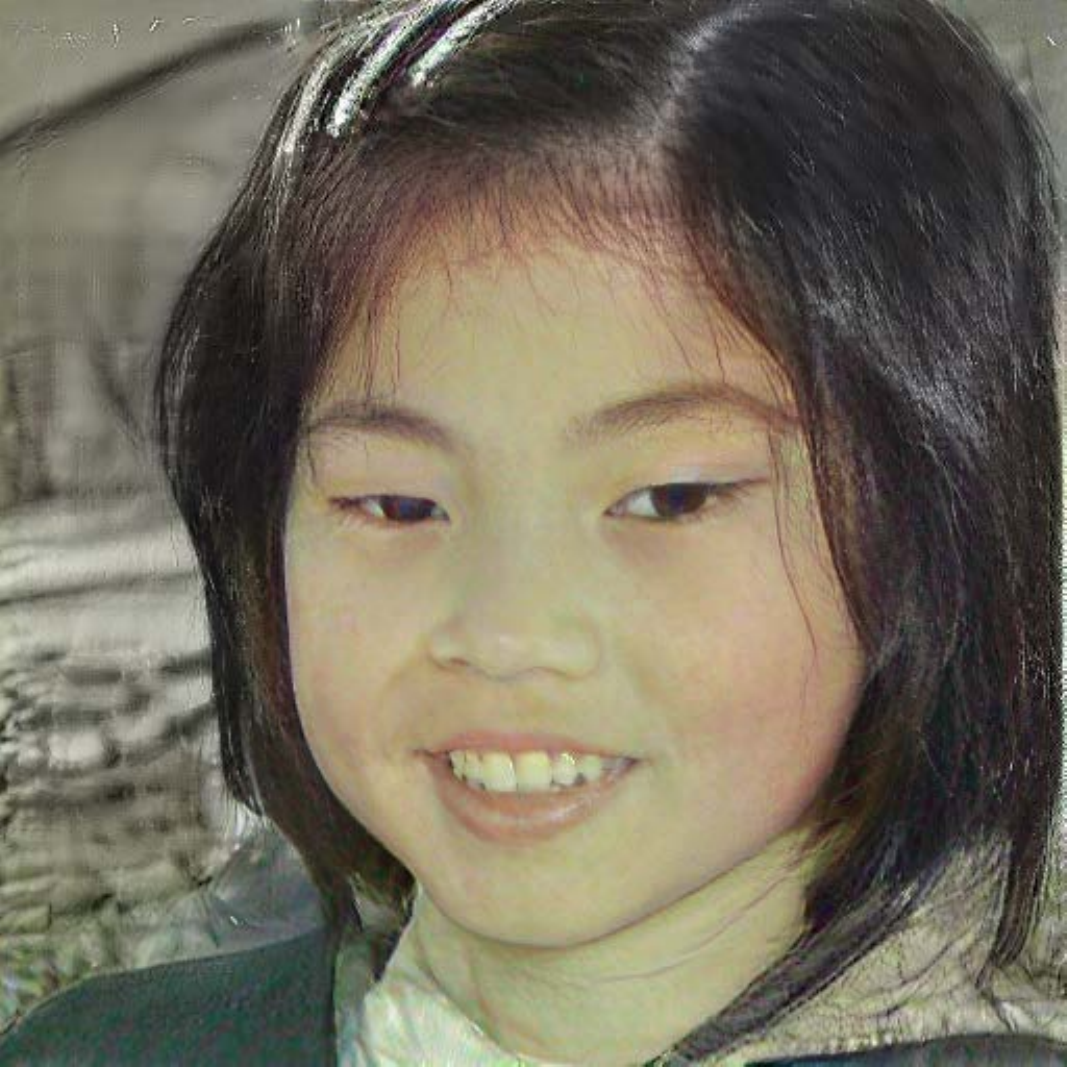}\\
    \includegraphics[width=\swfigfirst]{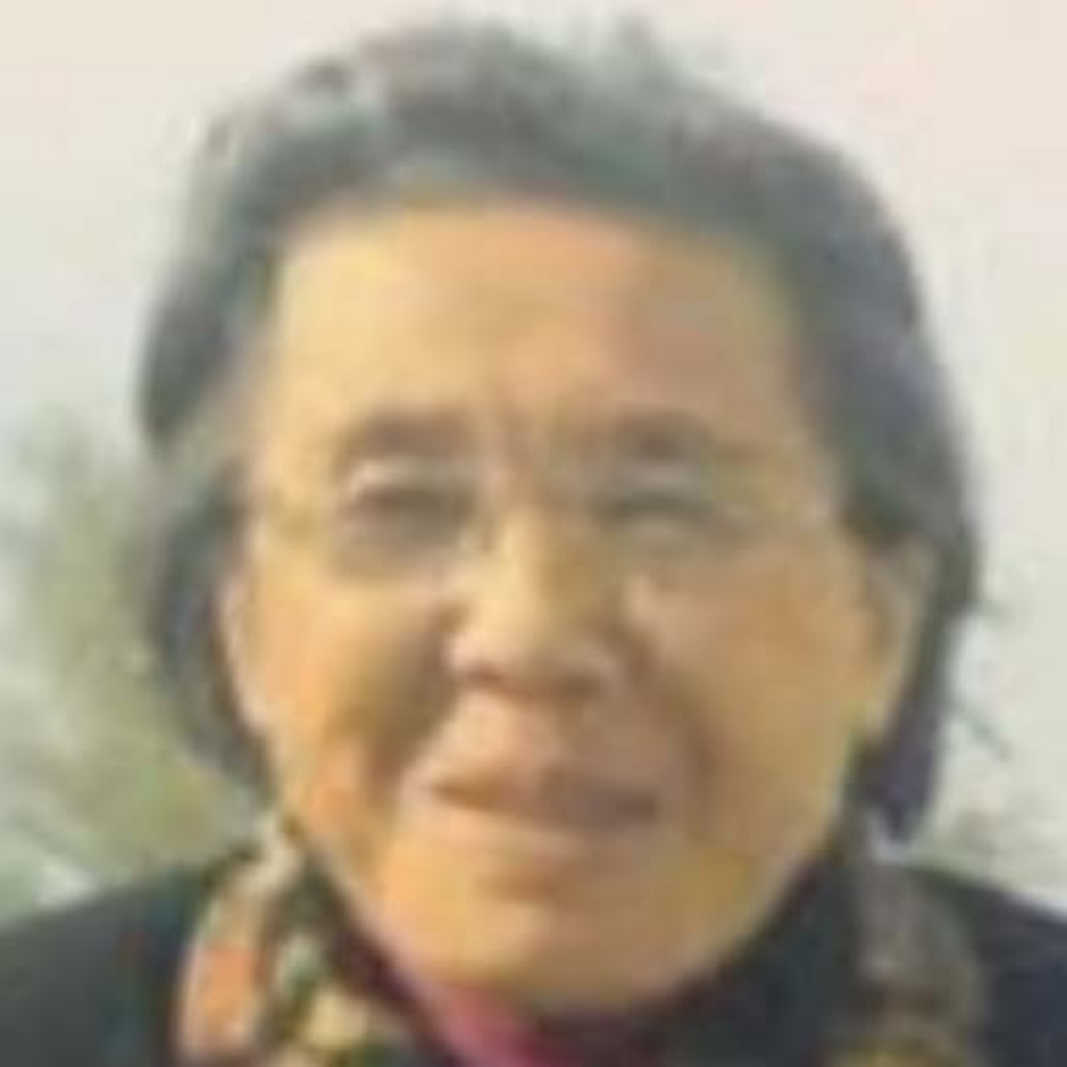}&
    \includegraphics[width=\swfigfirst]{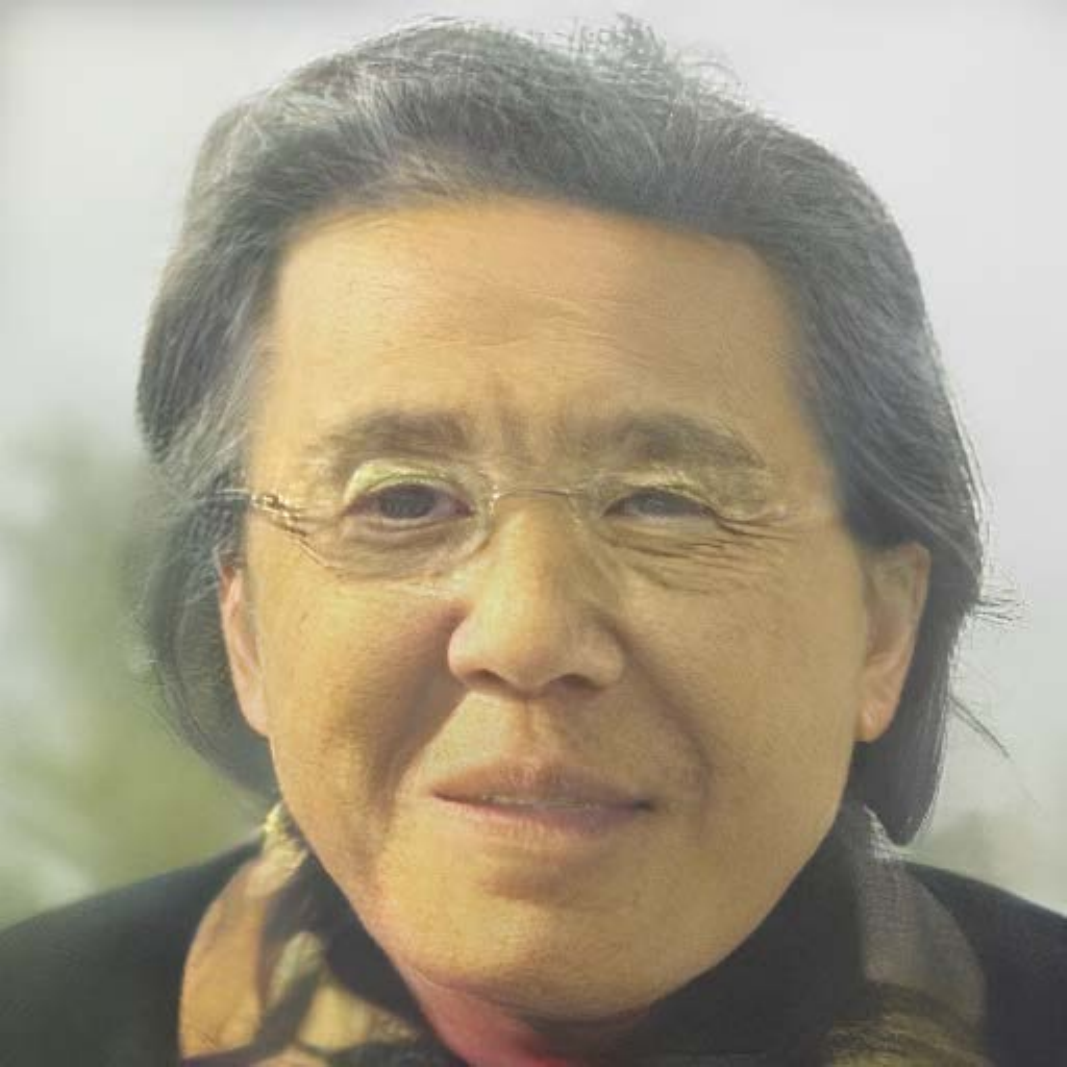}&
    \includegraphics[width=\swfigfirst]{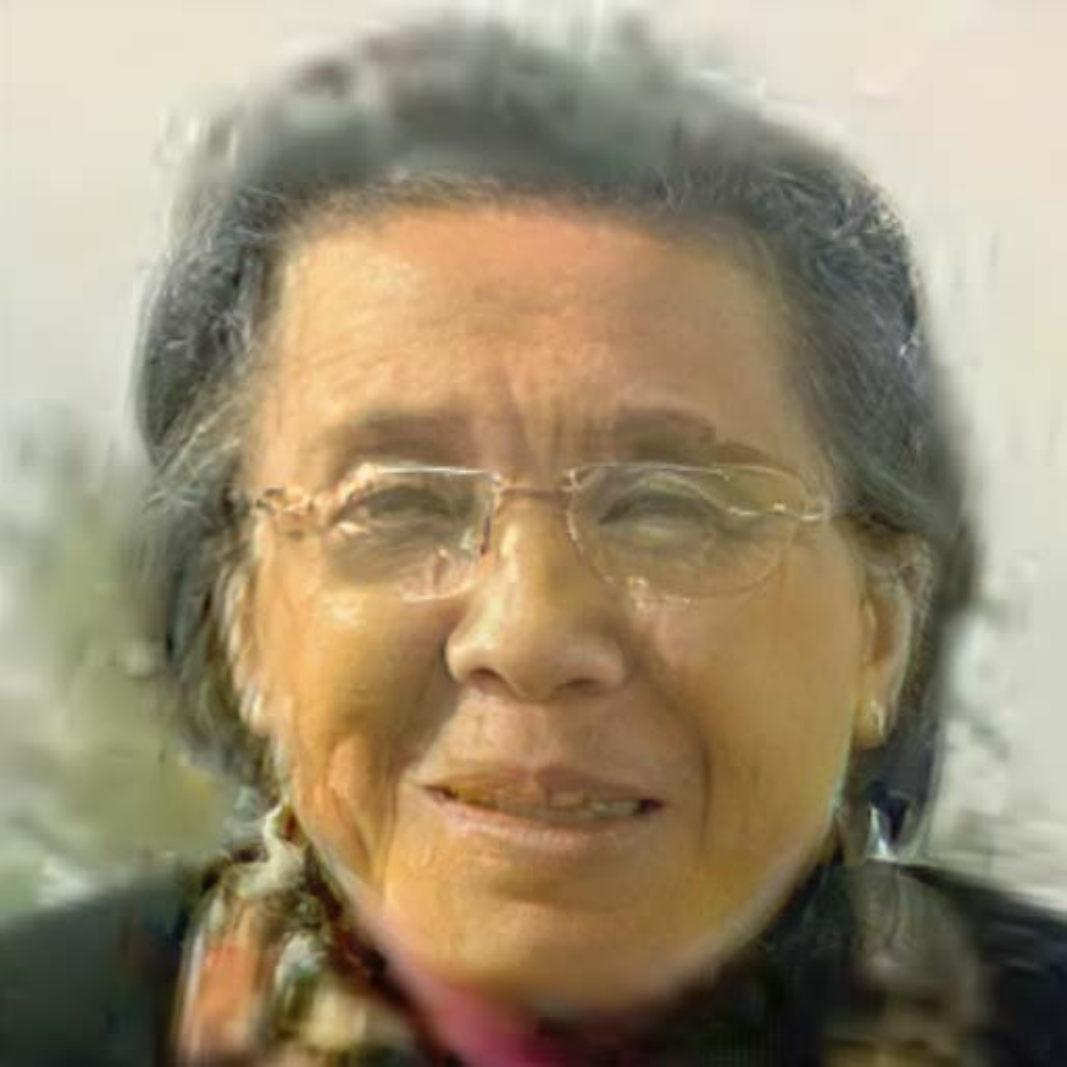}&
    \includegraphics[width=\swfigfirst]{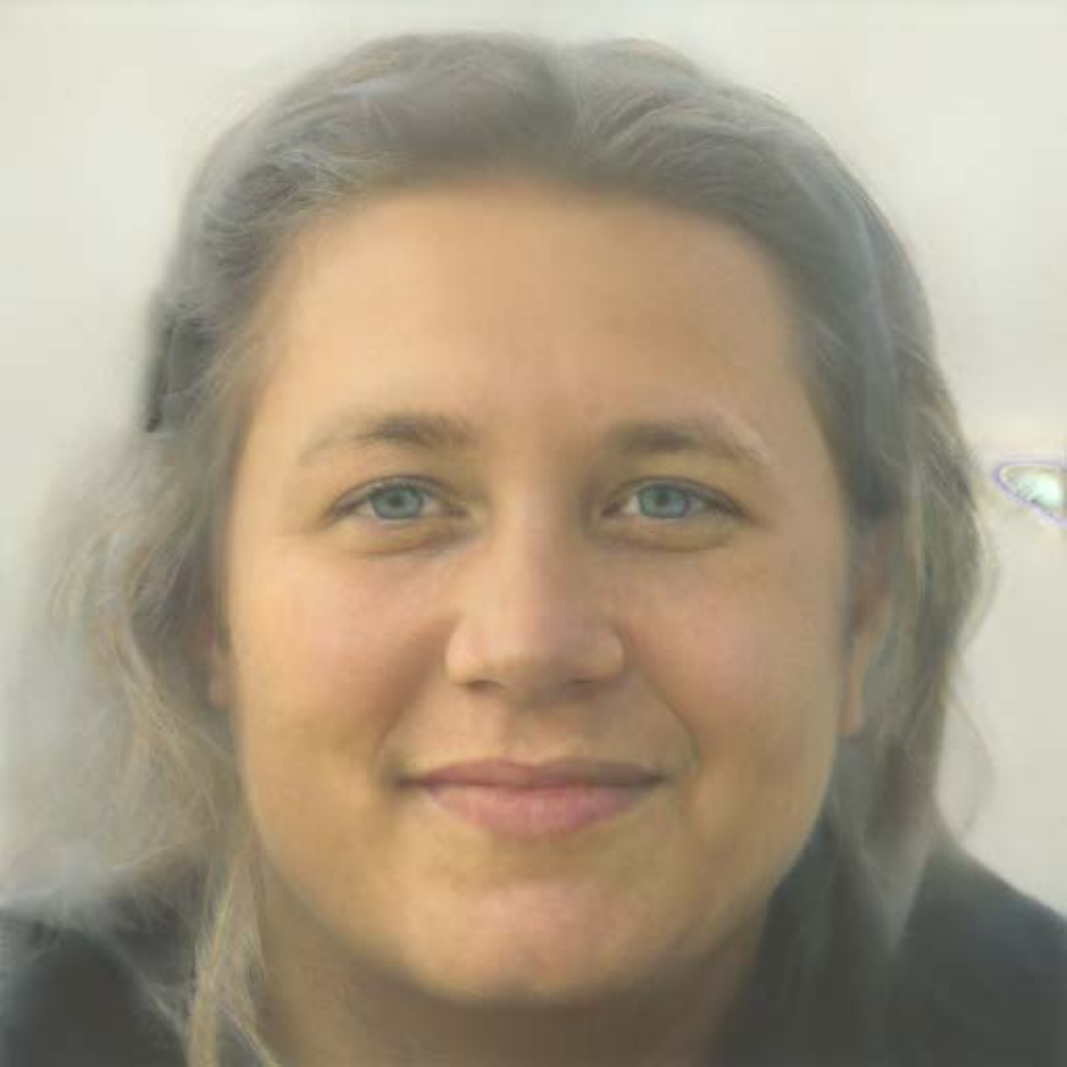}&
    \includegraphics[width=\swfigfirst]{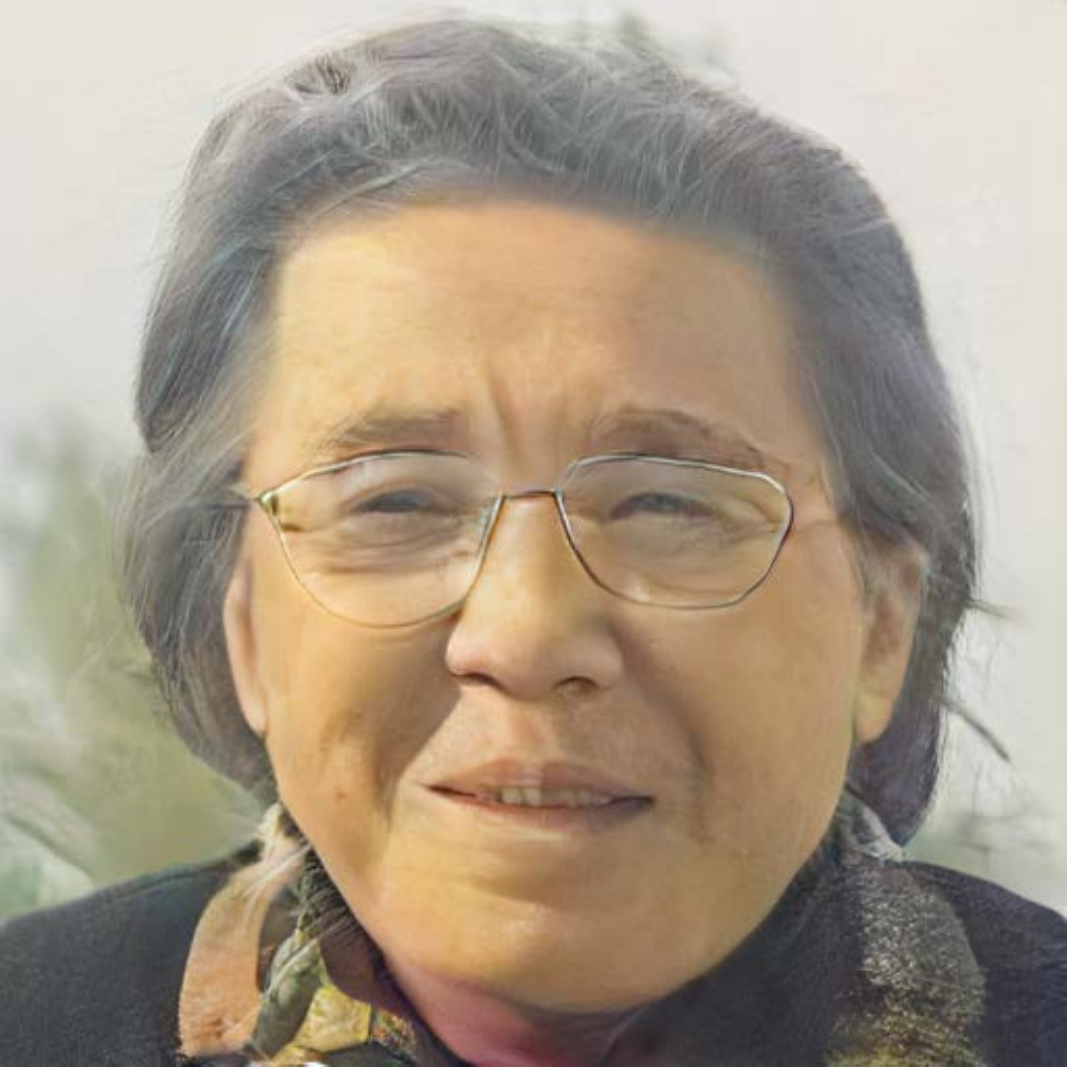}&
    \includegraphics[width=\swfigfirst]{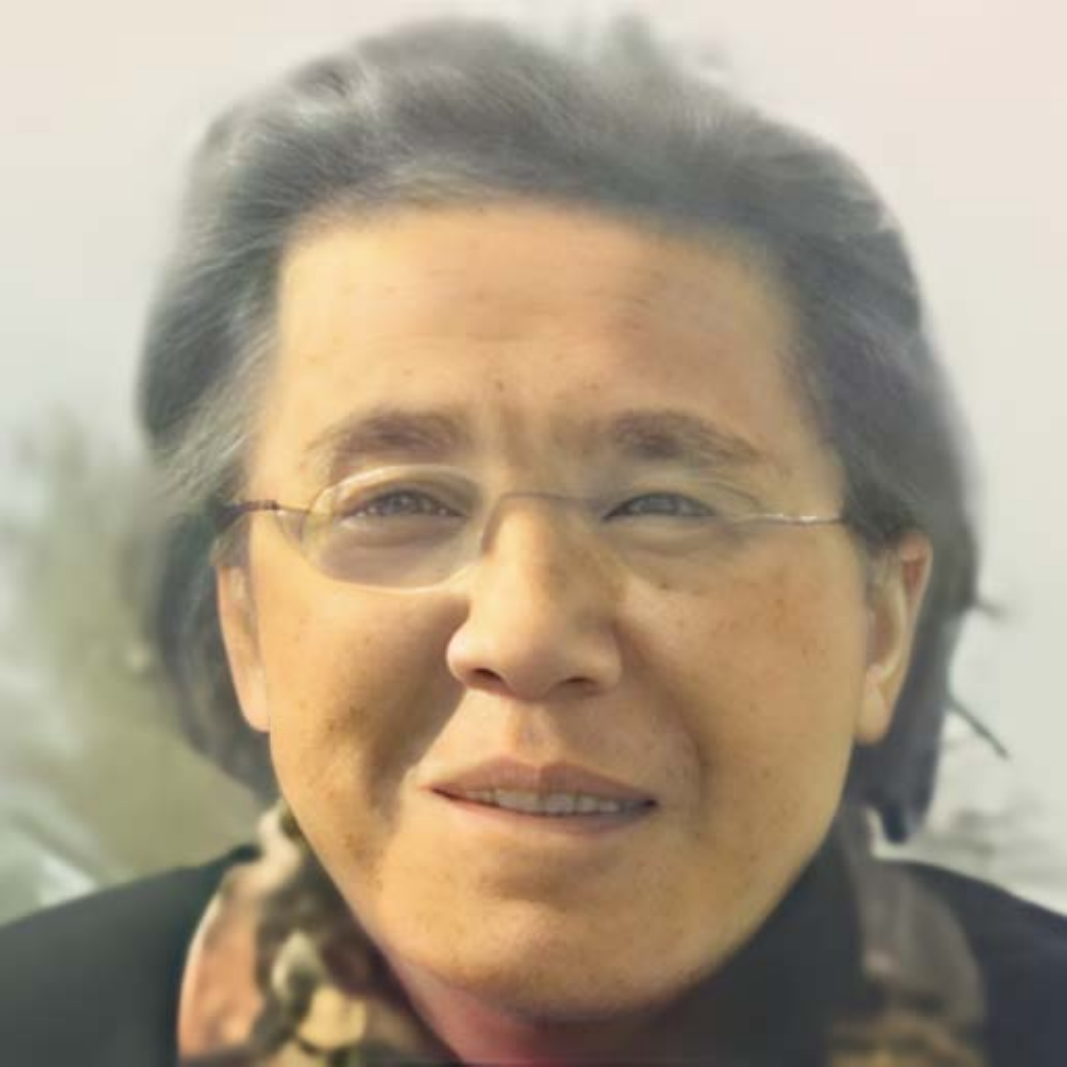}&
    \includegraphics[width=\swfigfirst]{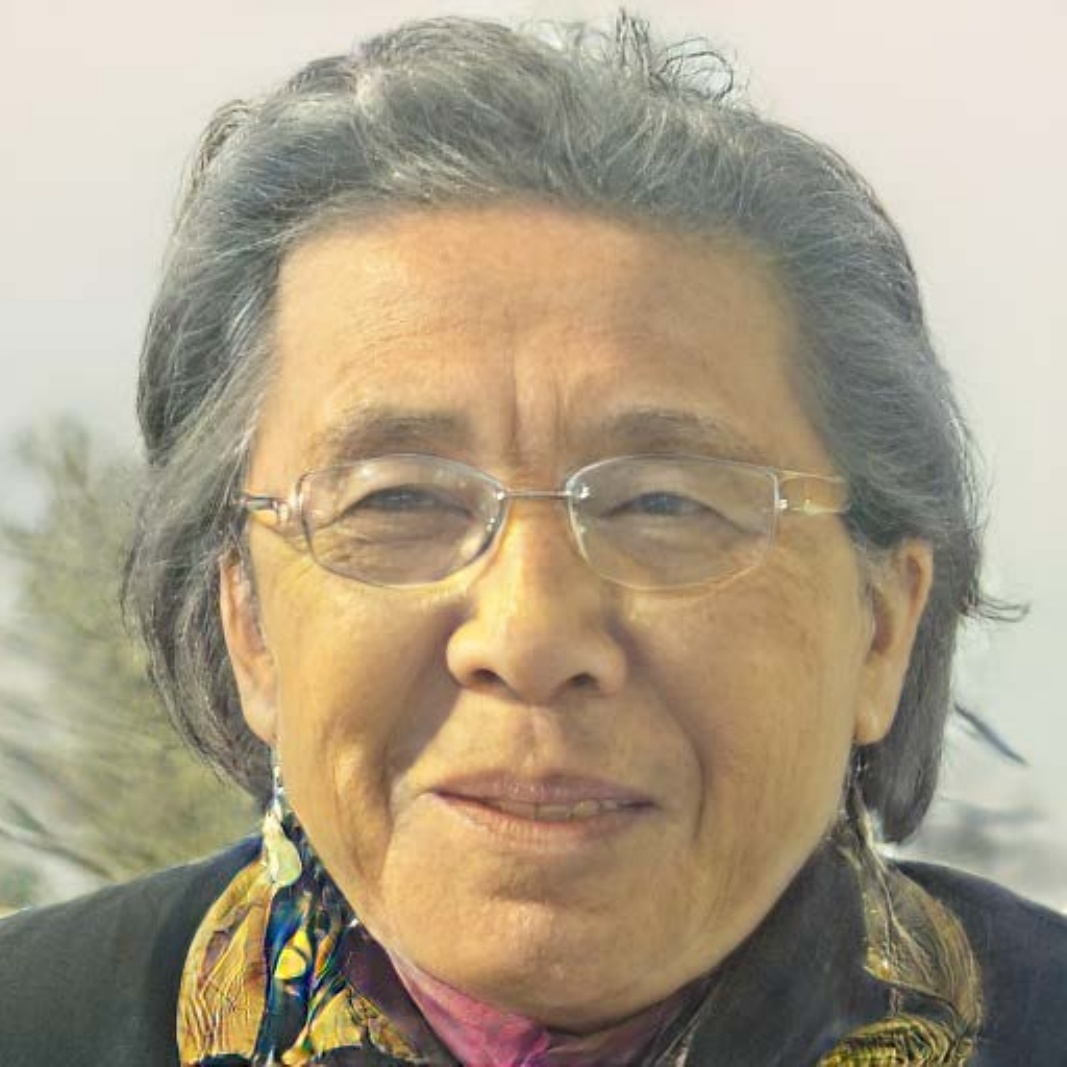}&
    \includegraphics[width=\swfigfirst]{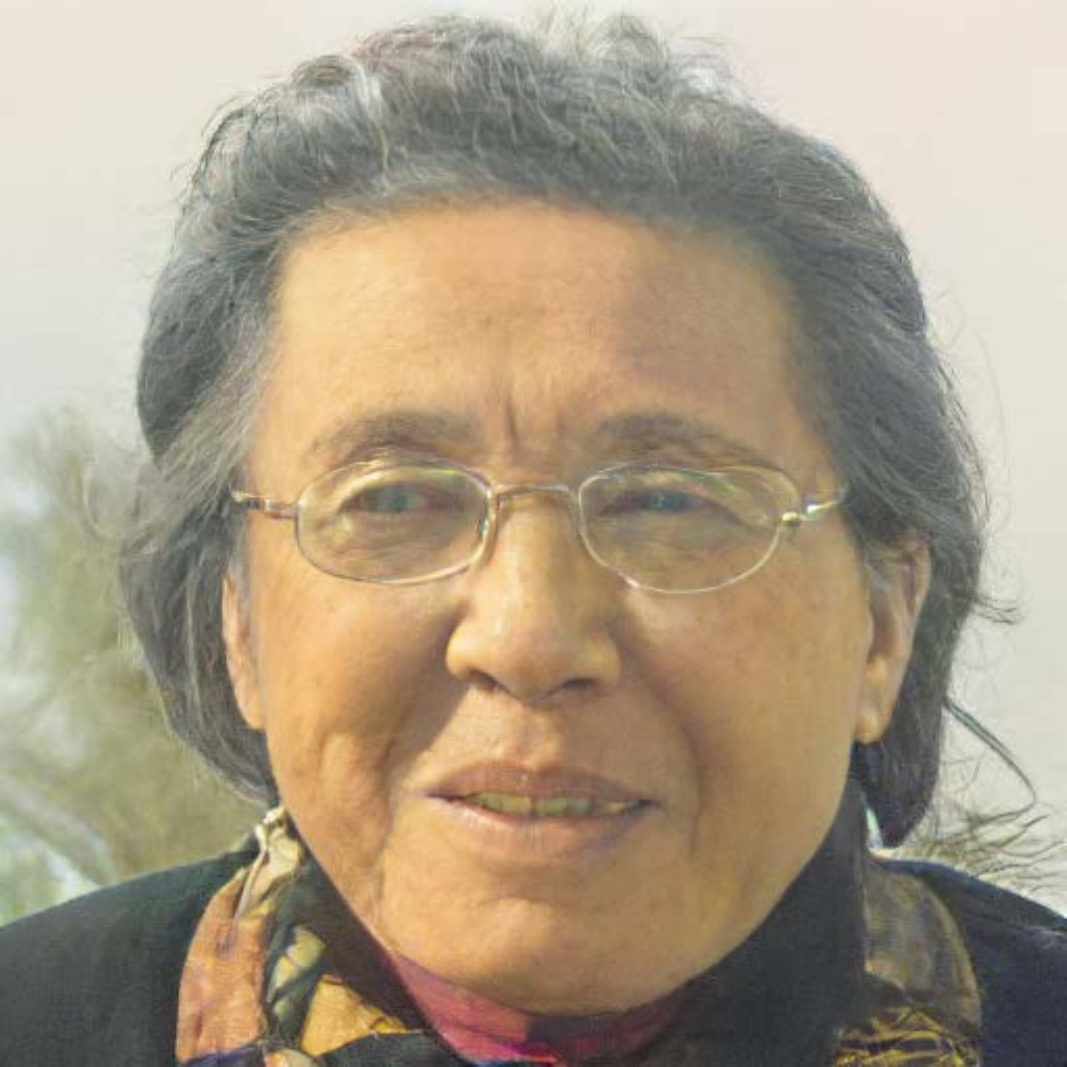}&
    \includegraphics[width=\swfigfirst]{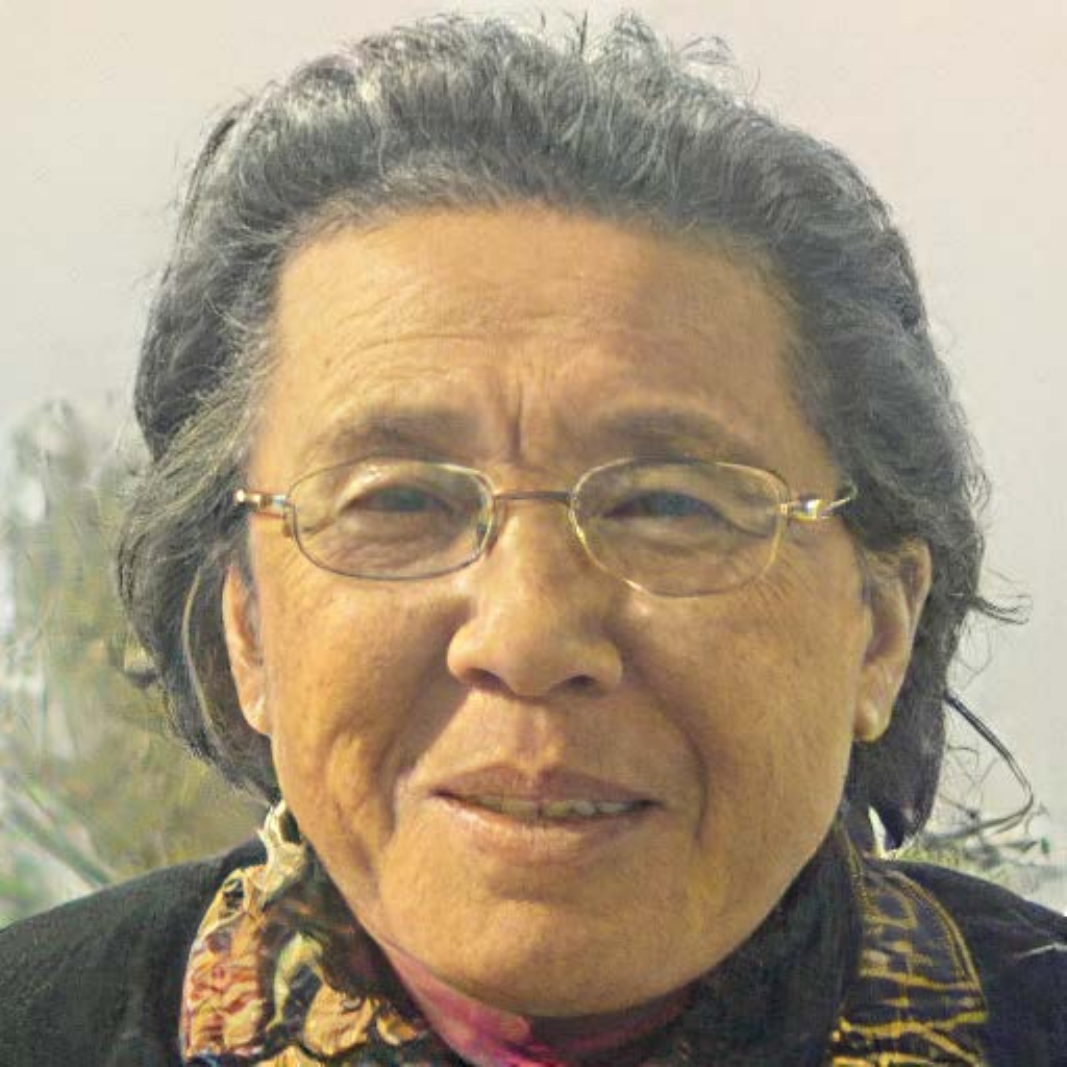}\\
    \includegraphics[width=\swfigfirst]{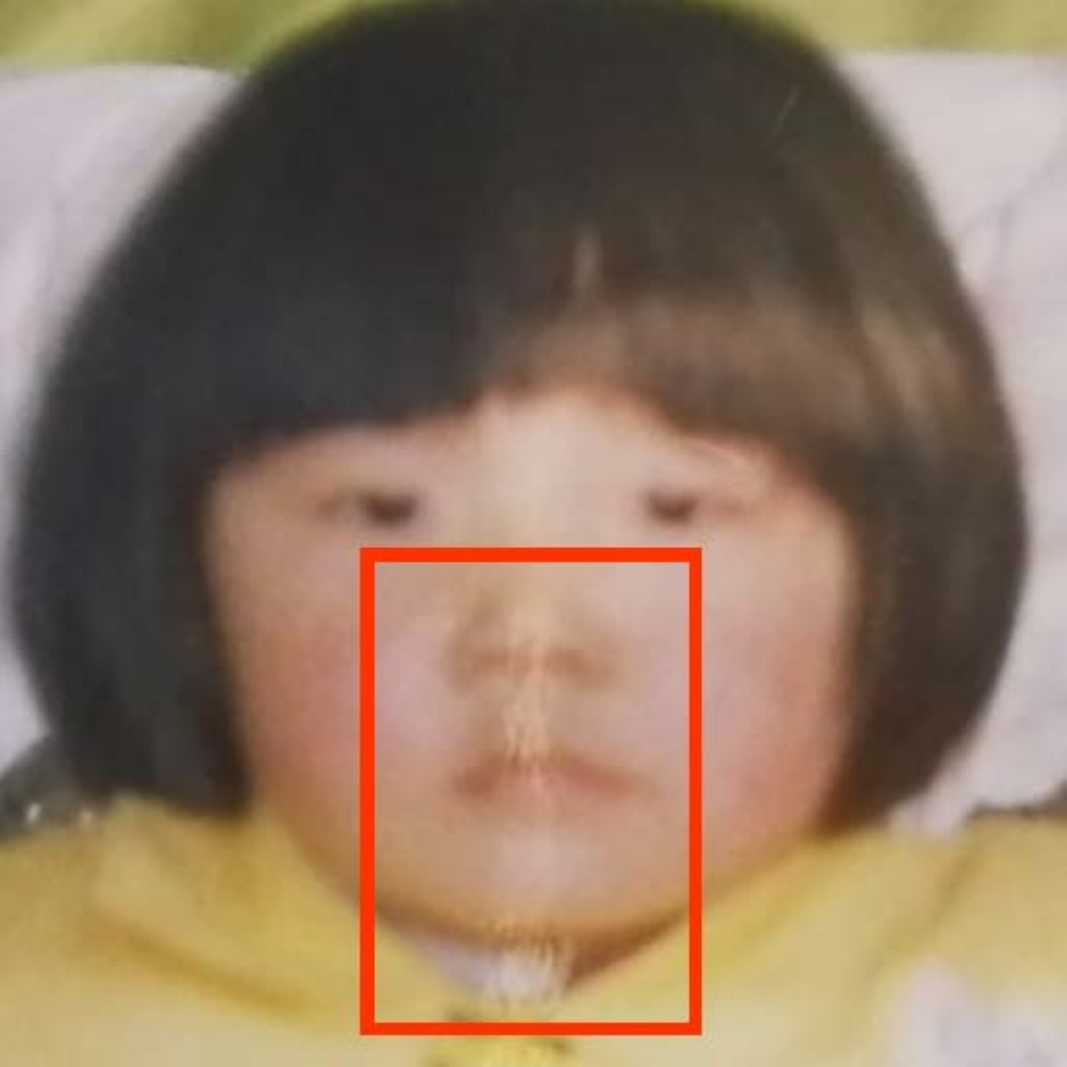}&
    \includegraphics[width=\swfigfirst]{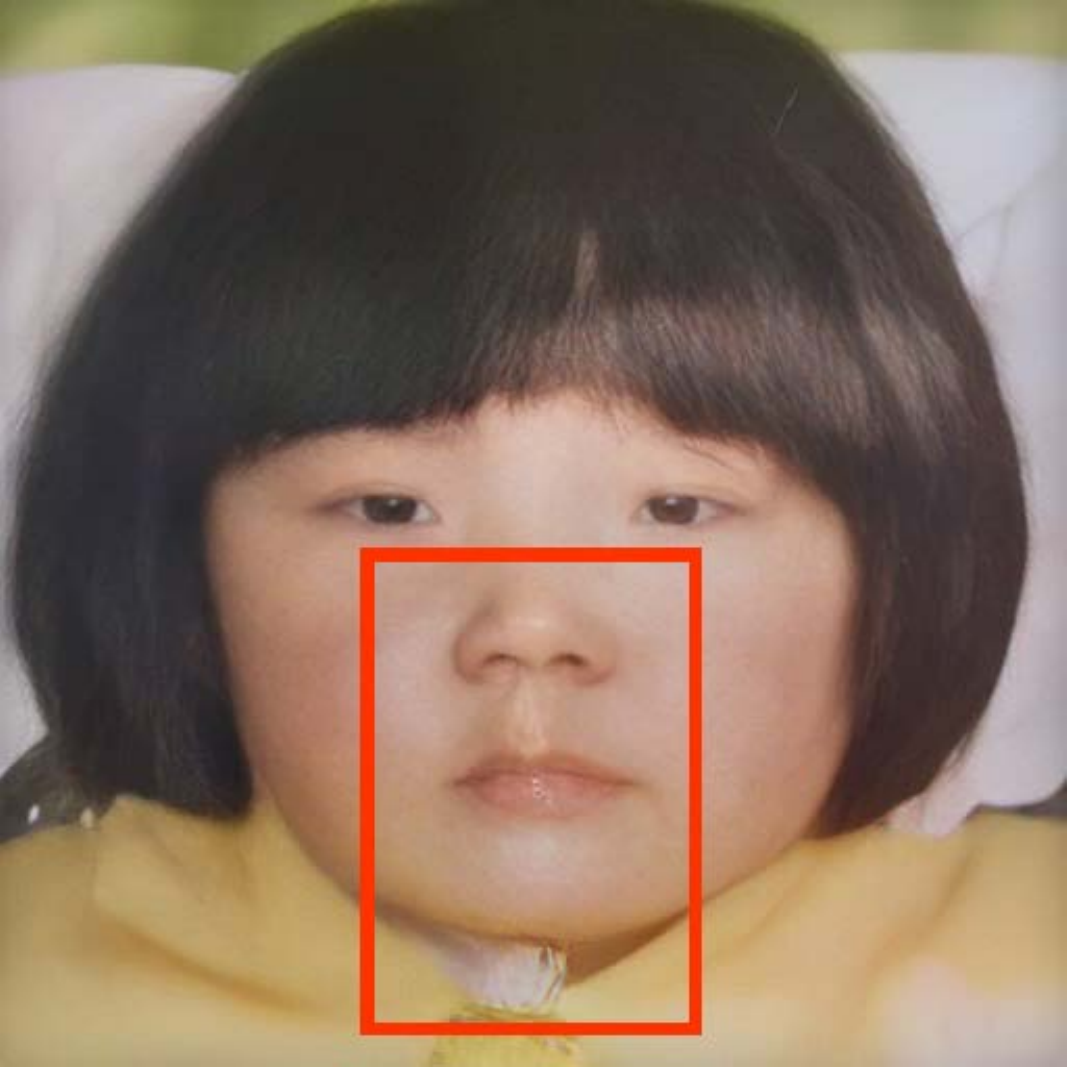}&
    \includegraphics[width=\swfigfirst]{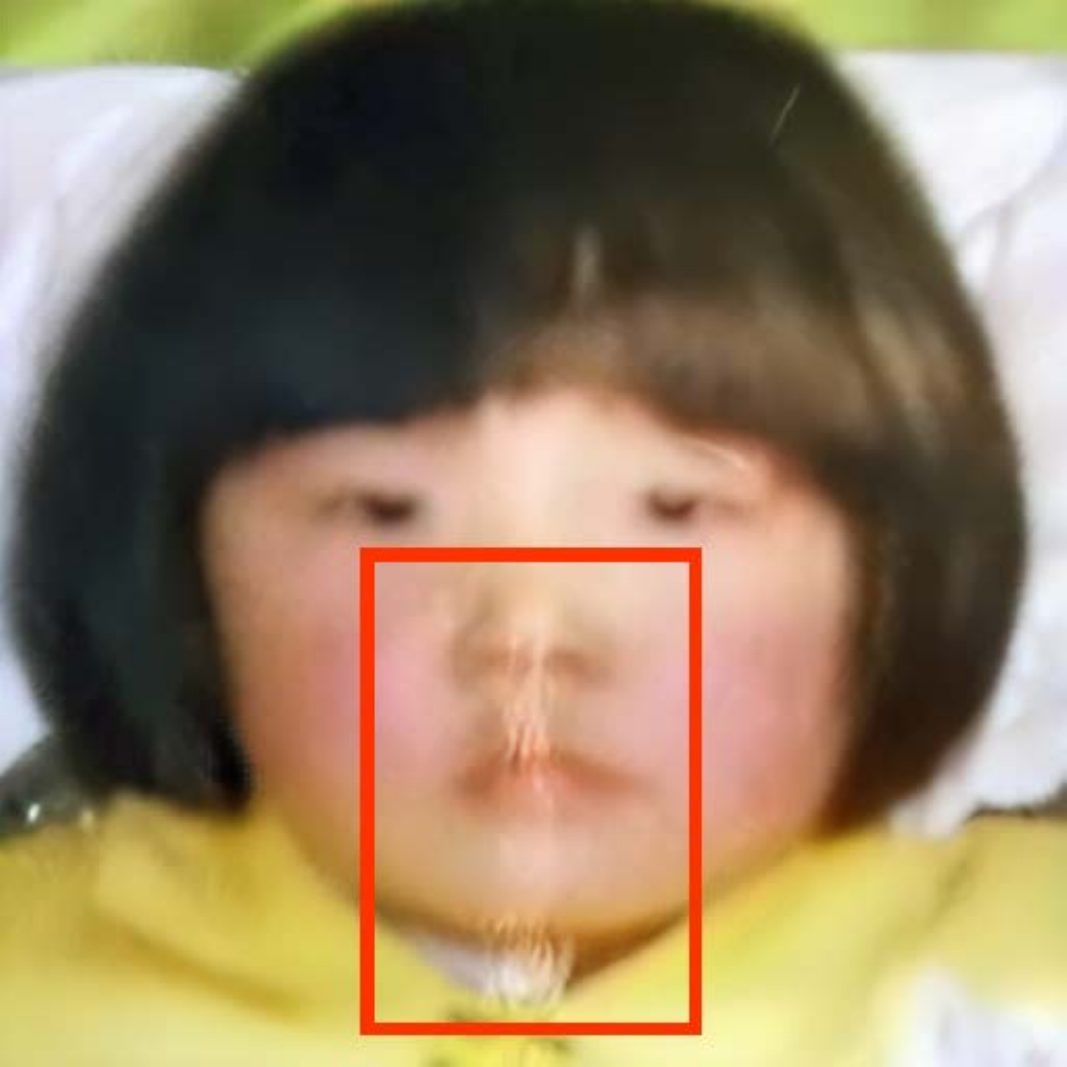}&
    \includegraphics[width=\swfigfirst]{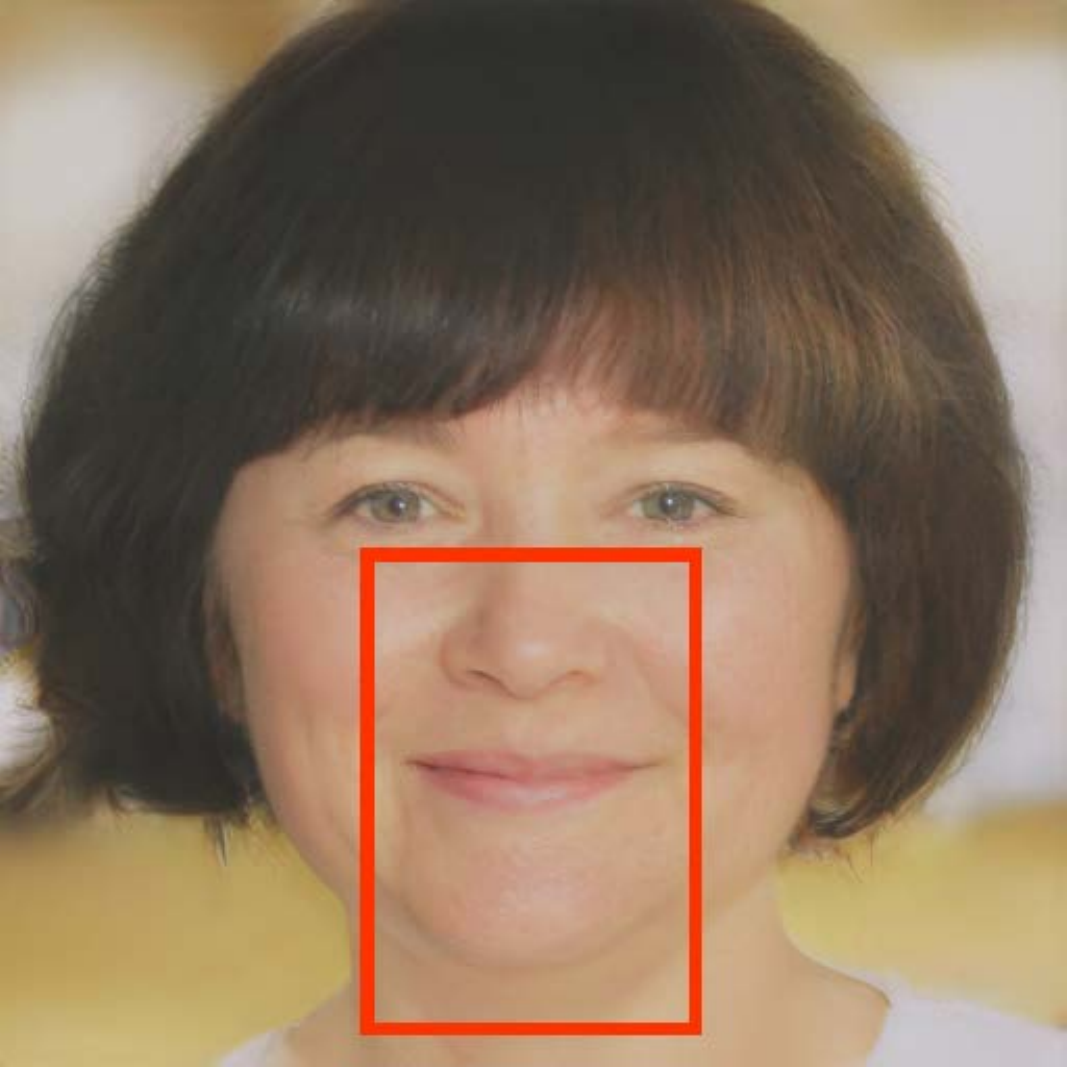}&
    \includegraphics[width=\swfigfirst]{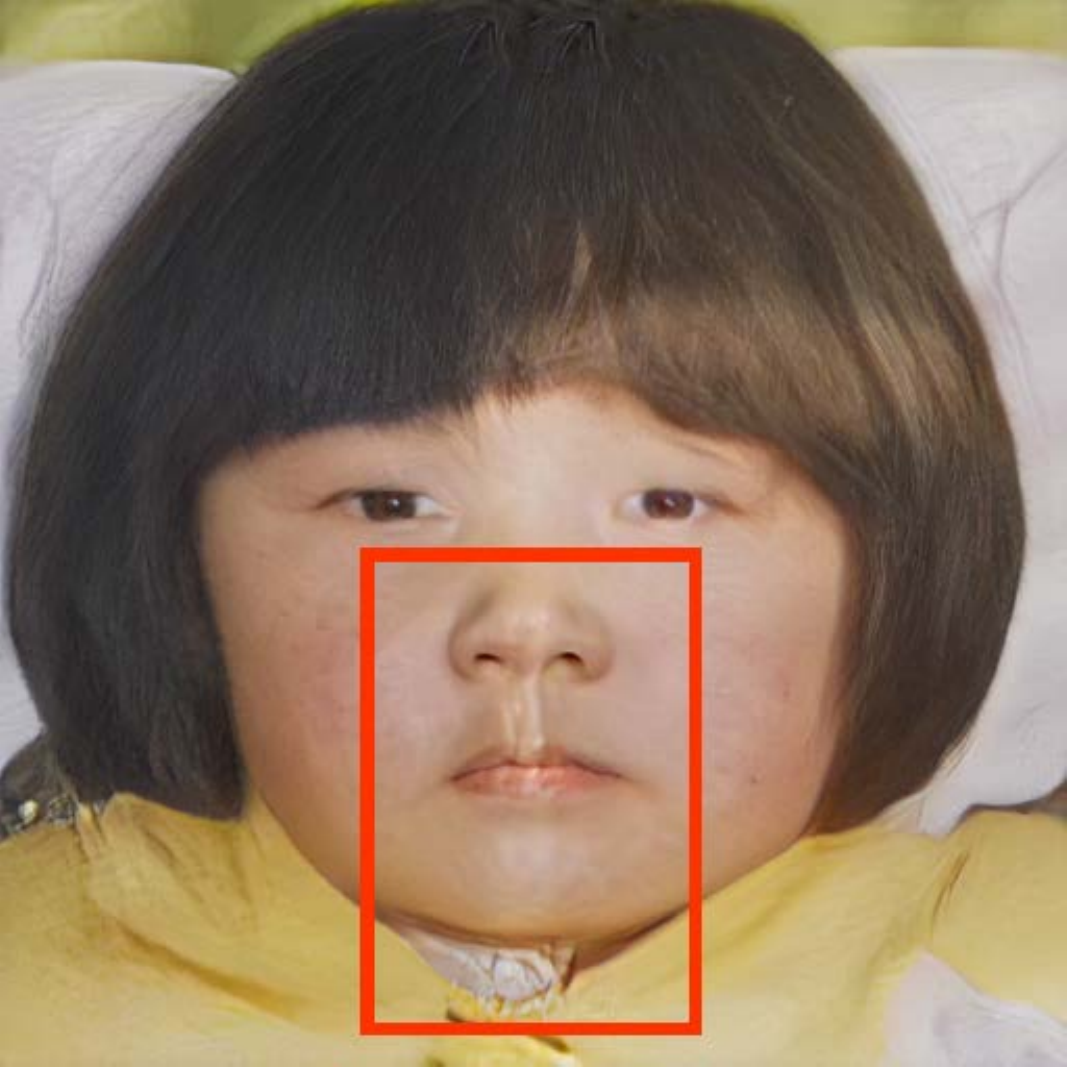}&
    \includegraphics[width=\swfigfirst]{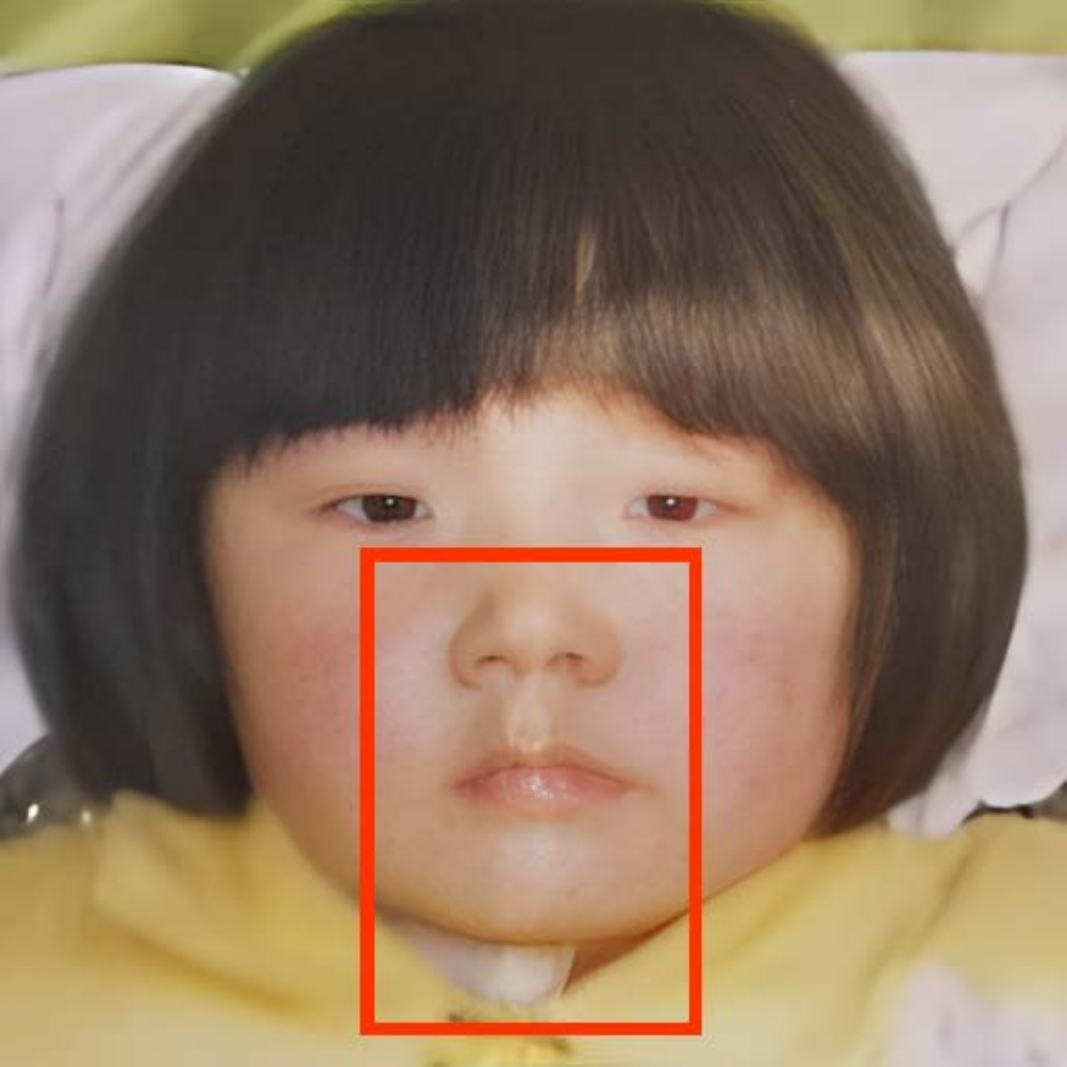}&
    \includegraphics[width=\swfigfirst]{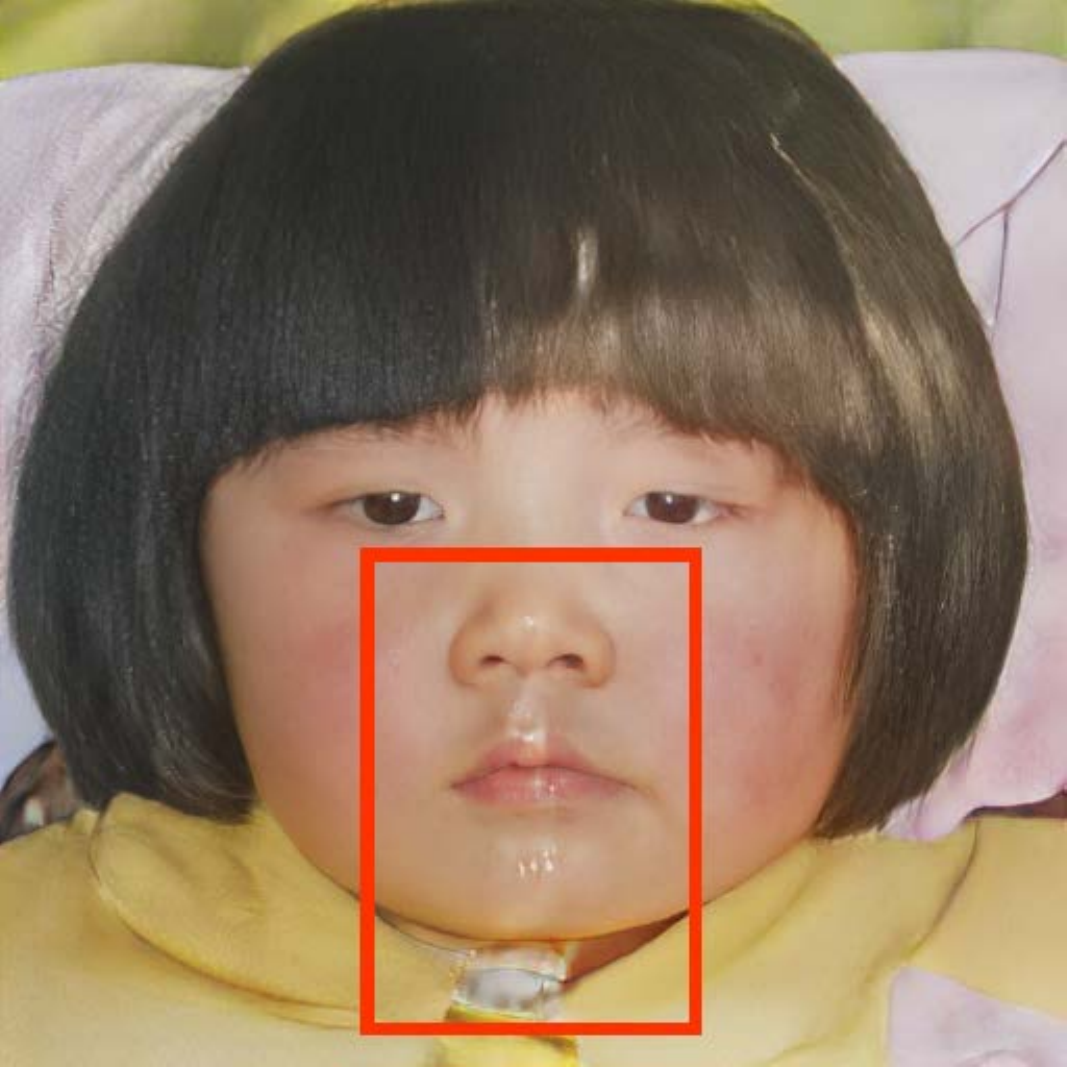}&
    \includegraphics[width=\swfigfirst]{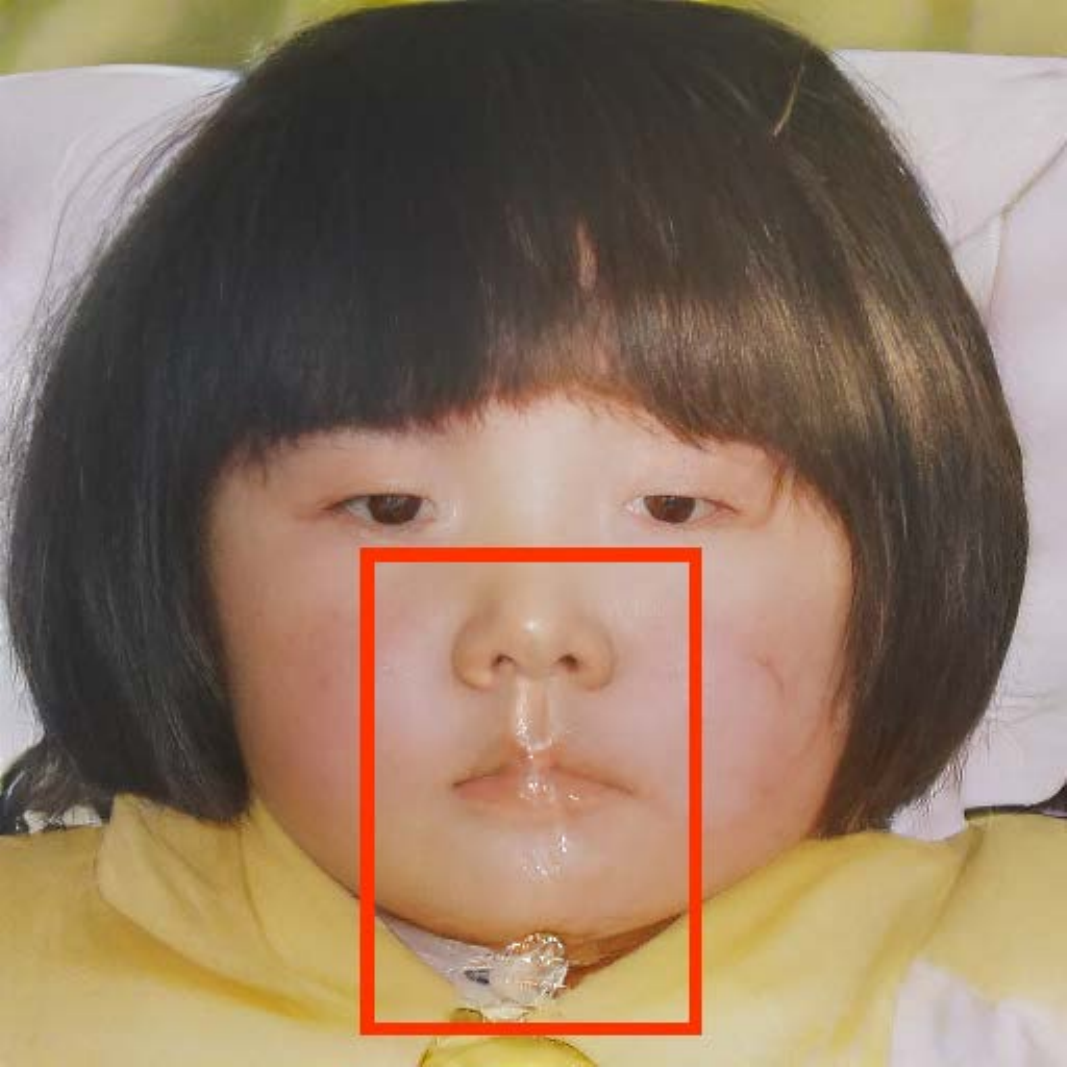}&
    \includegraphics[width=\swfigfirst]{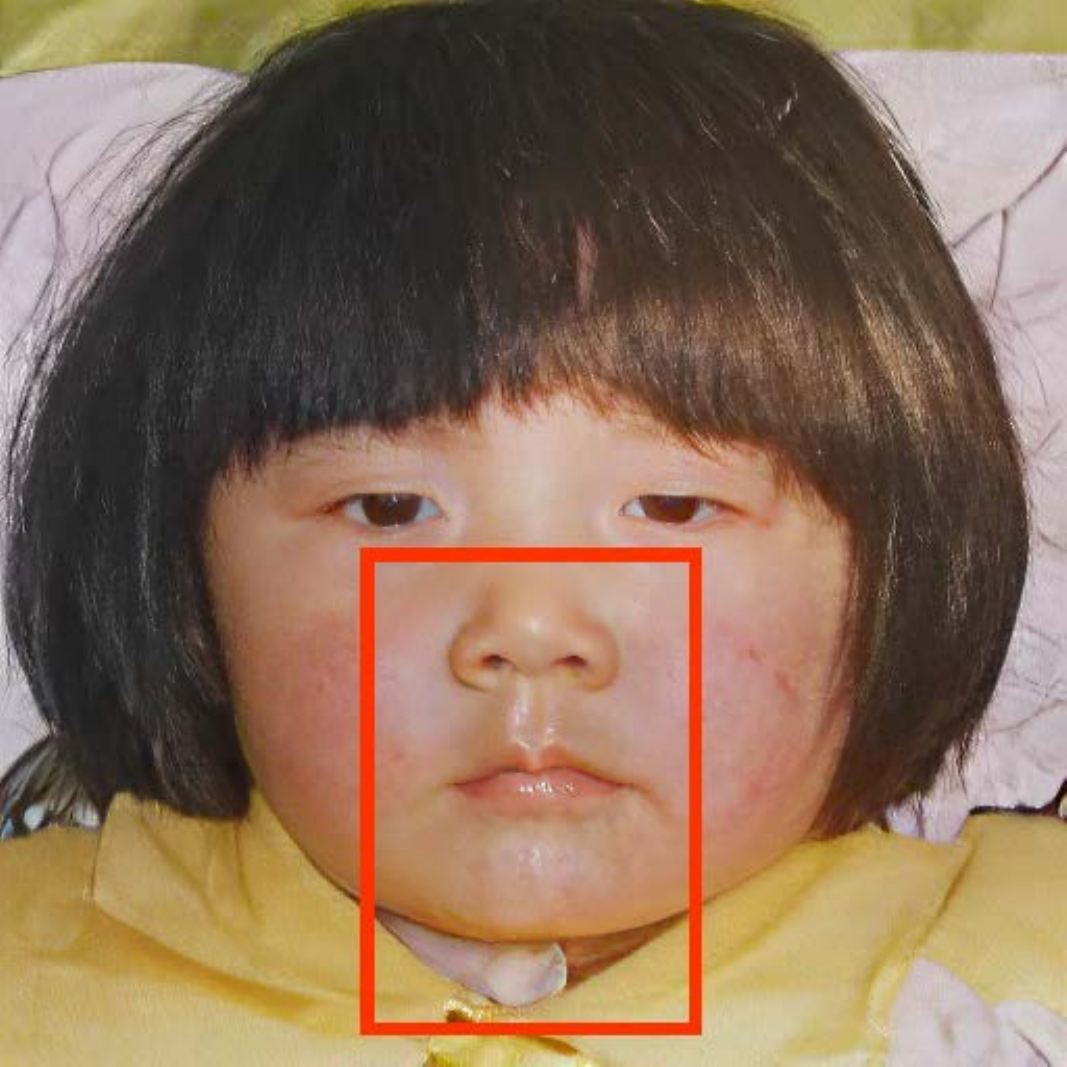}\\
   \label{fig:results}
  \scriptsize Input & \scriptsize DFDNet~\cite{li2020blind}  & \scriptsize Wan \textit{et~al.}~\cite{wan2020bringing} & \scriptsize PULSE~\cite{menon2020pulse} & \scriptsize PSFRGAN~\cite{chen2021progressive}  & \scriptsize GPEN~\cite{yang2021gan} & \scriptsize \textcolor{black}{VQFR}~\cite{gu2022vqfr} & \scriptsize{\textcolor{black}{RestoreFormer~\cite{wang2022restoreformer}}} & \scriptsize\textbf{RestoreFormer++} \\
  \scriptsize\textit{real-world} & \scriptsize\textit{ECCV 20} & \scriptsize\textit{CVPR 20} & \scriptsize\textit{CVPR 20} & \scriptsize\textit{CVPR 21}  & \scriptsize\textit{CVPR 21}  &\scriptsize \textit{ECCV 22} & \scriptsize\textit{\textcolor{black}{CVPR 22}} & \scriptsize\textit{\textbf{Ours}}
\end{tabular}
\end{center}
\caption{
\textcolor{black}{
Comparisons with state-of-the-art face restoration methods on some degraded real-world images.
Our conference version, RestoreFormer~\cite{wang2022restoreformer}, produces restored results with rich details and complete structures, making them more natural and authentic than the results of other methods. Our current version, RestoreFormer++, extends the multi-scale mechanism and EDM to remove haze from the degraded face images and process uneven degradation (highlighted with a red box in the third sample), resulting in a clearer and more pleasant look.
}
}
\label{fig:results}
\end{figure*}%

\IEEEPARstart{B}{lind} face restoration aims at recovering high-quality faces from a series of unknown degradations, such as blur, noise, downsampling, compression artifacts, \textit{etc}.
These degradations are complex and diverse in real-world scenarios, leaving limited information in the degraded face image.
Therefore, the face restored directly from its degraded one is not good enough, even with powerful DNN structures~\cite{krizhevsky2012imagenet,cao2017attention,huang2017wavelet,shi2019face,xu2017learning}. 
Introducing priors to complement additional high-quality details can effectively solve this issue~\cite{chen2021progressive,chen2018fsrnet,kim2019progressive,shen2018deep,yu2018face,yu2018super,zhu2016deep,li2020blind,wan2020bringing,yang2021gan,wang2021towards,gu2022vqfr}.

Despite the acknowledgment of progress, current prior-based algorithms mainly depend on geometric priors~\cite{chen2021progressive,chen2018fsrnet,kim2019progressive,shen2018deep,yu2018face,yu2018super,zhu2016deep} or recognition-oriented references~\cite{li2020blind}, which are not accordant to the restoration task and thus lead to sub-optimal performance. 
The geometric priors are landmarks~\cite{chen2018fsrnet,kim2019progressive}, facial parsing maps~\cite{chen2021progressive,shen2018deep}, or facial component heatmaps~\cite{yu2018face} that mainly provide shape information to aid face restoration. 
Recognition-oriented references like the facial component dictionaries in DFDnet~\cite{li2020blind} are extracted with a recognition model and only cover limited facial components, such as eyes, mouth, and nose.
Therefore, the restored faces of these algorithms tend to lack details.
For example, in Fig.~\ref{fig:results}, the results of PSFRGAN~\cite{chen2021progressive}, whose priors are facial parsing maps, and DFDnet~\cite{li2020blind} fail to recover facial details, especially in hair areas.
Although the generative priors encapsulated in a face generation network aim at face reconstruction and achieve superior performance compared to the previous two kinds of priors, their restored results still fail to yield fine-grained facial details or exist obvious artifacts. 
Examples are the restored results of Wan~\textit{et~al.}~\cite{wan2020bringing} and GPEN~\cite{yang2021gan} in Fig.~\ref{fig:results}.

On the other hand, effectively integrating the identity information in the degraded face and high-quality details in the priors is a critical step to attaining face images in both realness and fidelity.
\textcolor{black}{
However, current methods either take the degraded faces as supervision, e.g., PULSE~\cite{menon2020pulse}, or locally combine these two kinds of information by pixel-wise concatenation~\cite{dogan2019exemplar,li2020enhanced,li2018learning}, spatial feature transform (SFT)~\cite{wang2018recovering,chen2021progressive,li2020blind,wang2021towards}, or deformable operation~\cite{gu2022vqfr}.
They ignore the useful contextual information in the face image and its interplay with priors, and thus most of them cannot trade off the fidelity and realness of their restored results well.}
A typical example is PULSE~\cite{menon2020pulse}. As shown in Fig.~\ref{fig:results}, their restored results perform well in realness, \textcolor{black}{but their identities cannot be preserved.}

In this work, we propose RestoreFormer++, which introduces fully-spatial attention mechanisms to model the contextual information in the face image and its interplay with priors matched from a reconstruction-oriented dictionary.
Unlike the existing ViT methods~\cite{carion2020end,chen2021pre,dosovitskiy2020image,zhu2020deformable} that achieve fully-spatial attentions with multi-head self-attention mechanism (MHSA) \textcolor{black}{(Fig.~\ref{fig:framework}~(a))}, our RestoreFormer++ is equipped with multi-head cross-attention mechanism (MHCA) \textcolor{black}{(Fig.~\ref{fig:framework}~(b))} whose queries are the features of degraded face image while key-value pairs are high-quality facial priors.
%
%
\textcolor{black}{In addition, }\textcolor{black}{
MHCAs are applied to multi-scale features, enabling RestoreFormer++ to model the contextual information based on both semantic and structural information and effectively improve the restored performance in both realness and fidelity.
}
It is also worth mentioning that the priors adopted in our work have better quality since they are from a reconstruction-oriented high-quality dictionary (ROHQD).
Its elements are learned from plenty of uncorrupted faces by a high-quality face generation neural network implemented with the idea of vector quantization~\cite{oord2017neural}.
They are rich in high-quality facial details specifically aimed at face reconstruction (see Fig.~\ref{fig:dictionary} for a more intuitive comparison with the recognition-oriented dictionary).

In addition, RestoreFomer++ contains an extending degrading model (EDM) to generate more realistic degraded face images for alleviating the synthetic-to-real-world gap and further improving its robustness and generalization toward real-world scenarios.
\textcolor{black}{
Observations show that in the real world, besides blur, noise, downsampling, and compression artifacts, haze and uneven degradation are also common. Relevant examples are shown in Fig.~\ref{fig:results}.
However, existing methods cannot handle these degradations well.
Therefore, we introduce haze and uneven degradation into our EDM, which enables RestoreFormer++ to effectively remove the haze covered in the degraded face images and avoid the artifacts raised by uneven degradation.
%
Besides, EDM applies a spatial shift operation on the high-quality face before synthesizing the degraded face to reduce the effect introduced by inaccurate face alignment.
Due to the specificity of face structure, aligning the degraded face to a reference face (in this work, the reference face is from FFHQ~\cite{karras2019style}, and its landmarks are shown as green points in Fig.~\ref{fig:real} and Fig.~\ref{fig:EDM}) is helpful for the restoration of face images~\cite{li2020blind,chen2021progressive,wang2021towards}.
However, misalignment caused by severe degradation will lead to errors while restoring with existing methods.
For example, as shown in the second sample in Fig.~\ref{fig:EDM}, its left eyebrow is aligned with the left eye of the reference image, and the existing methods, such as PSFGAN~\cite{chen2021progressive}, GFP-GAN~\cite{wang2021towards}, and our conference version~\cite{wang2022restoreformer}, tend to restore the left eye near the eyebrow area instead of its original area in the degraded face image.
The small spatial shift adopted in EDM can improve the tolerance of RestoreFormer++ for face alignment error, thus improving its restoration performance as in Fig.~\ref{fig:EDM}~(g).
}

This work is an extension of our conference version~\cite{wang2022restoreformer}. 
In this version, we strengthen the work from three aspects.
\textcolor{black}{
First, we extend our multi-head attention mechanisms used for fusing the degraded facial features and their corresponding high-quality facial priors from single-scale to multi-scale. 
This enables RestoreFormer++ to model contextual information based on both semantic and structural information, effectively improving the restored performance in both realness and fidelity.
}
Second, we proposed an extending degrading model (EDM) to alleviate the synthetic-to-real-world gap and further improve the robustness and generalization of our RestoreFormer++ toward real-world scenarios.
Finally, we conduct more experiments and analyses to verify the superiority of RestoreFormer++ against existing methods and the contributions of each component in RestoreFormer++.

In conclusion, our main contributions are as follows: 
\begin{itemize}
    \item We propose RestoreFormer++, which on the one hand introduces multi-head cross-attention mechanisms to model the fully-spatial interaction between the degraded face and its corresponding high-quality priors and on the other hand, explores an extending degrading model to synthesize more realistic degraded face images for model training. It can restore face images with higher realness and fidelity for both synthetic and real-world scenarios.
    \item We introduce a reconstruction-oriented high-quality dictionary learning algorithm to generate priors that are more accordant to the face restoration task and thus provide suitable priors to RestoreFormer++ to restore faces with better realness.
    \item The extending degrading model contains more kinds of realistic degradations and simulates the face misaligned situation to further alleviate the synthetic-to-real-world gap. It improves the robustness and generalization of RestoreFormer++.
    \item Extensive experiments show that RestoreFormer++ outperforms current leading competitors on both synthetic and real-world datasets. We also conduct detailed ablation studies to analyze the contribution of each component to give a better understanding of RestoreFormer++.
\end{itemize}

The \textcolor{black}{remaining} of this work is organized as follows. 
We review the most related works in Sec.~\ref{sec:related_works} and detailedly introduce the RestoreFormer++ in Sec.~\ref{sec:method}.
We then present experiments with comparison and analysis in Sec.~\ref{sec:experiments}.
Finally, conclusions are in Sec.~\ref{sec:conclusion}.

\section{Related Works}
\label{sec:related_works}

\subsection{Blind Face Restoration}

Blind face restoration aims to recover high-quality faces from face images that have undergone unknown and complex degradations. Owing to the effectiveness of Deep Neural Networks (DNN)~\cite{krizhevsky2012imagenet, chen2022knowledge, chen2021cross}, researchers~\cite{cao2017attention, huang2017wavelet, xu2017learning, shi2019face} have attempted to restore high-quality faces directly from degraded ones using DNN-based approaches. However, since the information contained in degraded faces is limited, researchers have sought assistance from additional priors, such as geometric priors~\cite{chen2021progressive, chen2018fsrnet, kim2019progressive, li2019recovering, shen2018deep, yu2018face, yu2018super, zhu2016deep, hu2021face}, reference priors~\cite{dogan2019exemplar, li2020enhanced, li2018learning, li2020blind}, and generative priors~\cite{gu2020image, menon2020pulse, wan2020bringing, wang2021towards}.
Most geometric priors are predicted from the degraded faces, and the quality of these priors is significantly constrained by the degree of degradation in the face images, which further impacts the final restoration results. Reference priors, which are high-quality faces distinct from degraded ones, alleviate the limitations of geometric priors. However, exemplars~\cite{dogan2019exemplar, li2018learning, li2020enhanced} with the same identity as the degraded face are not always available, and facial component dictionaries extracted from high-quality face images are partial and recognition-oriented, restricting the performance of reference-based methods.
Recent studies~\cite{menon2020pulse, wan2020bringing, wang2021towards, yang2021gan} have suggested that generative priors encapsulated in well-trained high-quality face generators possess considerable potential for blind face restoration, and works~\cite{gu2022vqfr, zhou2022towards, zhao2022rethinking}, published concurrently or after our conference version, propose obtaining high-quality priors from a codebook similar to our ROHQD.
\textcolor{black}{
However, most of these previous studies employ pixel-wise concatenation~\cite{dogan2019exemplar,li2020enhanced,li2018learning}, spatial feature transform (SFT)~\cite{wang2018recovering,chen2021progressive,li2020blind,wang2021towards}, or deformable operation~\cite{gu2022vqfr} to fuse the degraded feature and priors. Both SFT~\cite{wang2018recovering} and deformable networks~\cite{dai2017deformable} are implemented with convolutional layers, and their receptive fields limit the attentive areas, leading to the neglect of useful contextual information when fusing degraded information and its corresponding priors.}

In contrast, our RestoreFormer++ is a unified framework for globally modeling the contextual information in the face with fully-spatial attention while fusing the features of the degraded face and their corresponding priors matched from a reconstruction-oriented dictionary. 
Due to the rich contextual information and high-quality priors, RestoreFormer++ performs better than previous related methods in both realness and fidelity.

\subsection{Vision Transformer}
These years, transformer~\cite{vaswani2017attention} designed with attention mechanism performs pretty well on natural language processing areas~\cite{brown2020language,devlin2018bert} and researchers turn to explore the potential possibility of transformer on computer vision.
The first attempt is ViT~\cite{dosovitskiy2020image}, a pure transformer that takes sequences of image patches as input. It achieves high performance on image classification tasks.
Then more works extend the transformer to object detection~\cite{carion2020end,zhu2020deformable}, segmentation~\cite{wang2021max}, and even low-level vision\cite{chen2021pre,esser2021taming,parmar2018image,yang2020learning,zhao2021improved,zhusketch}, which may suffer from more difficulties on efficiency. 
In the low-level vision, Chen \textit{et~al.}~\cite{chen2021pre} take the advantages of transformer on a large scale pre-training to build a model that covers many image processing tasks. 
Esser \textit{et~al.}~\cite{esser2021taming} apply the transformer on codebook-indices directly to make the generation of a very high-resolution image possible. 
Zhu \textit{et~al.}~\cite{zhusketch} exploit the global structure of the face extracted by the transformer to help the synthesis of photo-sketch.
Most of these works tend to search the global information in the patches of an image with a self-attention mechanism.
To model the interplay between the degraded face and its corresponding priors cooperating with contextual information, RestoreFormer++ adopts multi-scale multi-head cross-attention mechanisms whose queries are the features of the corrupted face and key-value pairs are the priors.

\subsection{Face Degrading Model}
Since there is no real training pair in blind face restoration, most previous works synthesize the training pairs with a degrading model.
The degrading model proposed in~\cite{xu2017learning} mainly consists of blur kernels, downsampling, and Gaussian noise. In this version, Gaussian noise is added before downsampling.
Li \textit{et~al.}~\cite{li2018learning} find that adding Gaussian noise after downsampling can better simulate the long-distance image acquisition. They further upgrade the degrading model with JPEG compression.
Most of the later methods follow this degrading model for degraded face synthesis except the work proposed by Wan \textit{et~al.}~\cite{wan2020bringing} that mainly focuses on the old photos that suffer from scratch texture.
To further diminish the gap between the synthetic and real-world datasets, our EDM extends the degrading model proposed in~\cite{li2018learning} with additional commonly existing degradations: haze and uneven degradation.
It also applies a spatial shift to high-quality face images while synthesizing the degraded face to alleviate the inherent bias introduced by face alignment.

\begin{figure*}[t]
\begin{center}
\begin{tabular}{ccccc}
\includegraphics[width=0.125\linewidth]{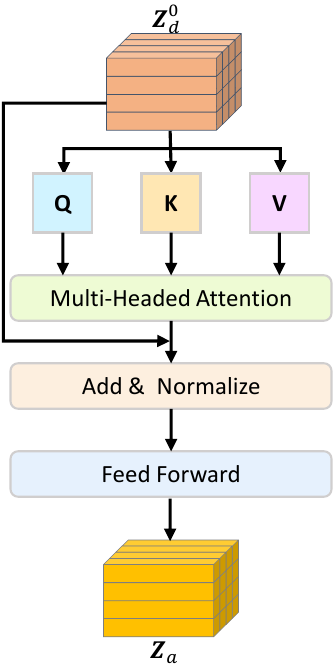} & \includegraphics[width=0.125\linewidth]{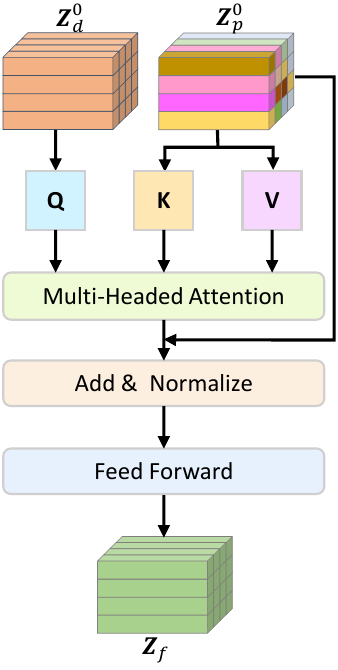} & \multicolumn{3}{c}{\includegraphics[width=0.7\linewidth]{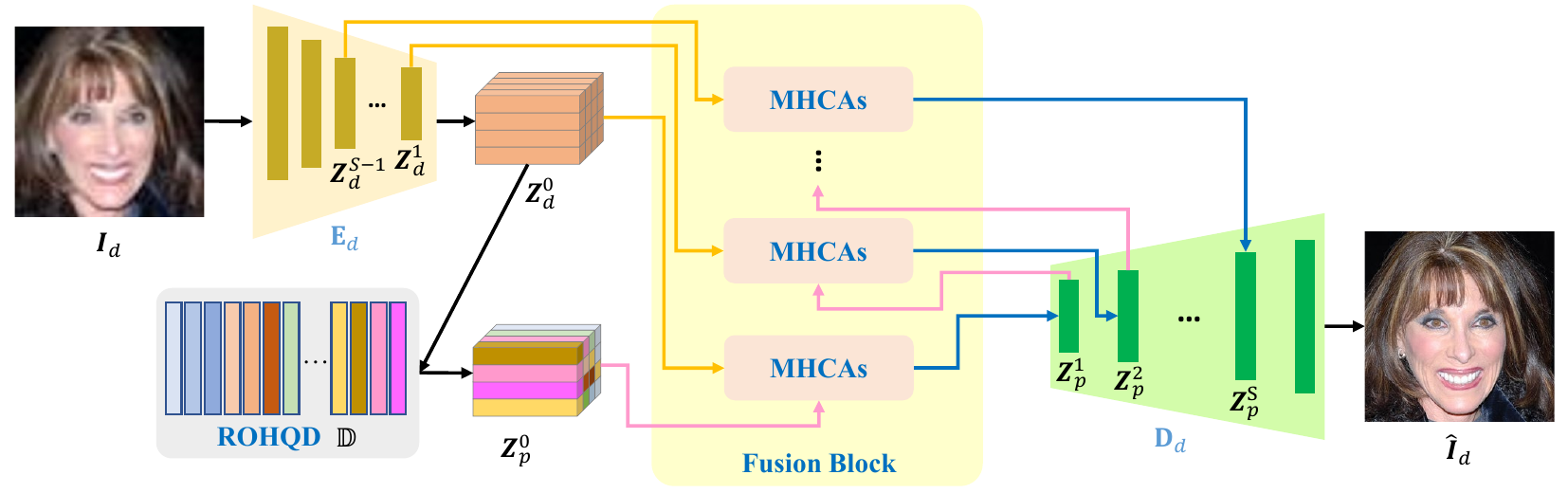}} \\
(a) \textbf{MHSA} & (b) \textbf{MHCA} & \multicolumn{3}{c}{(c) \textbf{RestoreFormer++}} \\
\end{tabular}
\end{center}
\caption{\textbf{Framework of RestoreFormer++}.
(a) MHSA is a transformer with multi-head self-attention used in most of the previous ViTs~\cite{carion2020end,chen2021pre,dosovitskiy2020image,zhu2020deformable}. Its queries, keys, and values are from the degraded information $\bm{Z}_d^0$.
(b) MHCA is a transformer with a multi-head cross-attention used in the proposed RestoreFormer++. 
It globally fuses the degraded information $\bm{Z}_d^0$ and the corresponding high-quality priors $\bm{Z}_p^0$ by taking $\bm{Z}_d^0$ as queries while $\bm{Z}_p^0$ as key-value pairs. 
(c) The whole pipeline of RestoreFormer++. 
First, a degraded face image $\bm{I}_d$ is sent to $\mathbf{E}_d$ for \textcolor{black}{multi-scale} feature extraction ($\bm{Z}_d^s, s \in \{0,1,\dots, S-1\}$, $S$ is the number of scales used for \textcolor{black}{fusion}).
Then, the degraded feature $\bm{Z}_d^s$ interacts with its corresponding priors $\bm{Z}_p^0$ matched from ROHQD $\mathbb{D}$ or previous fused output $\bm{Z}_p^s$ with MHCAs.
%
\textcolor{black}{Finally, a high-quality face $\bm{\hat{I}}_d$ is restored from the final fused result $\bm{Z}_p^S$ by the decoder $\mathbf{D}_d$.}
}
\label{fig:framework}
\end{figure*}

\section{RestoreFormer++}
\label{sec:method}
In this section, we will introduce the proposed RestoreFormer++ with the whole restored pipeline shown in Fig.~\ref{fig:framework}~(c).
The pipeline consists of four components: an encoder $\mathbf{E}_d$, a reconstruction-oriented high-quality dictionary $\mathbb{D}$ (ROHQD), a fusion block consisting of several Multi-Head Cross-Attention blocks (MHCAs), and a decoder $\mathbf{D}_d$.
First, a degraded face image $\bm{I}_d$ is sent to $\mathbf{E}_d$ for feature extraction ($\bm{Z}_d^s, s \in \{0,1,\dots, S-1\}$, $S$ is the number of scales used for fusing).
Then, the degraded feature $\bm{Z}_d^s$ fuses with its corresponding priors $\bm{Z}_p^0$ matched from ROHQD $\mathbb{D}$ or previous fused output $\bm{Z}_p^s$ with MHCAs.
%
\textcolor{black}{Finally, a high-quality face $\bm{\hat{I}}_d$ is restored from the final fused result $\bm{Z}_p^S$ by the decoder $\mathbf{D}_d$.}

We will introduce the details of the \textcolor{black}{restoration} process in Sec.~\ref{subsec: restoration} and describe the learning of the reconstruction-oriented high-quality dictionary (ROHQD) in Sec.~\ref{subsec: ROHQD}.
Besides, we will explain our extending degraded model (EDM) used for synthesizing degraded face images in Sec.~\ref{subsec: EDM}.

\subsection{Restoration}
\label{subsec: restoration}
RestoreFormer++ aims at globally modeling the contextual information in a face and the interplay with priors for restoring a high-quality face image with both realness and fidelity.
ViT (Vision Transformer)~\cite{vaswani2017attention} is such an effective method for modeling contextual information in computer vision.
However, most of the previous ViT-based methods~\cite{carion2020end,chen2021pre,dosovitskiy2020image,zhu2020deformable} model the contextual information with multi-head self-attention (MHSA) whose queries, keys and values are from different patches in the same image.
In this work, we propose to simultaneously model the contextual information and the interplay between the degraded face and its corresponding priors.
Therefore, our RestoreFormer++ adopts multi-head cross-attention (MHCA) mechanisms whose queries are from the features of degraded faces, while key-value pairs are from the corresponding priors.
To clarify the delicate design of our MHCA for blind face restoration, we will first describe MHCA by comparing it with MHSA before going deep into the restoration process.

\noindent\textbf{MHSA.} 
As Fig.~\ref{fig:framework}~(a) shown, MHSA aims at searching the contextual information in one source (for convenience, we set it as our degraded feature $\bm{Z}_d^0\in\mathbb{R}^{H'\times W'\times C}$, where $H', W'$ and $C$ are the height, width and the number of channels of the feature map, respectively). Its  queries $\bm{Q}$, keys $\bm{K}$, and values $\bm{V}$ can be formulated as:
\begin{equation}
\small{\bm{Q}=\bm{Z}_d^0\bm{W}_{q}+\bm{b}_{q}\ ,\  
\bm{K}=\bm{Z}_d^0\bm{W}_{k}+\bm{b}_{k}\ ,\   
\bm{V}=\bm{Z}_d^0\bm{W}_{v}+\bm{b}_{v},}
\end{equation}
where $\bm{W}_{q/k/v}\in\mathbb{R}^{C \times C}$ and  $\bm{b}_{q/k/v}\in\mathbb{R}^{C}$ are learnable weights and bias.

Multi-head attention is a mechanism for attaining powerful representations. It is implemented by separating the $\bm{Q}$, $\bm{K}$, and $\bm{V}$ into $N_h$ blocks along the channel dimension and gets $\{\bm{Q}_1, \bm{Q}_2, \dots, \bm{Q}_{N_h}\}$, $\{\bm{K}_1, \bm{K}_2, \dots, \bm{K}_{N_h}\}$, and $\{\bm{V}_1, \bm{V}_2, \dots, \bm{V}_{N_h}\}$, where $\bm{Q}_i/\bm{K}_i/\bm{V}_i\in\mathbb{R}^{H'\times W'\times C_h }$, $C_h=\frac{C}{N_h}$, and $i\in[0,N_h-1]$.
Then the attention map is represented as:
\begin{equation}
    \bm{Z}_{i} = \operatorname{softmax}(\frac{\bm{Q}_i\bm{K}_i^\intercal}{\sqrt{C_h}})\bm{V}_i, i=0,1,\dots,N_h-1.
    \label{eq:softmax}
\end{equation}
By concatenating all $\bm{Z}_{i}$, we get the final output of multi-head attention:
\begin{equation}
   \bm{Z}_{mh} = \operatornamewithlimits{concat}_{i=0,...,N_h-1}\bm{Z}_{i}.
    \label{eq:concat}
\end{equation}
In the conventional transformer, the attention output is added back to the input before \textcolor{black}{sequentially processed} by a normalization layer and a feed-forward network\textcolor{black}{, which can be formulated as:}
%
\begin{equation}
    \bm{Z}_{a} = \operatorname{FFN}(\operatorname{LN}(\bm{Z}_{mh} + \bm{Z}_d^0)),
    \label{eq:shortcut_degraded}
\end{equation}
where $\operatorname{LN}$ is a layer normalization, $\operatorname{FFN}$ is a feed-forward network implemented with two convolution layers, and $\bm{Z}_{a}$ is the final output of MHSA.

\noindent\textbf{MHCA.} 
\textcolor{black}{
As shown in Fig.~\ref{fig:framework} (b), since the MHCA adopted in our Restoreformer++ aims at modeling the contextual information in the face images and simultaneously attaining identity information in the degraded face and high-quality facial details in the priors, it takes both the degraded feature $\bm{Z}_d^0$ and the corresponding priors $\bm{Z}_p^0$ as inputs.
}
In MHCA, the queries $\bm{Q}$ are from the degraded feature $\bm{Z}_d^0$ while the keys $\bm{K}$ and values $\bm{V}$ are from the priors $\bm{Z}_p^0$:
\begin{equation}
\small{
\bm{Q}=\bm{Z}_d^0\bm{W}_{q}+\bm{b}_{q}\ ,\ 
\bm{K}=\bm{Z}_p^0\bm{W}_{k}+\bm{b}_{k}\ ,\ 
\bm{V}=\bm{Z}_p^0\bm{W}_{v}+\bm{b}_{v},}
\end{equation}
Its following operations for attaining the multi-head attention output $\bm{Z}_{mh}$ are the same as Eq.~\ref{eq:softmax} and Eq.~\ref{eq:concat}.
Since high-quality priors play a more important role in blind face restoration, $\bm{Z}_{mh}$ is added with $\bm{Z}_{p}^0$ instead of $\bm{Z}_{d}^0$ in RestoreFormer++. 
The rest operations are:
\begin{equation}
    \bm{Z}_{f} = \operatorname{MHCA}(\bm{Z}_{d}^0, \bm{Z}_p^0)=\operatorname{FFN}(\operatorname{LN}(\bm{Z}_{mh} + \bm{Z}_p^0)).
    \label{eq:shortcut_prior}
\end{equation}

\noindent\textbf{Restoration.}
As described before, the restored pipeline consists of four components.
The first component $\mathbf{E}_d$ is used for extracting \textcolor{black}{multi-scale} features $\bm{Z}_d^s$ ($s=\{0, 1, \dots, S-1\}$, $S$ means the number of scales) from the degraded face image $\bm{I}_d$.
Then, we \textcolor{black}{ can get} the priors $\bm{Z}_p^0$ of $\bm{Z}_d^0$ from ROHQD $\mathbb{D}=\{\bm{d}_m\}_{m=0}^{M-1}$ ($\bm{d}_m\in\mathbb{R}^{C}$, $M$ is the number of elements in $\mathbb{D}$) with minimum Euclidean distance:
\begin{equation}
\bm{Z}_p^0(i,j) = \mathop{\arg\min}_{\bm{d}_m \in \mathbb{D}} \|\bm{Z}_d^0(i,j)-\bm{d}_m\|_2^2,
\end{equation}
where $(i,j)$ is the spatial position of map $\bm{Z}_p^0$ and $\bm{Z}_d^0$ and $||\cdot||_2$ means the L2-norm. 
After attaining the degraded features $\bm{Z}_d^s$ ($s=\{0, 1, \dots, S-1\}$) and $\bm{Z}_p^0$, these two kinds of information are fused in the Fusion Block.
In this block, for each scale, the degraded features and priors or previous fused results (for convenience, we denote the fused results of each scale as $\bm{Z}_p^s$ ($s=\{1, \dots, S\}$)) are fused with MHCAs, which consists of $K$ MHCA.
We formula this procedure as follows:
\begin{equation}
\begin{aligned}
    \bm{Z}_p^{s+1} &= \operatorname{MHCAs}(\bm{Z}_d^{s},\bm{Z}_p^{s}) \\
                   &= \operatorname{MHCA}(\bm{Z}_d^{s}, \dots, \operatorname{MHCA}(\bm{Z}_d^s, \operatorname{MHCA}(\bm{Z}_d^s, \bm{Z}_p^s))), \\
                   & s=\{0,1,\dots,S-1\}.
\end{aligned}
\end{equation}
Finally, $\bm{Z}_p^{S}$ is fed into the rest layers of the decoder  $\mathbf{D}_d$ for recovering the high-quality face image $\bm{\hat{I}}_d$.

\noindent\textbf{Learning.}
For attaining high-quality faces with both realness and fidelity, we design the objective functions from three aspects: content learning, realness learning, and identity learning.

\noindent\textbf{Content learning.}
We adopt $L1$ loss and perceptual loss~\cite{johnson2016perceptual,ledig2017photo} for ensuring the content consistence between the restored face image $\bm{\hat{I}}_d$ and its ground truth $\bm{I}_h$:
\begin{equation}
   \mathcal{L}_{l1} = |\bm{I}_h - \bm{\hat{I}}_d|_1\ ;\ \mathcal{L}_{per} = \|\phi(\bm{I}_h) - \phi(\bm{\hat{I}}_d)\|_2^2,
\label{eq: rf_l1_per}
\end{equation}
where $\bm{I}_h$ is the ground truth high-quality image,
$\phi$ is the pretrained VGG-19~\cite{simonyan2014very}, and the feature maps are extracted from $\lbrace conv1, \dots, conv5 \rbrace$.
Besides, for improving the accuracy of the matched priors, we tend to guide the extracted features $\bm{Z}_d^0$ to approach their selected priors $\bm{Z}_p^0$ and the corresponding objective function is:
\begin{equation}
   \mathcal{L}_{p} = \|\bm{Z}_p^0 - \bm{Z}_d^0\|_2^2.
\end{equation}

\noindent\textbf{Realness learning.}
We adopt adversarial losses for the learning of realness in this work.
Since some crucial facial components, such as the eyes and mouth, play an important role in face presenting~\cite{wang2021towards}, our adversarial losses are not only applied to the whole face image but also applied to these key facial components independently.
These losses are expressed as:
\begin{equation}
\begin{split}
\mathcal{L}_{adv} &= [\log D(\bm{I}_h) + \log (1-D(\bm{\hat{I}}_d))], \\
\mathcal{L}_{comp} &= \sum_{r}[\log D_r(R_{r}(\bm{I}_h)) + \log (1 - D_r(R_{r}(\bm{\hat{I}}_d)))],
\end{split}
\label{eq: rf_adv_comp}
\end{equation}
where $D$ and $D_r$ are the discriminators for the whole face image and a certain region $r$ ($r\in$\{left eye, right eye, mouth\}), respectively.
The region $r$ is attained with $R_{r}$ implemented with ROI align~\cite{he2017mask}.

\noindent\textbf{Identity learning.}
In this work, except extracting the identity information from the degraded face by fusing it with the high-quality priors, we also adopt an identity loss~\cite{wang2021towards} to attain the identity supervision from the ground truth:
\begin{equation}
\mathcal{L}_{id} = \|\eta(\bm{I}_h) - \eta(\bm{\hat{I}}_d)\|_2^2,
\label{eq: identity}
\end{equation}
where $\eta$ denotes the identity feature extracted from ArcFace~\cite{deng2019arcface} which is a well-trained face recognition model.

Therefore, the overall objective function is:
\begin{equation}
\begin{aligned}
\mathcal{L}_{RF} &= \mathcal{L}_{l1} + \lambda_{per} \mathcal{L}_{per} + \lambda_{p} \mathcal{L}_{p} + \lambda_{adv} \mathcal{L}_{adv} \\
&+ \lambda_{comp} \mathcal{L}_{comp} + \lambda_{id} \mathcal{L}_{id},\\
\end{aligned}
\end{equation}
where $\lambda_{\dots}$ are the weighting factors for different losses.

\begin{figure}
\begin{center}
\begin{tabular}{cc}
\includegraphics[width=0.4\linewidth]{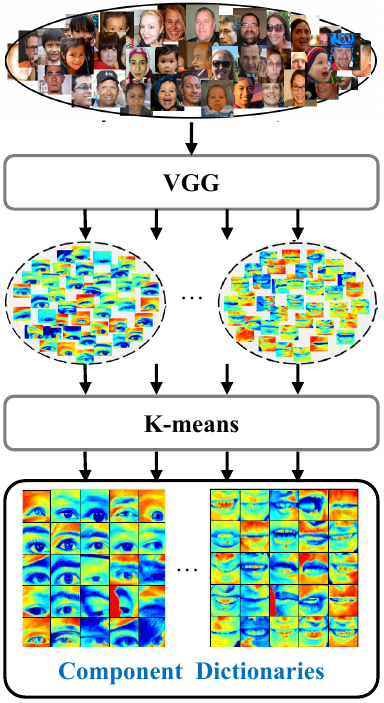} & \includegraphics[width=0.4\linewidth]{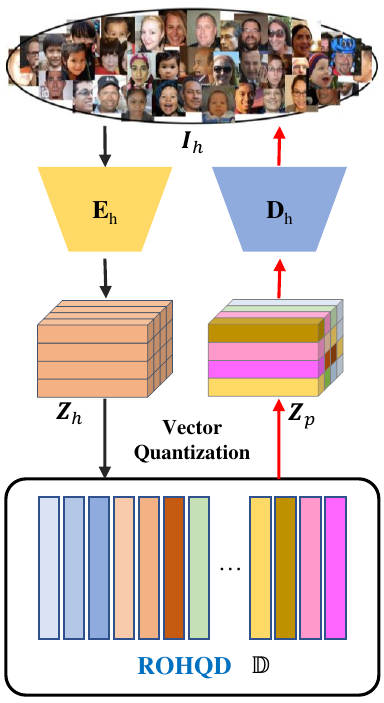} \\
(a) Component Dictionaries & (b) ROHQD
\end{tabular}
\end{center}
\caption{\textbf{Recognition-Oriented Dictionary v.s. Reconstruction-Oriented Dictionary}.
(a) Component Dictionaries, proposed in DFDNet~\cite{li2020blind}, are recognition-oriented dictionaries since they are extracted with an off-line image recognition model (VGG~\cite{simonyan2014very}). 
(b) ROHQD, proposed in this paper, is a reconstruction-oriented dictionary since it is learned with a high-quality face generation network incorporating the idea of vector quantization~\cite{oord2017neural}. 
Priors from ROHQD contain more facial details specifically aimed at face restoration.
}
\label{fig:dictionary}
\end{figure}

\begin{figure*}
\begin{center}
\includegraphics[width=0.9\linewidth]{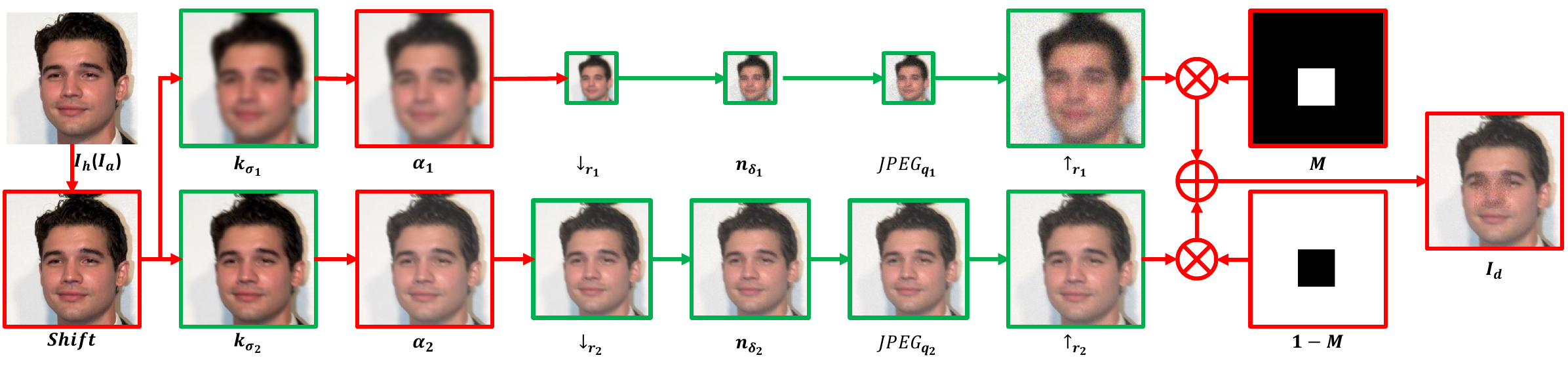}
\end{center}
\caption{\textbf{The whole pipeline of the extending degrading model (EDM)}.
The degradations represented in \textcolor{green}{GREEN} are the operations contained in the traditional degrading model (Eq.~\ref{eq:degrading_model}) while the degradations in \textcolor{red}{RED} are the additional operations extended by EDM (Eq.~\ref{eq:shift} to \ref{eq:uneven}).
\textcolor{black}{
Specifically, a high-quality face image $\bm{I}_a$ is first shifted with an operator $\operatorname{Shift}$.
Then, it is sequentially degraded with blur, haze, downsampling, noise, and JPEG compression.
The degraded face images will be upsampled back to the size of the original image.
The degraded faces attained after ${\uparrow_r}_1$ and ${\uparrow_r}_2$ are degraded from the same high-quality face image but with two different degraded parameters: $\alpha_1$ and $\alpha_2$, $\sigma_1$ and $\sigma_2$, $r_1$ and $r_2$, $\delta_1$ and $\delta_2$, and $q_1$ and $q_2$. They are \textbf{independently} and \textbf{randomly} sampled from their own uniform distributions.
Combining these two degraded faces with a mask $\bm{M}$, the final unevenly degraded face image $\bm{I}_d$ is attained.
}
}
\label{fig:degrading_process}
\end{figure*}

\subsection{Reconstruction-Oriented High-Quality Dictionary}
\label{subsec: ROHQD}
In this subsection, we introduce the learning of the Reconstruction-Oriented High-Quality Dictionary (ROHQD) $\mathbb{D}=\{\bm{d}_m\}_{m=0}^{M-1} ( d_m\in\mathbb{R}^{C}$, $M$ is the number of elements) used in RestoreFormer++.

Different from the facial component dictionaries~\cite{li2020blind} (Fig.~\ref{fig:dictionary}~(a)) whose elements are high-quality facial details of specific facial components extracted with an off-line recognition-oriented model (VGG~\cite{simonyan2014very}), our ROHQD provides richer high-quality facial details specifically aimed at face reconstruction.
We achieve this goal by deploying a high-quality face encoder-decoder network with the idea of vector quantization~\cite{oord2017neural}.
As shown in Fig.~\ref{fig:dictionary}~(b), this encoder-decoder network takes a high-quality face image $\bm{I}_h\in \mathbb{R}^{H \times W \times 3}$ as input and encodes it to feature $\bm{Z}_h\in \mathbb{R}^{H' \times W' \times C}$ with encoder $\mathbf{E}_h$.
Then, instead of decoding $\bm{Z}_h$ directly back to the high-quality face with decoder $\mathbf{D}_h$, it quantizes feature vectors in $\bm{Z}_h$ with the index of the nearest vectors in $\mathbb{D}$ and attains $\bm{Z}_{p}\in \mathbb{R}^{H' \times W' \times C}$:
\begin{equation}
\bm{Z}_p(i,j) = \mathop{\arg\min}_{\bm{d}_m \in \mathbb{D}} \|\bm{Z}_h(i,j)-\bm{d}_m\|_2^2,
\label{eq:quantization}
\end{equation}
where $(i,j)$ is the spatial position of map $\bm{Z}_p$ and $\bm{Z}_h$.
%
\textcolor{black}{Finally, a high-quality face image $\bm{\hat{I}}_h$ is restored from $\bm{Z}_p$ by the decoder $\mathbf{D}_h$.}

\noindent\textbf{Learning.}
The whole pipeline shown in Fig.~\ref{fig:dictionary} (b) is essentially a high-quality face generation network.
Therefore, we apply an $L1$ loss, a perceptual loss, and an adversarial loss to the final result $\bm{\hat{I}}_h$ with the supervision from its high-quality input $\bm{I}_h$:
\begin{equation}
\begin{aligned}
\mathcal{L'}_{l1} &= \|\bm{I}_h - \bm{\hat{I}}_h\|_1,\\
\mathcal{L'}_{per} &= \|\phi(\bm{I}_h ) - \phi(\bm{\hat{I}}_h)\|_2^2, \\
\mathcal{L'}_{adv} &= [\log D(\bm{I}_h) + \log (1-D(\bm{\hat{I}}_h))].
\end{aligned}
\end{equation}
The definitions of $\phi$ and $D$ are same as Eq.~\ref{eq: rf_l1_per} and Eq.~\ref{eq: rf_adv_comp}.
It is worth noting that since Eq~\ref{eq:quantization} is non-differentiable, the gradients back-propagated from $\bm{\hat{I}_h}$ reach $\bm{Z}_h$ by copying the gradients of $\bm{Z}_p$ to $\bm{Z}_h$ directly~\cite{oord2017neural}.

%
The ultimate goal of ROHQD in this work is to optimize $\mathbb{D}$ to attain high-quality facial details used for face restoration.
Therefore, we update the elements $\bm{d}_m$ constructed $\bm{Z}_p$ (Eq.~\ref{eq:quantization}) by forcing them to be close to their corresponding high-quality features $\bm{Z}_h$ with $L2$ loss:
\begin{equation}
   \mathcal{L'}_{d} = \|\operatorname{sg}[\bm{Z}_h] - \bm{Z}_p\|_2^2,
\end{equation}
where $\operatorname{sg}[\cdot]$ denotes the stop-gradient operation.
Besides, as described in~\cite{oord2017neural}, to avoid collapse, a commitment loss is needed to adjust the learning pace of the encoder $\mathbf{E}_h$ and dictionary $\mathbb{D}$. The commitment loss is represented as:
\begin{equation}
   \mathcal{L'}_{c} = \|\bm{Z}_h - \operatorname{sg}[\bm{Z}_p]\|_2^2.
\end{equation}

Finally, the objective function for learning ROHQD is:
\begin{equation}
\mathcal{L}_{ROHQD} = \mathcal{L'}_{l1} + \lambda_{per}\mathcal{L'}_{per} + \lambda_{adv}\mathcal{L'}_{adv} + \lambda_d\mathcal{L'}_{d} + \lambda_c \mathcal{L'}_{c},
\end{equation}
where $\lambda_{\dots}$ are the weighting factors.

\subsection{Extending Degrading Model}
\label{subsec: EDM}
To diminish the distance between the synthetic training data and the real-world data and further improve the robustness and generalization of RestoreFormer++, EDM extends the degrading model~\cite{li2020enhanced,li2018learning,wang2021towards} whose original expression is:
\begin{equation}
\bm{I}_d = \{[(\bm{I}_h \otimes \bm{k}_\sigma) \downarrow_r + \bm{n}_\delta]_{{JPEG}_q}\}\uparrow_r,
\label{eq:degrading_model}
\end{equation}
where $\bm{I}_h$ is a high-quality face image and $\bm{I}_d$ is the final synthetic degraded face image.
$\bm{I}_h$ is first blurred by a Gaussian blur kernel $\bm{k}_\sigma$ with sigma $\sigma$.
Then, it is downsampled by $r$ with bilinear interpolation and added with a white Gaussian noise $\bm{n}_\delta$ whose sigma is $\delta$.
Next, the intermediate degraded result is further compressed with JPEG compression, whose quality is $q$.
After that, it is upsampled back to the size of $\bm{I}_h$ with scale $r$.
Then we get the final synthetic degraded face image $\bm{I}_d$.
These operations are sequentially described in Fig.~\ref{fig:degrading_process} with \textcolor{green}{GREEN} color.

Excepting the common degradations described in Eq.~\ref{eq:degrading_model}, EDM adds haze and uneven degradation with a certain probability since they also obviously exist in the real-world degraded faces (examples are in Fig.~\ref{fig:EDM}).
In addition, EDM also attempts to ease the error introduced by face alignment in real-world data (the third sample in Fig.~\ref{fig:real} and the second sample in Fig.~\ref{fig:EDM}) by disturbing the perfect alignment in the synthetic training set with a spatial shift operation.
The EDM is expressed as:
\begin{equation}
\bm{I}_h = \mathop{Shift}(\bm{I}_{a}, s_h, s_w),
\label{eq:shift}
\end{equation}
\begin{equation}
\bm{I}_d^e = \{[(\alpha(\bm{I}_h \otimes \bm{k}_\sigma)+(1-\alpha)\bm{I}_{haze}) \downarrow_r + \bm{n}_\delta]_{{JPEG}_q}\}\uparrow_r,
\label{eq:haze}
\end{equation}
\begin{equation}
\begin{split}
\bm{I}_d=&\bm{M}\odot\bm{I}_d^e(\alpha_1, \sigma_1, r_1, \delta_1, q_1) + \\
&(1-\bm{M})\odot\bm{I}_d^e(\alpha_2, \sigma_2, r_2, \delta_2, q_2).
\label{eq:uneven}
\end{split}
\end{equation}
$\bm{I}_a$ is the well aligned high-quality face image \textcolor{black}{($\bm{I}_a$ is equal to $\bm{I}_h$ in Eq.~\ref{eq:degrading_model})} and $\mathop{Shift}(\cdot)$ means spatially shifting $\bm{I}_a$ with $s_h$ and $s_w$ pixels in height and width dimensions, respectively.
%
\textcolor{black}{Then the degraded face image is synthesized from the shifted high-quality face image $\bm{I}_h$.}
\textcolor{black}{
We synthesize haze in Eq.~\ref{eq:haze}. 
Before downsampled, the blurry face image will be combined with $\bm{I}_{haze}$ with ratio $\alpha : (1-\alpha), \alpha\in[0, 1]$. $\bm{I}_{haze}$ is a globally white image.
In Eq.~\ref{eq:haze}, the degraded result $\bm{I}_d^e$ is a globally evenly degraded face image. 
To attain an unevenly degraded face image $\bm{I}_d$, we first synthesize two evenly degraded faces, $\bm{I}_d^e(\alpha_1, \sigma_1, r_1, \delta_1, q_1)$ and $\bm{I}_d^e(\alpha_2, \sigma_2, r_2, \delta_2, q_2)$, whose parameters: $\alpha_1$ and $\alpha_2$, $\sigma_1$ and $\sigma_2$, $r_1$ and $r_2$, $\delta_1$ and $\delta_2$, and $q_1$ and $q_2$, are independently and randomly sampled from uniform distributions (the experimental setting of the uniform distribution of each parameter in this paper is described in Subsec.~\ref{subsec:settings}).
%
Then we combine these two unevenly degraded face images with a mask map $\bm{M}$ whose size is the same as $\bm{I}_d^e$. 
The whole map of $\bm{M}$ is set to $0$ except that a random $L\times L$ patch of it is set to $1$ ($L$ is smaller than both the height and width of $\bm{I}_d^e$).
$\odot$ is an element-wise multiplication operation.
%
}
The whole pipeline of EDM is described in Fig.~\ref{fig:degrading_process}, and the operations in \textcolor{red}{RED} are the additional degradations extended by EDM.

\section{Experiments and Analysis}
\label{sec:experiments}

\subsection{Datasets}
\noindent\textbf{Training Datasets.}
ROHQD is \textcolor{black}{trained on} FFHQ~\cite{karras2019style}, which contains 70000 high-quality face images resized to $512 \times 512$.
RestoreFormer++ is also trained on synthesized data attained by applying EDM to the high-quality face images in FFHQ.

\noindent\textbf{Testing Datasets.}
We evaluate RestoreFormer++ on one synthetic dataset and three real-world datasets.
The synthetic dataset, CelebA-Test~\cite{liu2015deep}, contains 3000 samples and is attained by applying EDM on the testing set of CelebA-HQ~\cite{liu2015deep}.
The three real-world datasets include LFW-Test~\cite{huang2008labeled}, CelebChild-Test~\cite{wang2021towards}, and WebPhoto-Test~\cite{wang2021towards}.
Specifically, LFW-Test contains 1711 images and is built with the first image of each identity in the validation set of LFW~\cite{huang2008labeled}.
Both CelebChild-Test and WebPhoto-Test are collected from the Internet by Wang \textit{et~al.}~\cite{wang2021towards}. They respectively own 180 and 407 degraded face images.

\subsection{Experimental Settings and Metrics}
\label{subsec:settings}
\noindent\textbf{Settings.}
The encoder and decoder in the RestoreFormer++ and ROHQD are constructed with 12 residual blocks and 5 nearest downsampling/upsampling operations.
Each MHCAs contains $K=3$ MHCA. 
%
The input size of the model is $512 \times 512 \times 3 $. After encoding, the size of $\bm{Z}_d$ is $16\times16\times256$.
ROHQD contains $M=1024$ elements whose length is $256$.
As for EDM, $s_h$, $s_w$, $\alpha$, $\sigma$, $r$, $\delta$, $q$, and $L$ are randomly sampled from$\lbrace0:32\rbrace$, $\lbrace0:32\rbrace$, $\lbrace0.7:1.0\rbrace$, $\lbrace0.2:10\rbrace$, $\lbrace1:8\rbrace$, $\lbrace0:20\rbrace$, $\lbrace60:100\rbrace$, and $\lbrace128:256\rbrace$, respectively.
While training, the batch size is set to $16$ and the weighting factors of the loss function are $\lambda_{per}=1.0$, $\lambda_{p}=0.25$, $\lambda_{adv}=0.8$, $\lambda_{comp}=1.0$, $\lambda_{id}=1.0$, $\lambda_{d}=1.0,$ and $\lambda_{c}=0.25$.
Both RestoreFormer++ and ROHQD are optimized by Adam~\cite{kingma2014adam} with learning rate $0.0001$.
Noted that we do not update the elements of the ROHQD while training RestoreFormer++.

\noindent\textbf{Metrics.}
In this paper, we evaluate the state-of-the-art methods and our RestoreFormer++ objectively and subjectively.
From the objective aspect, we adopt the widely-used non-reference metric FID~\cite{heusel2017gans} to evaluate the realness of the restored face images and introduce an identity distance (denoted as IDD) to judge the fidelity of the restored face images.
IDD is the angular distance between the features of the restored face image and its corresponding ground truth. Features are extracted with a well-trained face recognition model ArcFace~\cite{deng2019arcface}.
Besides, we adopt PSNR, SSIM, and LPIPS~\cite{zhang2018unreasonable} to build a more comprehensive comparison.
From the subjective aspect, we deploy a user study to evaluate the quality of the restored results from the \textcolor{black}{perspective} of humans.

\renewcommand{\tabcolsep}{.5pt}
\begin{figure*}[t]
\begin{center}
\begin{tabular}{cccccccc}
    \includegraphics[width=\swceleba]{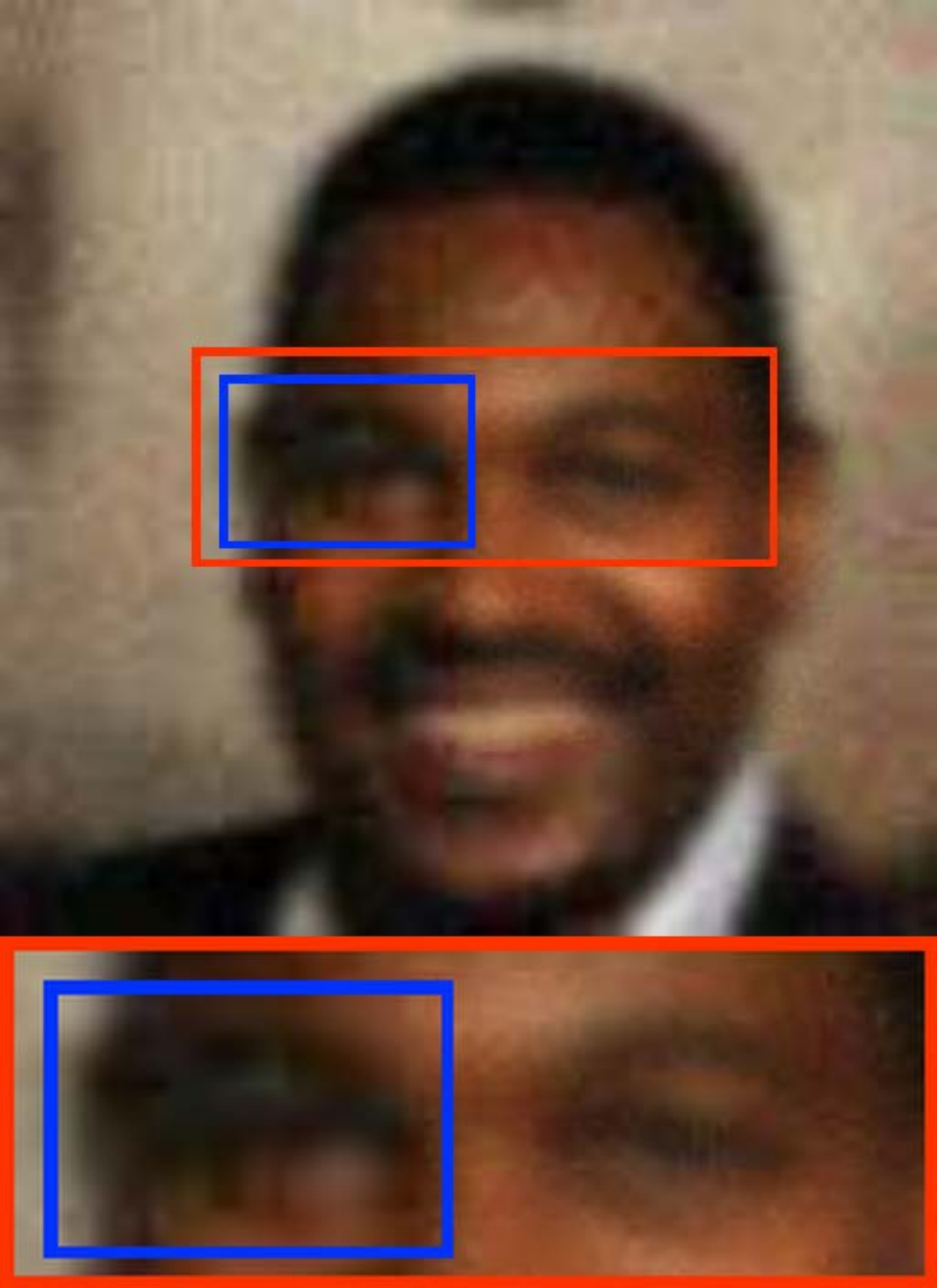}&
   \includegraphics[width=\swceleba]{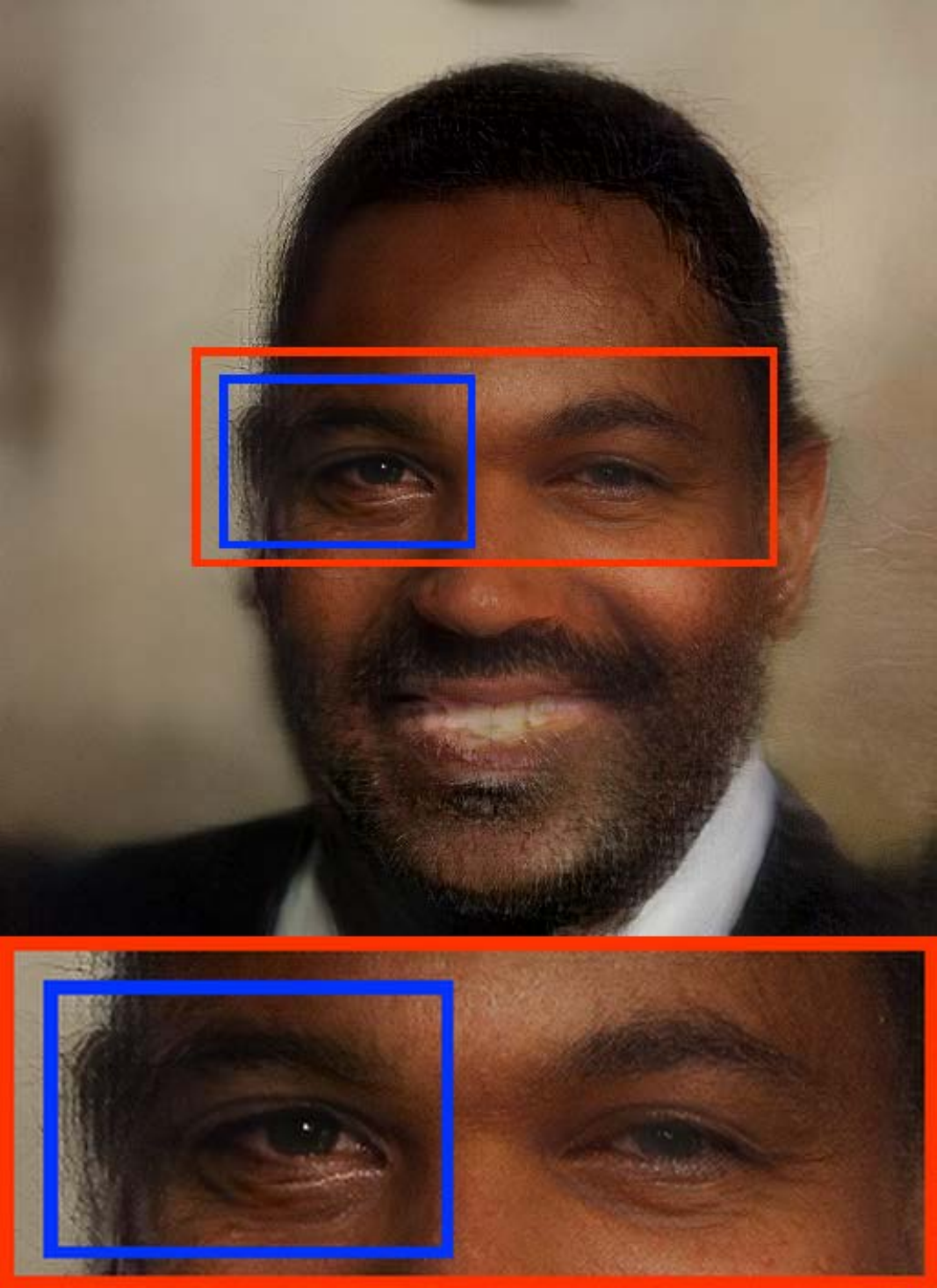}&
   \includegraphics[width=\swceleba]{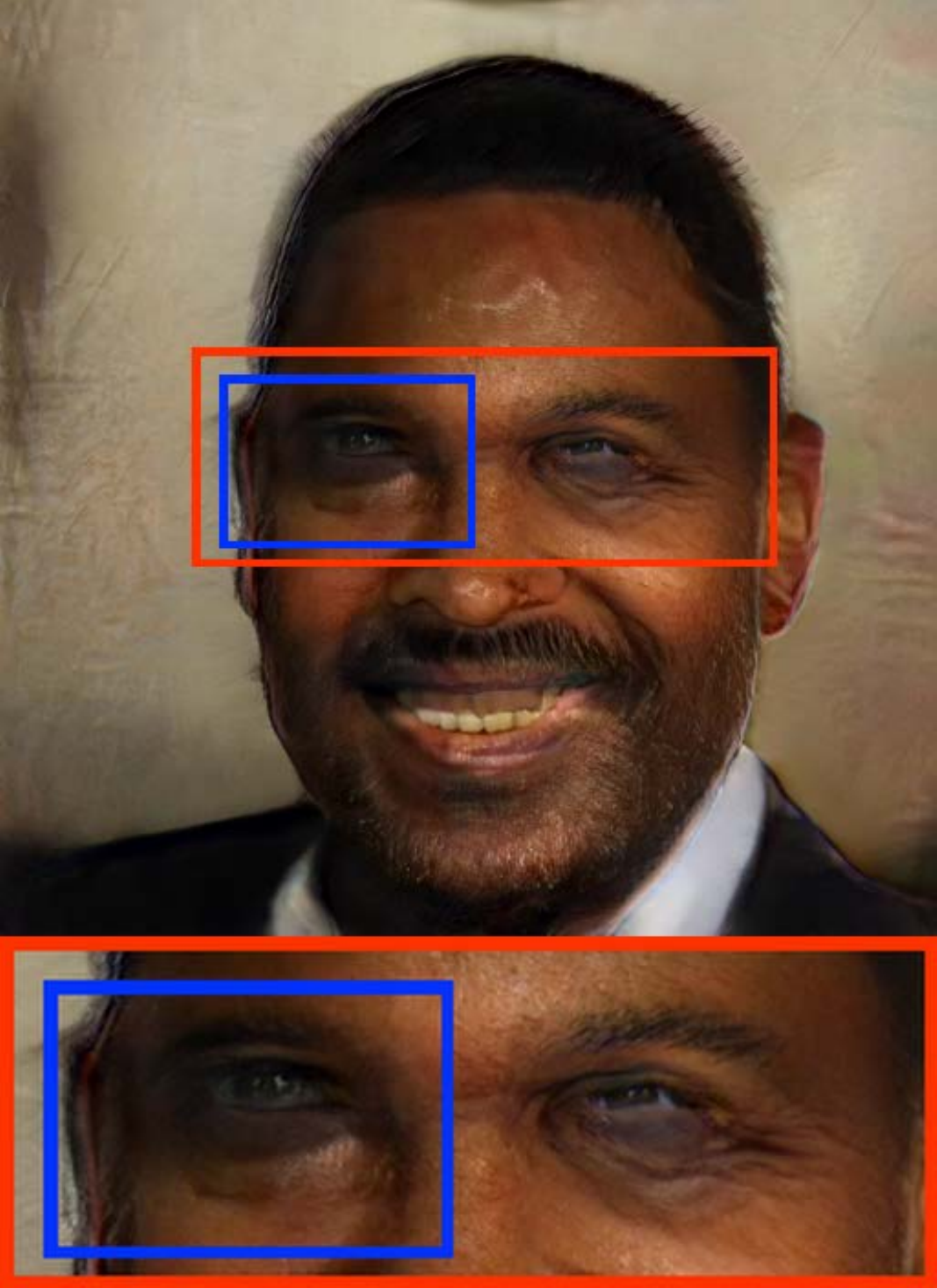}&
   \includegraphics[width=\swceleba]{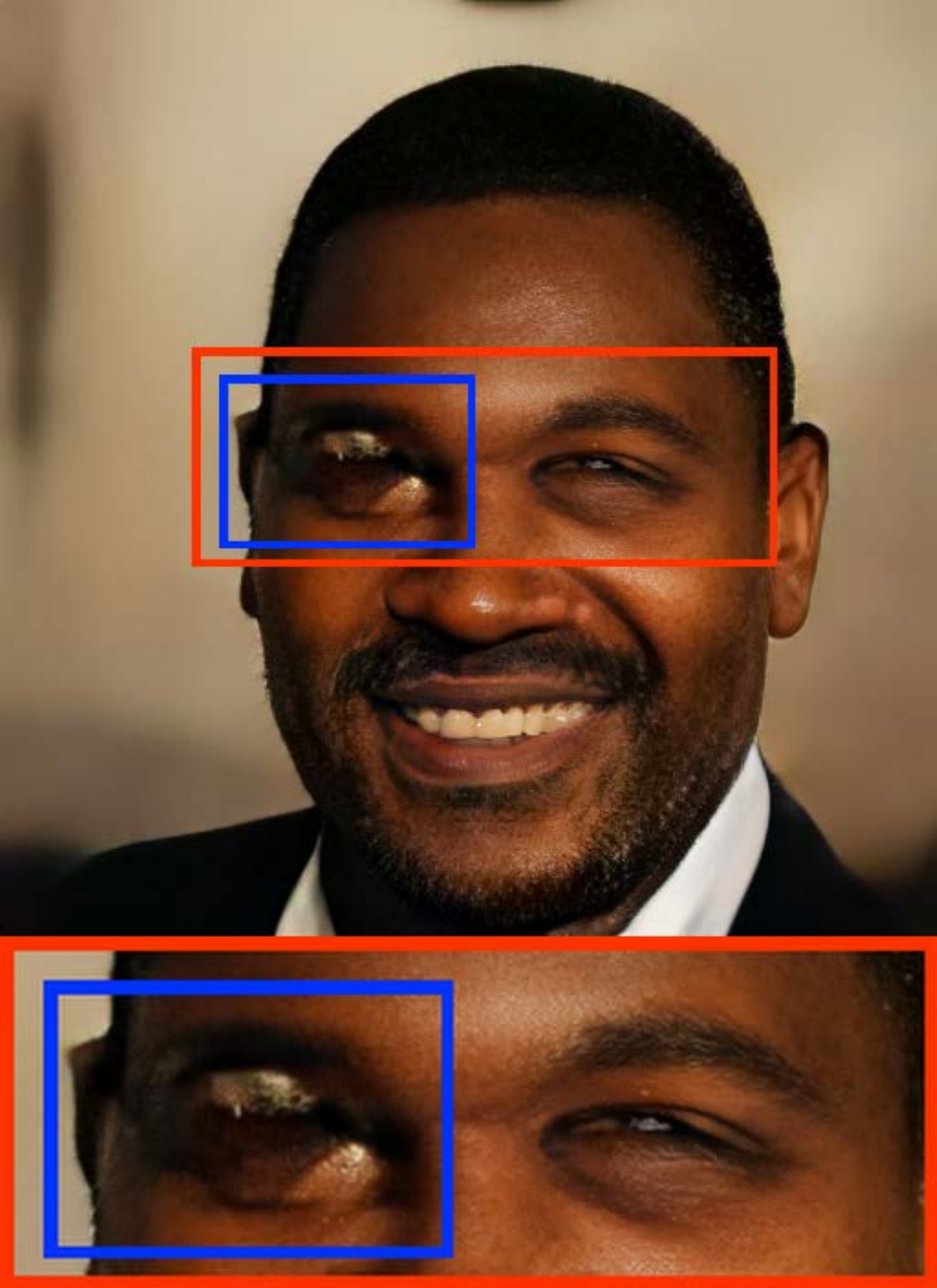}&
   \includegraphics[width=\swceleba]{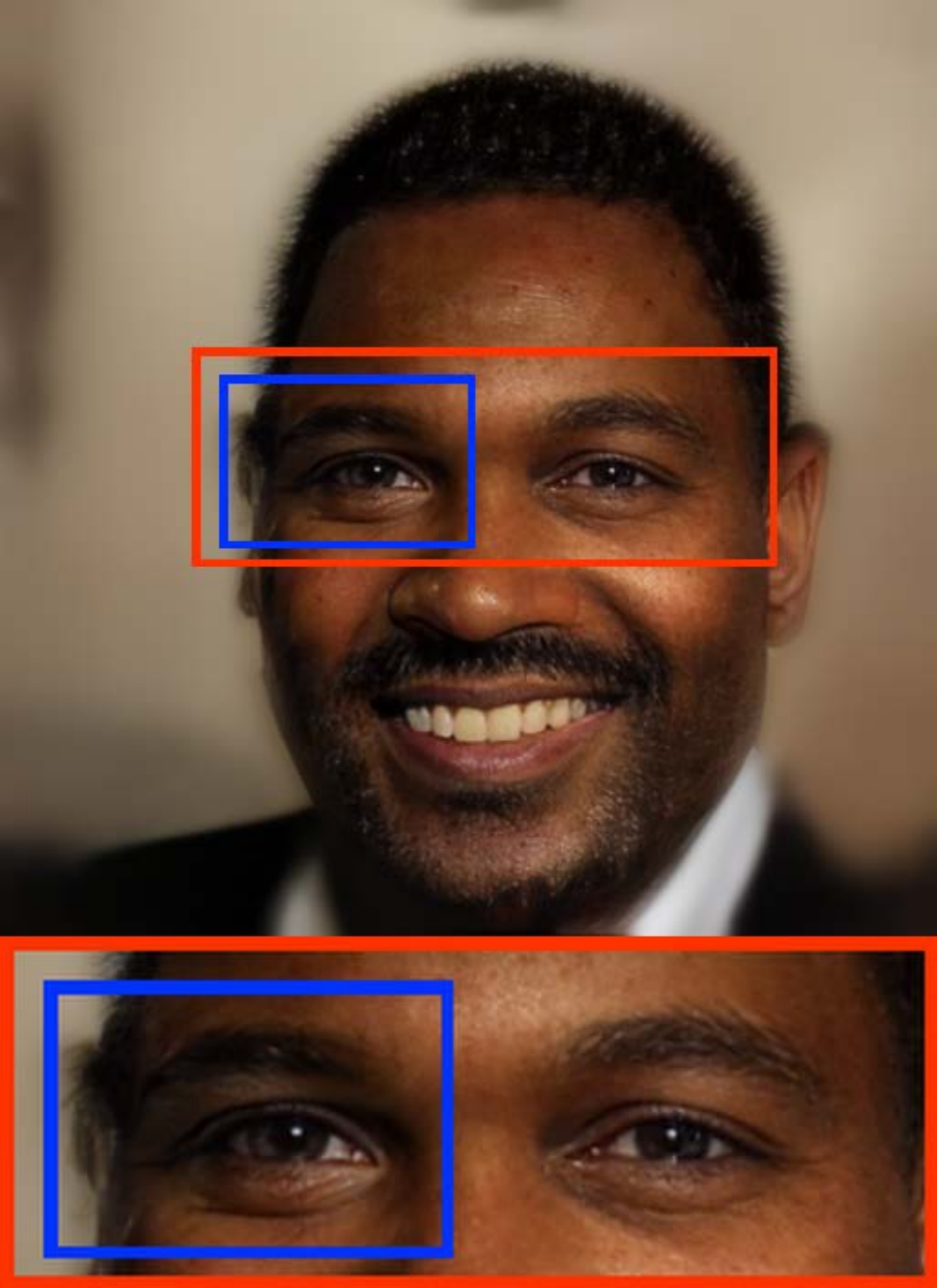}&
   \includegraphics[width=\swceleba]{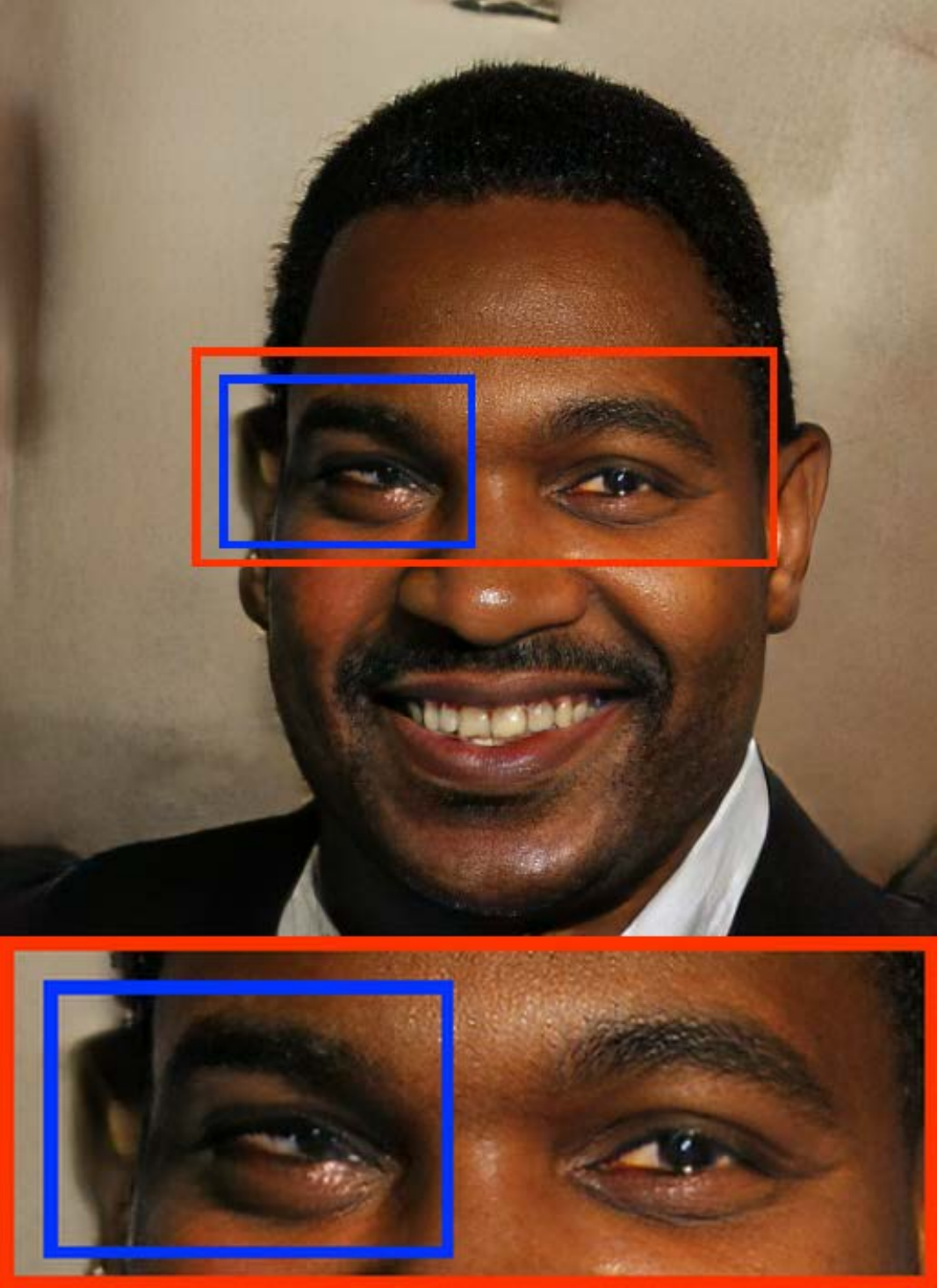} &
   \includegraphics[width=\swceleba]{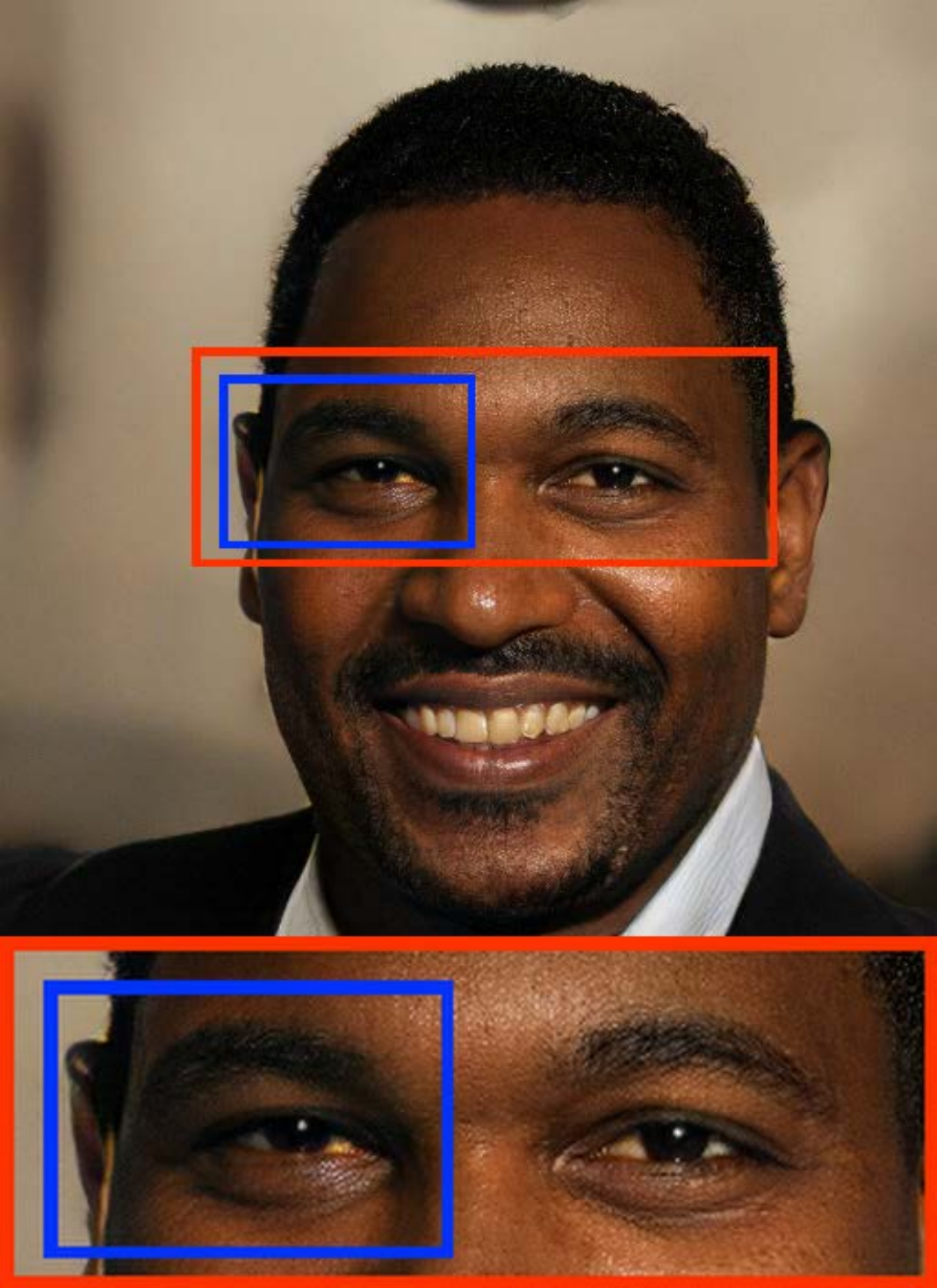} &
   \includegraphics[width=\swceleba]{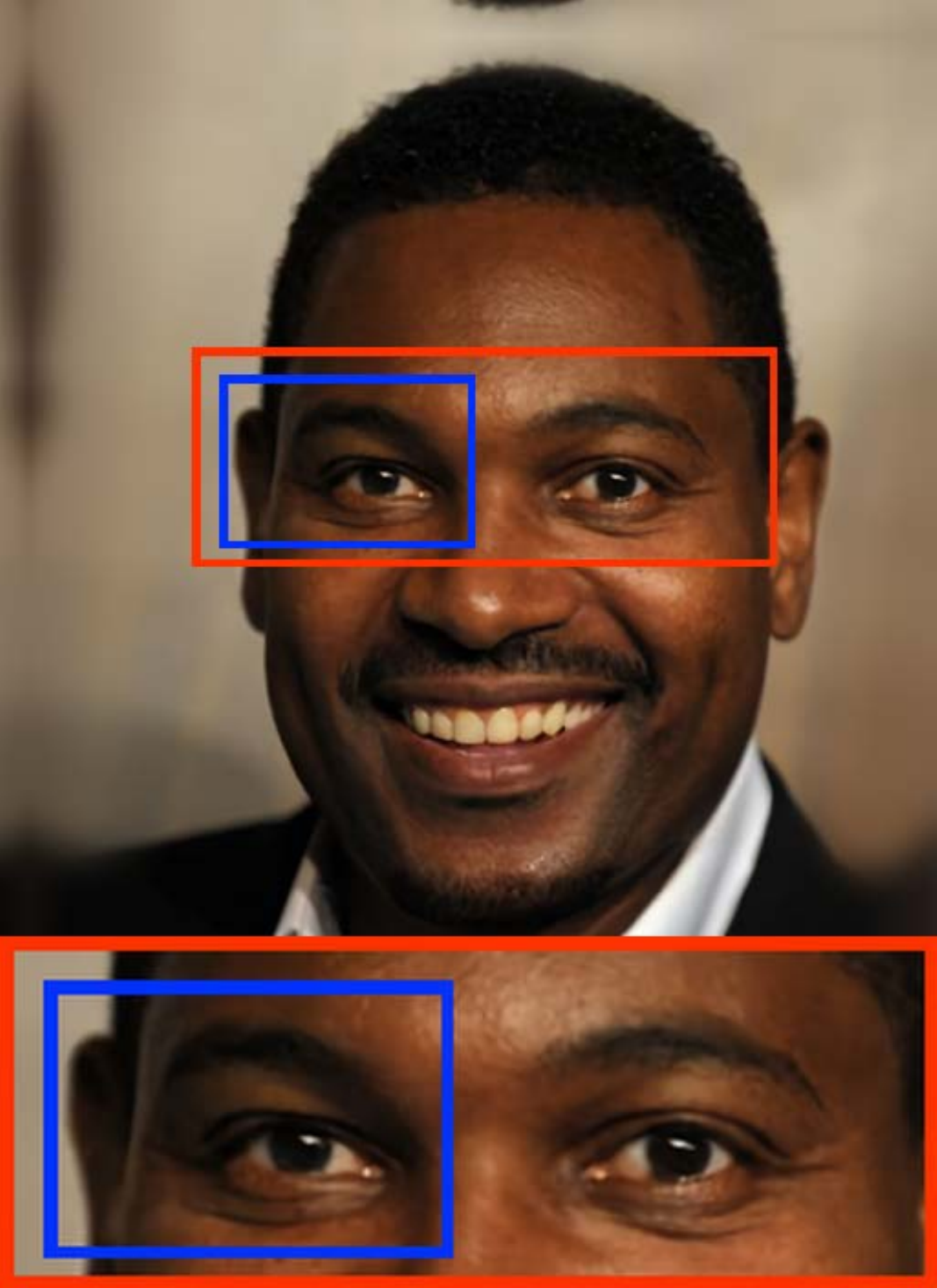} \\
   \includegraphics[width=\swceleba]{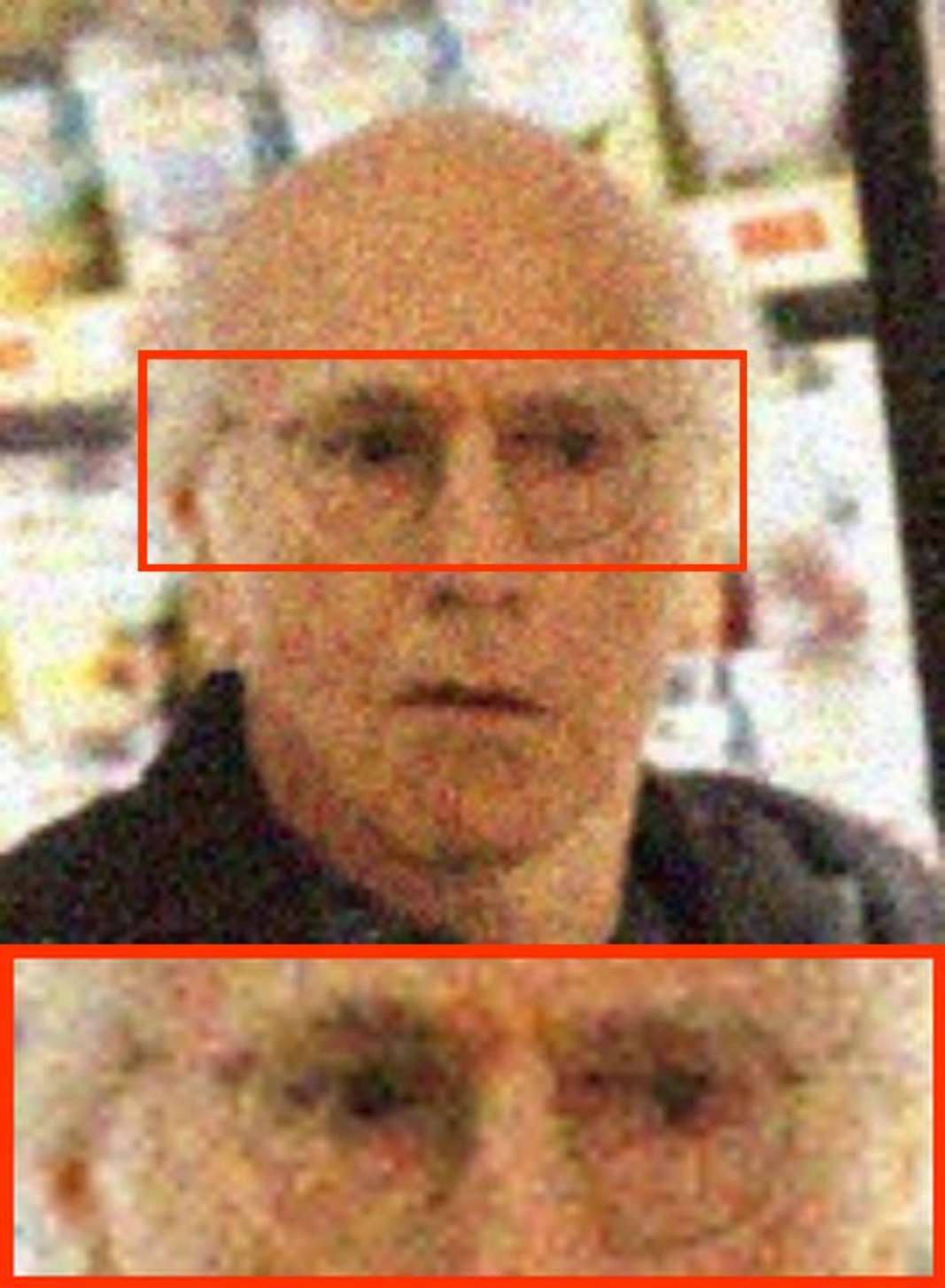}&
   \includegraphics[width=\swceleba]{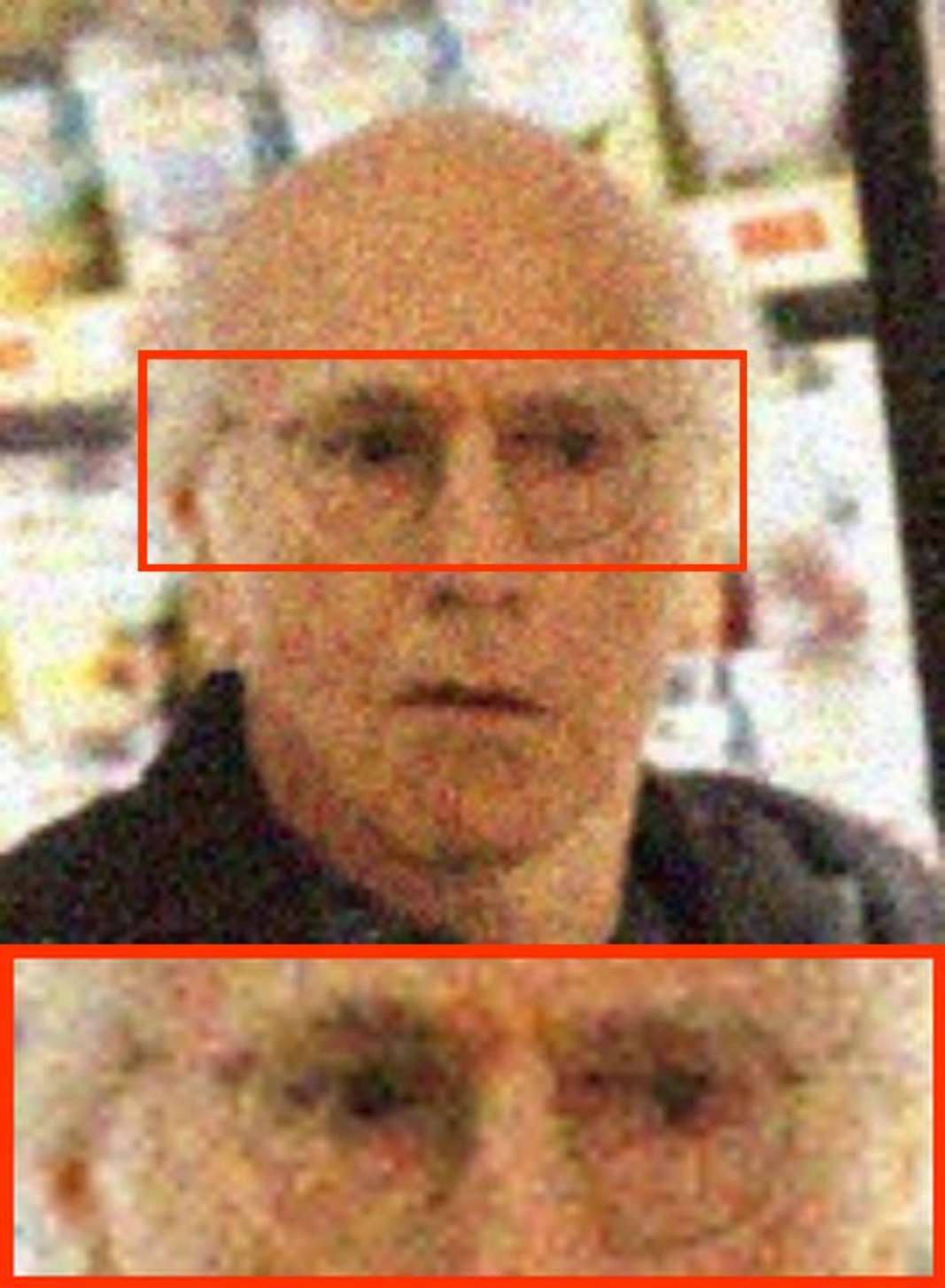}&
   \includegraphics[width=\swceleba]{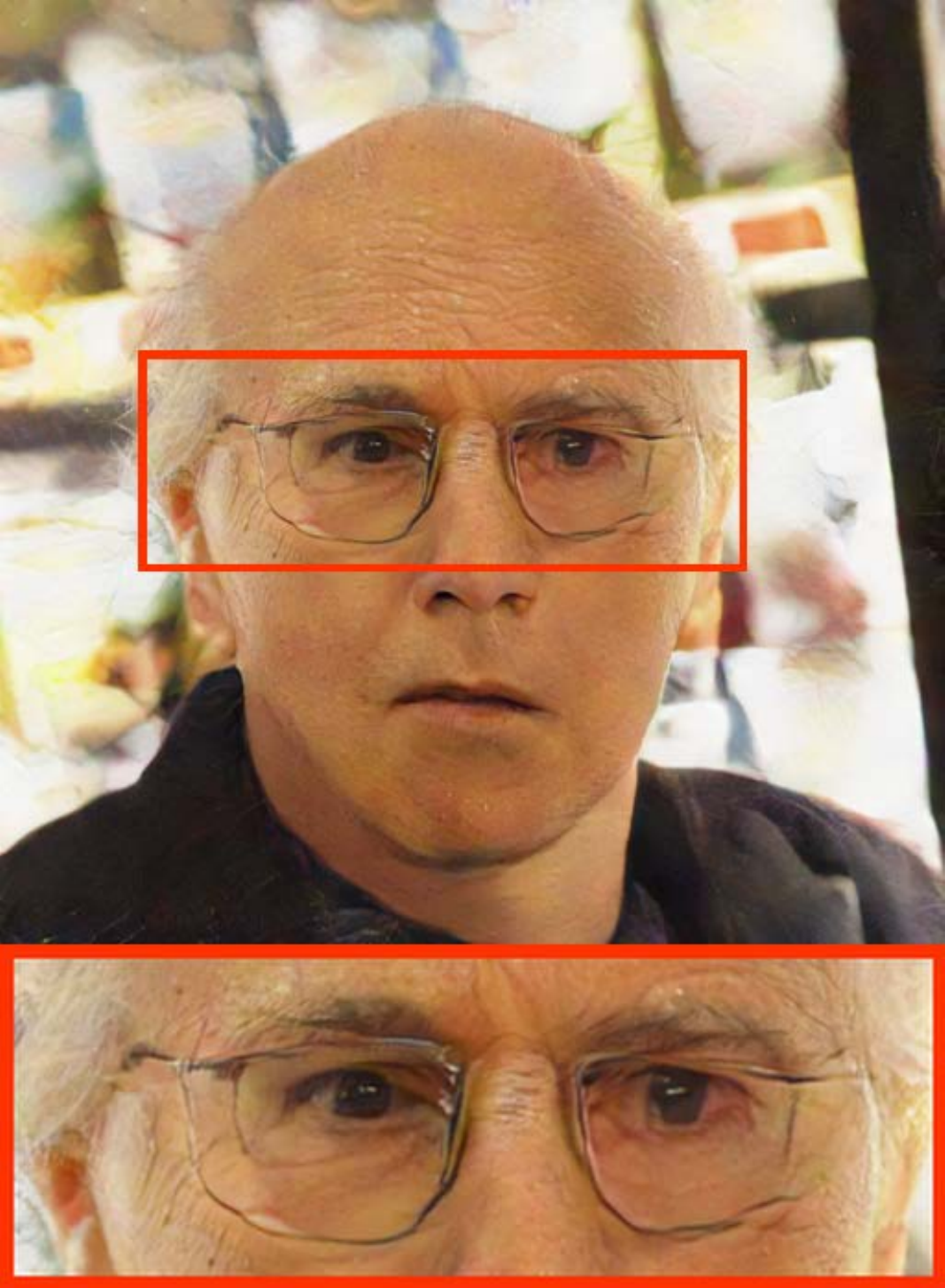}&
   \includegraphics[width=\swceleba]{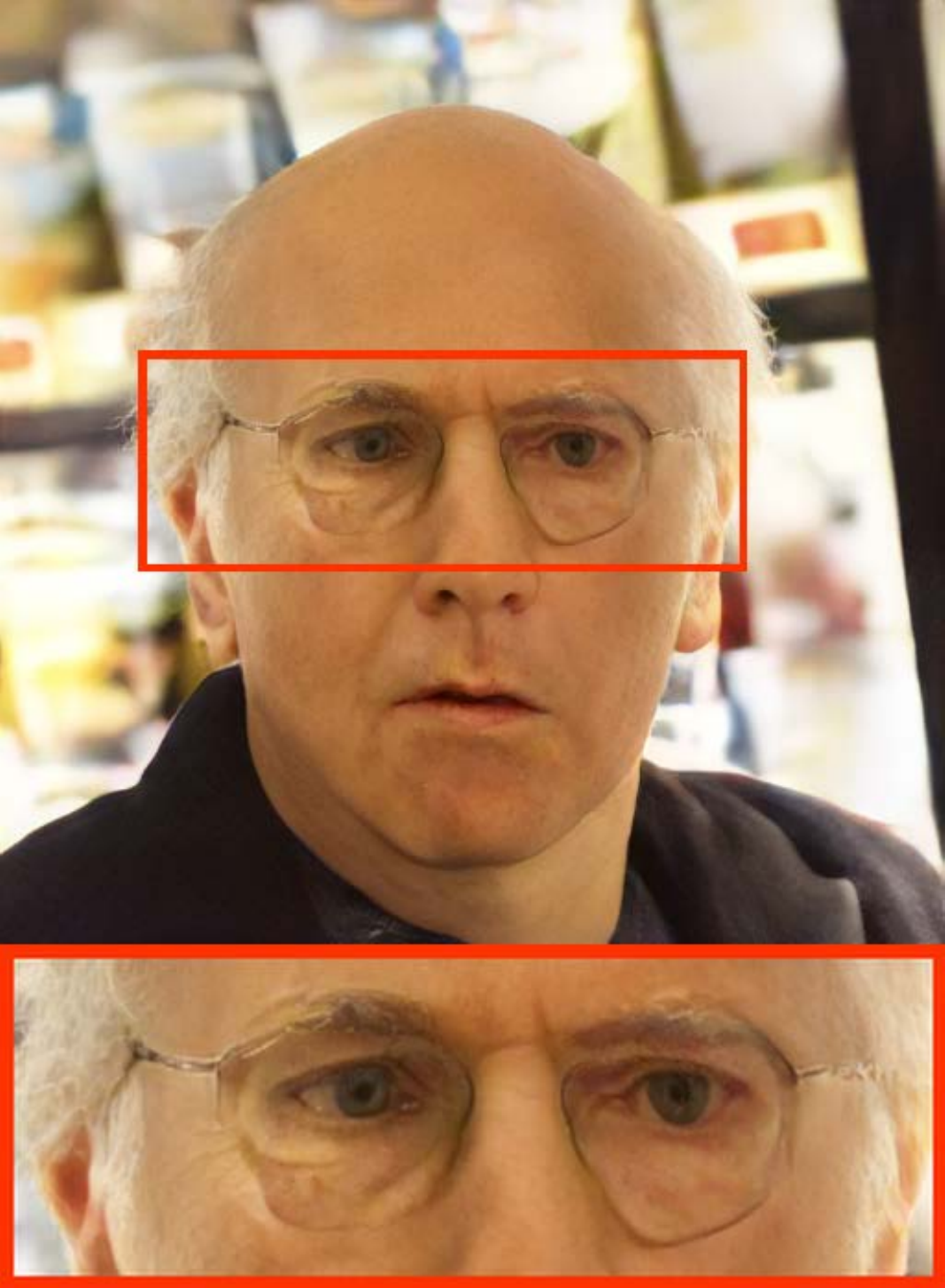}&
   \includegraphics[width=\swceleba]{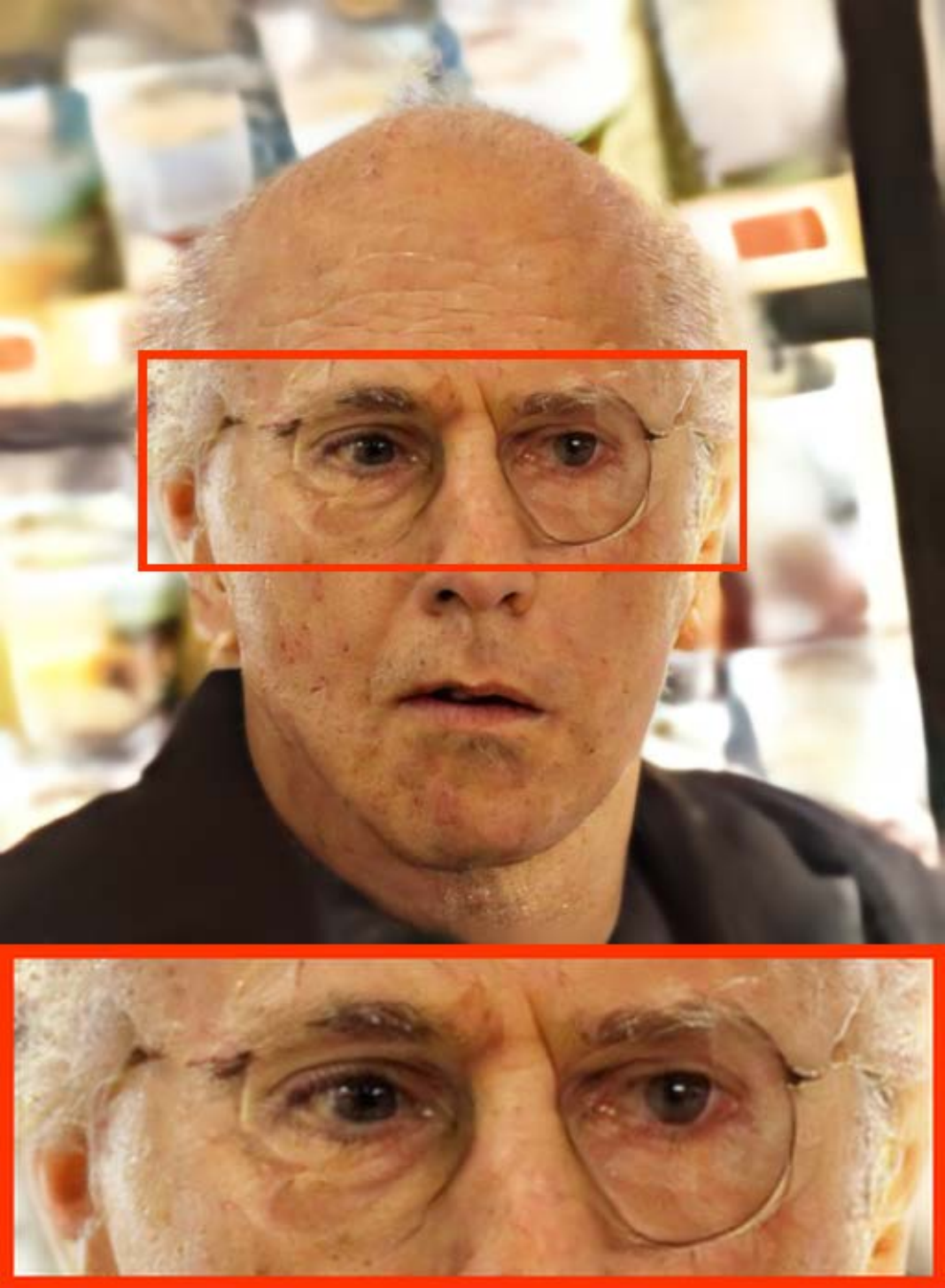}&
   \includegraphics[width=\swceleba]{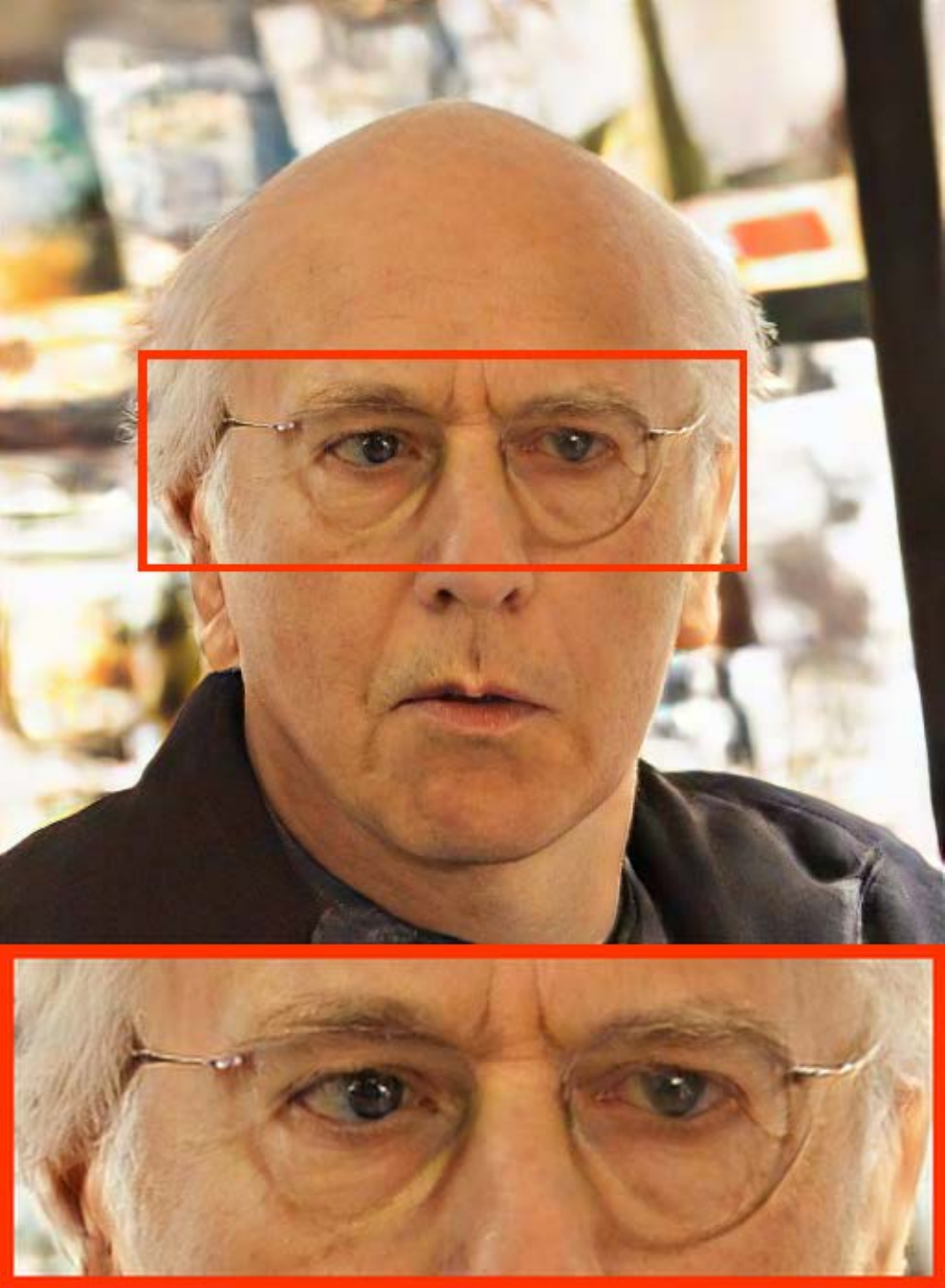} &
   \includegraphics[width=\swceleba]{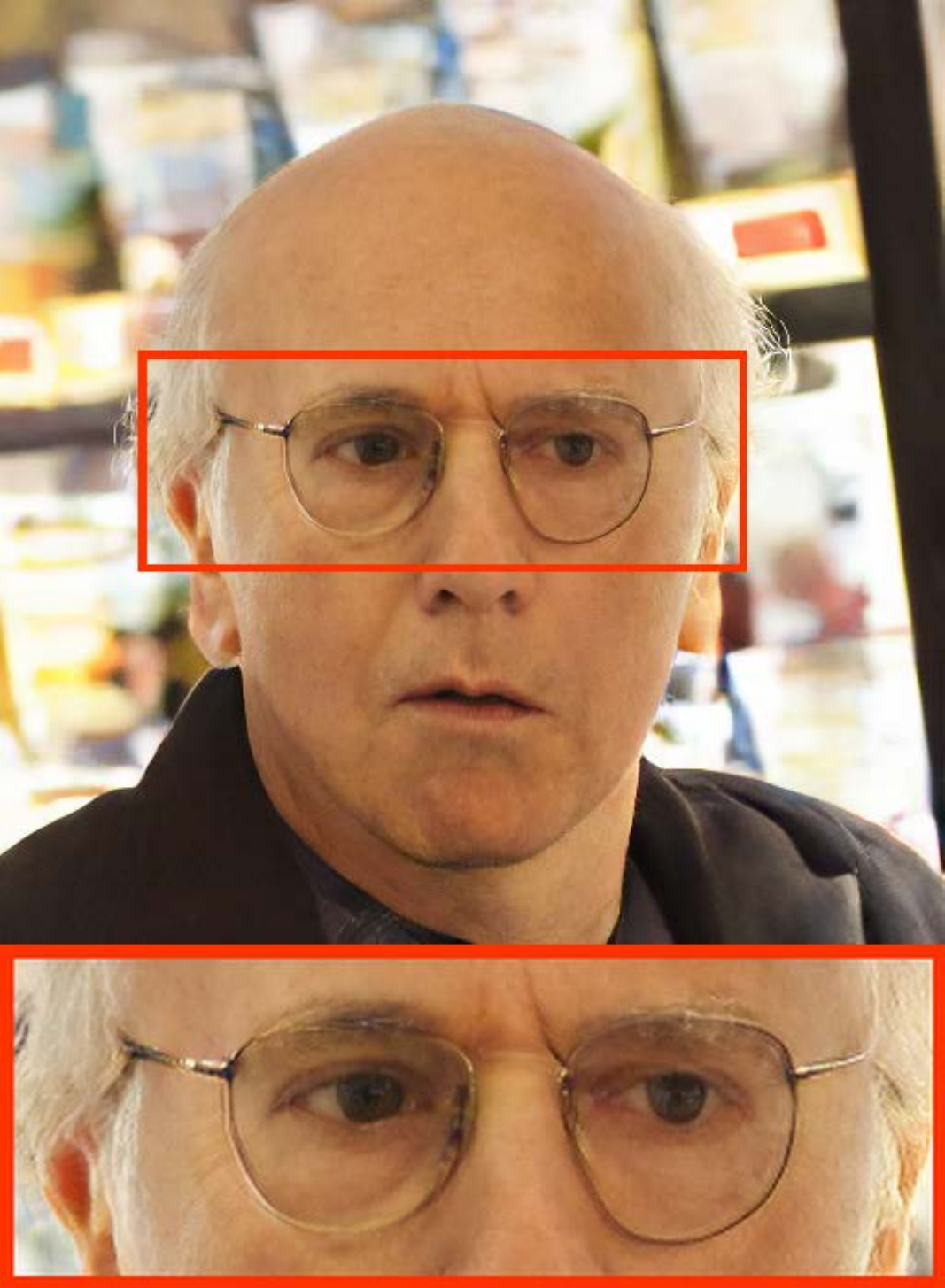} &
   \includegraphics[width=\swceleba]{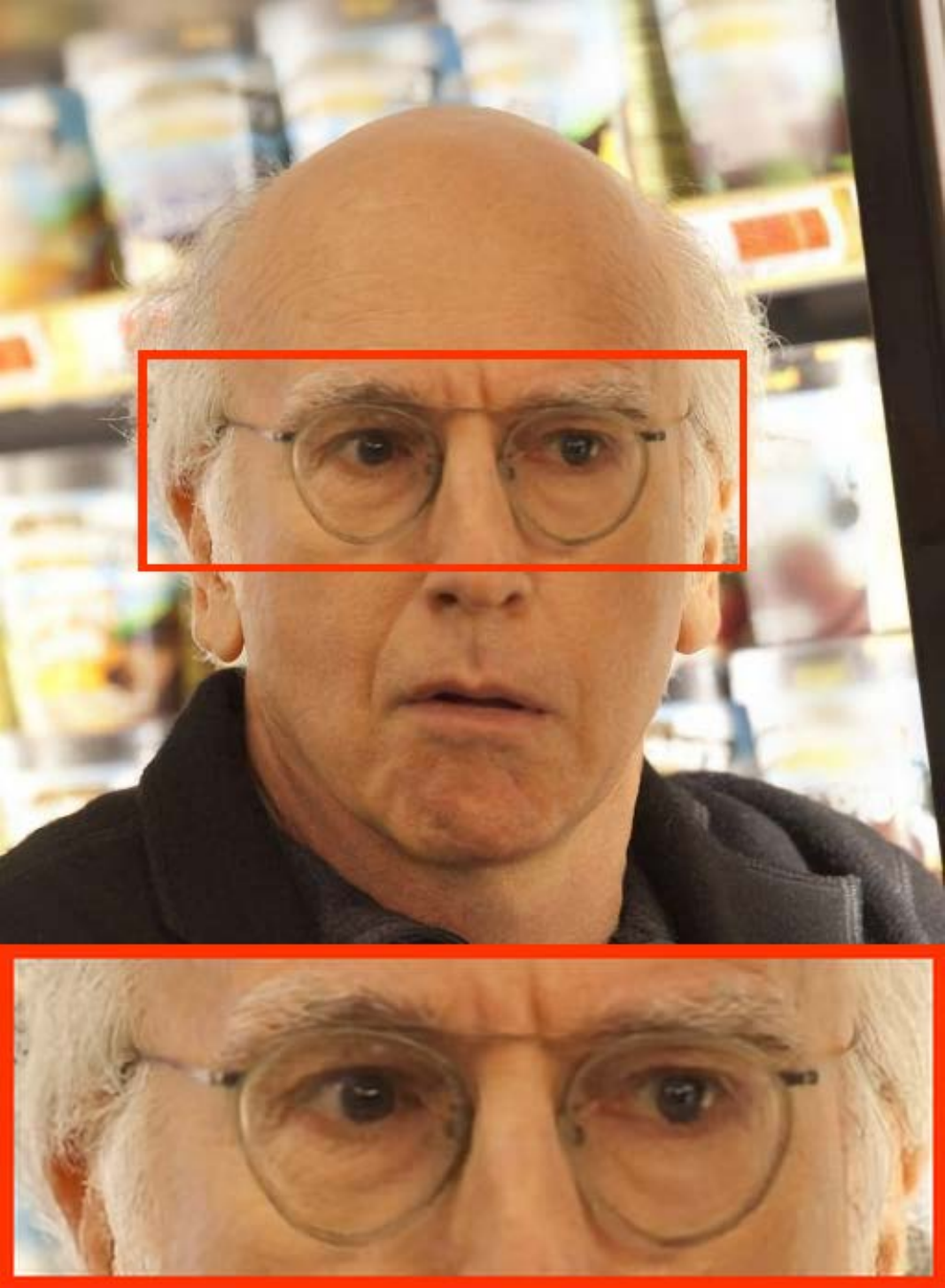} \\
  Input & DFDNet~\cite{li2020blind}  & PSFRGAN~\cite{chen2021progressive} & GFP-GAN~\cite{wang2021towards} &  GPEN~\cite{yang2021gan} & VQFR~\cite{gu2022vqfr} & \textbf{Ours} & GT
\end{tabular}
\end{center}
\caption{
\textcolor{black}{Qualitative comparison on the \textbf{CelebA-Test}~\cite{liu2015deep}. The results of our RestoreFormer++ have a more natural and complete overview and contain more details in the areas of eyes, mouth, and glasses. Note that DFDNet~\cite{li2020blind} relies on dlib~\cite{king2009dlib} for facial detection while matching priors from its facial component dictionaries, and failure in detection results in no restoration, as seen in the second result.}
\textbf{Zoom in for a better view}.
}
\label{fig:celeba}
\end{figure*}%

\renewcommand{\tabcolsep}{.5pt}
\begin{figure*}
\hsize=\textwidth
\vspace{-0.3cm}
\begin{center}
\begin{tabular}{cccccccc}
    \includegraphics[width=\swceleba]{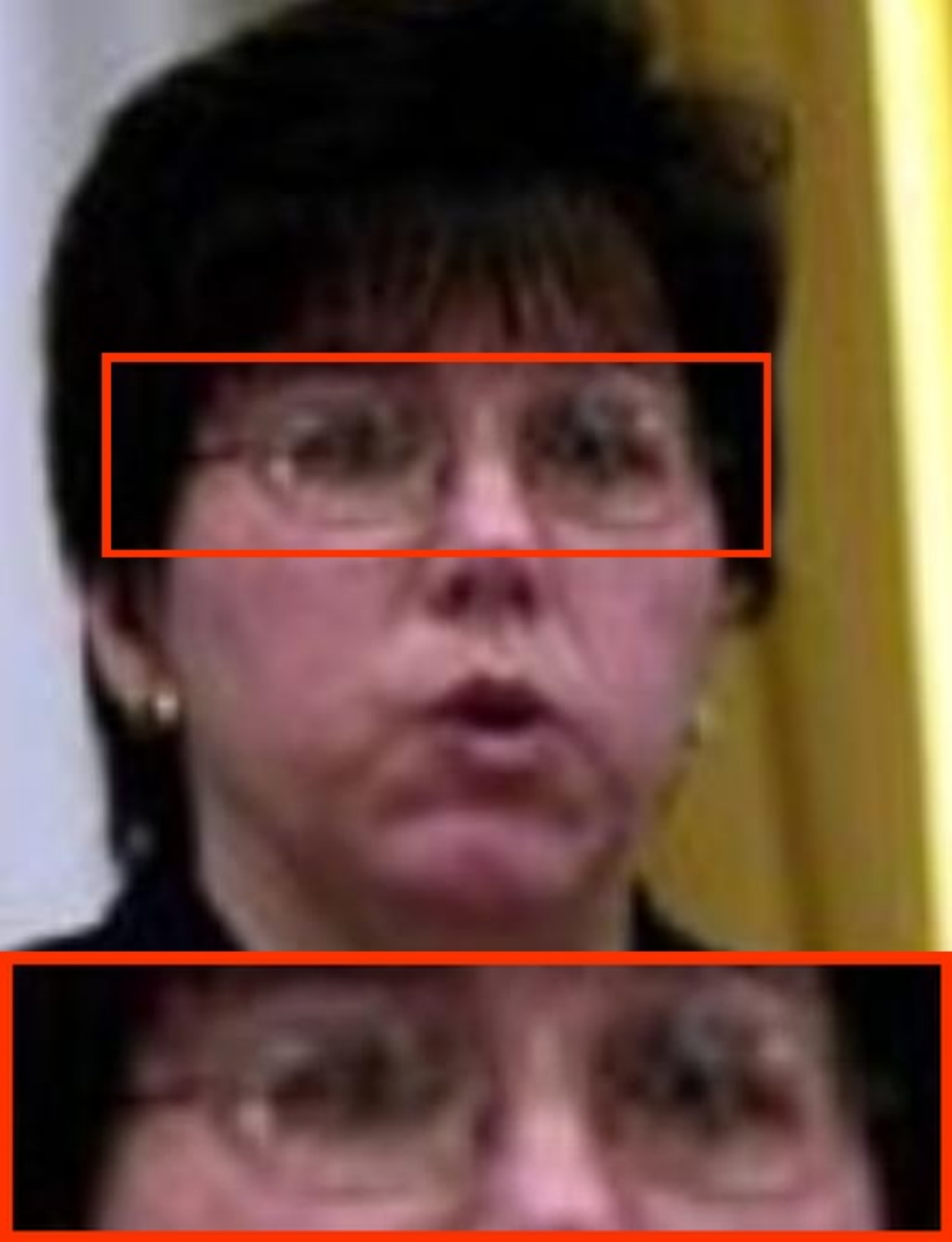}&
    \includegraphics[width=\swceleba]{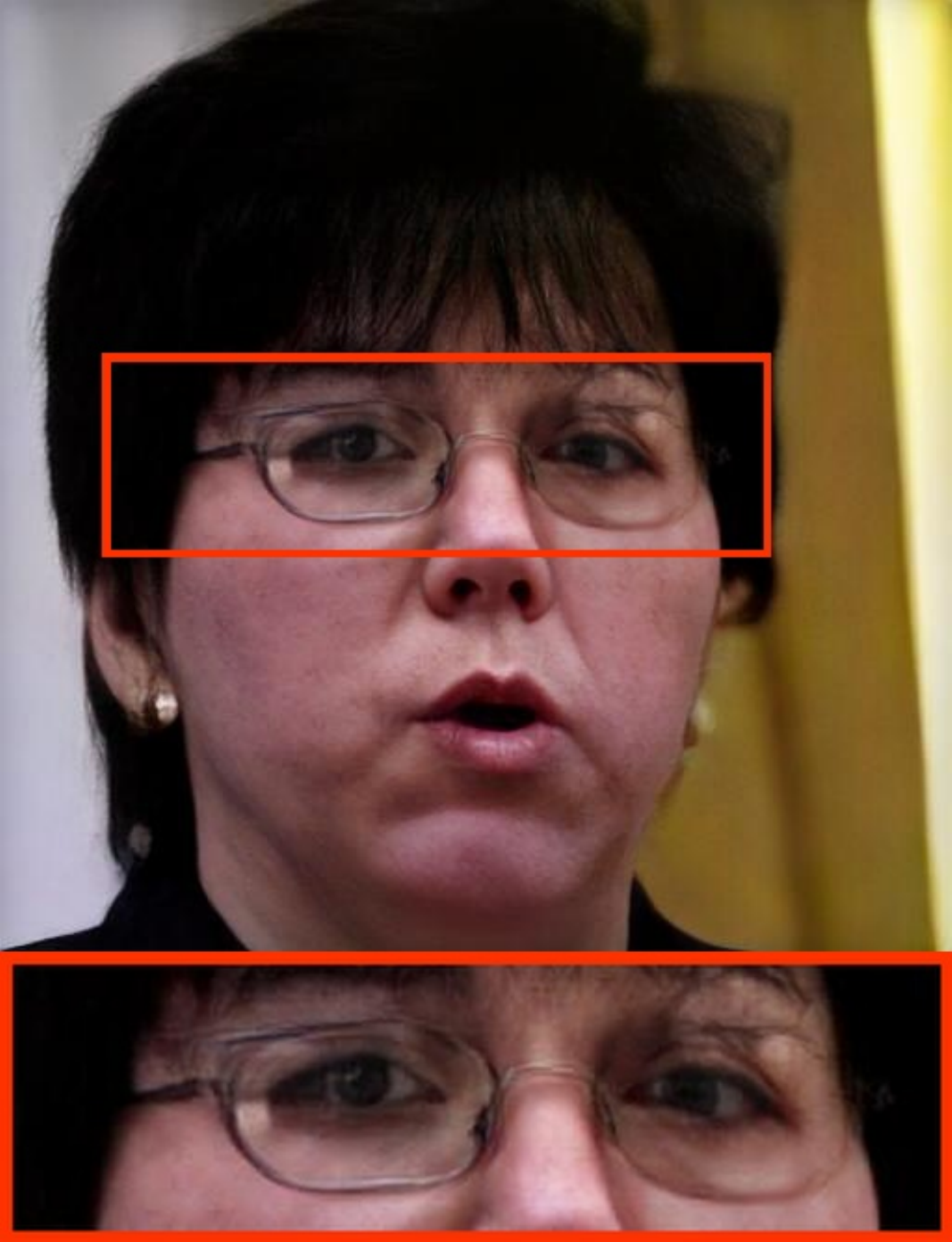}&
    \includegraphics[width=\swceleba]{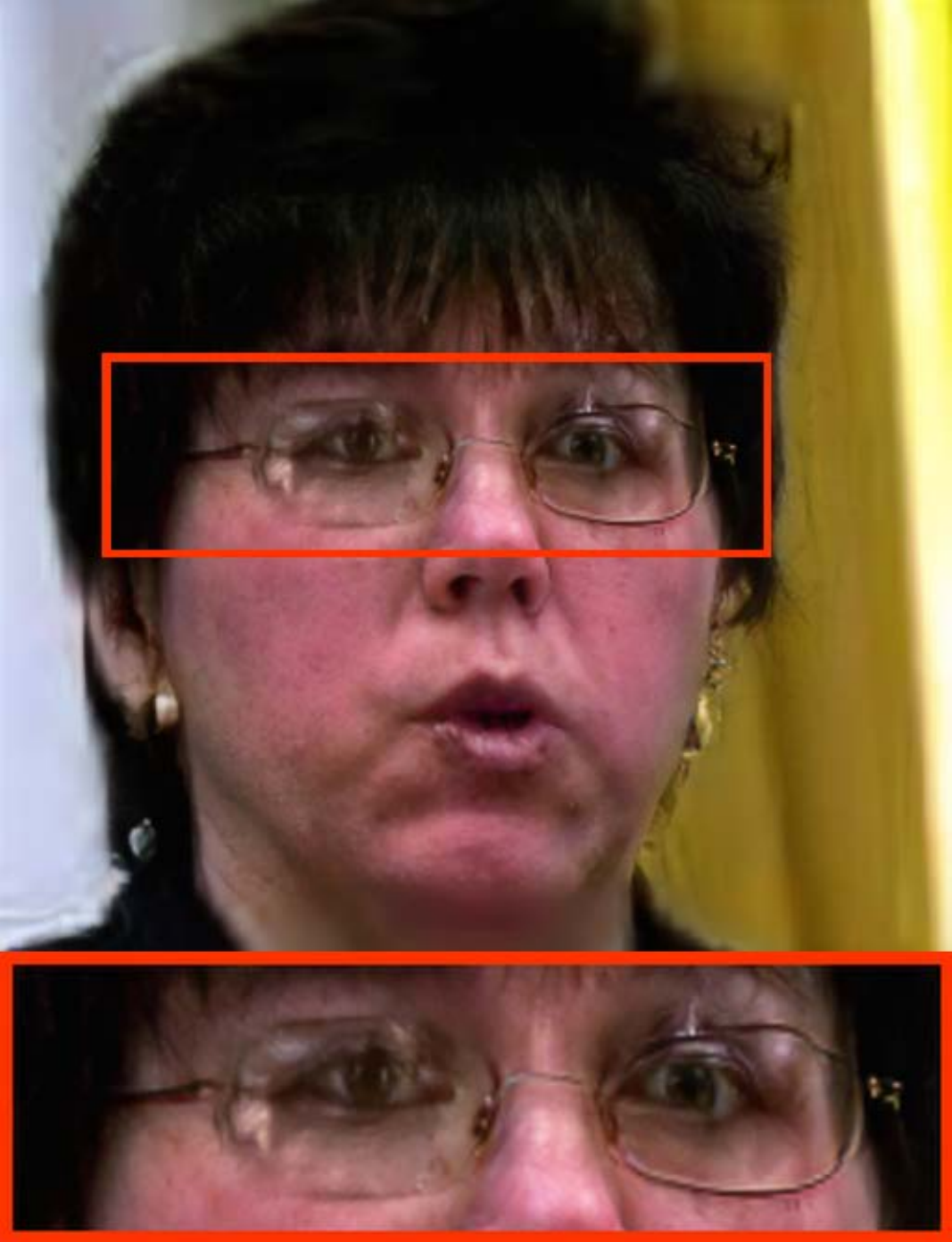}&
    \includegraphics[width=\swceleba]{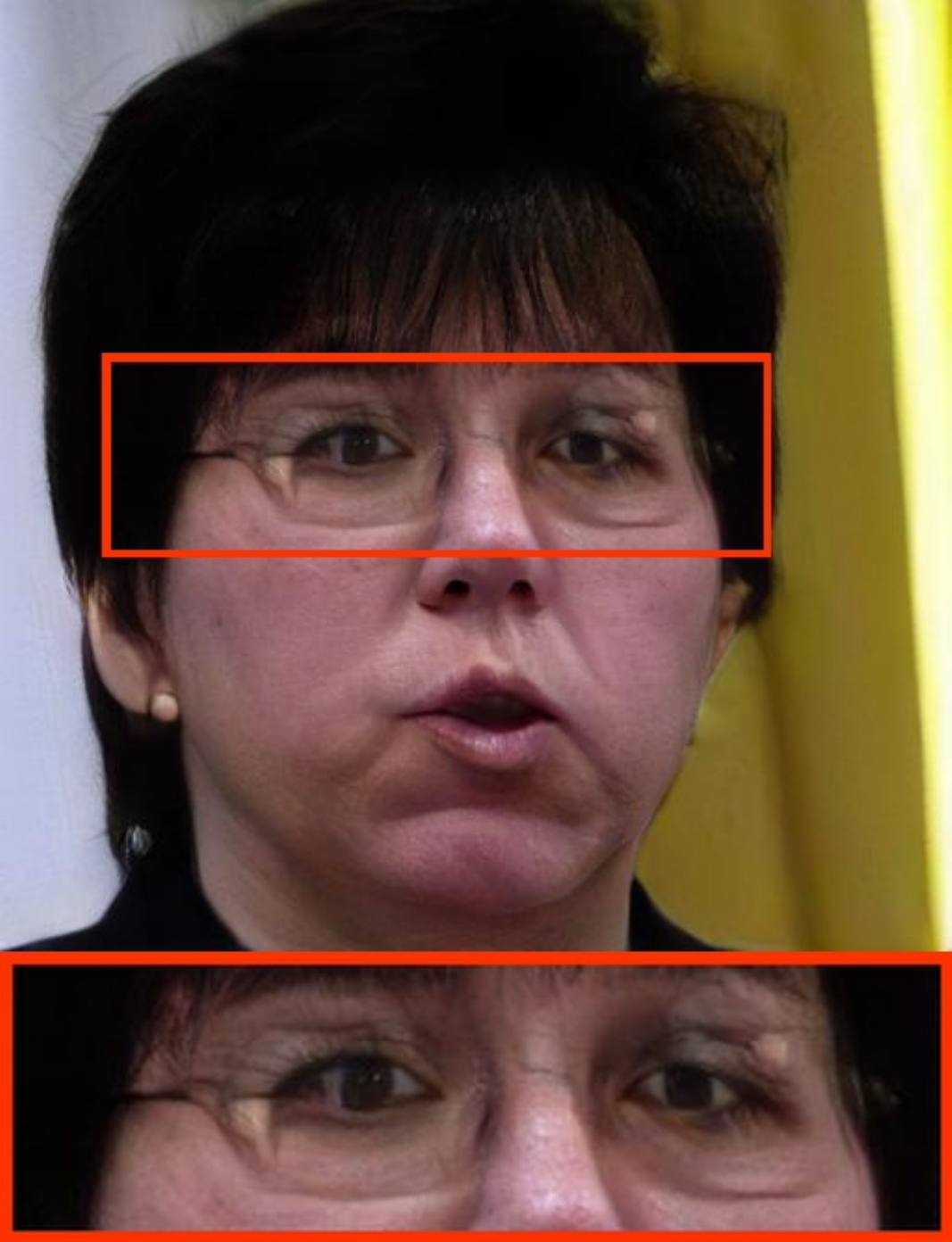}&
    \includegraphics[width=\swceleba]{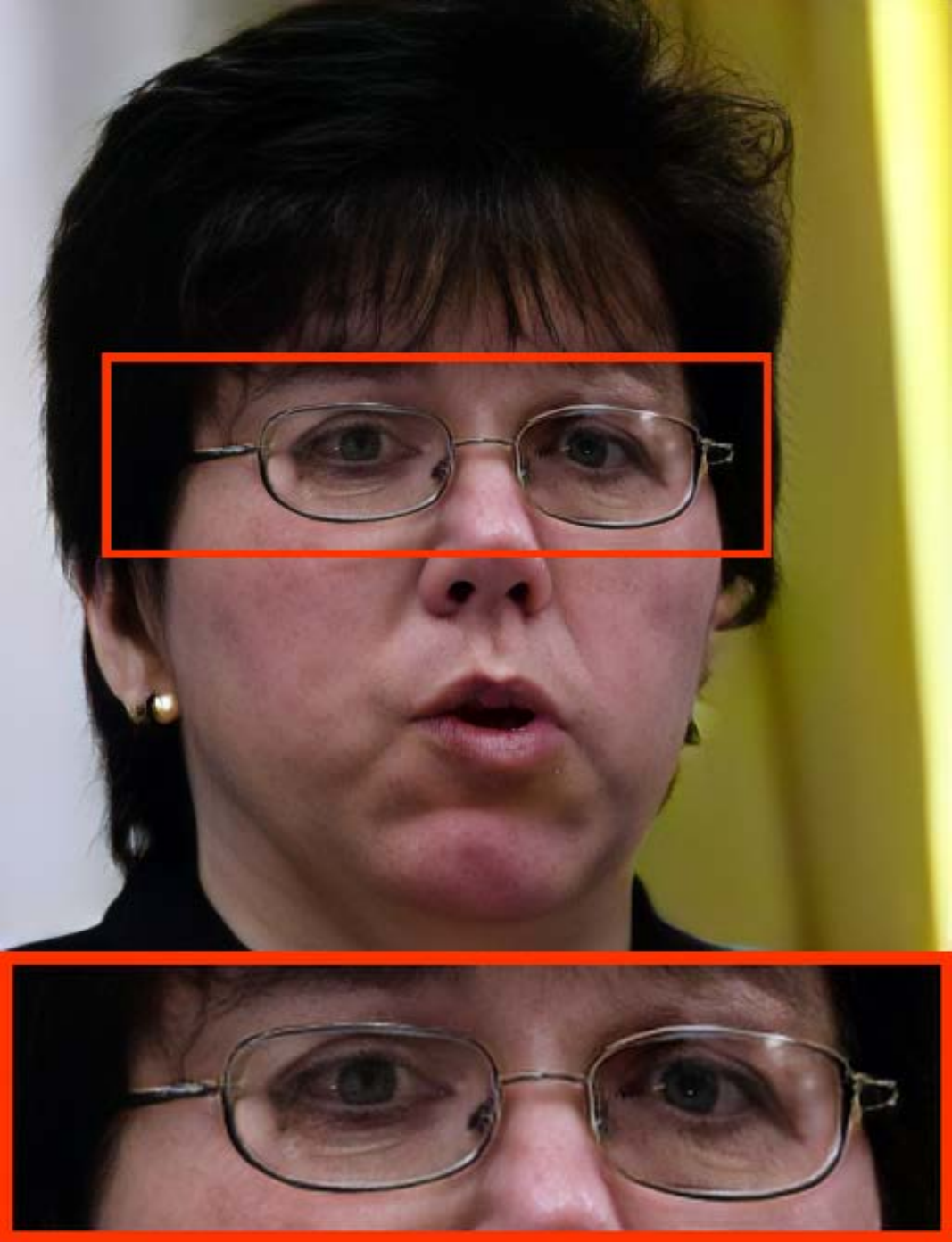}&
    \includegraphics[width=\swceleba]{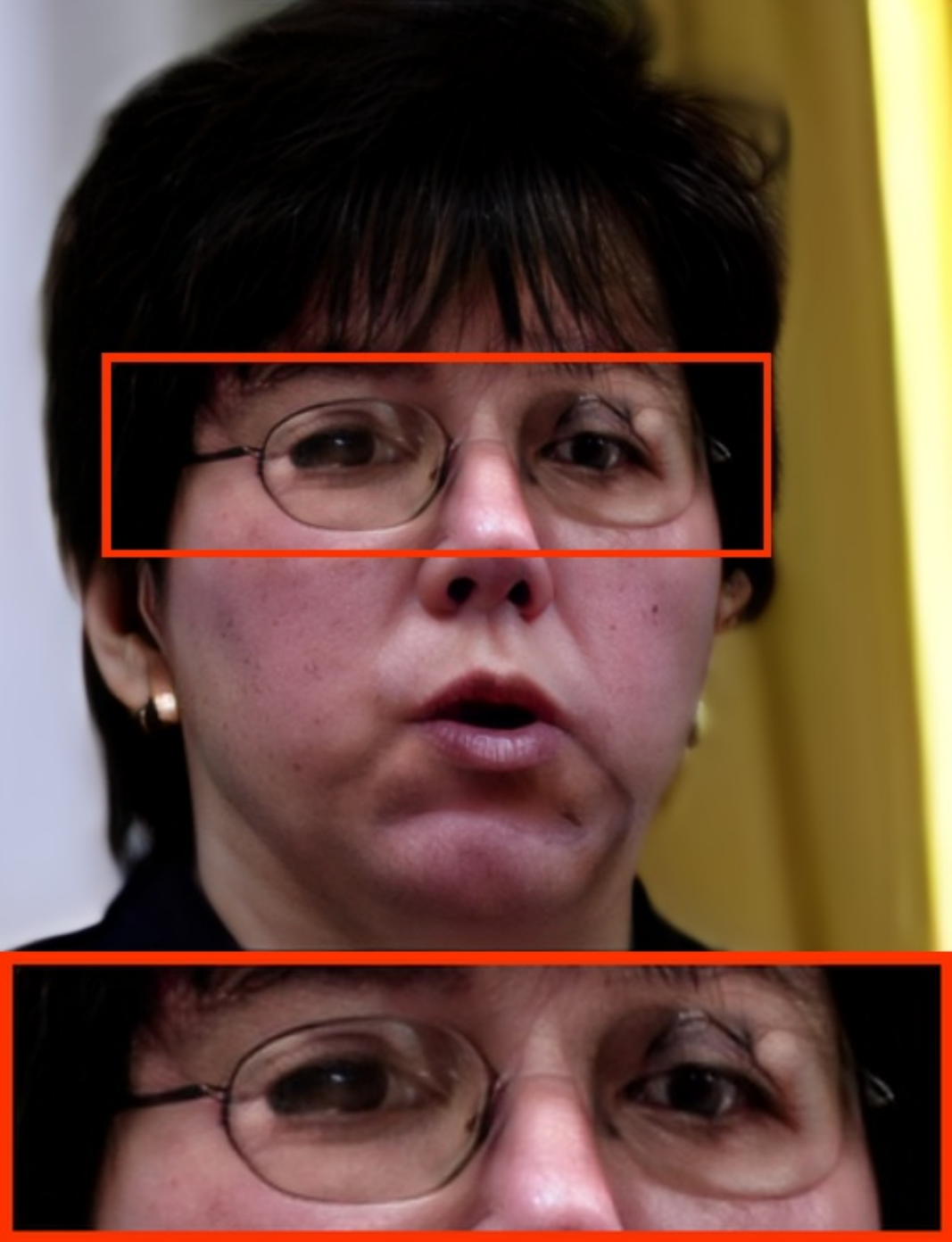}&
    \includegraphics[width=\swceleba]{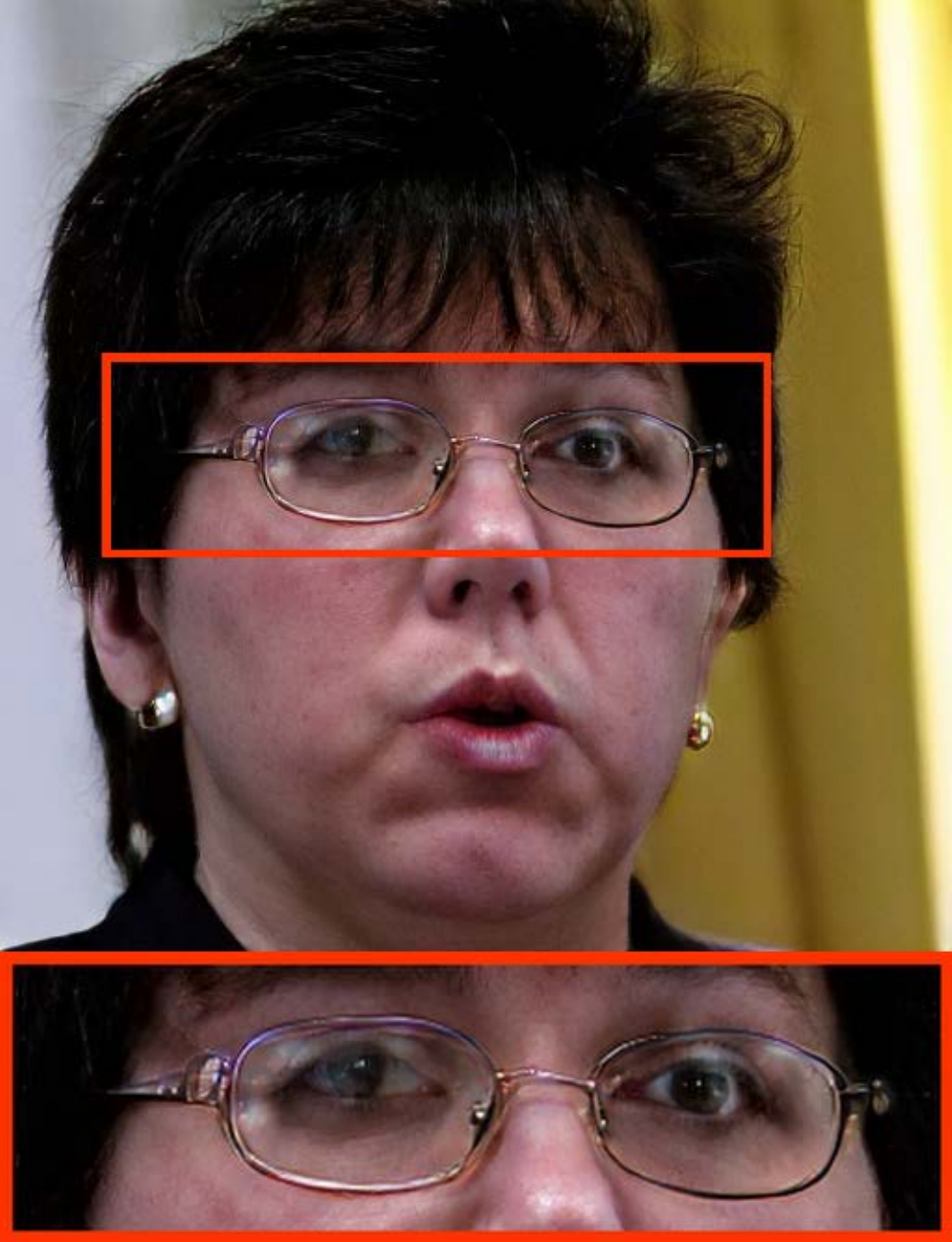}&
    \includegraphics[width=\swceleba]{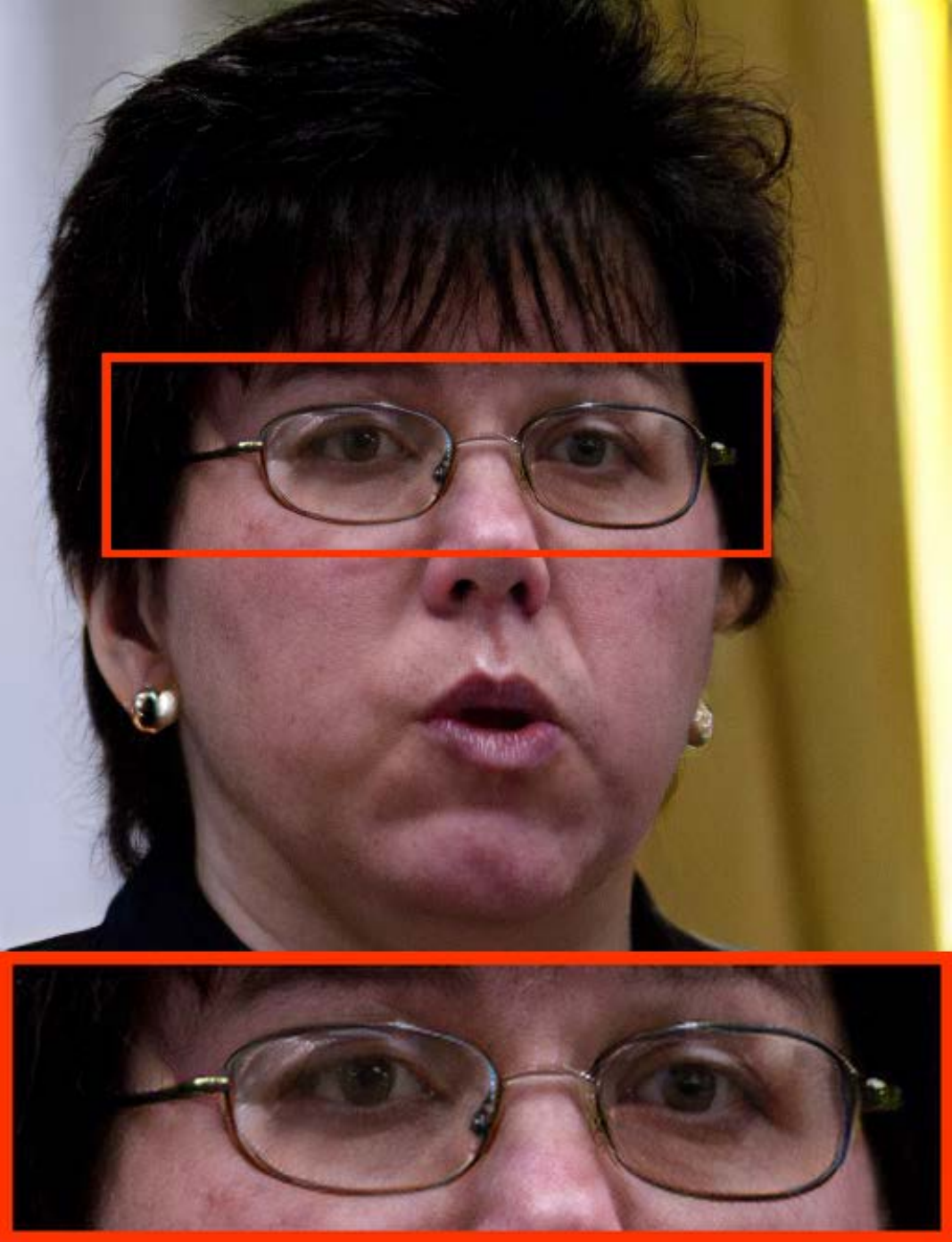} \\
    \includegraphics[width=\swceleba]{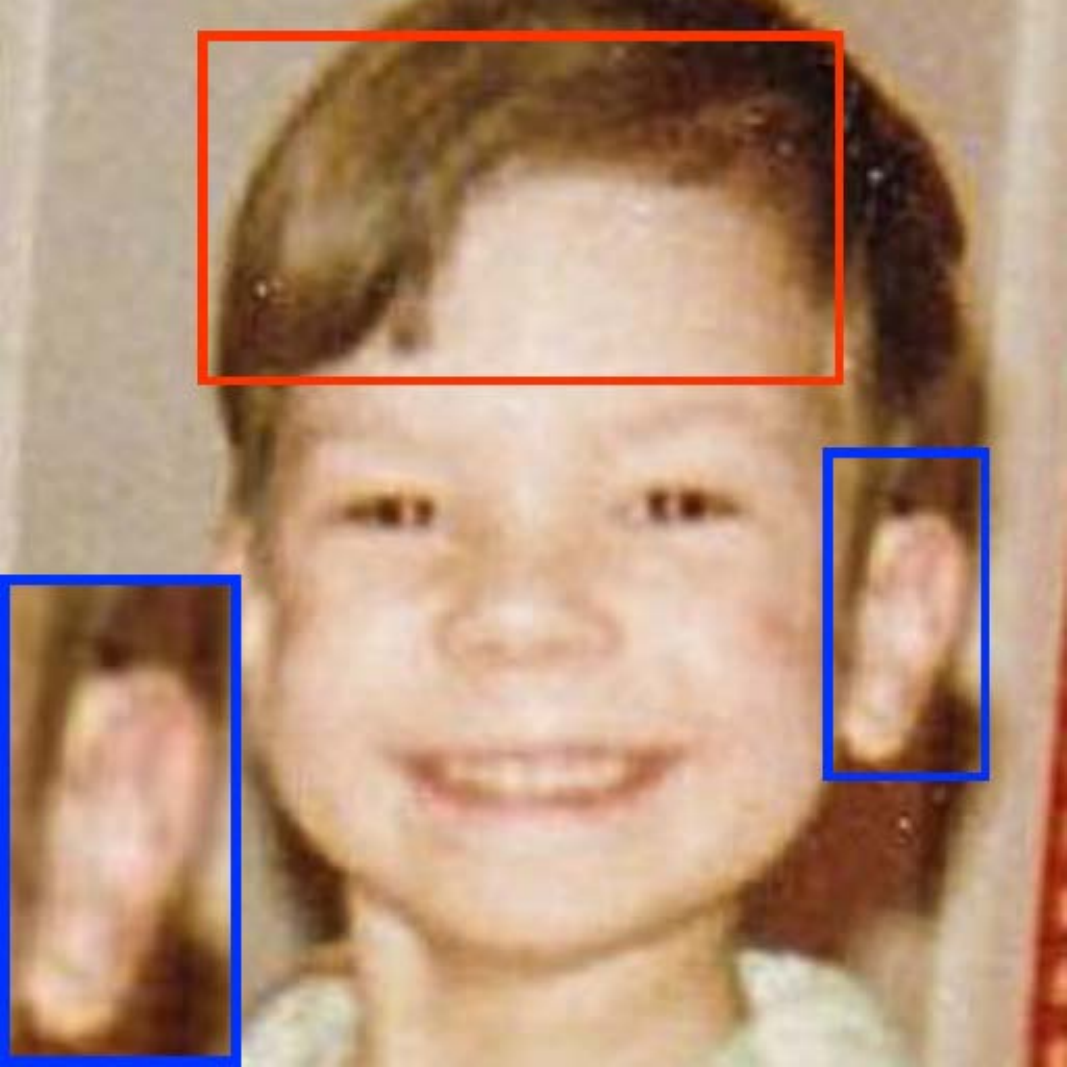}&
    \includegraphics[width=\swceleba]{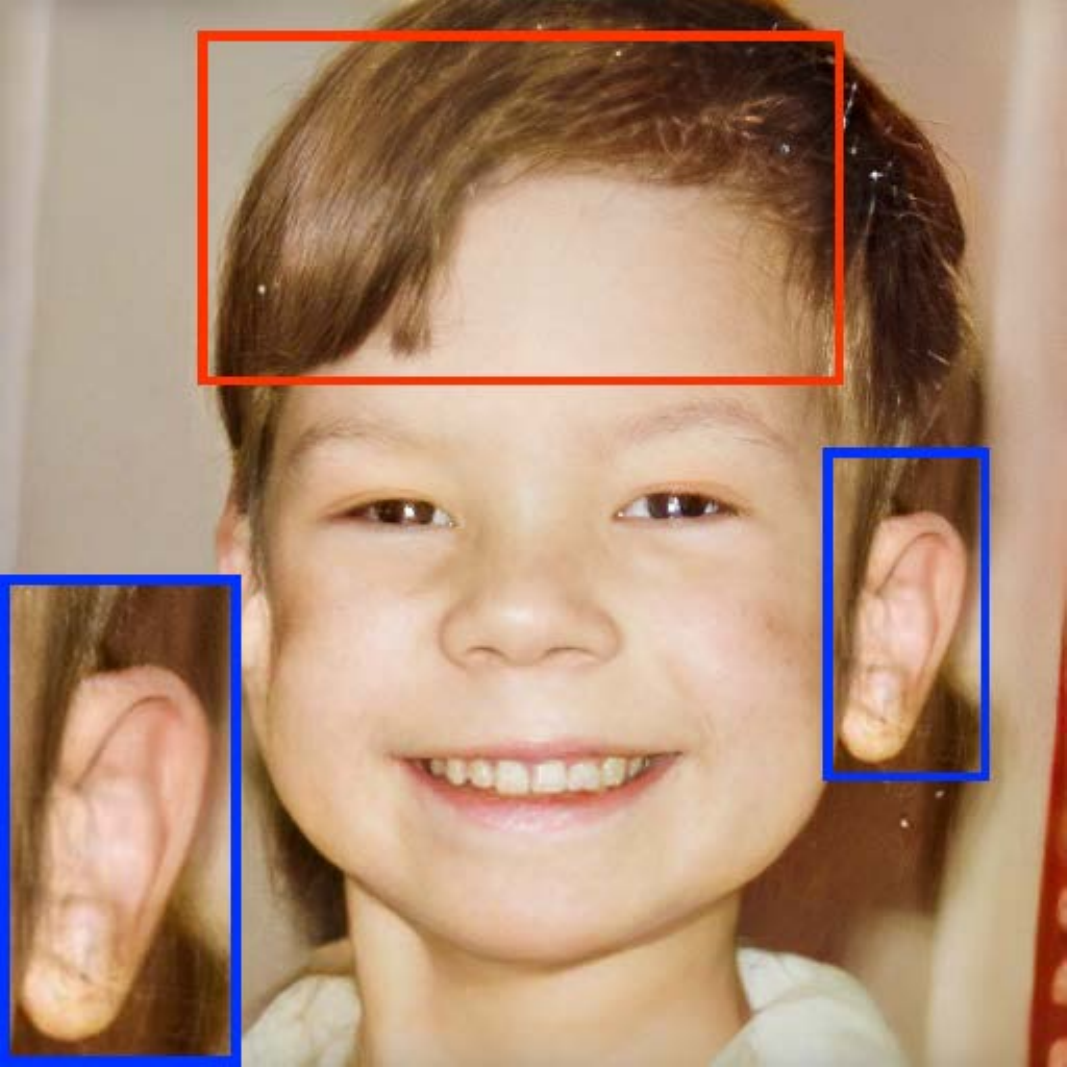}&
    \includegraphics[width=\swceleba]{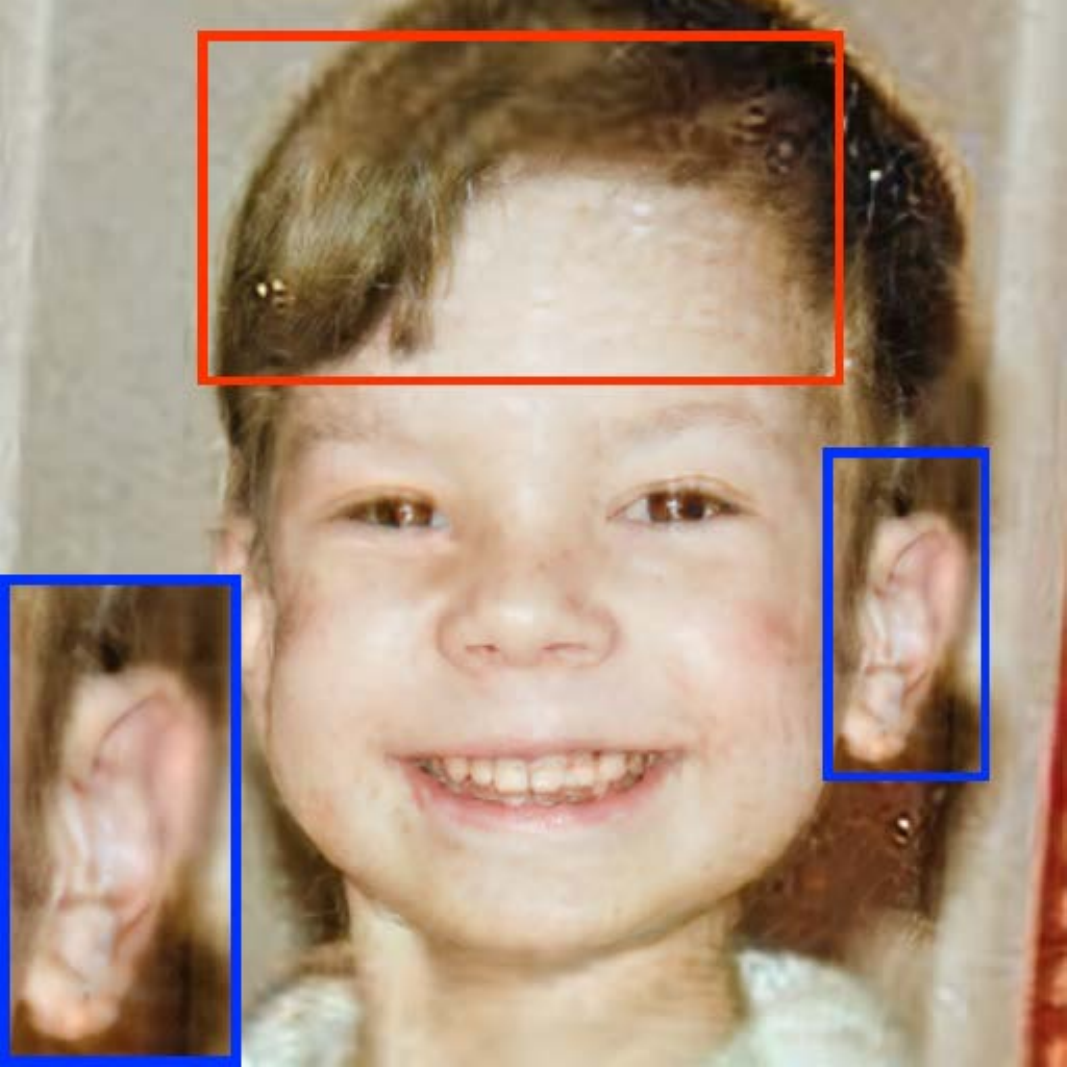}&
    \includegraphics[width=\swceleba]{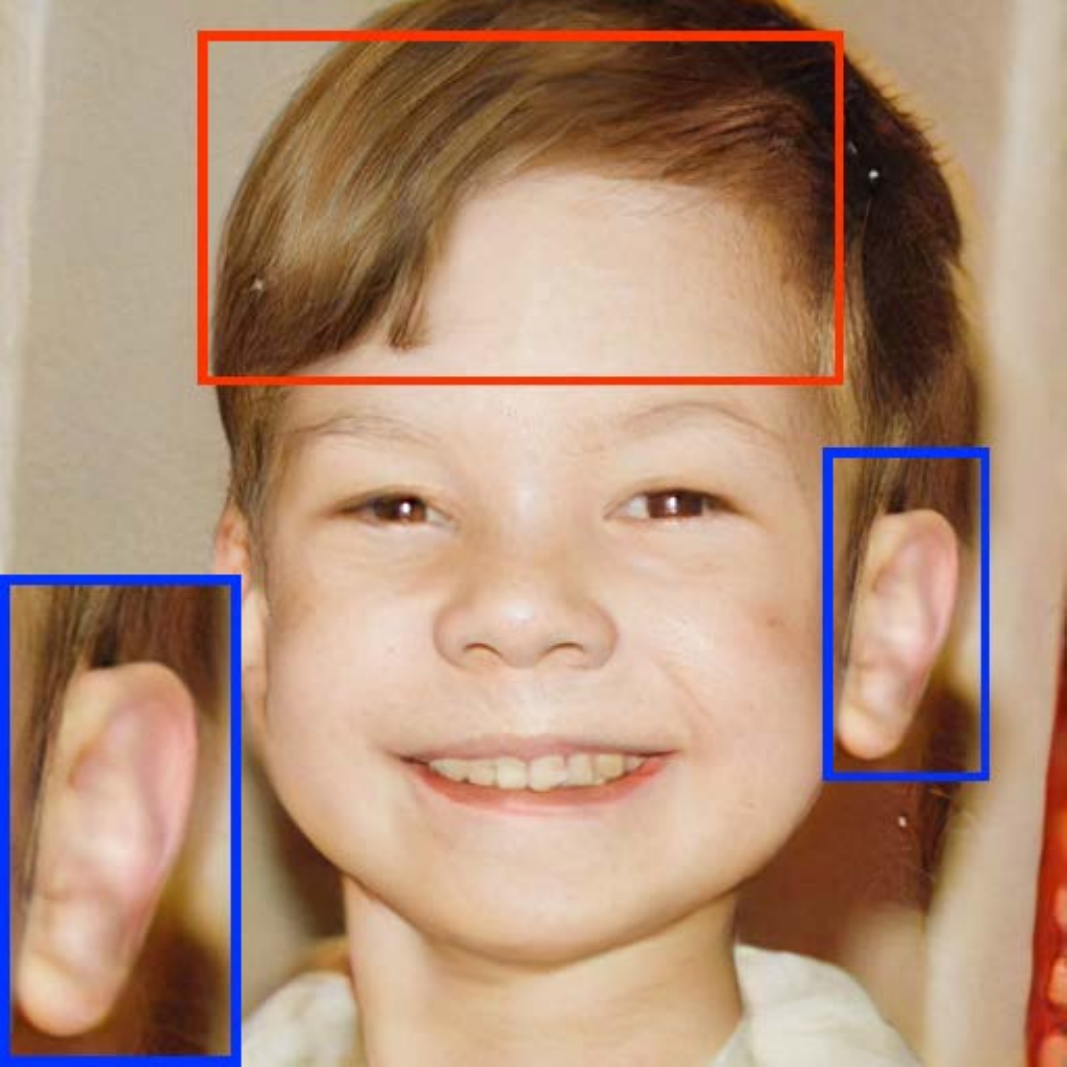}&
    \includegraphics[width=\swceleba]{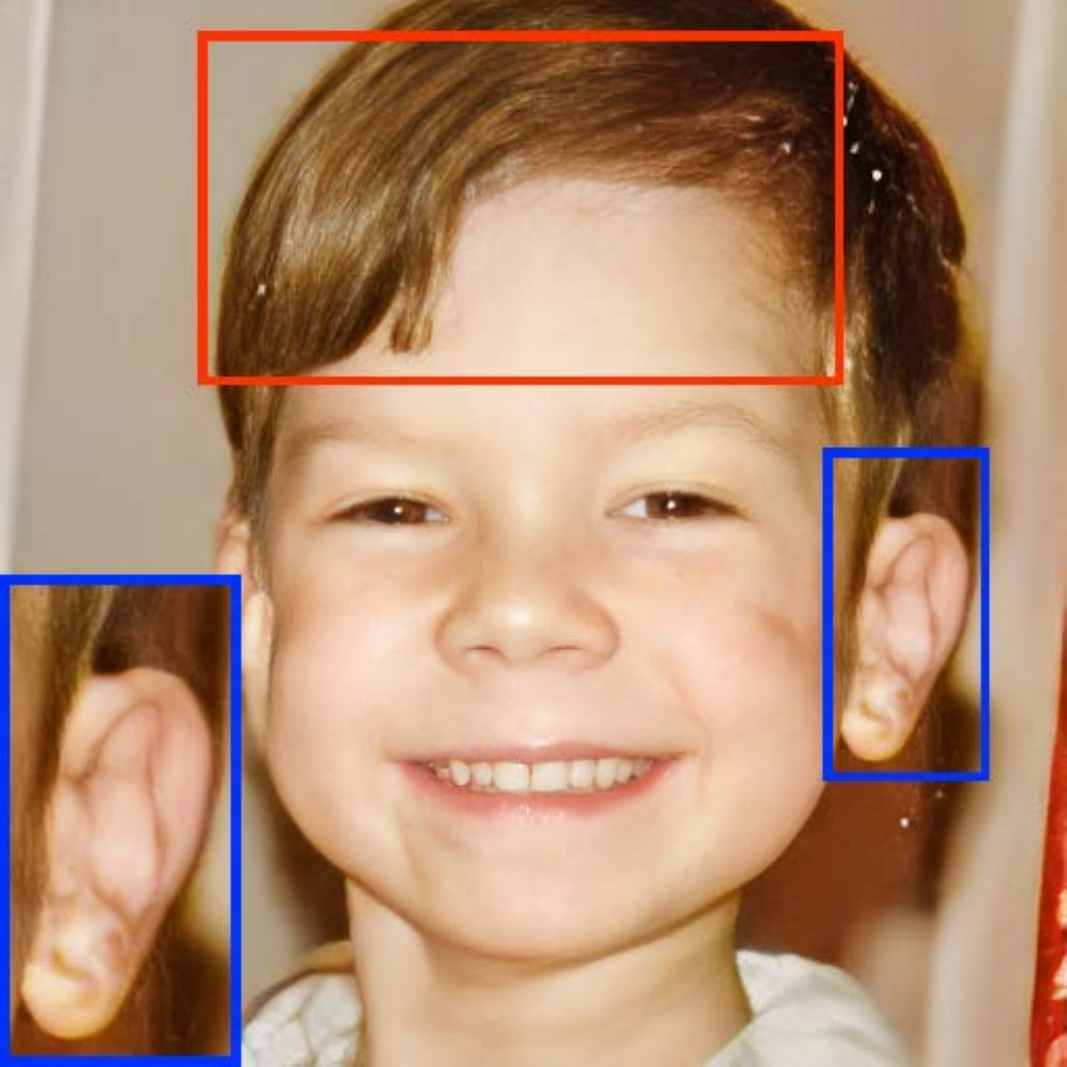}&
    \includegraphics[width=\swceleba]{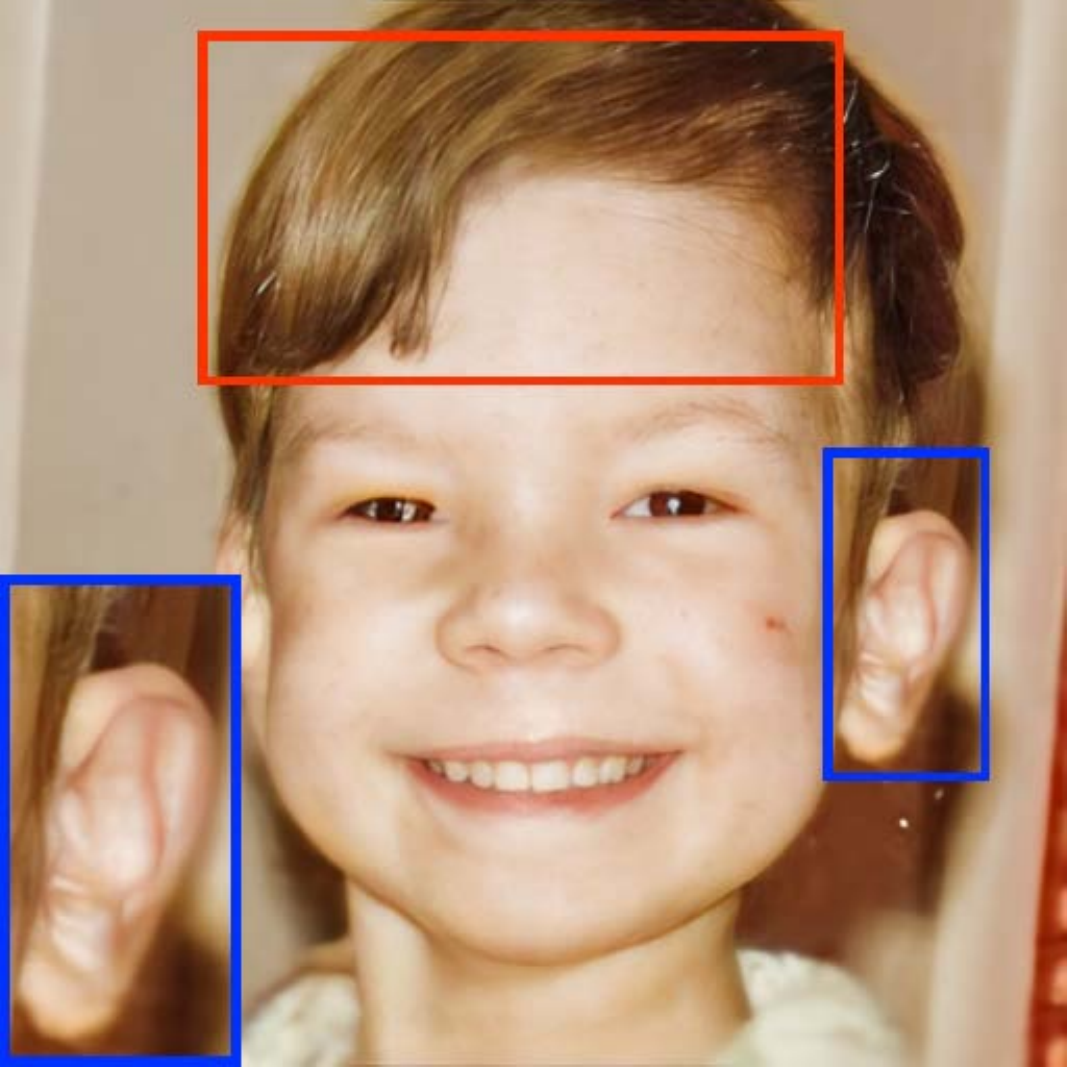}&
    \includegraphics[width=\swceleba]{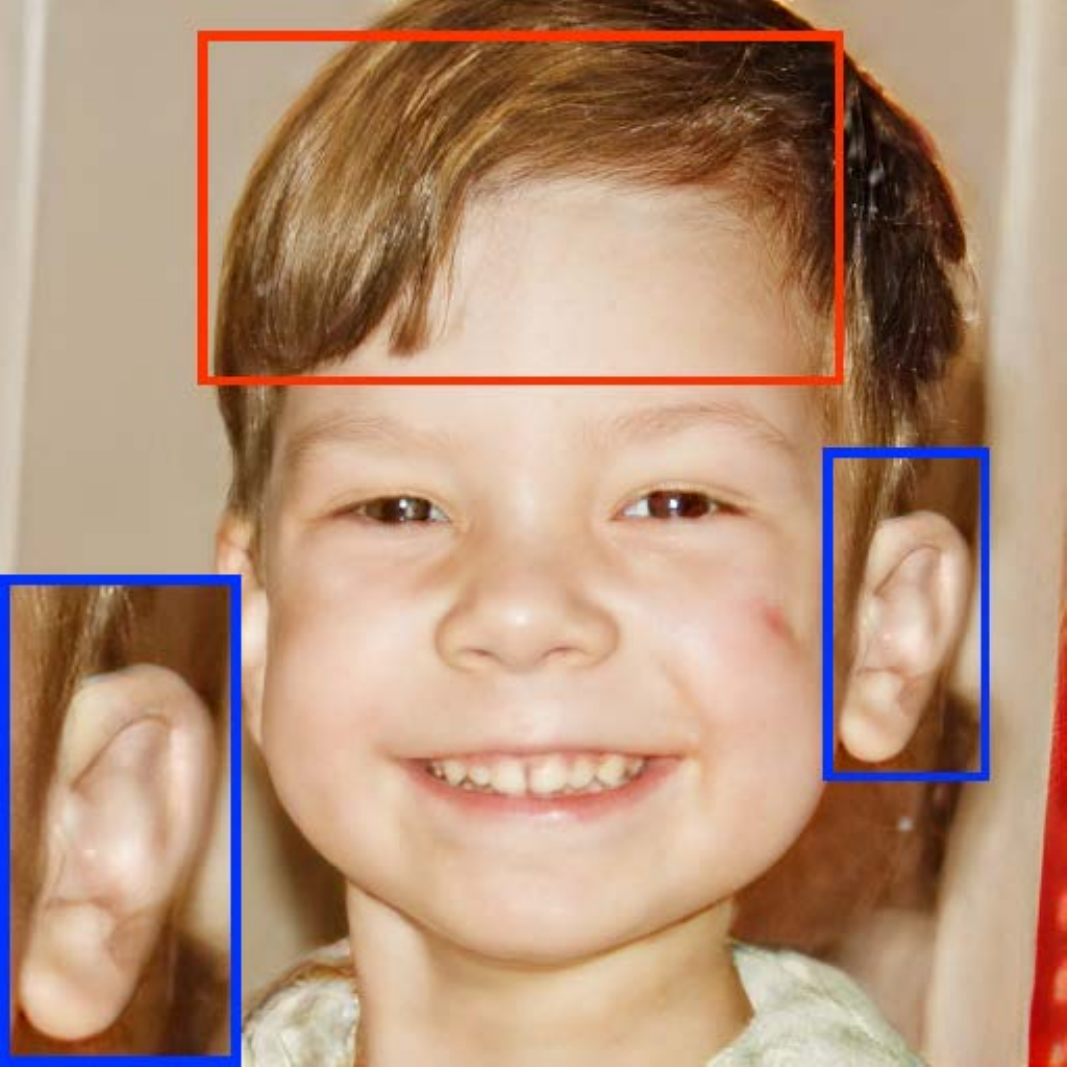}&
    \includegraphics[width=\swceleba]{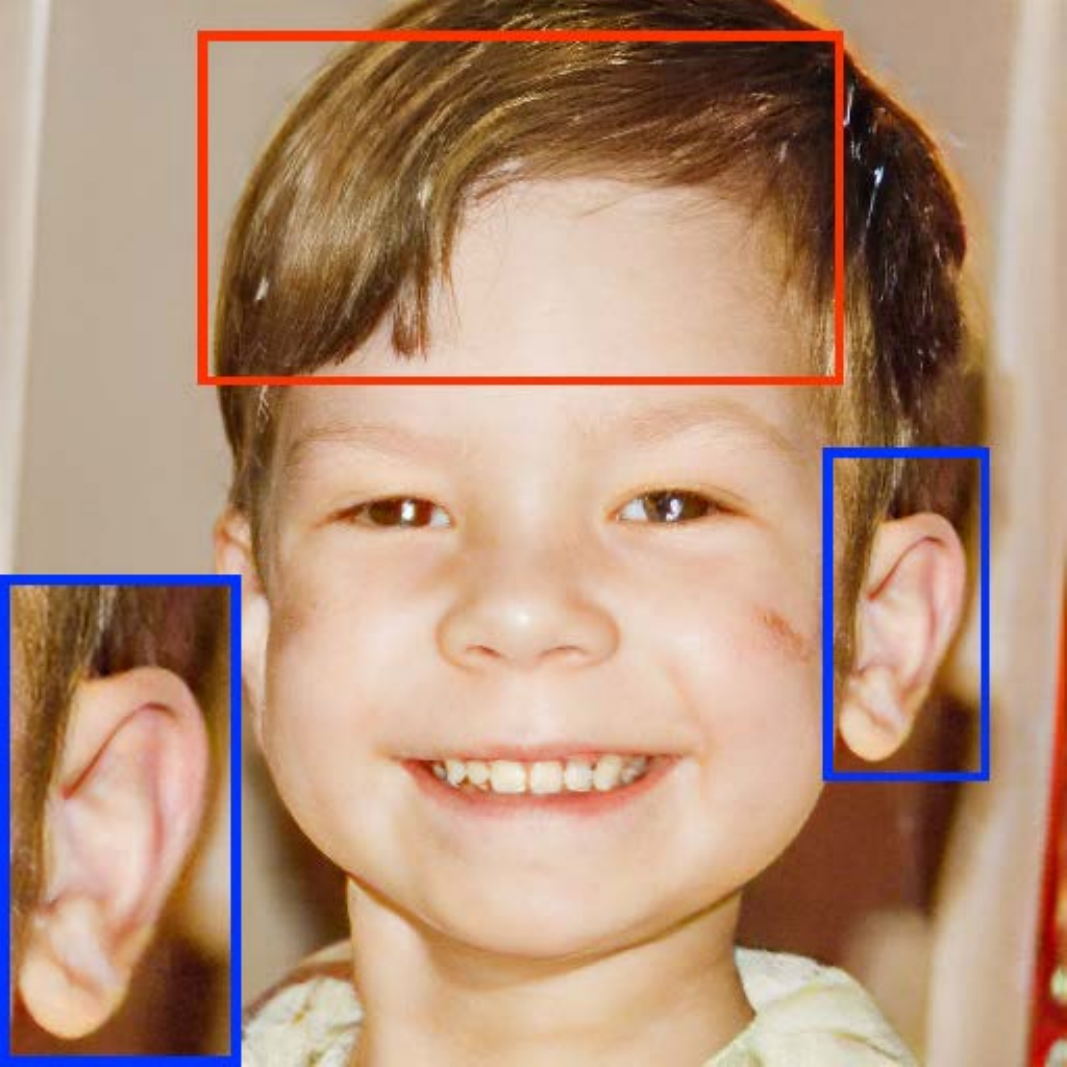} \\
    \includegraphics[width=\swceleba]{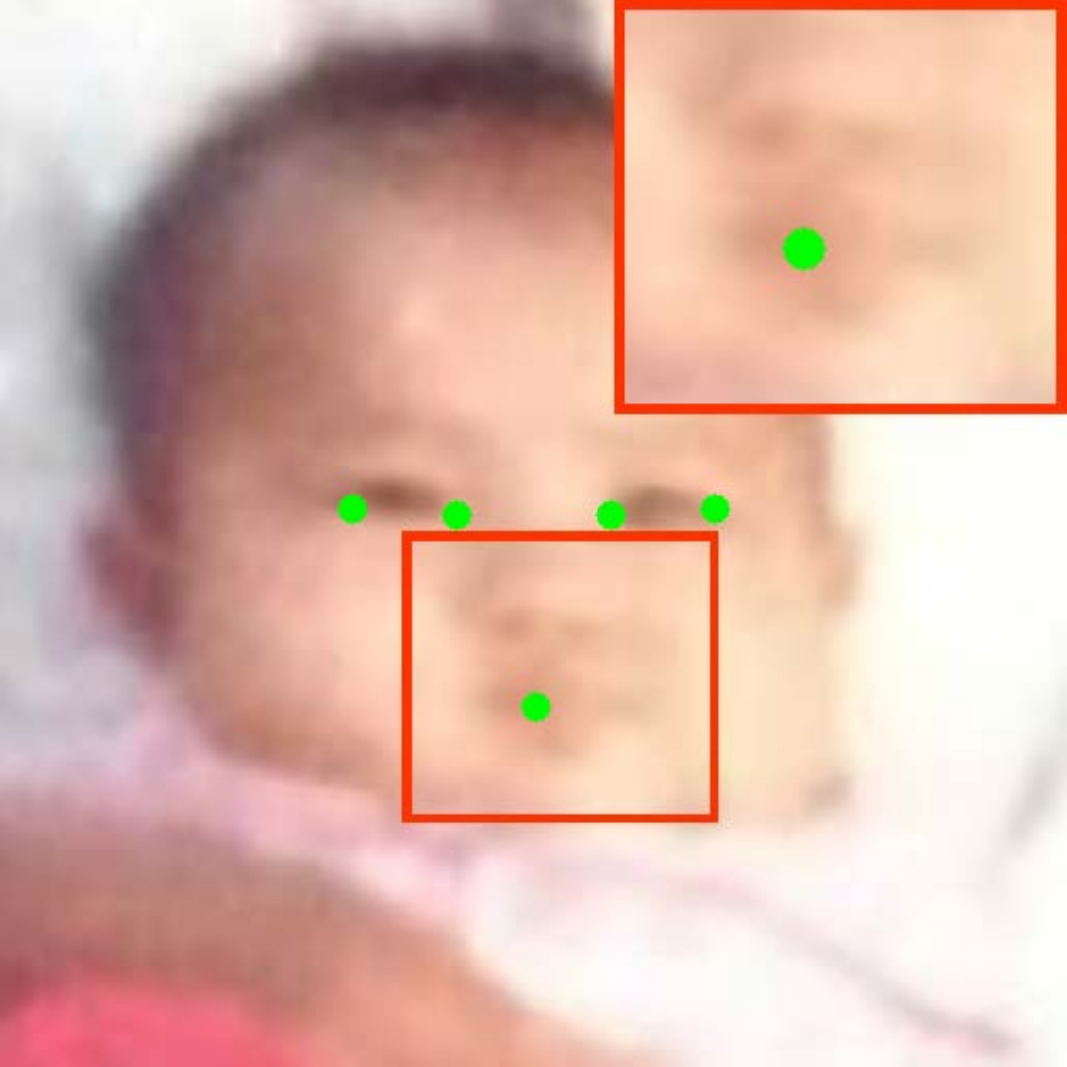}&
    \includegraphics[width=\swceleba]{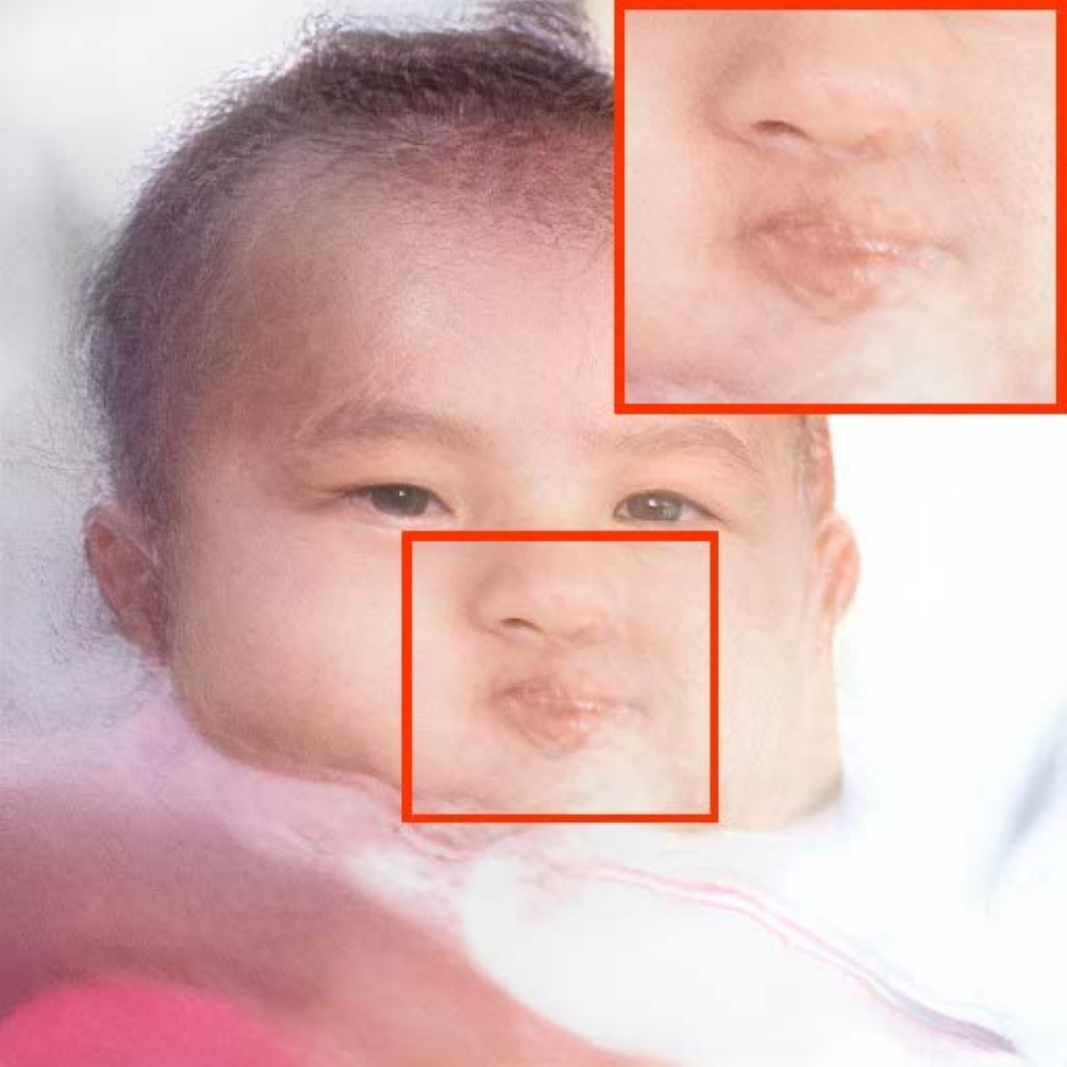}&
    \includegraphics[width=\swceleba]{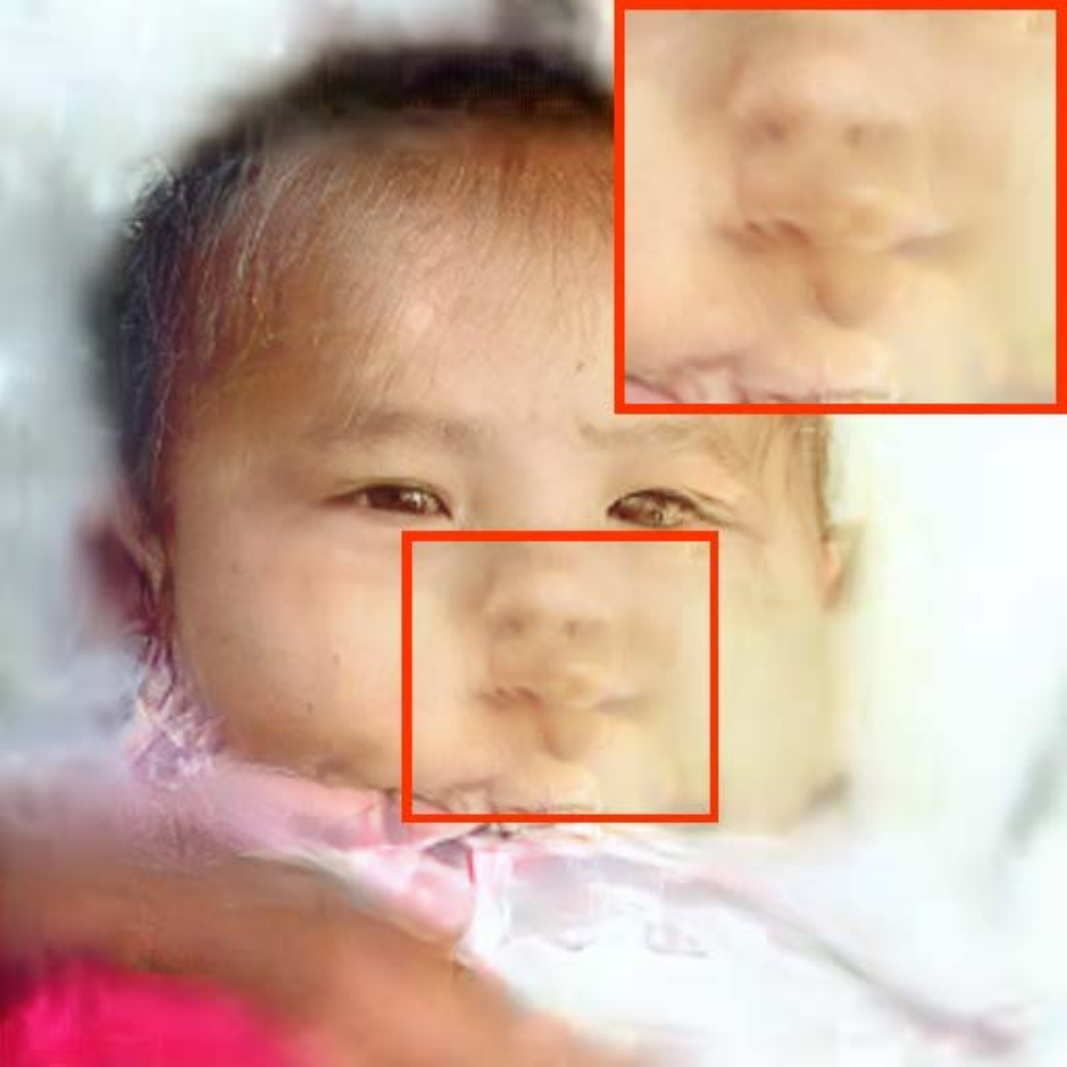}&
    \includegraphics[width=\swceleba]{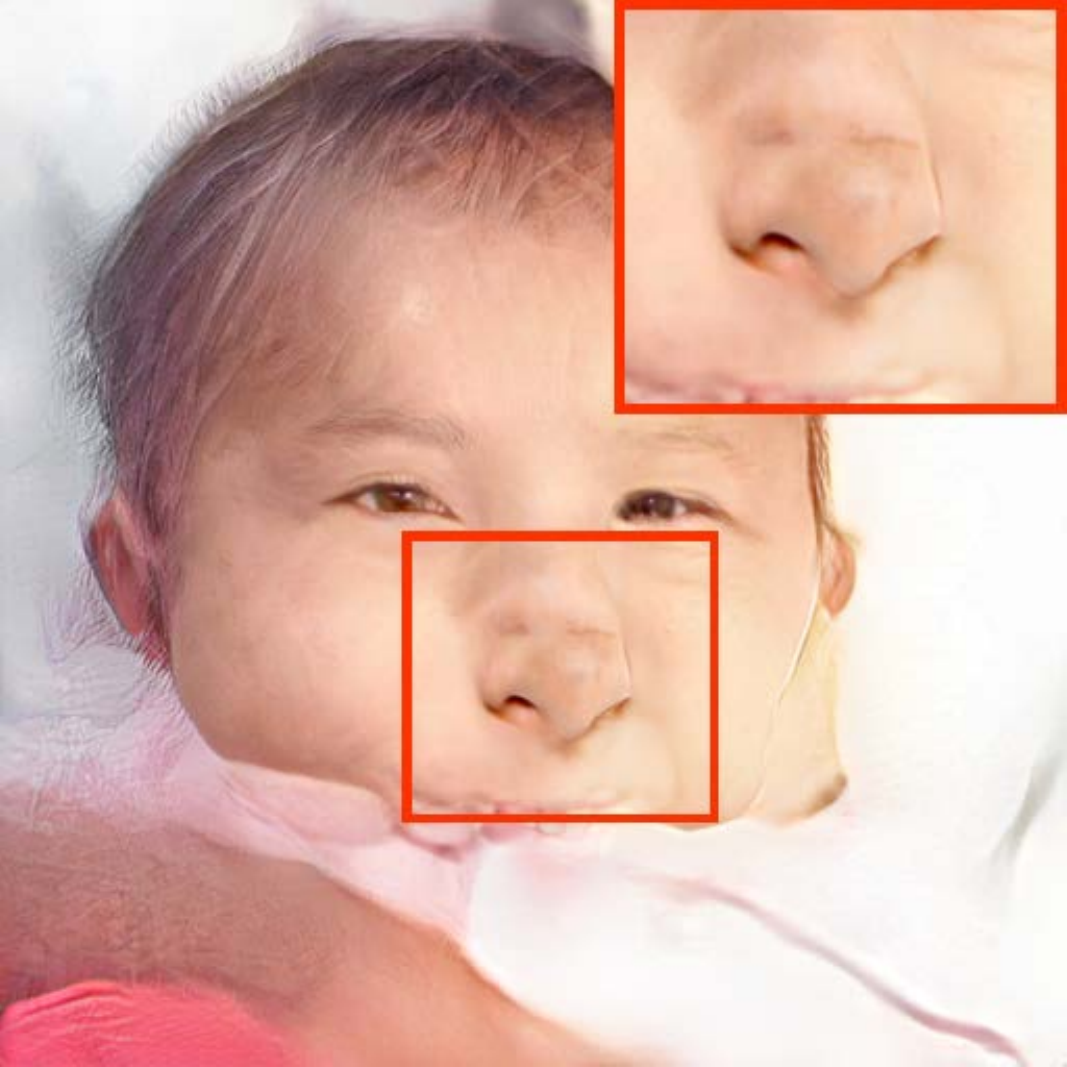}&
    \includegraphics[width=\swceleba]{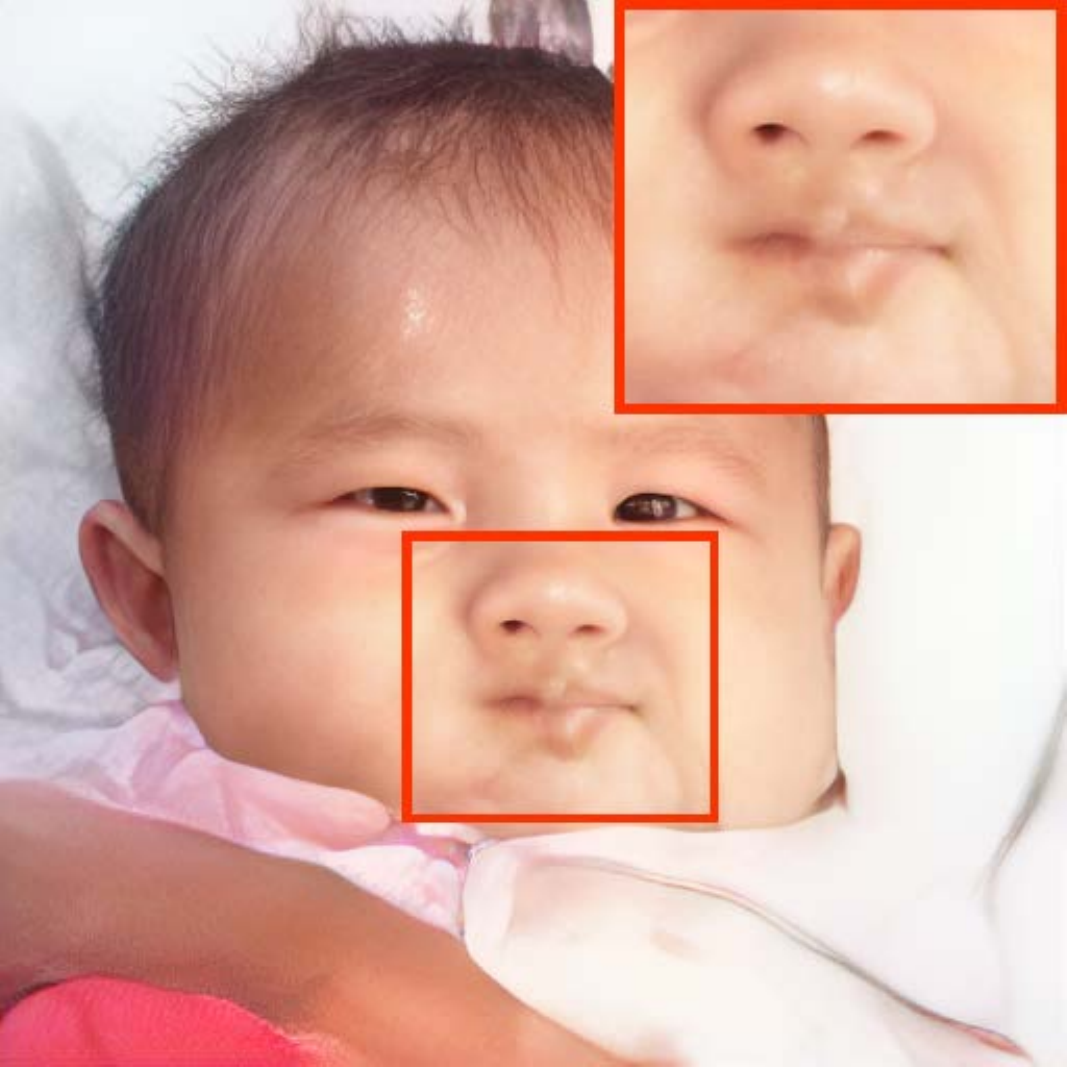}&
    \includegraphics[width=\swceleba]{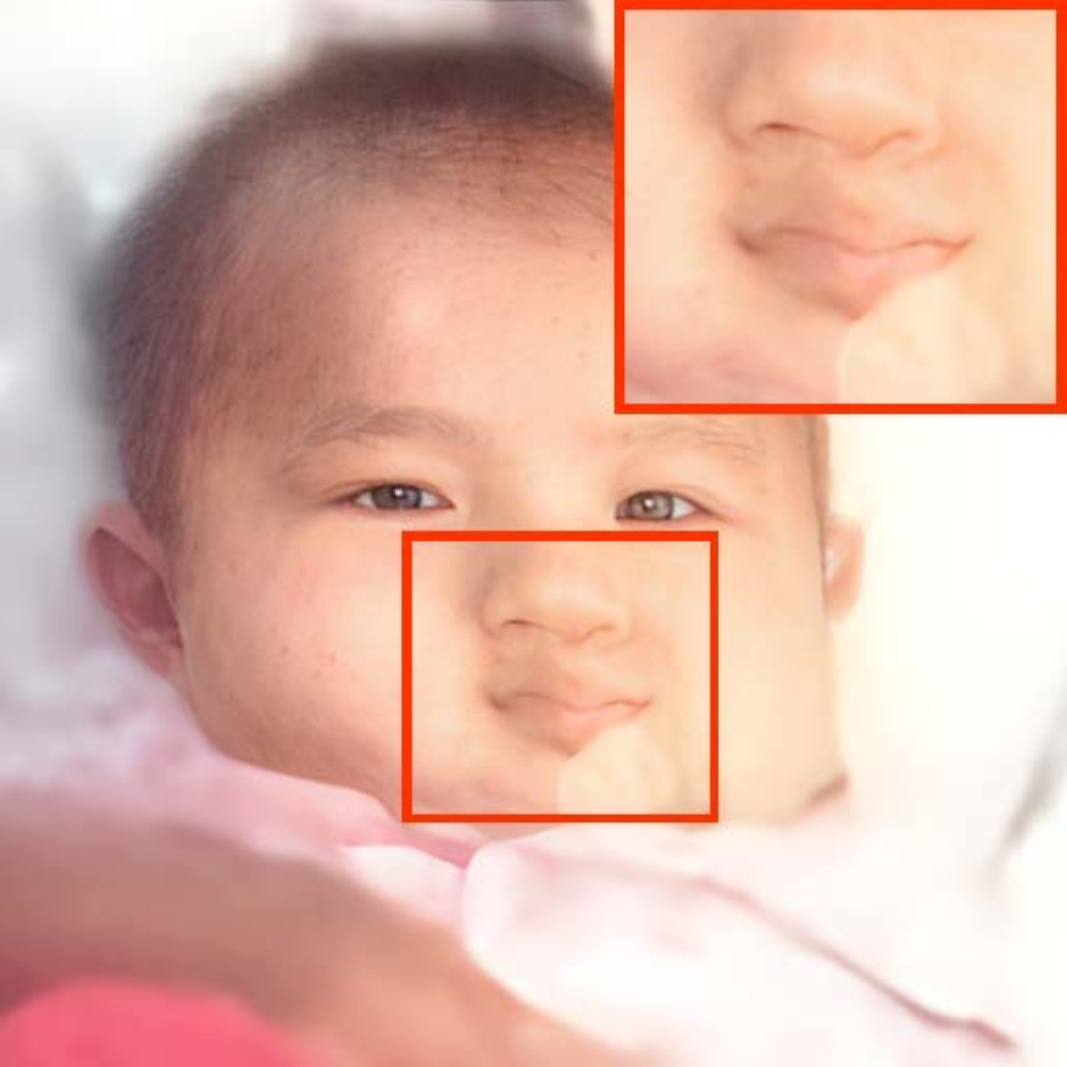}&
    \includegraphics[width=\swceleba]{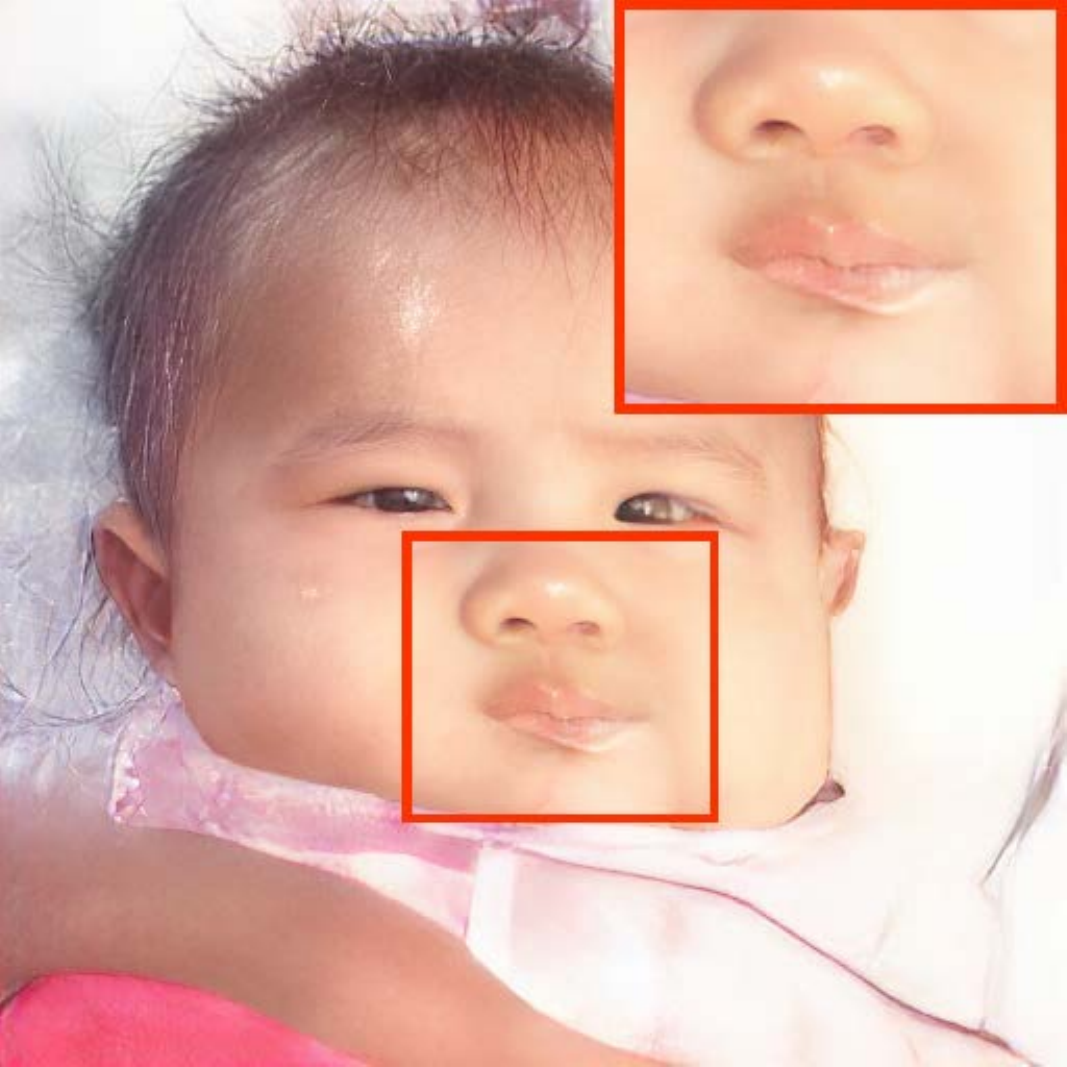}&
    \includegraphics[width=\swceleba]{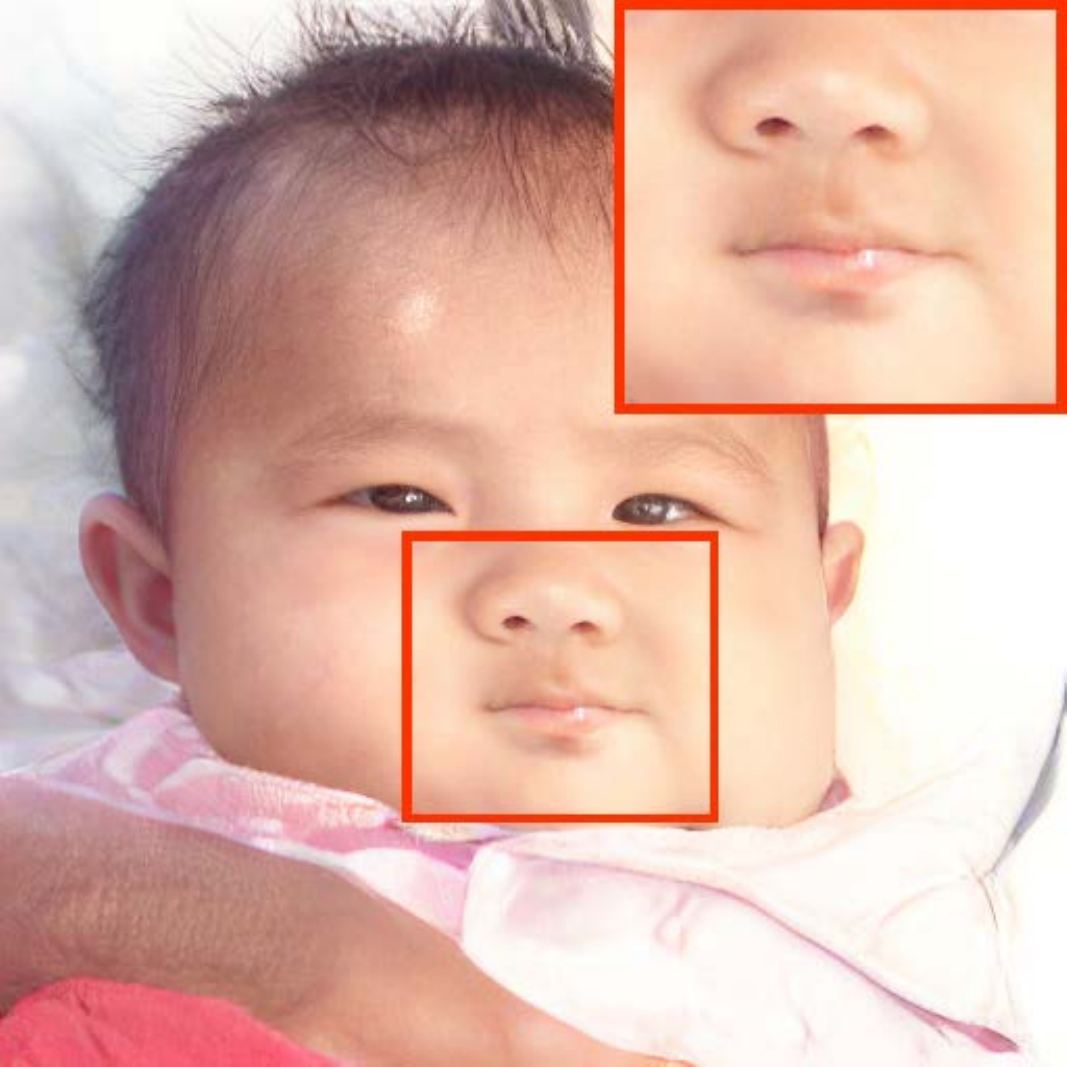} \\
   Input & DFDNet~\cite{li2020blind}  & Wan \textit{et~al.}~\cite{wan2020bringing} & PSFRGAN~\cite{chen2021progressive} & GFP-GAN~\cite{wang2021towards} & GPEN~\cite{yang2021gan} & VQFR~\cite{gu2022vqfr} & \textbf{Ours} \\
\end{tabular}
\end{center}
\caption{
\textcolor{black}{Qualitative comparison on three \textbf{real-world} datasets: \textbf{LFW-Test}~\cite{huang2008labeled}, \textbf{CelebChild-Test}~\cite{wang2021towards}, and \textbf{WebPhoto-Test}~\cite{wang2021towards} (from top to down, respectively). 
The results of our RestoreFormer++ have a more natural and complete overview and contain more details in the areas of eyes, glasses, hair, and mouth. 
The green points on the third degraded face image are the reference landmarks used for face alignment.
In this sample, its mouth is aligned to the landmark of the nose, and the existing methods, e.g., \cite{wan2020bringing, chen2021progressive}, restore the mouth with a nose-like shape.
Benefiting from the spatial shift adopted in EDM, our restored result looks more natural.
}
\textbf{Zoom in for a better view}.
}
\label{fig:real}
\end{figure*}%

\subsection{Comparison with State-of-the-art Methods}{\label{sec:sofa}}

In this subsection, we compare our RestoreFormer++ with state-of-the-art prior-based methods, including DFDNet~\cite{li2020blind} based on component dictionaries, PSFRGAN~\cite{chen2021progressive} implemented with facial parsing maps, Wan \textit{et~al.}~\cite{wan2020bringing}, PULSE~\cite{menon2020pulse}, GPEN~\cite{yang2021gan}, and GFP-GAN~\cite{wang2021towards} restored with generative priors, and VQFR~\cite{gu2022vqfr} utilized codebook.
We also compare RestoreFormer++ with our conference version, RestoreFormer.
Compared to RestoreFormer++, RestoreFormer is trained with synthetic data attained with the traditional degrading model rather than EDM and its fusion between the degraded face and priors only involves one scale.
Comparisons between these methods and our proposed method are conducted on synthetic and real-world datasets.

\subsubsection{Performance on Synthetic Dataset}
The quantitative results of the aforementioned state-of-the-art methods and our RestoreFormer++ on the synthetic dataset CelebA-Test~\cite{liu2015deep} are in TABLE~\ref{tab:celeba}. 
We can see that RestoreFormer++ performs better than other methods on FID and IDD, which means that the restored faces of RestoreFormer++ are more real and their identities are closer to the degraded faces.
Our RestoreFormer++ also achieves comparable performance in terms of PSNR, SSIM, and LIPIS, which are pixel-wise and perceptual metrics.
These metrics have been proved not that consistent with the subjective judgment of human beings~\cite{blau20182018,ledig2017photo}. We also find that the visualized results of GPEN~\cite{yang2021gan} which performs better on PSNR, SSIM, and LIPIS are over-smooth and lack details.
Visualized results are shown in Fig.~\ref{fig:celeba}.
Compared to other methods, the restored results of our RestoreFormer++ have a more natural look and contain more details, especially in the eyes, mouth, and glasses.
Besides, our method can restore a more complete face, such as the left eye in the first sample and the glasses in the second sample.
Due to severe degradations, most existing methods fail to restore the left eye and glasses, although they can \textcolor{black}{properly} restore the right eye and part of the glasses.
On the contrary, since our RestoreFormer++ can model the contextual information in the face, its restored left eye and glasses are more natural and complete by utilizing the related information in the right eye area and the clear part of the glasses.
The quantitative results in TABLE~\ref{tab:celeba} show that RestoreFormer++ attains an obvious improvement compared to the conference version, RestoreFormer, due to the participation of EDM and multi-scale mechanism.
More detailed analyses of the contributions of these components are discussed in Subsec.~\ref{subsub: multi-scale} and  Subsec.~\ref{subsub: EDM}, and more visualized results are in the supplementary materials.

\begin{table}[!t]
\renewcommand{\arraystretch}{1.3}
\caption{Quantitative comparisons on \textbf{CelebA-Test}~\cite{liu2015deep}. Our RestoreFormer++ performs better in terms of FID and IDD, which indicates the realness and fidelity of the restored results of our method. It also gets comparable results on PSNR, SSIM, and LPIPS.}
\label{tab:celeba}
\centering
  \begin{tabular}{c|c|c|c|c|c} 
    \hline
    Methods & ~~~FID$\downarrow$~~~ & ~~~PSNR$\uparrow$~~~ & ~~~SSIM$\uparrow$~~~ & ~~~LPIPS$\downarrow$~~~ & ~~~IDD$\downarrow$~~~ \\
    \hline
    \hline
    Input & 132.05 & \textcolor{blue}{24.91} & 0.6637 & 0.4986 & 0.9306 \\
    \hline
    DFDNet~\cite{li2020blind} & 50.88 & 24.09 & 0.6107 & 0.4516 & 0.7700  \\
    Wan \textit{et~al.}~\cite{wan2020bringing} & 67.13 & 23.01 & 0.6174 & 0.4789 & 0.8058 \\
    PSFRGAN~\cite{chen2021progressive} & 40.69 & 24.30 & 0.6273 & 0.4220 & 0.7284 \\
    PULSE~\cite{menon2020pulse} &  84.03 & 20.73 & 0.6151 & 0.4745 & 1.2267 \\
    GPEN~\cite{yang2021gan} & 48.97 & \textcolor{red}{25.44} & \textcolor{red}{0.6965} &  \textcolor{blue}{0.3562} & 0.6434 \\
    GFP-GAN\cite{wang2021towards} &40.87 & 24.39 & \textcolor{blue}{0.6671} & 0.3575 & 0.6127 \\
    VQFR~\cite{gu2022vqfr} & \textcolor{blue}{38.51} & 23.82 & 0.6379 & \textcolor{red}{0.3544} & 0.6354 \\
    \hline
    \textbf{RestoreFormer} & 39.90 & 24.19 & 0.6232 & 0.3716 &  \textcolor{blue}{0.5677} \\
    \textbf{RestoreFormer++} & \textcolor{red}{38.41} & 24.40 & 0.6339 & 0.3619 &  \textcolor{red}{0.5375} \\
    \hline
    GT & 41.66 & $\infty$ & 1 & 0 & 0 \\
    \hline
  \end{tabular}
\end{table}

\subsubsection{Performance on Real-world Datasets }
The quantitative and qualitative results of our Restoreformer++ and the compared methods on three real-world datasets are in TABLE~\ref{tab:real_fid} and Fig.~\ref{fig:real}, respectively.
According to TABLE~\ref{tab:real_fid}, RestoreFormer++ performs superiorly on FID compared to other methods.
The qualitative results in Fig.~\ref{fig:real} also reveal that although most of the current methods can attain clear faces from the corrupted face images with slight degradations (the first two samples), RestoreFormer++ attains more details on the crucial areas, such as the eyes with glasses, hair, and ear. That mainly benefits from the contextual information in the face and our learned reconstruction-oriented high-quality dictionary.
Besides, since our RestoreFormer++ is further enhanced with EDM, it can remove the haze covered on the face image and avoid \textcolor{black}{restoration artifacts} caused by misalignment, thus attaining more natural and pleasant results.
For example, after face alignment, the mouth of the last sample in Fig.~\ref{fig:real} is aligned to the reference landmark of the nose, which leads to the restored mouth of Wan~\textit{et~al.}~\cite{wan2020bringing} and PSFRGAN~\cite{chen2021progressive} is nose-like.
Although the restored results of other \textcolor{black}{existing} methods look better, they still look weird.
With EDM, the restored result of RestoreFormer++ looks more natural.

In addition, as shown in TABLE~\ref{tab:real_fid}, in the real-world datasets, the performance of RestoreFormer++ is better or comparable to our conference version, RestoreFormer.
RestoreFormer is slightly superior to RestoreFormer++ on LFW-Test~\cite{huang2008labeled} since the degree of the degradation in this dataset is generally slight, and the delicate design in RestoreFormer is enough for attaining high-quality restored results.
However, since the degradation in CelebChild-Test~\cite{wang2021towards} and WebPhoto-Test~\cite{wang2021towards} are more severe, RestoreFormer++, with additional EDM and multi-scale mechanism, can handle these two datasets better compared to RestoreFormer.
More visualizations are in the supplementary materials.

Besides, a user study is adopted to collect the subjective judgment of human beings on the real-world dataset WebPhto-Test~\cite{wang2021towards}.
Specifically, we randomly select 100 samples from the real-world dataset and conduct pair comparisons between our conference version RestoreFormer and three other methods: DFDNet~\cite{li2020blind}, PSFRGAN~\cite{chen2021progressive}, and GFP-GAN\cite{wang2021towards}.
Subjective comparisons between RestoreFormer++, RestoreFormer, and VQFR~\cite{gu2022vqfr} are also conducted.
We invite 100 volunteers to make their subjective selection on these pair comparisons.
The statistic results are in Tab~\ref{tab:real_us}. It shows that a high percentage of volunteers vote for the results of our RestoreFormer and RestoreFormer++ as the more natural and pleasant restored results compared to other methods, and the restored results of RestoreFormer++ are better than those of RestoreFormer.

\begin{table}[!t]
\renewcommand{\arraystretch}{1.3}
\caption{Quantitative comparisons on three \textbf{real-world dataset} in terms of FID. RestoreFormer++ performs better.}
\label{tab:real_fid}
\centering
  \begin{tabular}{c|c|c|c} 
    \hline
    Methods & ~~\textbf{LFW-Test}~~ & ~\textbf{CelebChild-Test}~ & ~~\textbf{WebPhoto-Test}~~ \\   
    \hline
    \hline
    Input & 126.12 &  144.36 &  170.46  \\
    \hline
    DFDNet~\cite{li2020blind} & 72.87 & 110.85 & 100.45  \\
    PSFRGAN~\cite{chen2021progressive} & 53.17 & 105.65 & 83.50 \\
    Wan \textit{et~al.}~\cite{wan2020bringing} & 71.24 & 115.15 & 99.91 \\
    PULSE~\cite{menon2020pulse} & 66.08 & 104.06 & 86.39 \\
    GPEN~\cite{yang2021gan} & 55.52 & 107.57 & 86.07 \\
    GFP-GAN~\cite{wang2021towards} & 50.30 & 111.78 & 87.82 \\
    VQFR~\cite{gu2022vqfr} & 50.22 & \textcolor{blue}{103.96} & \textcolor{blue}{74.22} \\
    \hline
    \textbf{RestoreFormer} & \textcolor{red}{48.11} & 104.01 & 75.49 \\
    \textbf{RestoreFormer++} & \textcolor{blue}{48.48} & \textcolor{red}{102.66} & \textcolor{red}{74.21} \\
    \hline
  \end{tabular}
\end{table}

\begin{table}[!t]
\renewcommand{\arraystretch}{1.3}
\caption{User study results on \textbf{WebPhoto-Test}~\cite{wang2021towards}.
  For ``a/b", a is the percentage where our RestoreFormer or RestoreFomer++ is better than the compared method, and b is the percentage where the compared method is considered better than our RestoreFormer or RestoreFomer++.}
\label{tab:real_us}
\centering
  \begin{tabular}{c|c|c|c} 
    \hline
     Methods & DFDNet~\cite{li2020blind} & PSFRGAN\cite{chen2021progressive} & GFP-GAN~\cite{wang2021towards}  \\
    \hline
    RestoreFormer & ~~\textbf{89.60\%}/10.40\%~~ & ~~\textbf{68.81\%}/31.19\%~~ &  ~~\textbf{79.21\%}/20.79\%~~ \\
    \hline
    \hline
    Methods & VQFR~\cite{gu2022vqfr} & \multicolumn{2}{c}{RestoreFormer} \\
    \hline
    RestoreFormer++ & \textbf{67.82\%}/32.18\% & \multicolumn{2}{c}{\textbf{66.91\%}/33.19\%} \\
    \hline
  \end{tabular}
\end{table}

\renewcommand{\tabcolsep}{.5pt}
\begin{figure*}
\hsize=\textwidth
\vspace{-0.3cm}
\begin{center}
\begin{tabular}{cccccccc}
  \includegraphics[width=\swceleba]{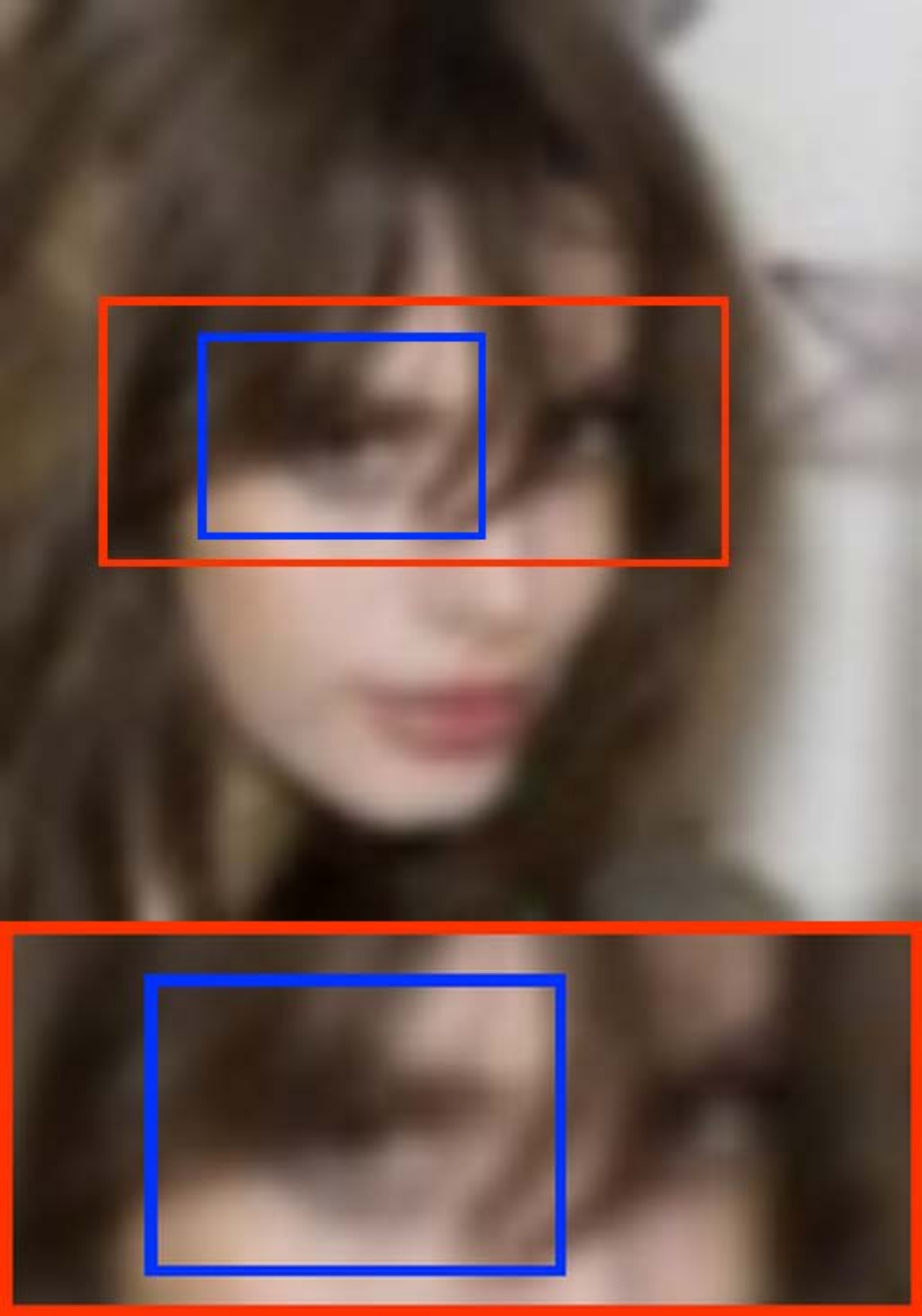}&
  \includegraphics[width=\swceleba]{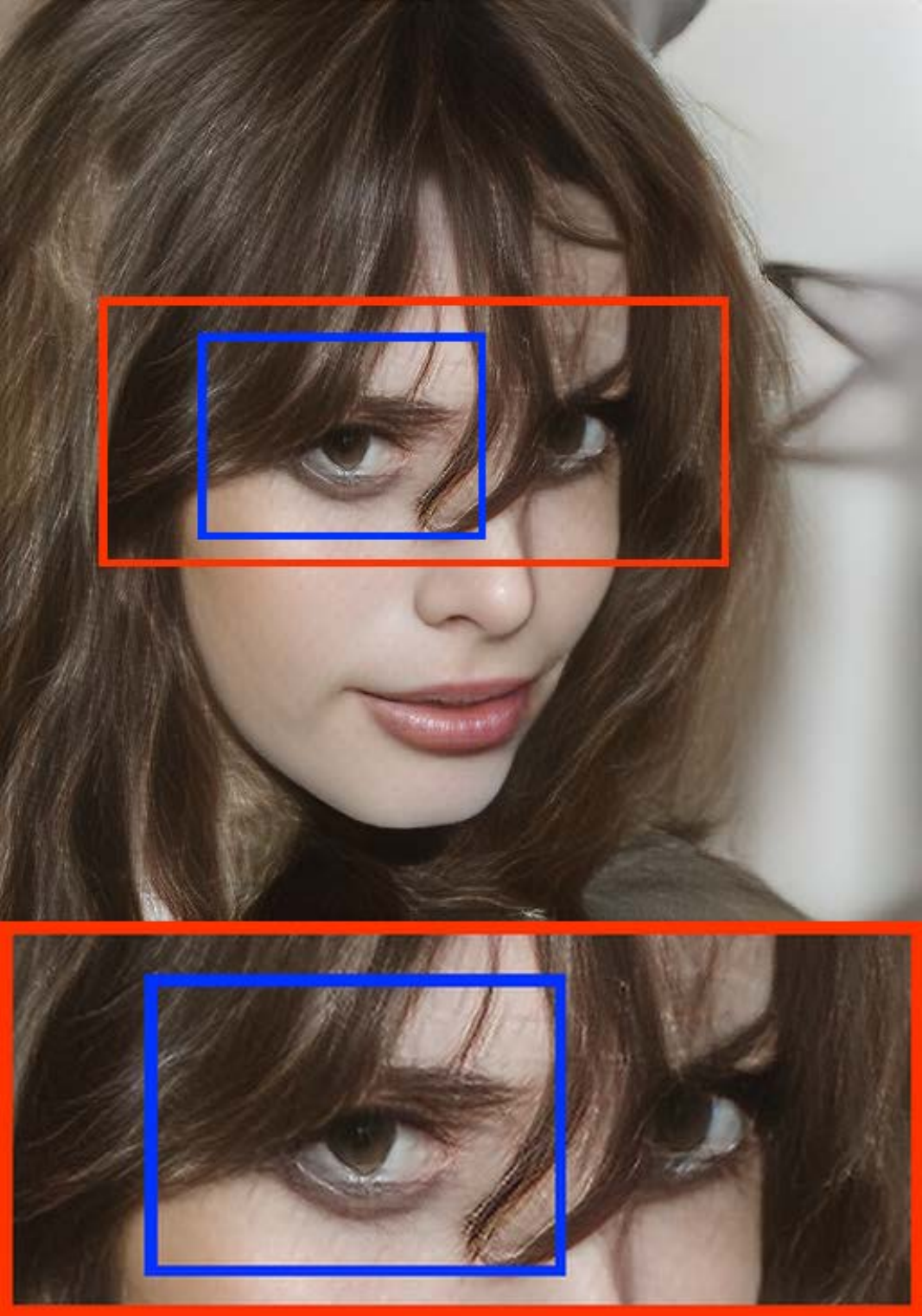}&
  \includegraphics[width=\swceleba]{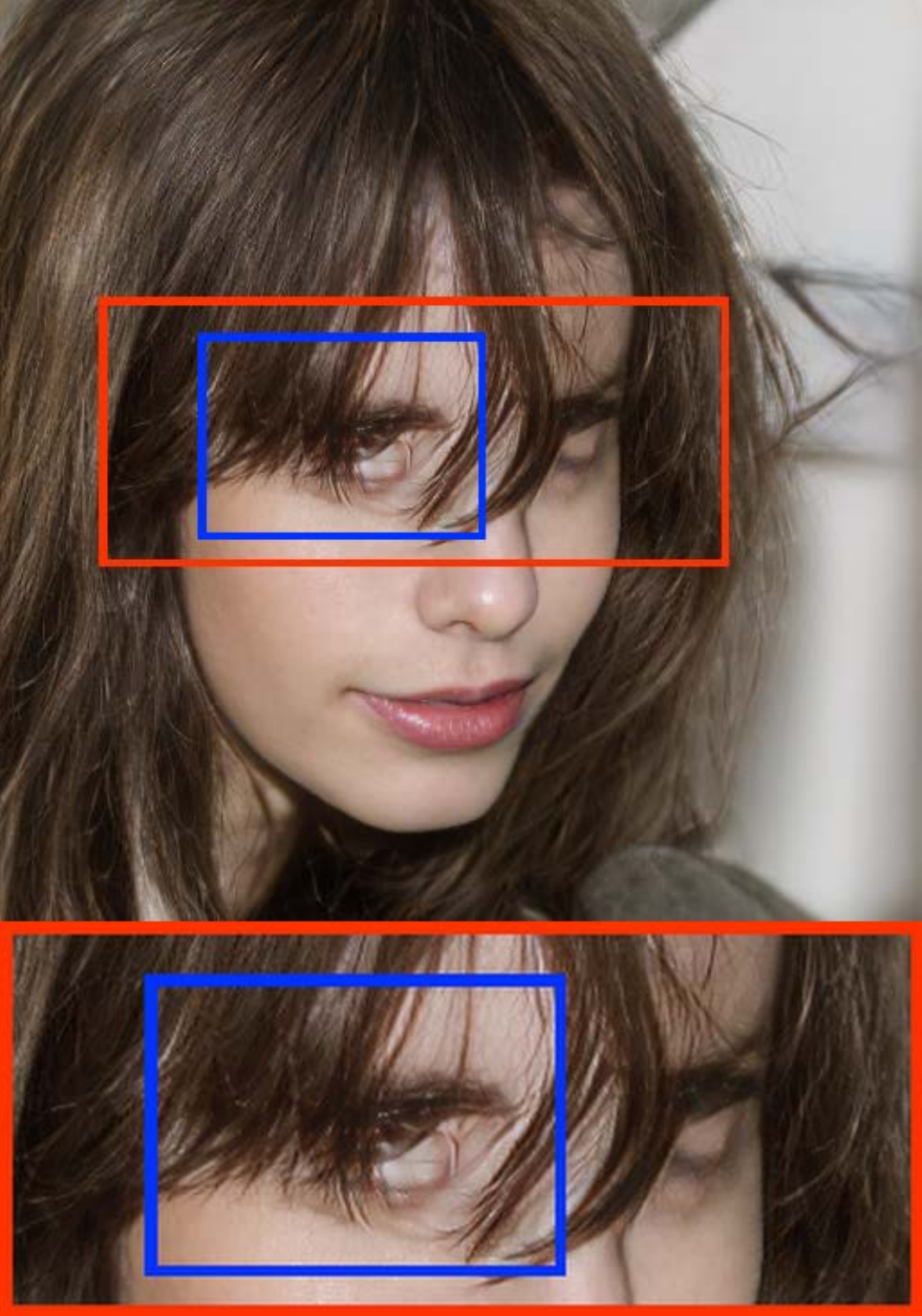}&
  \includegraphics[width=\swceleba]{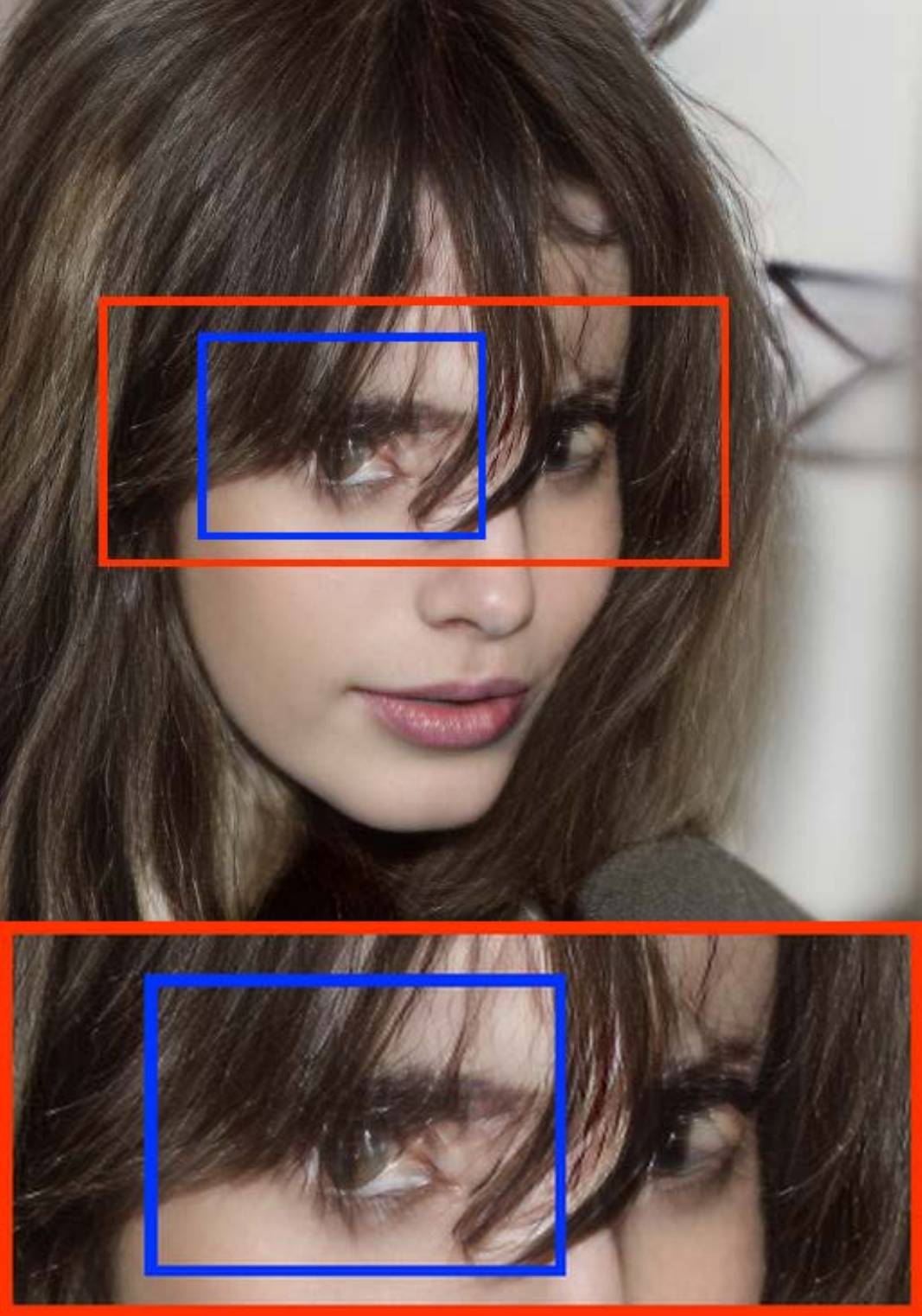}&
  \includegraphics[width=\swceleba]{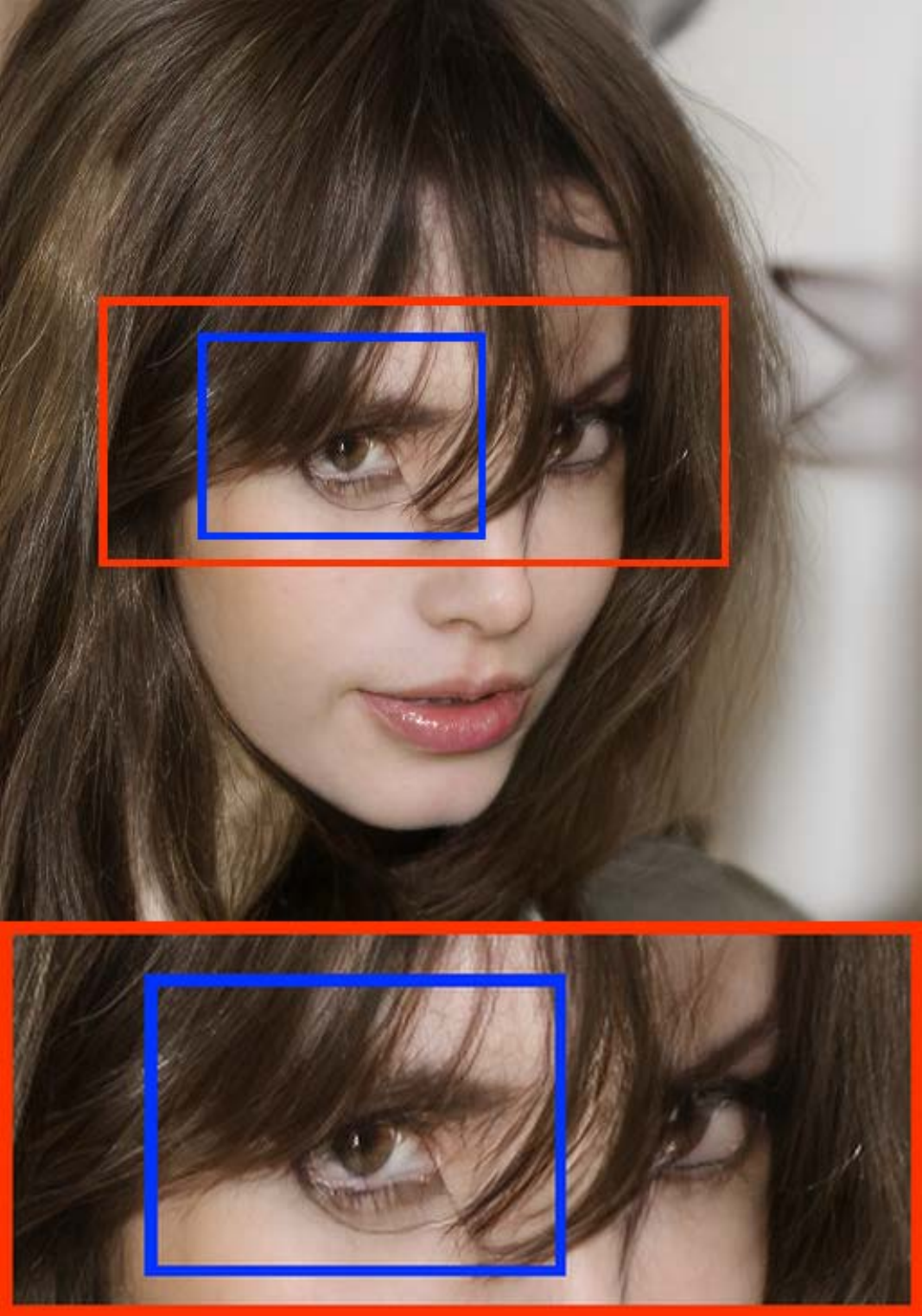}&
  \includegraphics[width=\swceleba]{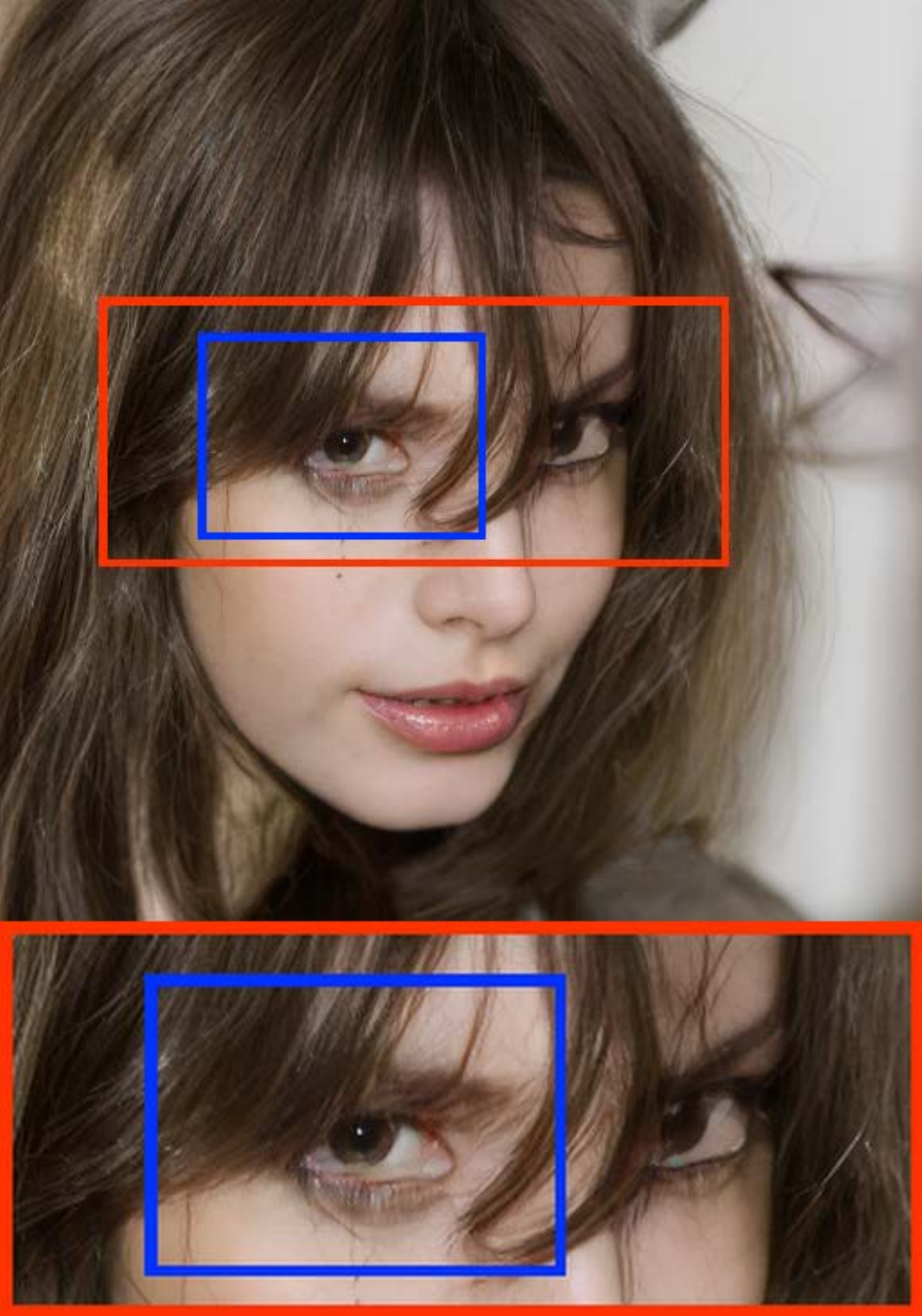}&
  \includegraphics[width=\swceleba]{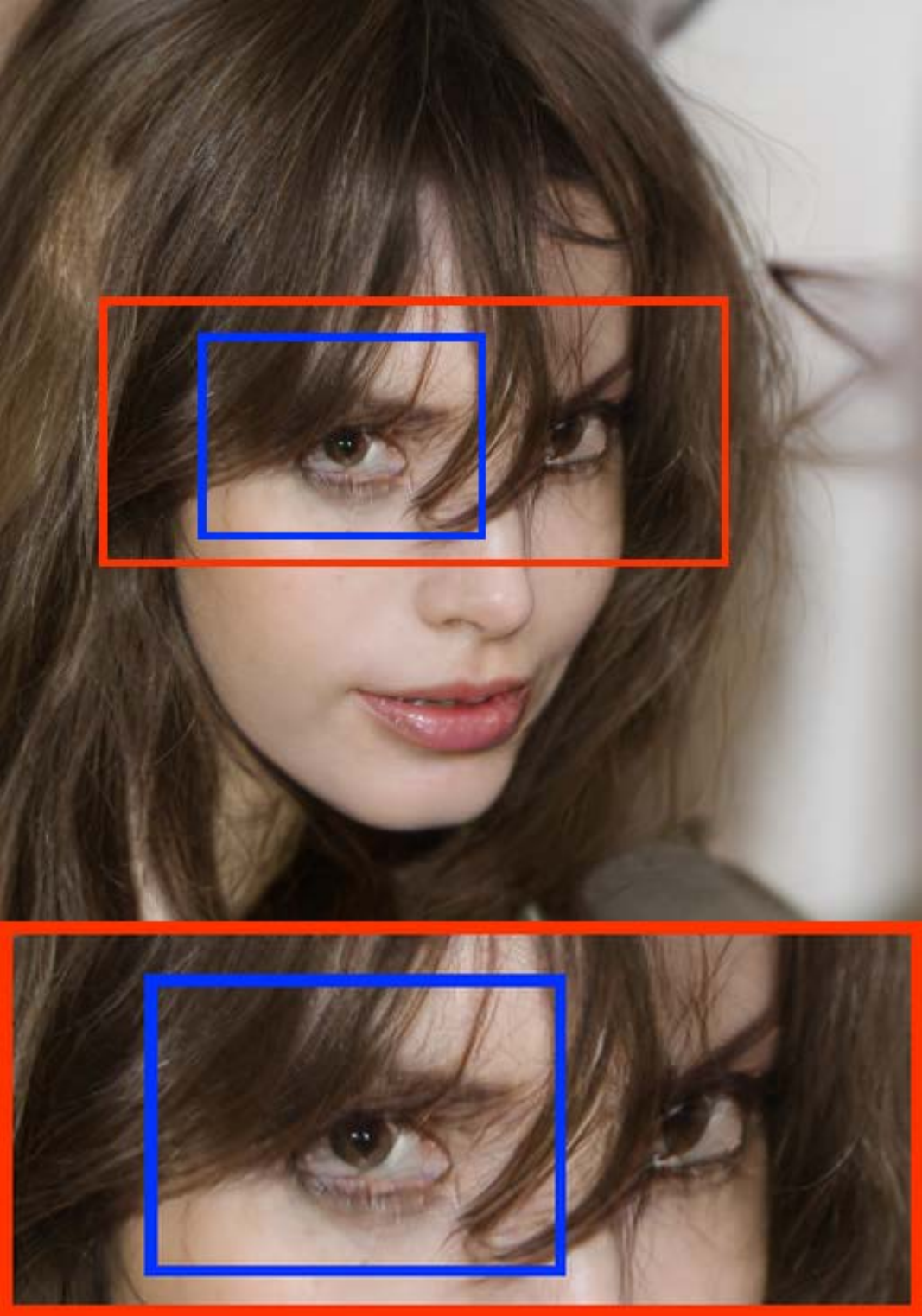}&
  \includegraphics[width=\swceleba]{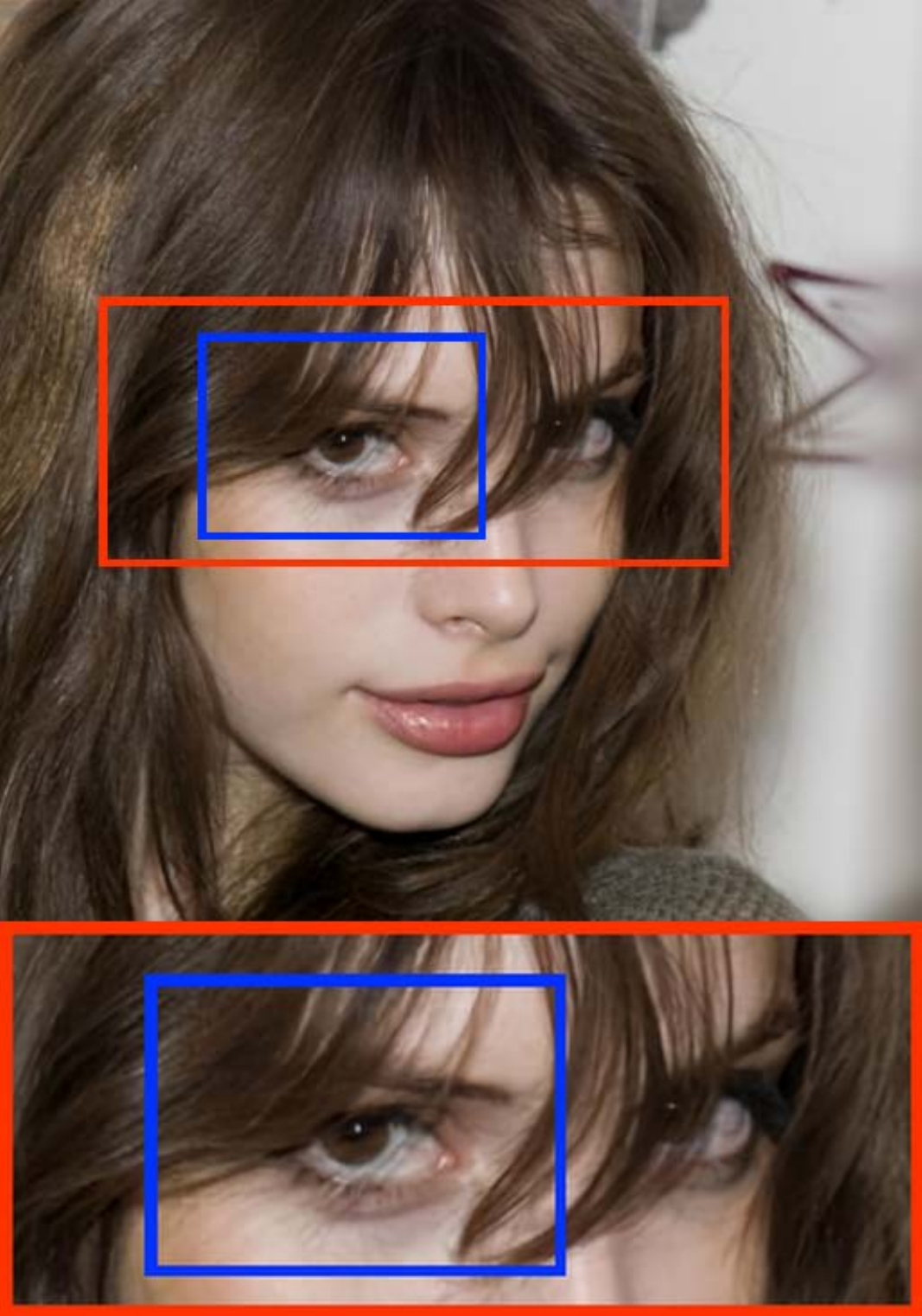} \\
   (a) Input & (b) exp2 & (c) exp3 & (d) exp4 & (e) exp6 & (f) exp7 & (g) \scriptsize RestoreFormer++ & (h) GT \\
PSNR: 26.21 & PSNR: 25.29 & PSNR: 23.83 & PSNR: 23.98 & PSNR: 25.40 & PSNR: 25.97 & PSNR: 26.56 & PSNR: $\infty$ \\
IDD: 1.0689 & IDD: 0.5322 & IDD: 0.7981 & IDD: 0.6239 & IDD: 0.5220 & IDD: 0.4560 & IDD: 0.4689 & IDD: 0 
   \end{tabular}
   \begin{tabular}{cccccc}
  \includegraphics[width=\swceleba]{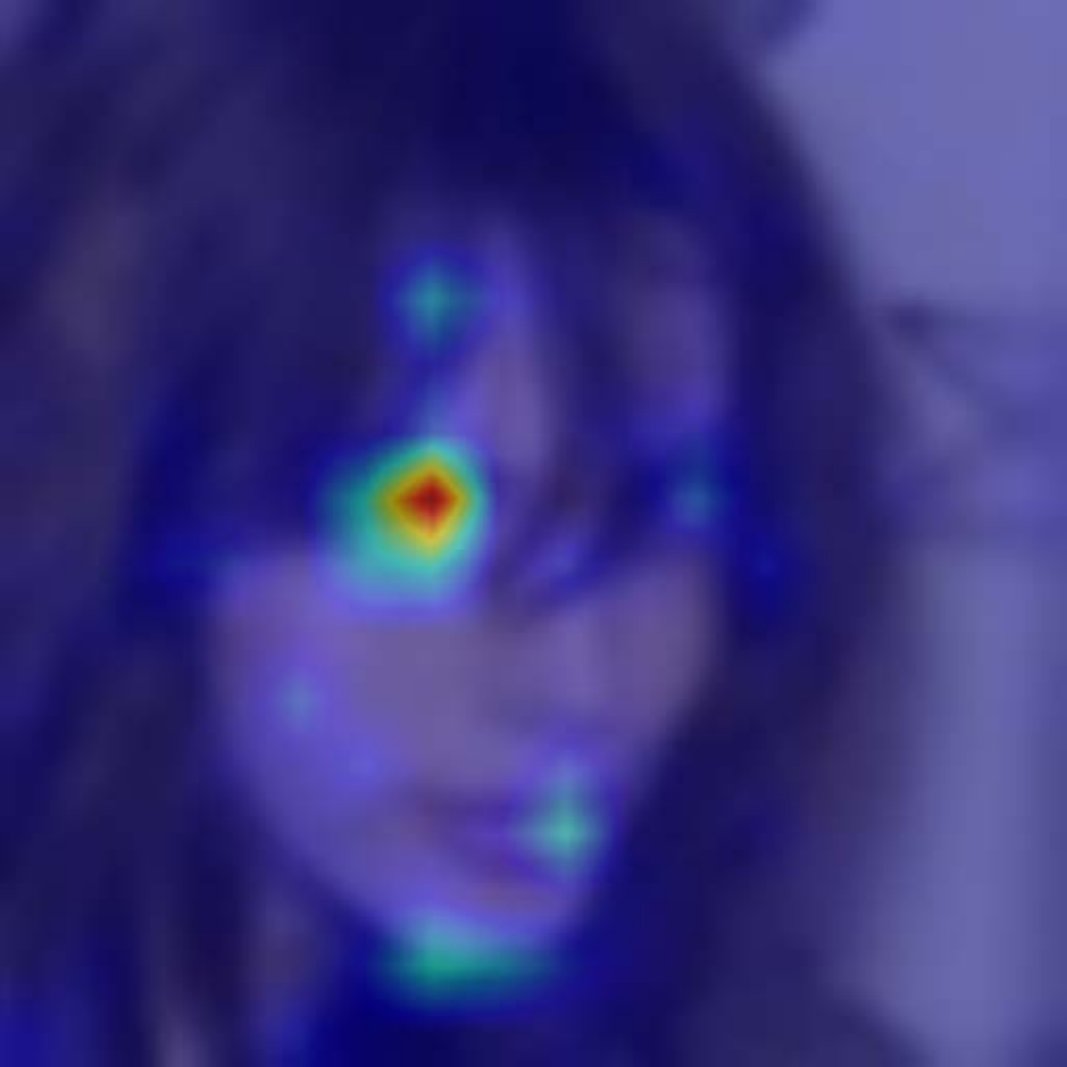}&
  \includegraphics[width=\swceleba]{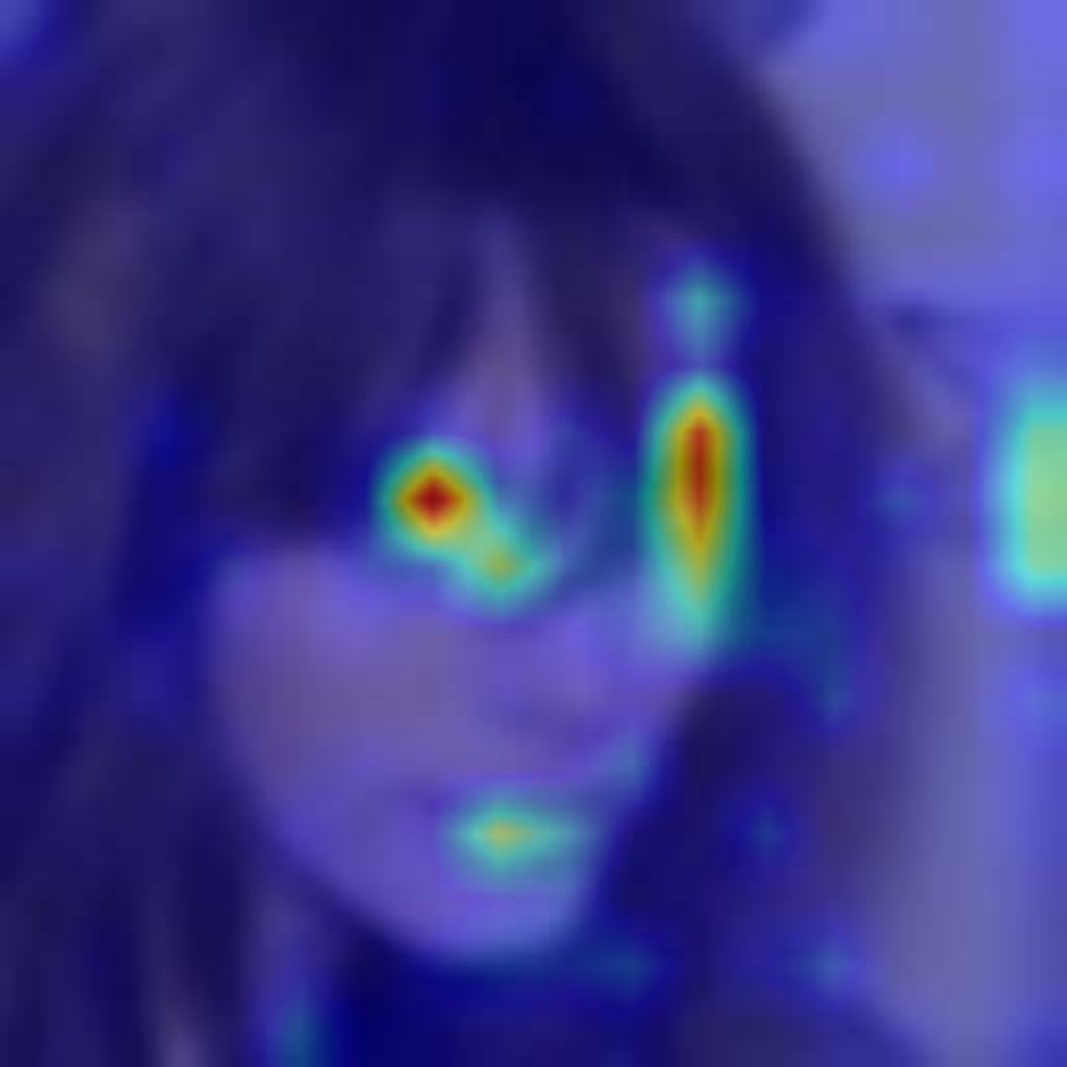}&
  \includegraphics[width=\swceleba]{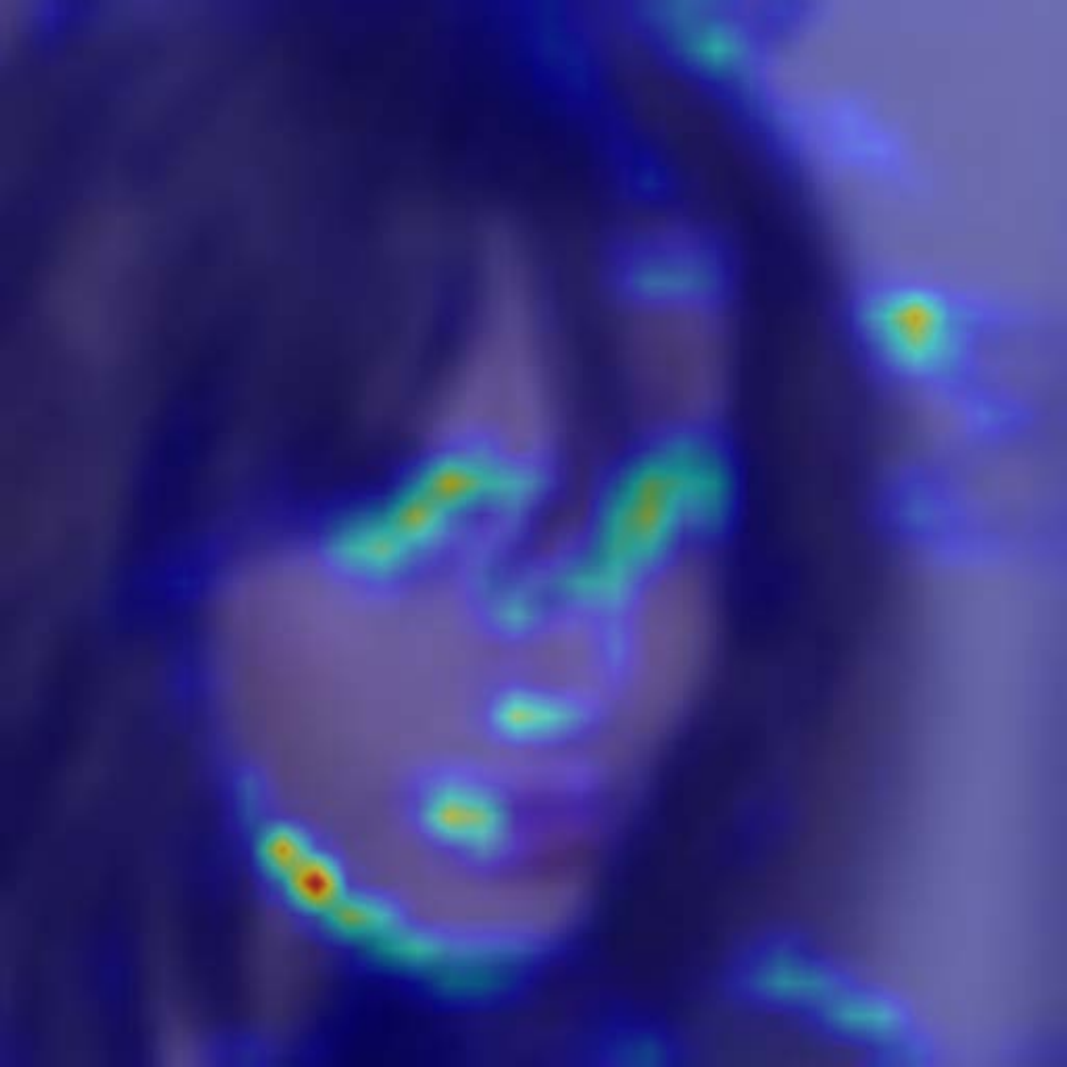}&
  \includegraphics[width=\swceleba]{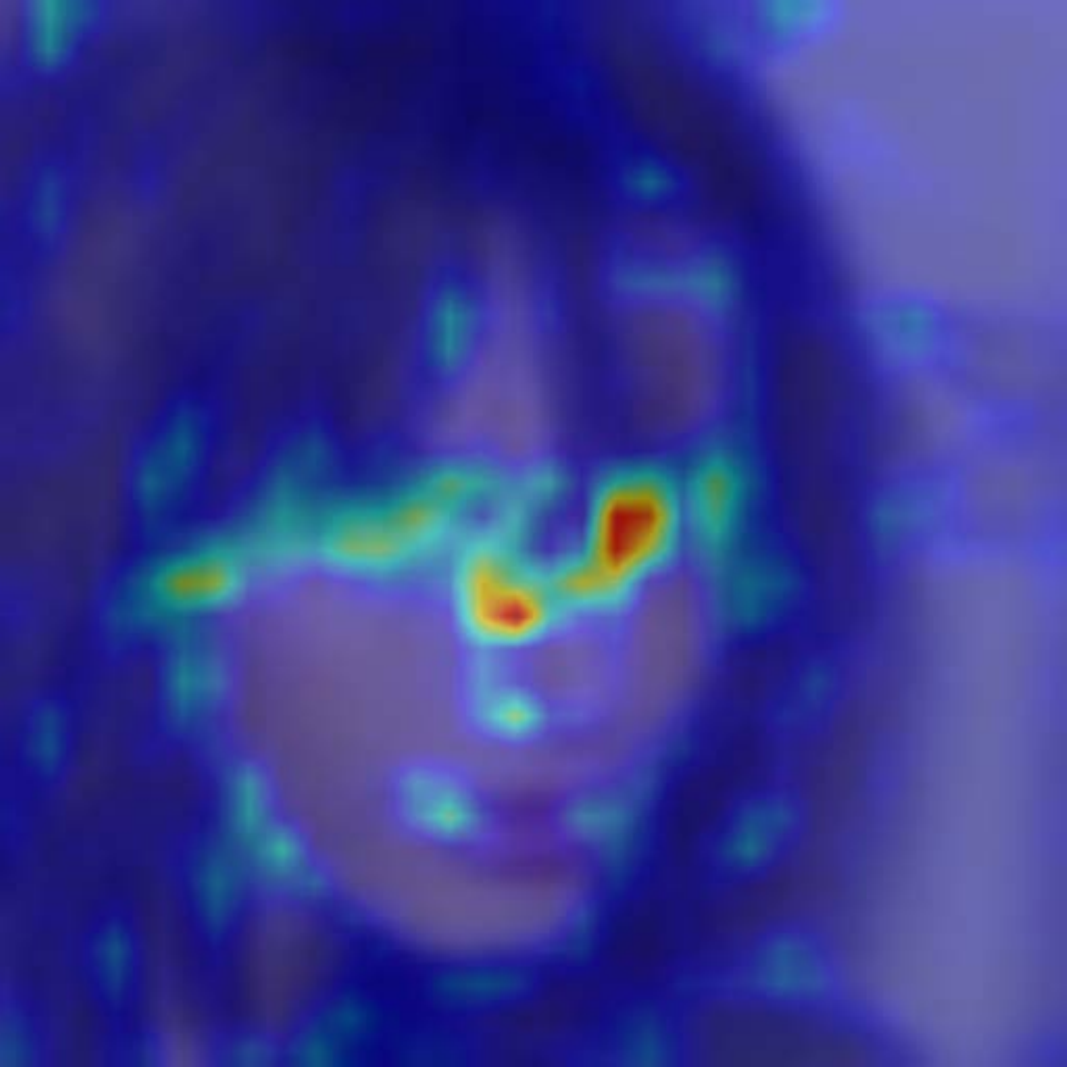}&
  \includegraphics[width=\swceleba]{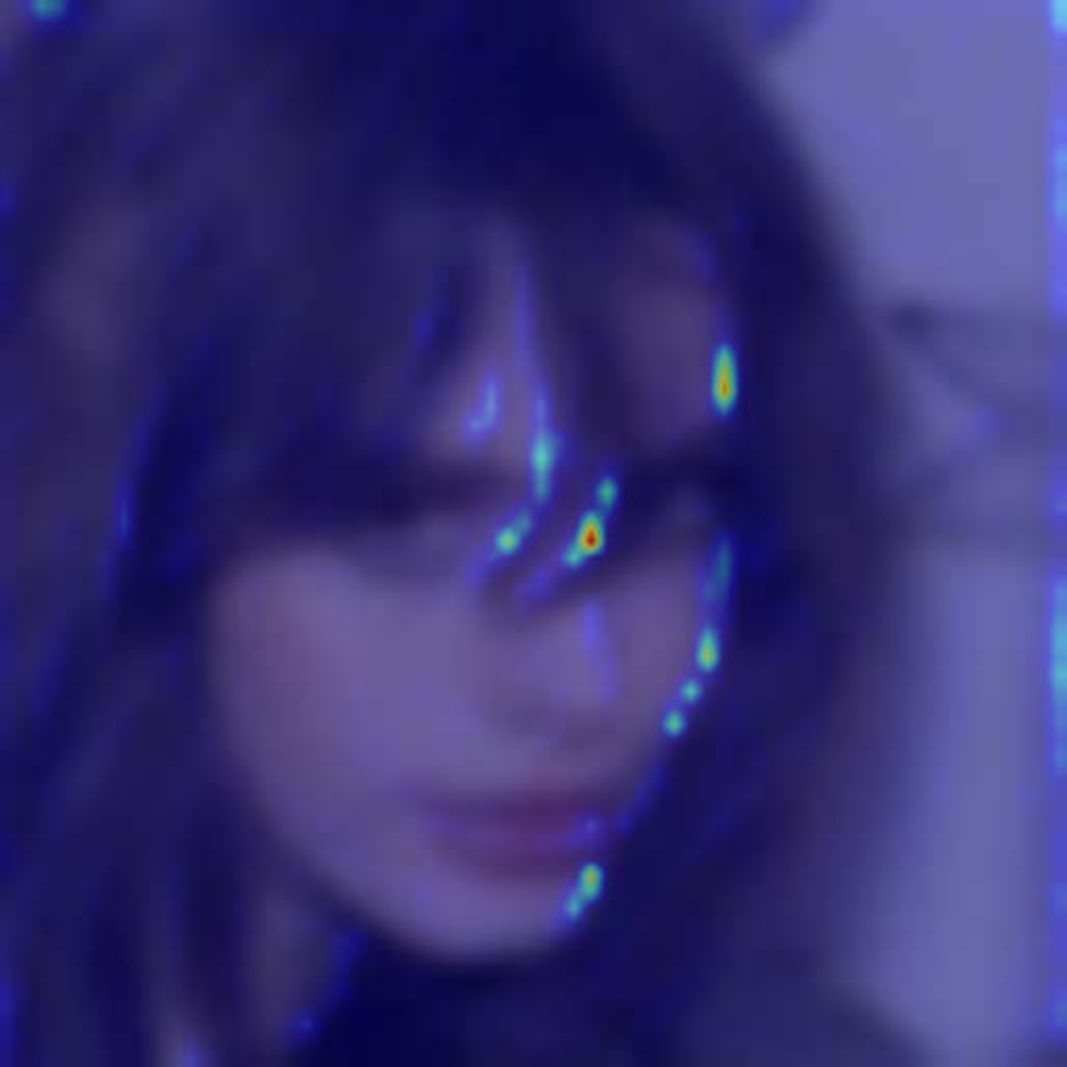} &
  \includegraphics[width=\swceleba]{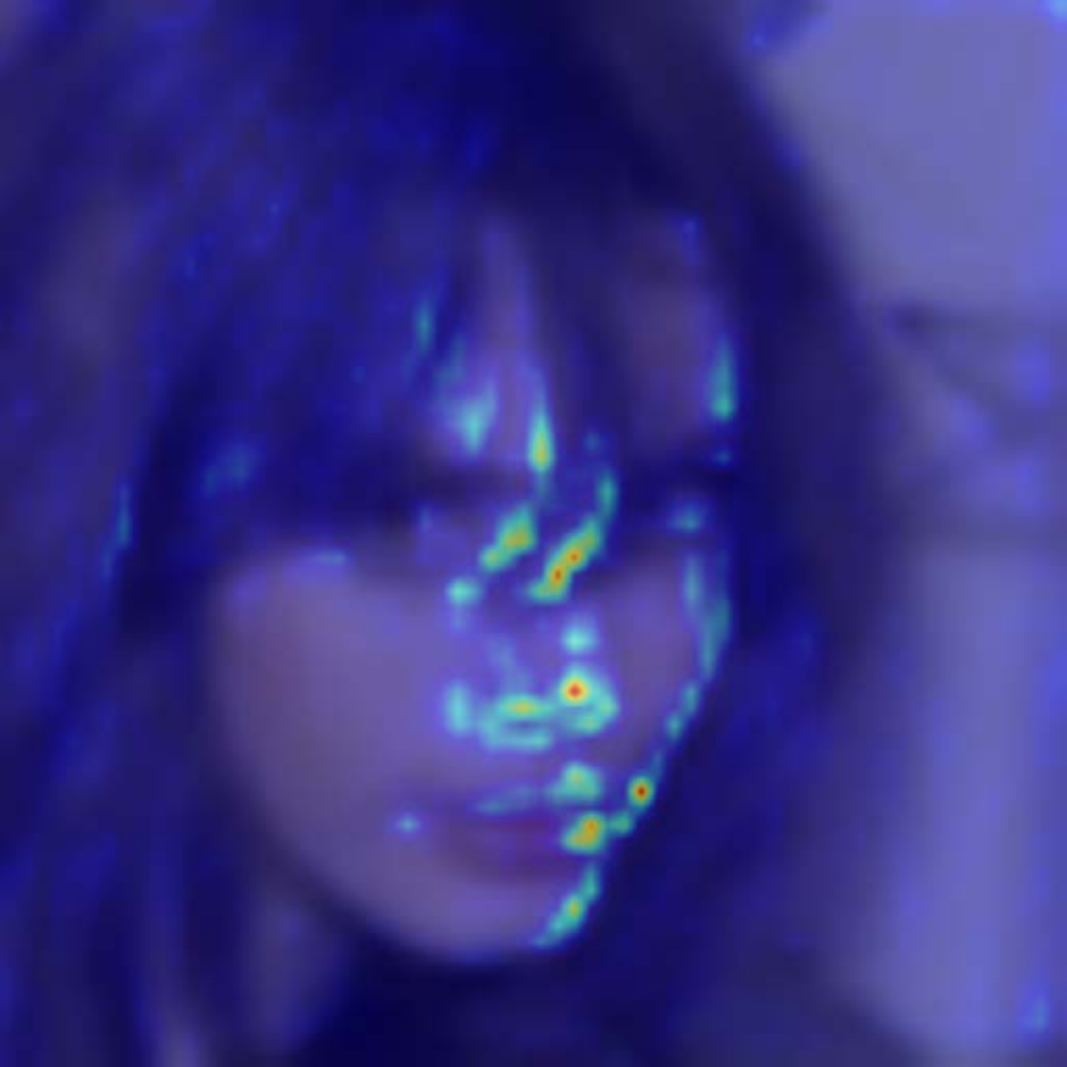} \\
  (1) s16h0 & (2) s16h1 & (3) s32h0 & (4) s32h1 & (5) s64h0 & (6) s64h1 \\
  
\end{tabular}
\end{center}
\caption{
\textcolor{black}{The qualitative visualizations from (b) to (g) are results of the experiments whose settings correspond to those in TABLE~\ref{tab:ablation}.
The result of (e) exp6, which takes the degraded face and priors as inputs, is better than the results of (b) exp2 and (c) exp3 in both realness and fidelity since exp2 and exp3 only take one of these two kinds of resources as input.
By globally fusing the features of degraded faces and priors with MHCAs, (e) exp6 also performs better than (d) exp4 implemented with the local fusion algorithm SFT~\cite{wang2021towards}.
In (g), RestoreFormer++ with a two-scale setting can avoid the weird eye shape restored in (e) exp6 implemented with a single-scale setting, but while extending to a three-scale setting, the result of (f) exp7 cannot see an obvious improvement compared to (g).
Images from (1) to (6) are the heatmaps of the left eye area attained on different scales.
`s$x$h$y$' means the $y$-th ($y\in\{0,1\}$) head attention map in $x \times x$ ($x\in\{16,32,64\}$) resolution.
In the low resolution, heatmaps (1)-(2) mainly focus on the most related eye areas while in the middle resolution, heatmaps (3)-(4) expand to salient edges that help the restoration of the shape of the left eye.
In high resolution, heatmaps (5)-(6) focus on more detailed edges. They yield less further improvement on the final restoration, and thus our RestoreFormer++ adopts a two-scale setting.
}}
\label{fig:components}
\end{figure*}%

\subsection{Ablation Study}
Our proposed RestoreFormer++ consists of several components, including MHCA, EDM, ROHQD, multi-scale mechanism, and several losses. It also contains two kinds of inputs: the degraded face and high-quality priors.
Each component plays an important role in the whole \textcolor{black}{restoration} pipeline. 
The followings are \textcolor{black}{the} detailed analyses of the effectiveness of these components. A discussion about the efficiency of our proposed method is also included.
%

\begin{table*}[!t]
\renewcommand{\arraystretch}{1.3}
\caption{Quantitative results of ablation studies on CelebA-Test~\cite{liu2015deep}.
  `degraded' and `prior' mean fusion information from degraded input and ROHQD, respectively.
  `none' and `MHSA' respectively mean the network uses either `degraded' or `prior' information without or with a self-attention mechanism.
  `SFT', `MHCA-D' and `MHCA-P' use both `degraded' and `prior' information.
  `SFT' uses SFT~\cite{wang2021towards} to fuse the information, while `MHCA-D' and `MHCA-P' use multi-head cross attention.
  The difference between `MHCA-D' and `MHCA-P' is that `MHCA-D' fuses $\bm{Z}_{mh}$ with $\bm{Z}_d^s$ while `MHCA-P' fuses $\bm{Z}_{mh}$ with $\bm{Z}_p^s$.
  `$S$' is the number of feature scales used for fusion.
  $S=1$ means the fusion \textcolor{black}{only exists} in $16\times16$ resolution while $S=2$ means the fusion are involved in both $16\times16$ and $32\times32$ resolutions. $S=3$ means it is further extended to $64 \times 64$ resolution.
  The proposed RestoreFormer++ integrated with `MHCA-P' and set with more than one scale performs the best relative to other variants.
  }
  \label{tab:ablation}
  \centering
  \renewcommand{\arraystretch}{1.1}
  \begin{tabular}{c|c|c|c|c|c|c|c|c|c|c} 
    \hline
    &\multicolumn{2}{c|}{sources} & \multicolumn{6}{c|}{methods} & \multicolumn{2}{c}{metrics} \\
    \hline\hline
    ~~No. of exp.~~ & ~~degraded~~ & ~~prior~~ & ~~none~~ & ~~MHSA~~ & ~~SFT~~  & ~~MHCA-D~~ & ~~MHCA-P~~ & ~~$S$~~ & ~~FID$\downarrow$~~ & ~~IDD$\downarrow$~~\\
    \hline
    exp1 & \checkmark & & \checkmark &&&&&1&48.33&0.6520 \\ 
    \hline
    exp2 &\checkmark&&&\checkmark&&&&1&47.96&0.6461\\
    \hline
    exp3 &&\checkmark&&\checkmark&&&&1&42.53&0.7467\\
    \hline
    exp4 &\checkmark&\checkmark&&&\checkmark&&&1&44.67&0.6373\\
    \hline
    exp5 &\checkmark&\checkmark&&&&\checkmark&&1&42.25&0.6038\\
    \hline
    exp6&\checkmark&\checkmark&&&&&\checkmark&1& 39.31 & 0.5677\\
    \hline
    exp7&\checkmark&\checkmark&&&&&\checkmark&3& \textcolor{blue}{39.11} & \textcolor{red}{0.5355}\\
    \hline
    \textbf{RestoreFormer++}&\checkmark&\checkmark&&&&&\checkmark&2& \textcolor{red}{38.41} & \textcolor{blue}{0.5375}\\
    \hline
  \end{tabular}
\end{table*}

\subsubsection{Analysis of Spatial Attention Mechanism}
\textcolor{black}{In RestoreFormer++,} global spatial attention mechanism is used to model the \textcolor{black}{rich} facial contextual information in the face image and its interplay with priors for aiding the \textcolor{black}{face} restoration. 
To \textcolor{black}{validate} the effectiveness of the spatial attention mechanism, we compare our single-scale RestoreFormer++ with and without attention mechanisms.
As shown in TABLE~\ref{tab:ablation}, both exp1 and exp2 only get information from the degraded face image.
By adopting self-attention (MHSA) to model contextual information, exp2 performs better than exp1 which is without MHSA in terms of FID and IDD.
This conclusion is also valid when comparing exp4 to exp6, whose inputs include both degraded information and additional high-quality priors.
\textcolor{black}{
In exp4, we replace MHCA in RestoreFormer++ with SFT~\cite{wang2018recovering} for locally fusing these two kinds of information. Since it ignores the facial contextual information in the face image, its result in Fig.~\ref{fig:components} (d) fails to restore natural eyes.
Exp6 is a version of RestoreFormer++ implemented with a single-scale fusion mechanism. It uses MHCA for globally fusing degraded information and priors. (1)-(4) in Fig.~\ref{fig:components} are its multi-head (4 heads) attention maps of the left eye region in scale $16\times16$.
It shows that the highlighted areas not only occur in the left eye area but also in other regions of the face image, especially the more related right eye region.
It means that apart from the information in the left areas, our RestoreFormer++ with MHCA can also utilize the related information in other areas to restore the left eye with more natural appearance (Fig.~\ref{fig:components} (e)).
}

\subsubsection{Analysis of Degraded Information and Priors.}
In this subsection, we analyze the roles of the degraded information extracted from the degraded face image and its corresponding high-quality priors matched from ROHQD.
In exp2 and exp3 (TABLE~\ref{tab:ablation}), we replace the MHCA in our single-scale RestoreFormer++ with MHSA, whose queries, keys, and values are all from either the degraded information or the high-quality priors.
We can see that exp2 attains a better average IDD score which means it performs better in fidelity.
In contrast, exp3 has a better FID score, meaning its results \textcolor{black}{contain more realness}.
By globally fusing the degraded information and priors with MHCA in our single-scale RestoreFormer++ (exp6 in TABLE~\ref{tab:ablation}), it performs better than exp2 and exp3 in both IDD and FID, which means that our RestoreFormer++ can restore faces with both realness and fidelity.
The visualized results in Fig.~\ref{fig:components} show that the result of exp2 (Fig.~\ref{fig:components} (b)) is more similar to GT but contains fewer details compared to (c) and (e), which are the results of exp3 and exp6, respectively.
Although the details in (c) are richer, it looks less \textcolor{black}{similar to} the GT, especially in the eyes.
On the contrary, Our result shown in (e) is similar to GT and meanwhile contains rich details, and thus presents pleasantly.
Besides, according to Fig.~\ref{fig:framework} (b) and Eq.~\ref{eq:shortcut_prior}, we tend to add the attended feature $\bm{Z}_{mh}$ to $\bm{Z}_p^0$ rather than $\bm{Z}_d^0$ (corresponding to exp5 in TABLE~\ref{tab:ablation}), since we experimentally find that it can attain better performance.

\subsubsection{Analysis of Multi-scale Mechanism}
\label{subsub: multi-scale}
\textcolor{black}{
%
Our multi-scale mechanism aims to facilitate RestoreFormer++ by modeling contextual information based on both semantic and structural information, thereby improving the restoration performance in both realness and fidelity.
First, we apply MHCAs to fuse the degraded features and priors at a resolution of $16\times16$, which is the smallest resolution in our model (this setting corresponds to exp6 in TABLE~\ref{tab:ablation}).
The features of a face at this scale are semantic information of facial components, such as eyes, mouth, nose, etc.
The highlighted areas in the attention maps of the left eye in Fig.~\ref{fig:components}~(1)-(2) are eyes areas, which reveal that the restoration of the left eye in Fig.~\ref{fig:components}~(e) is achieved by leveraging contextual information from its semantic-related areas.
Compared with the results in (d) attained with SFT~\cite{wang2021towards}, a spatial-based fusion approach, the restored left eye of (e) is more complete and real.
However, its edge shape is not smooth enough, leading to a weird look.
Therefore, we extend MHCAs to features with a larger scale, $32\times32$ (corresponding to Restoreformer++ in TABLE~\ref{tab:ablation}), and attain a restored result with a more natural look as shown in Fig.~\ref{fig:components}~(g).
Its corresponding attention maps in Fig.~\ref{fig:components}~(3)-(4) show that apart from related eye areas, its highlighted areas diffuse to some salient edges that help reconstruct the smooth and natural shape of the left eye .
FID and IDD scores on CelebA-Test~\cite{liu2015deep} in TABLE~\ref{tab:ablation} indicate that increasing the number of scales from one to two can improve restoration performance in both realness and fidelity.
To make further exploration, we extend MHCAs to the features at a resolution of $64\times64$ (corresponding to exp7 in TABLE~\ref{tab:ablation}).
Its attention maps (Fig.~\ref{fig:components}~(5)-(6)) focus on more detailed structures such as hairs.
However, its restored result in Fig.~\ref{fig:components}~(f) does not show an obvious improvement compared to (g) attained with a two-scale setting.
Its quantitative results in TABLE~\ref{tab:ablation} show that it attains a better IDD score but worse FID score than RestoreFormer++ implemented with a two-scale setting.
%
%
Comprehensively considering efficiency, where the running time of the three-scale setting increases by about 17\% compared to the two-scale setting (TABLE~\ref{tab:efficiency}), we adopt a two-scale setting in RestoreFormer++.
}

\subsubsection{Analysis of ROHQD.}
Comparisons between our RestoreFormer++ and DFDNet~\cite{li2020blind}, whose priors are recognition-oriented, have \textcolor{black}{validated} the effectiveness of ROHQD.
To further evaluate the contribution of ROHQD in RestoreFormer++, we replace ROHQD with a recognition-oriented dictionary with the same learning process as ROHQD.
We implement it by replacing the encoders $\bm{E}_d$ and $\bm{E}_h$ with a VGG~\cite{simonyan2014very}. Similar to~\cite{li2020blind}, we initialize these encoders with weights attained with ImageNet~\cite{deng2009imagenet} and freeze them while training.
We conduct experiments on CelebA-Test~\cite{liu2015deep}. Its scores in terms of FID and IDD are 50.39 and 0.7572, which is worse than RestoreFormer++ implemented with ROHQD.
It indicates that the facial details in ROHQD that are accordant to reconstruction tasks are helpful for face restoration.

\begin{table}[!t]
\renewcommand{\arraystretch}{1.3}
\caption{Quantitative results of methods with or without EDM measured on FID$\downarrow$. Methods with EDM perform better than those without EDM on CelebChild-Test~\cite{wang2021towards} and WebPhoto-Test~\cite{wang2021towards} datasets whose degradations are more diverse and severe and perform comparably on LFW-Test~\cite{huang2008labeled} dataset with more common degradations. RestoreFormer++ is better than the other methods in both settings. }
\label{tab:real_edm}
\centering
\resizebox{\linewidth}{!}{
  \begin{tabular}{c|c|c|c} 
    \hline
    Methods & ~~\textbf{LFW-Test}~\cite{huang2008labeled}~~ & ~~\textbf{CelebChild-Test}~\cite{wang2021towards}~~ & ~~\textbf{WebPhoto-Test}~\cite{wang2021towards}~~  \\
    \hline
    \hline
    PSFRGAN~\cite{chen2020generative} & 53.17 & 105.65 & 83.50 \\
    \hline
    PSFRGAN w/ EDM & 53.20 & 104.22 & 82.28 \\
    \hline
    GFP-GAN~\cite{wang2021towards} & 50.30 & 111.78 & 87.82 \\
    \hline
    GFP-GAN w/ EDM &  50.72 & 109.08 & 86.17 \\
    \hline
    Ours w/o EDM &  48.10 & 103.86 & 75.42 \\
    \hline
    Ours  & 48.48 & 102.66 & 74.21  \\
    \hline
  \end{tabular}
  }
\end{table}

\renewcommand{\tabcolsep}{.5pt}
\begin{figure*}
\hsize=\textwidth
\vspace{-0.3cm}
\begin{center}
\begin{tabular}{ccccccc}
  \includegraphics[width=\swseven]{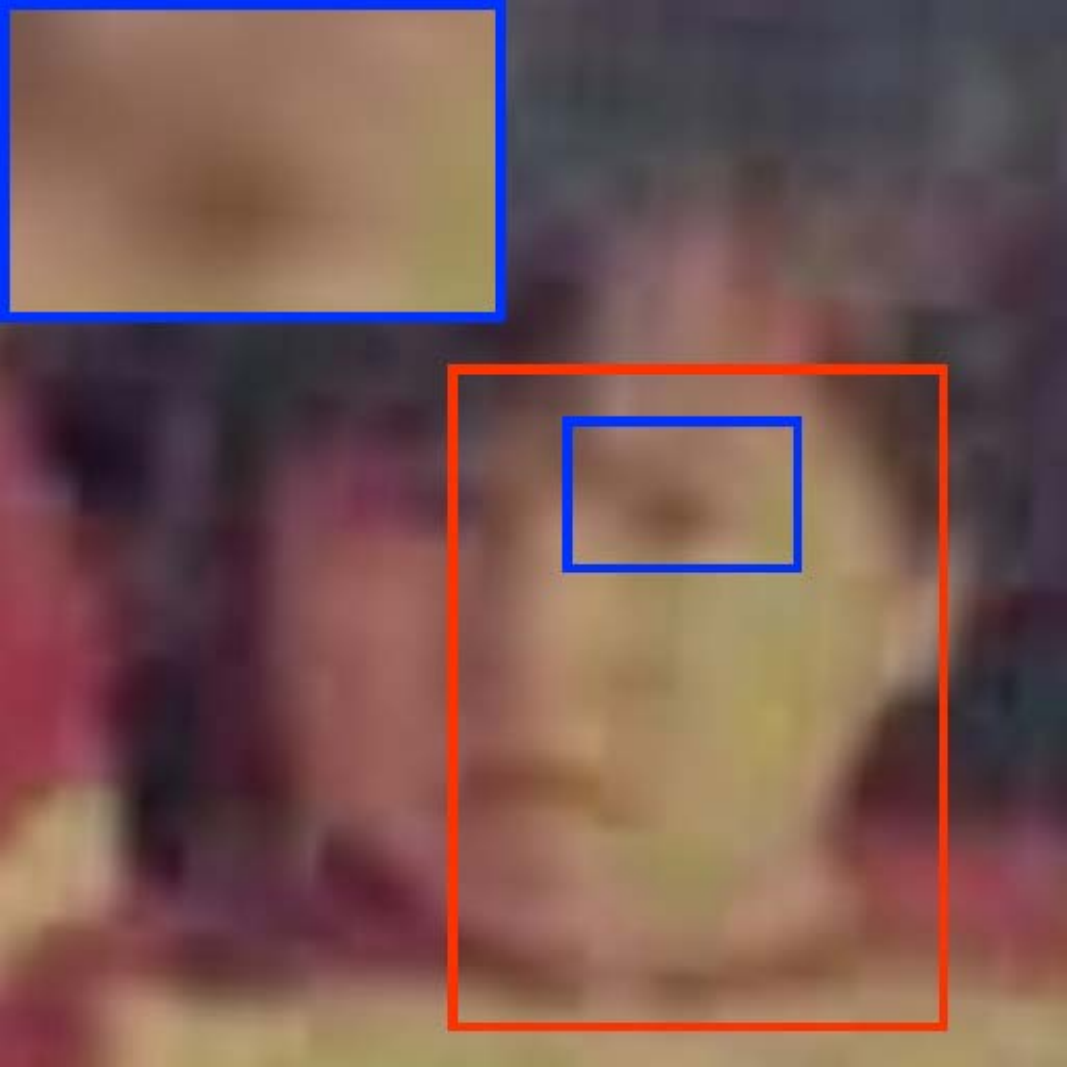}&
  \includegraphics[width=\swseven]{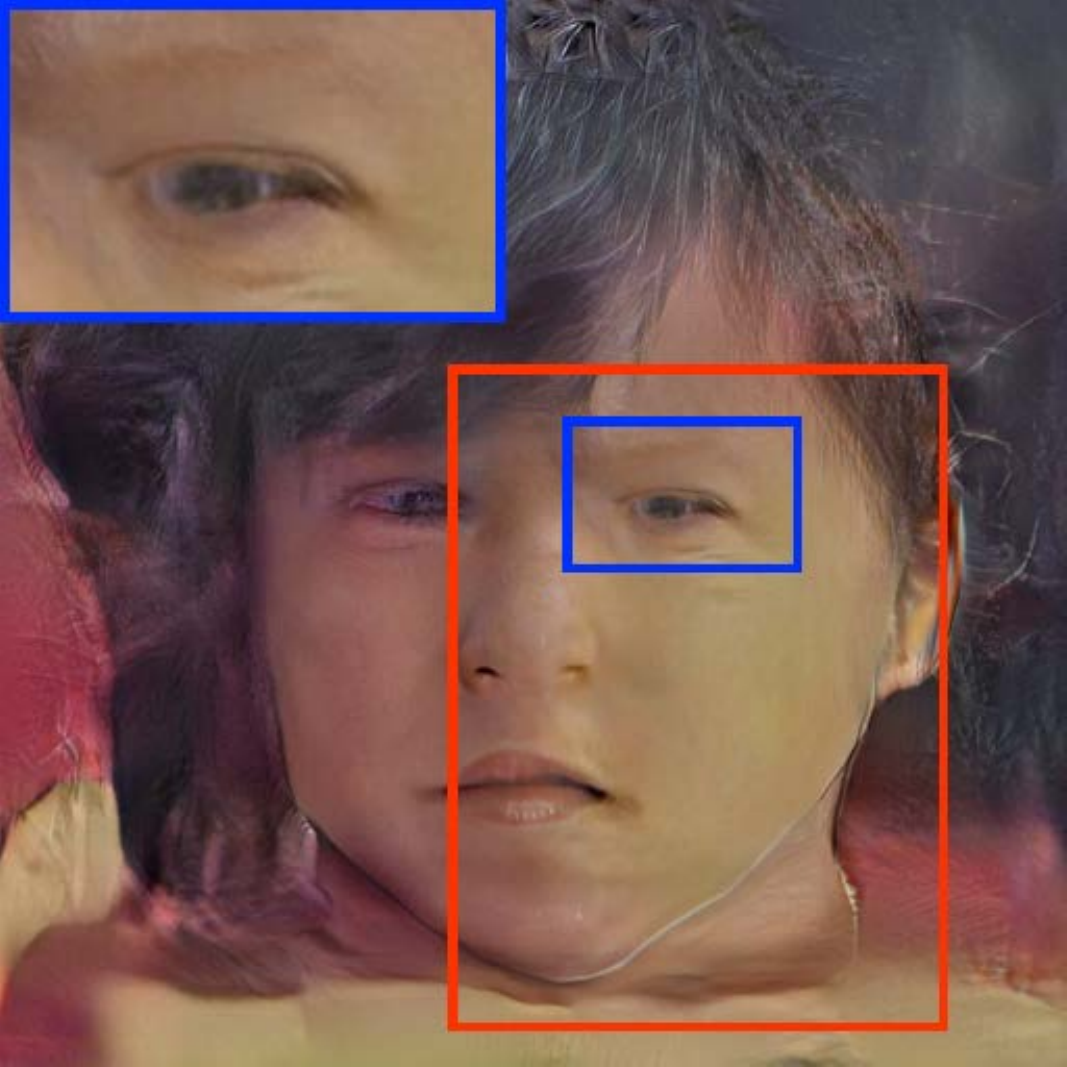}&
  \includegraphics[width=\swseven]{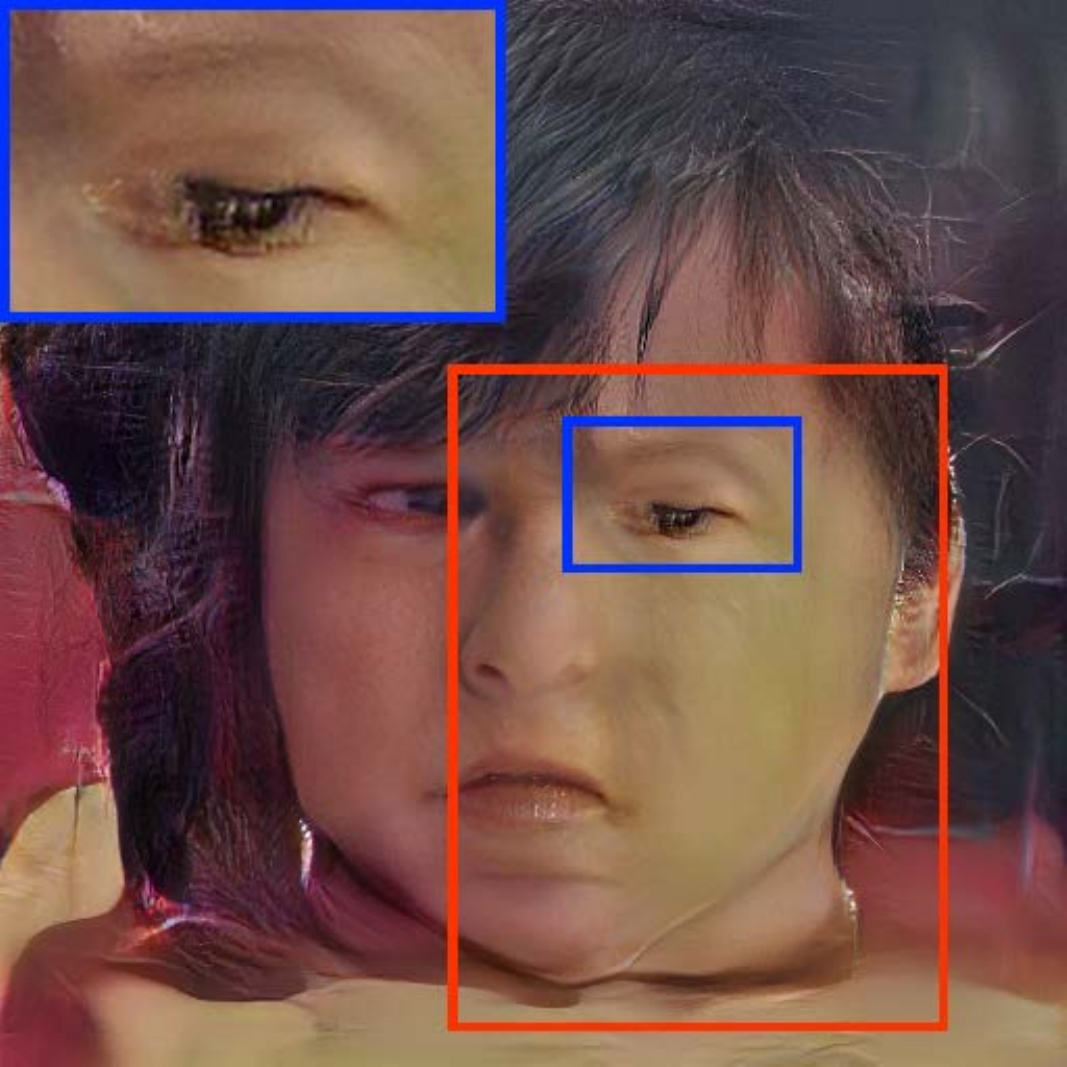}&
  \includegraphics[width=\swseven]{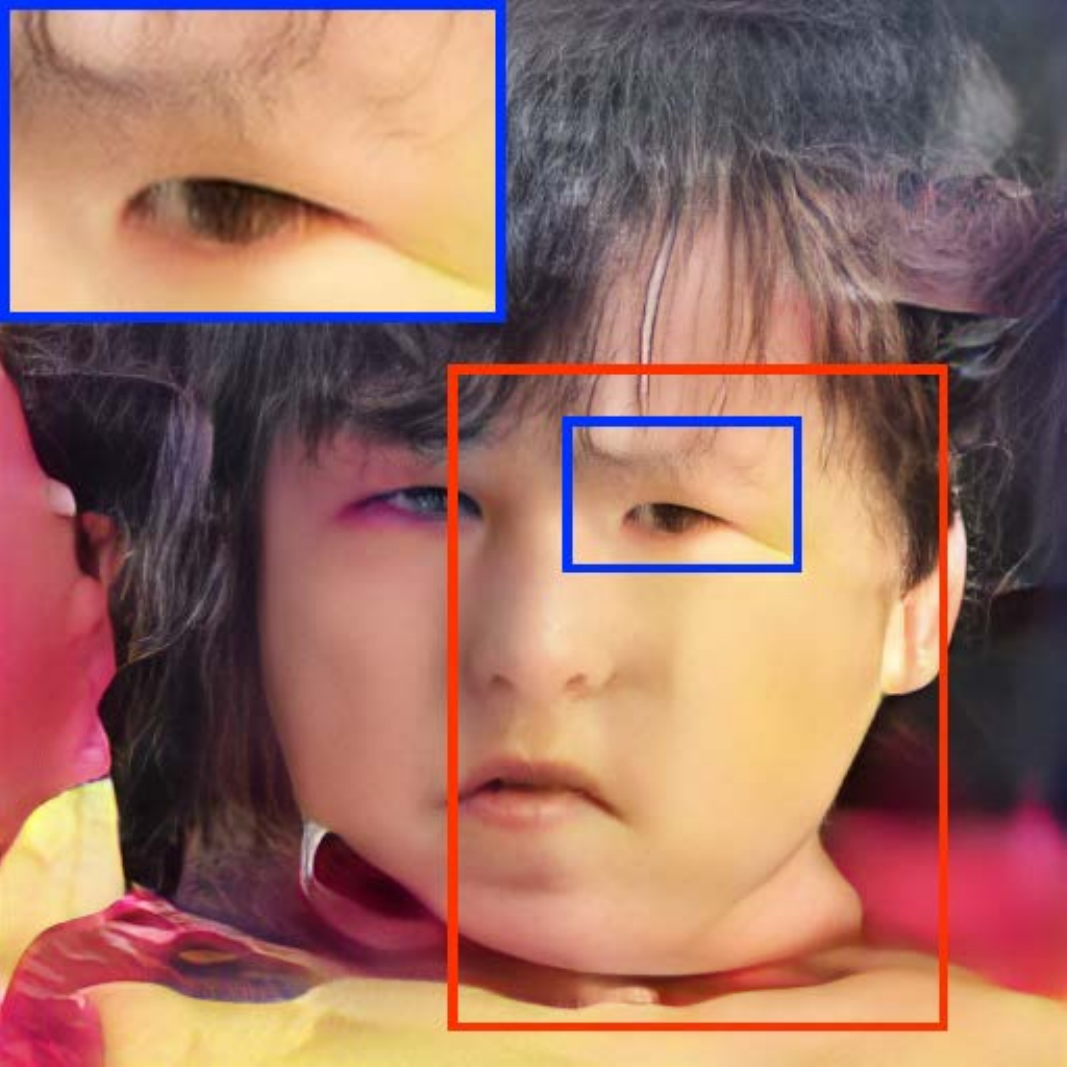}&
  \includegraphics[width=\swseven]{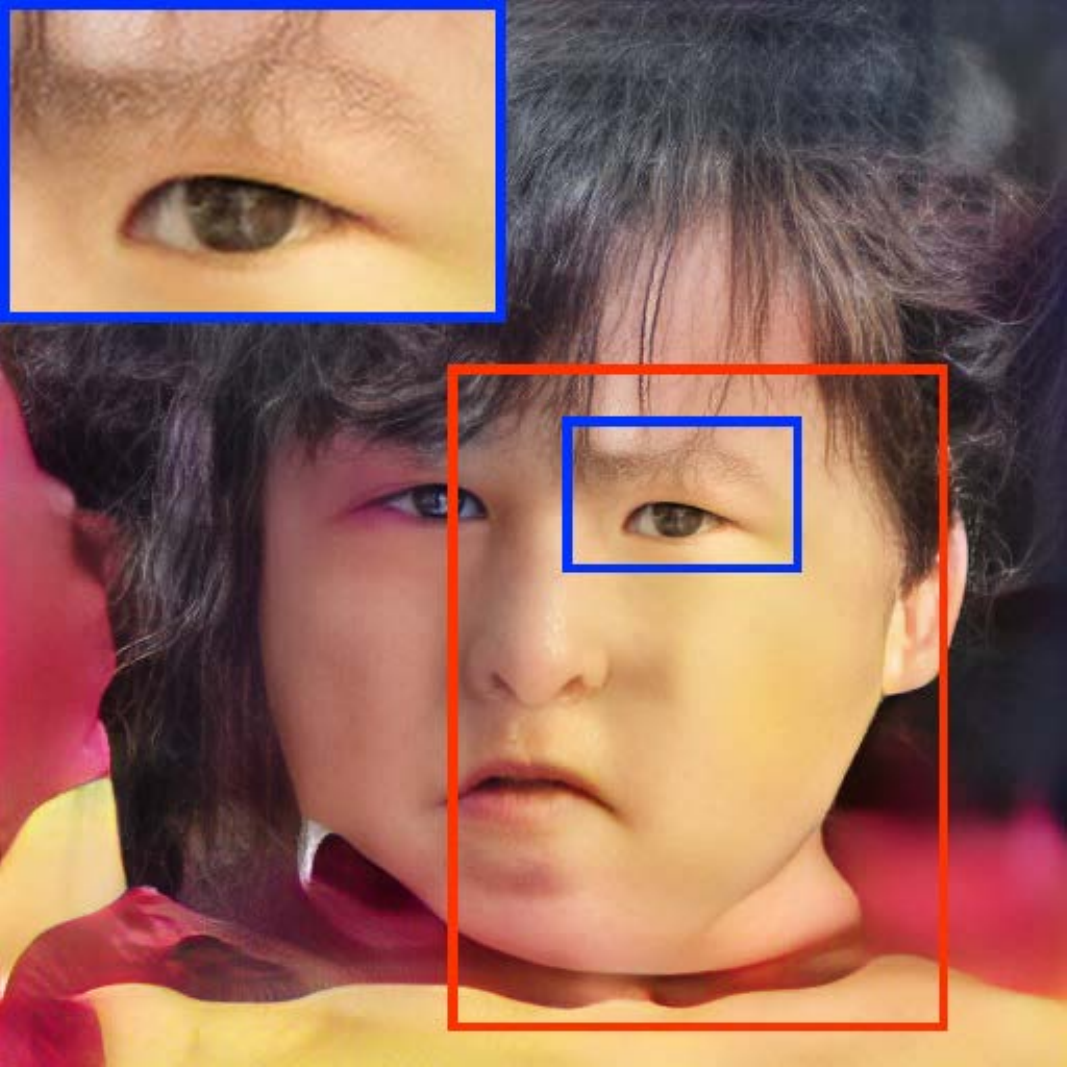}&
  \includegraphics[width=\swseven]{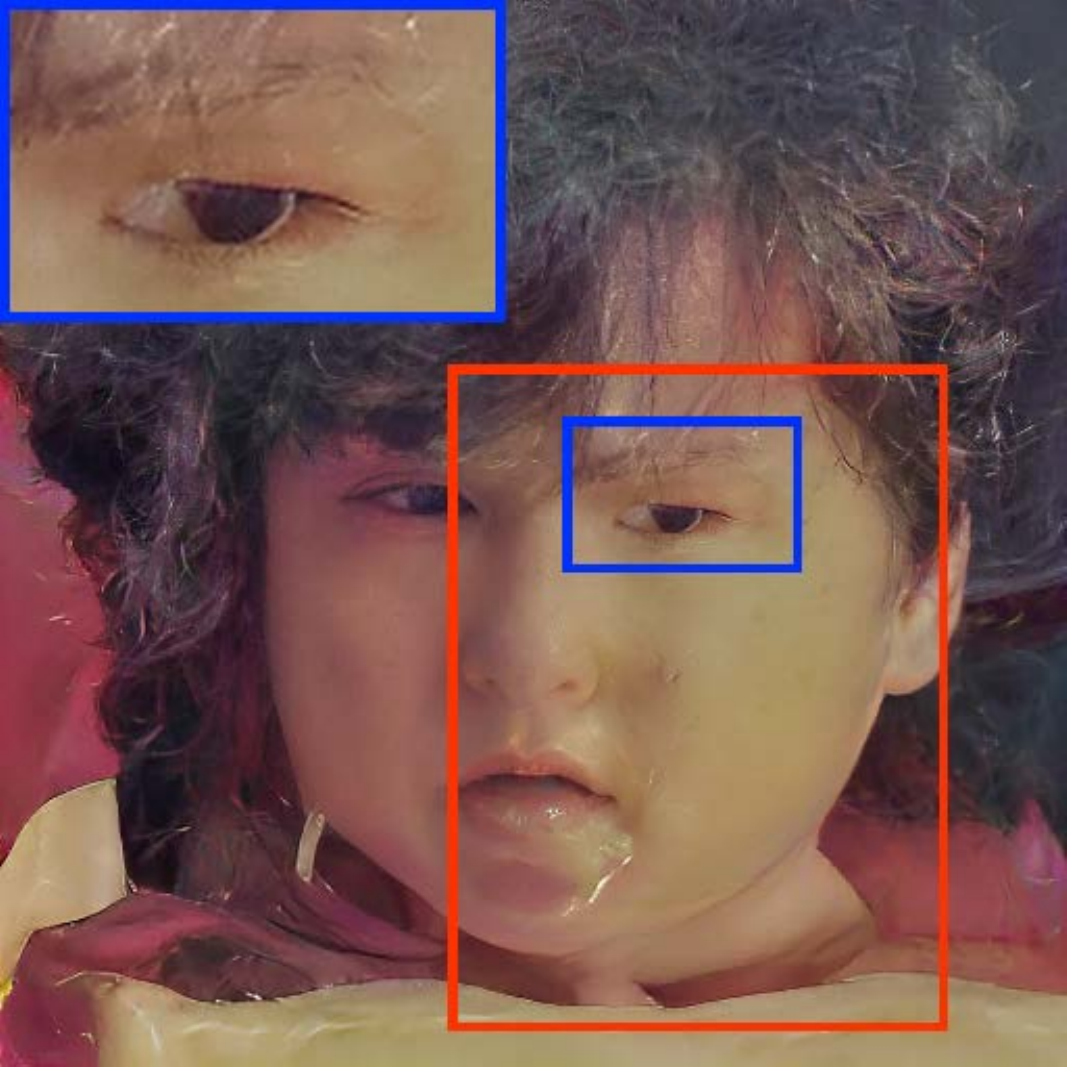}&
  \includegraphics[width=\swseven]{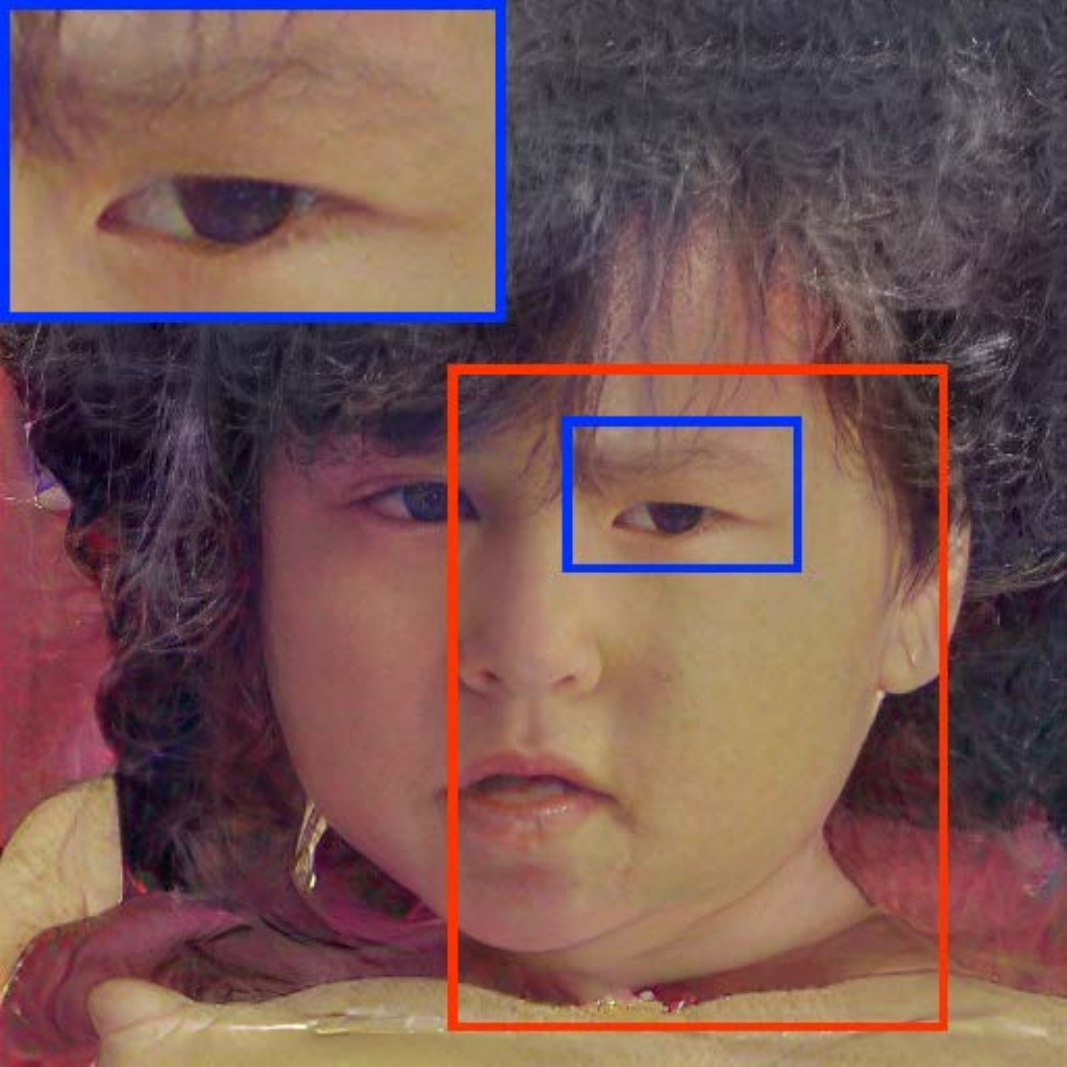} \\
  \includegraphics[width=\swseven]{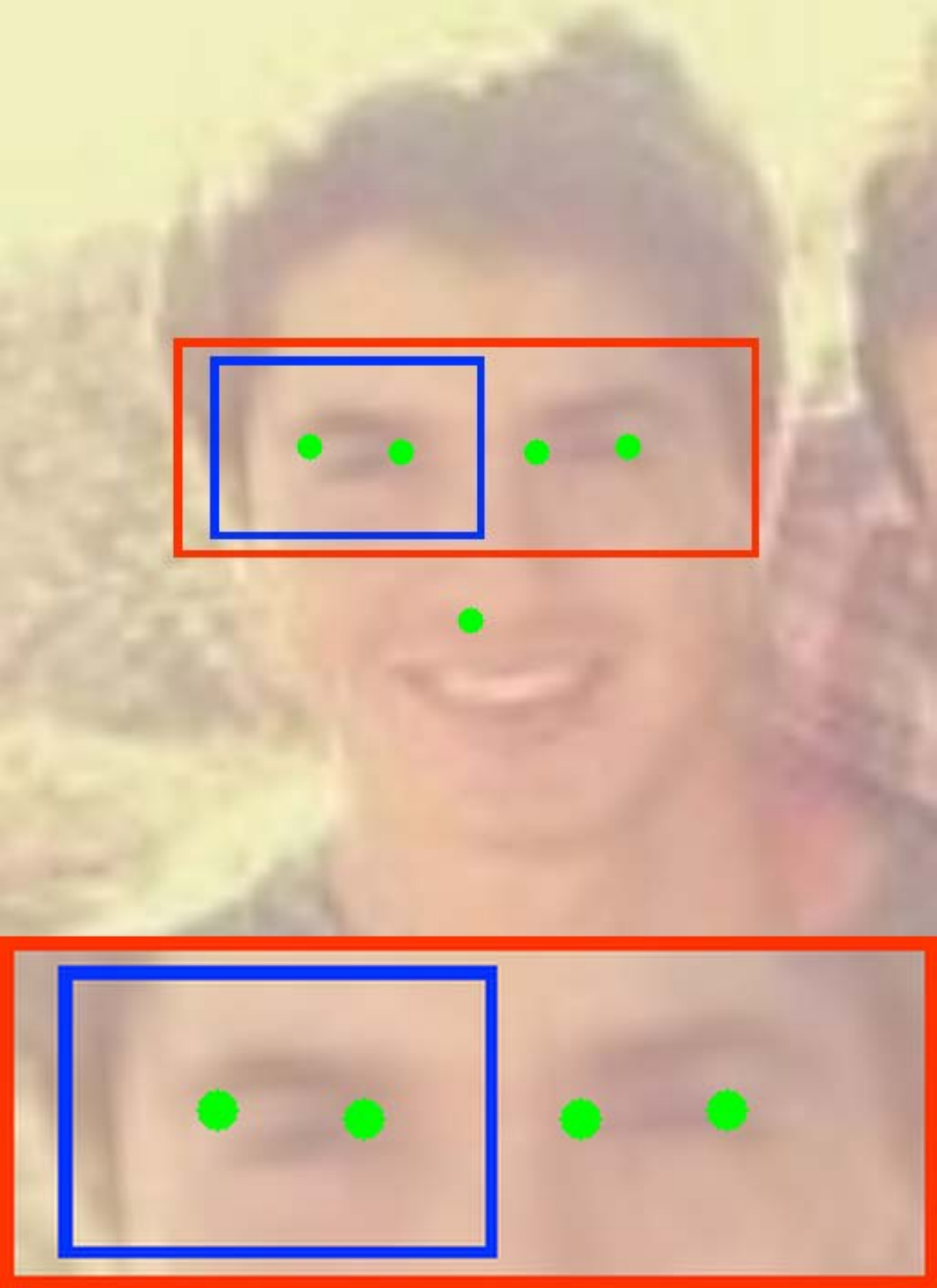}&
  \includegraphics[width=\swseven]{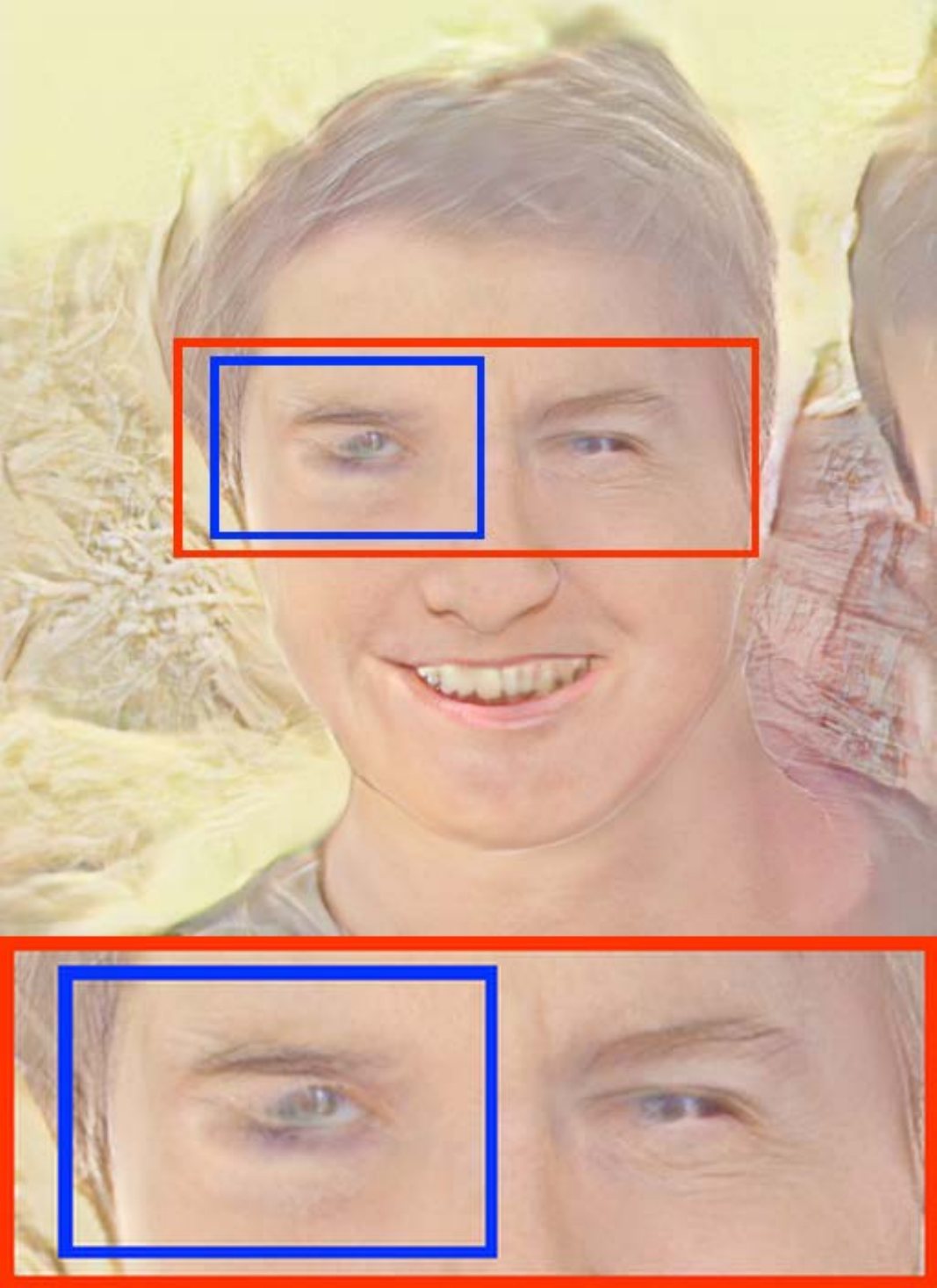}&
  \includegraphics[width=\swseven]{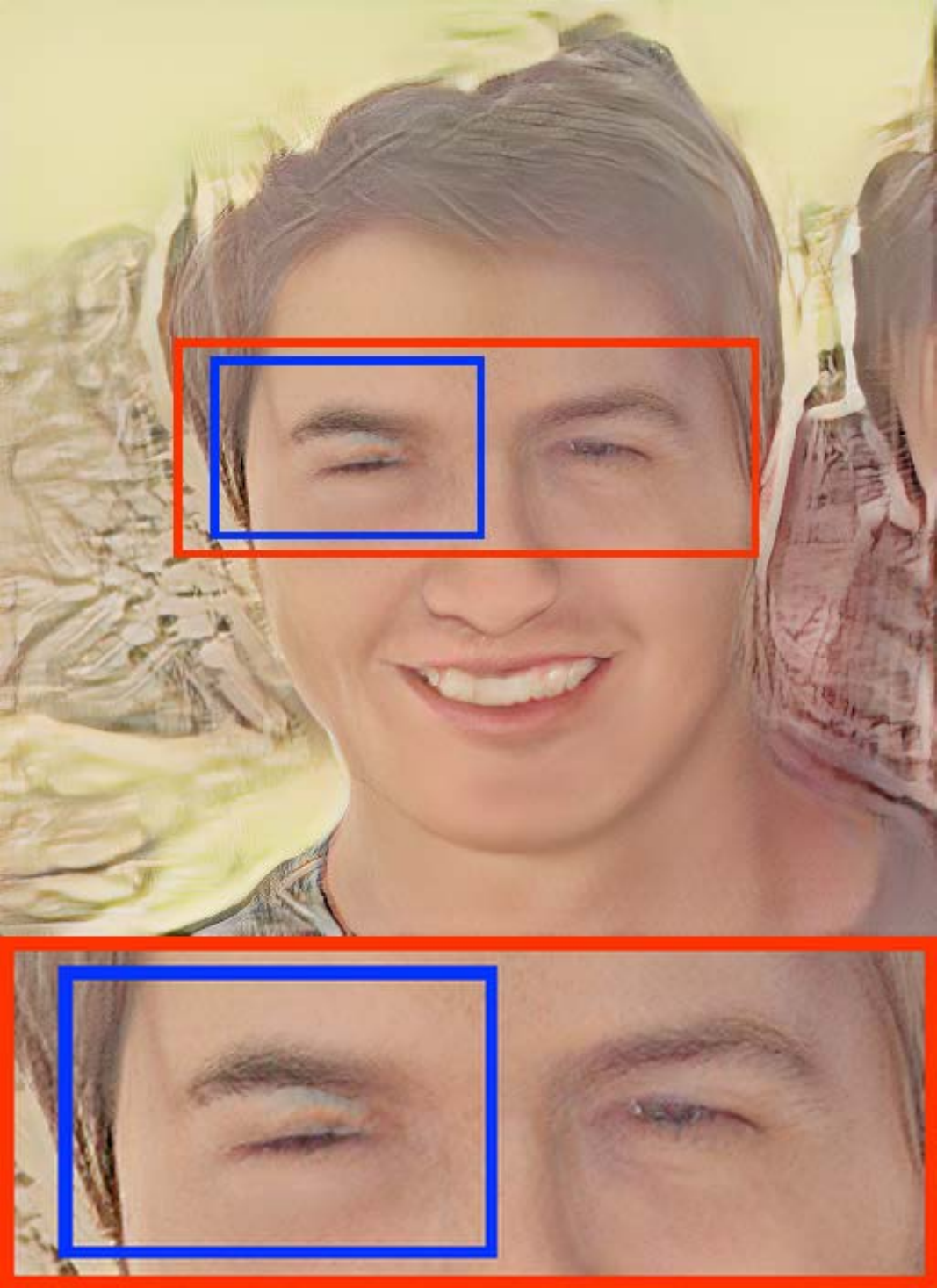}&
  \includegraphics[width=\swseven]{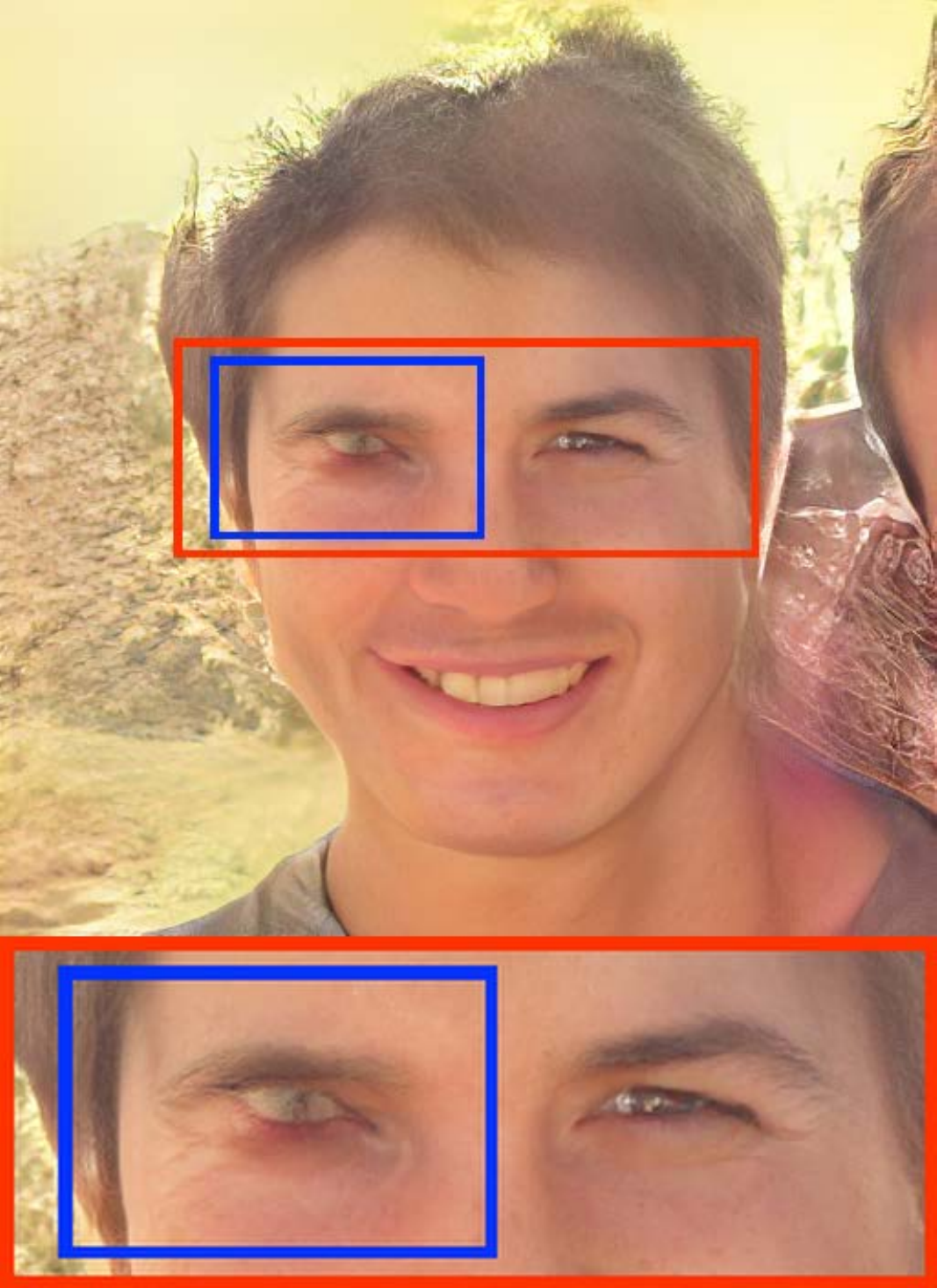}&
  \includegraphics[width=\swseven]{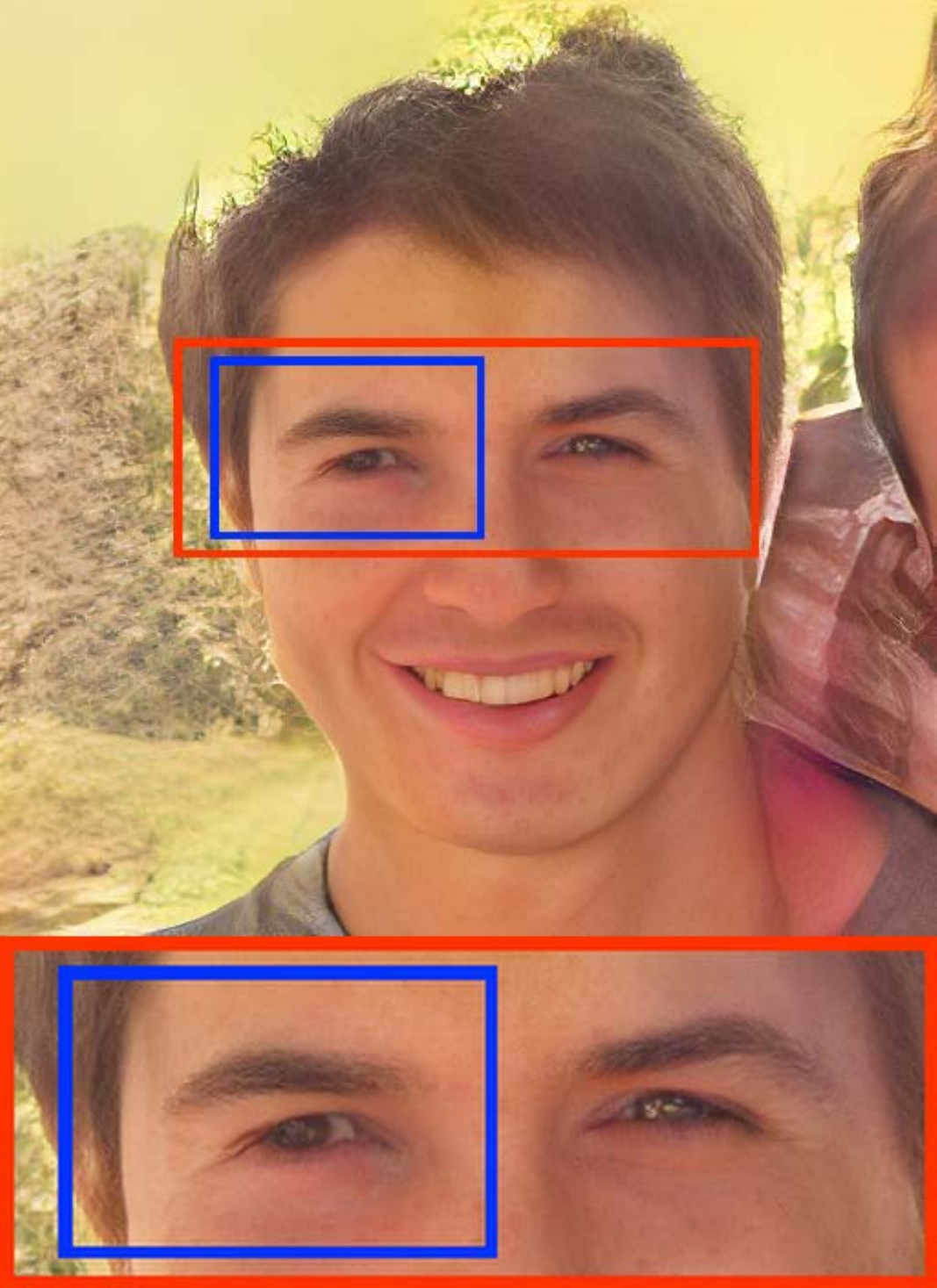}&
  \includegraphics[width=\swseven]{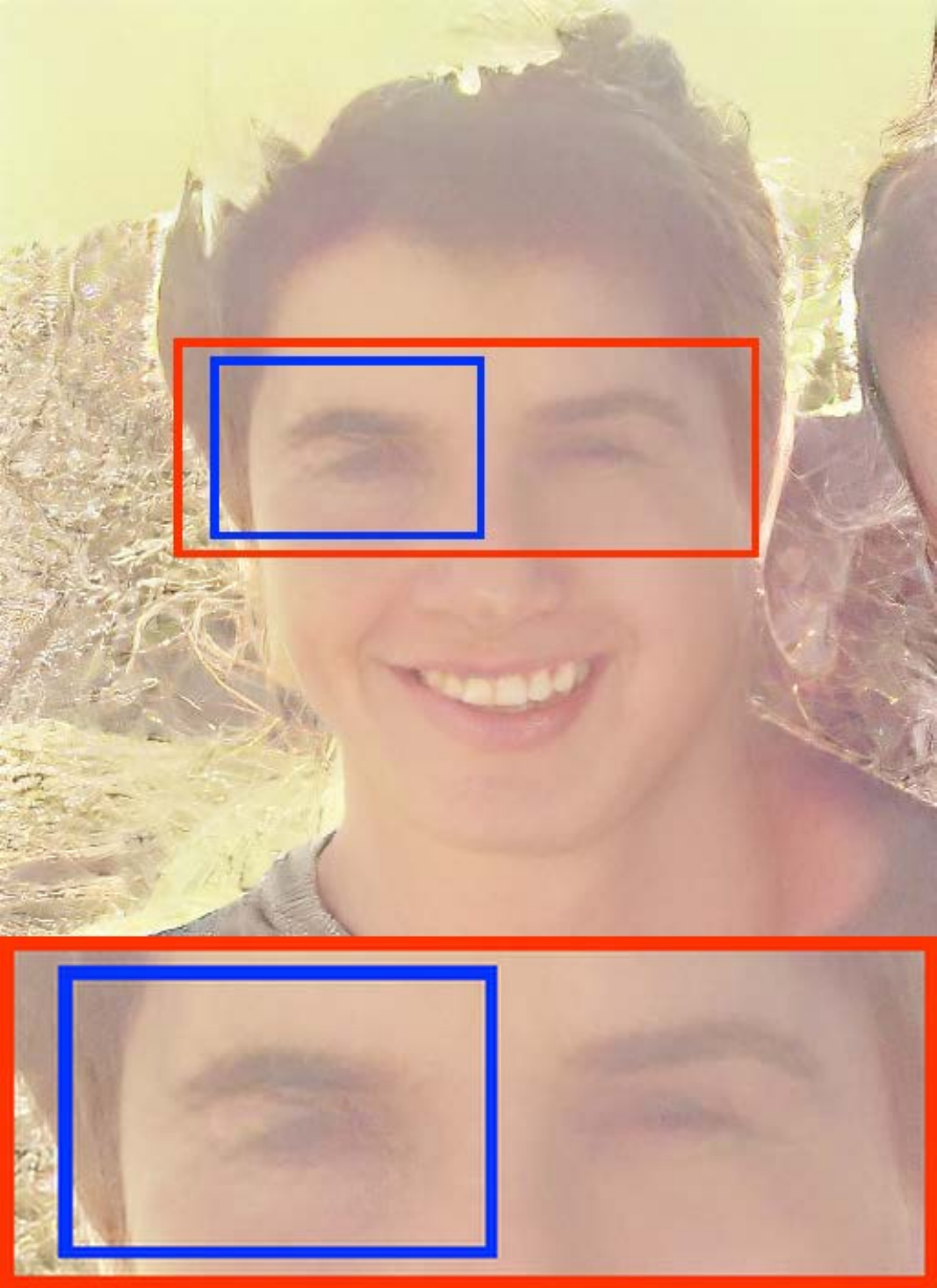}&
  \includegraphics[width=\swseven]{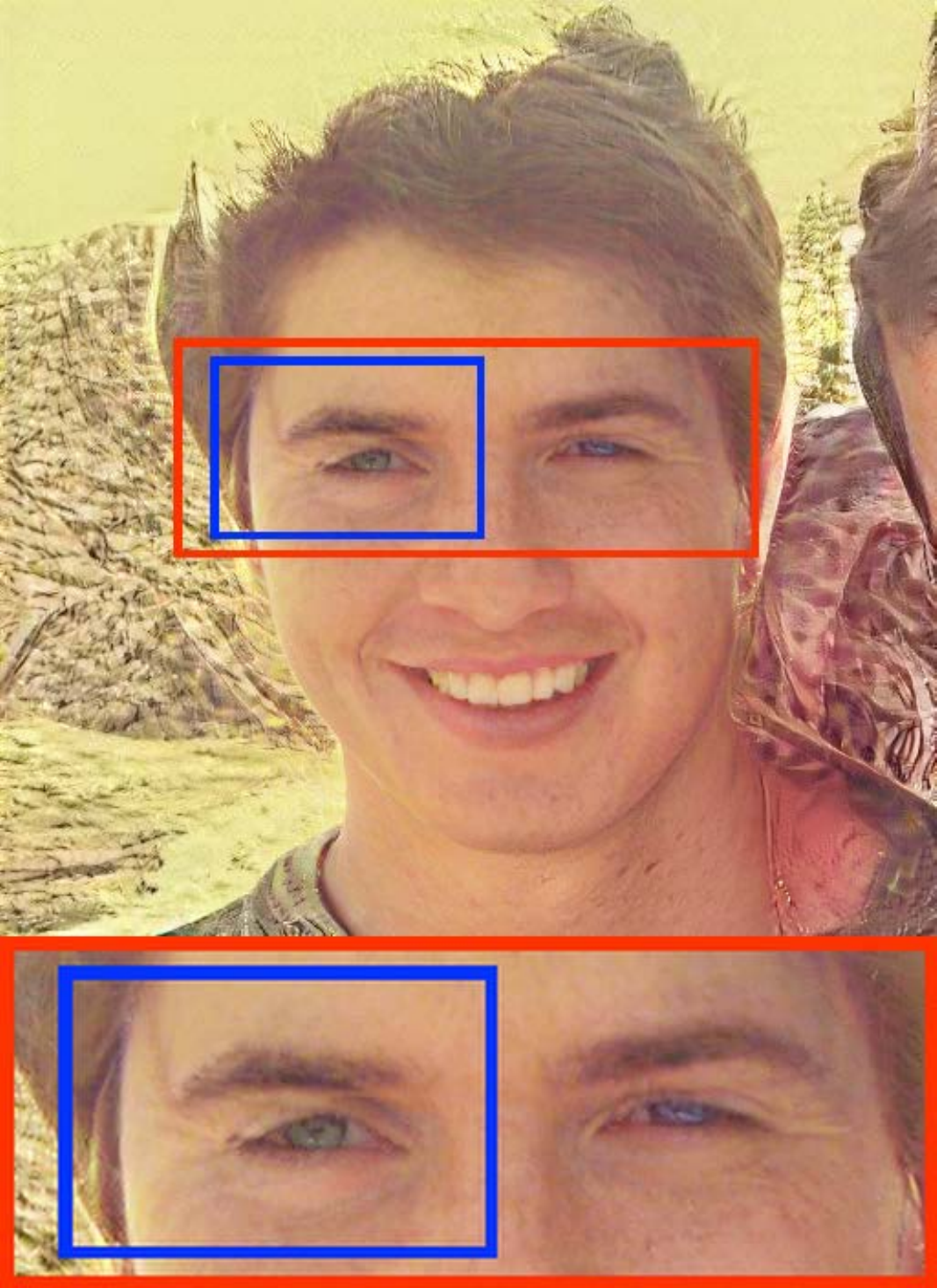} \\
   \scriptsize (a) Input & \scriptsize (b) PSFGAN~\cite{chen2020generative} & \scriptsize (c) PSFGAN w/ EDM & \scriptsize (d) GFP-GAN~\cite{wang2021towards} & \scriptsize (e) GFP-GAN w/ EDM & \scriptsize (f) Ours w/o EDM & \scriptsize (g) Restoreformer++ \\
\end{tabular}
\end{center}
\caption{\textcolor{black}{Qualitative results of methods with or without EDM. Methods with EDM can rectify the distortion (area in the blue box) introduced by uneven degradation (area in the red box in the first sample) or misalignment (areas in the red box in the second sample) and remove the haze covered on the face image (the second sample).
In the input of the second example, the green points are the reference landmarks defined in the training dataset. The reference landmark of the left eye is aligned to the left eyebrow of this sample, leading to the unnatural restored results of the methods without EDM. However, the methods with EDM can alleviate this issue. }
}
\label{fig:EDM}
\end{figure*}%

\subsubsection{Analysis of EDM.}
\label{subsub: EDM}
\begin{figure}[th]
\scriptsize
\begin{center}
\begin{tabular}{ccc}
    \includegraphics[width=\swedm1]{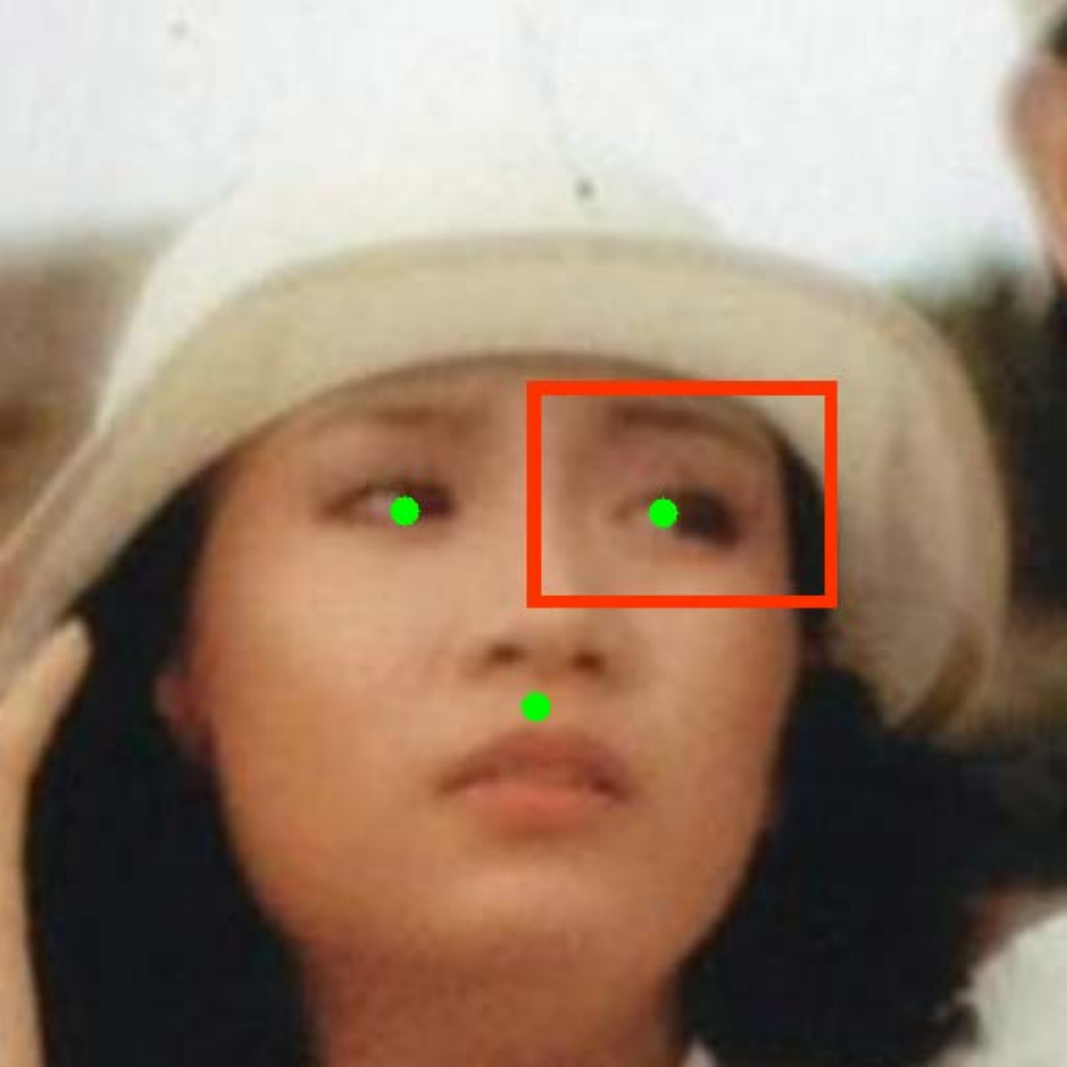}&
    \includegraphics[width=\swedm1]{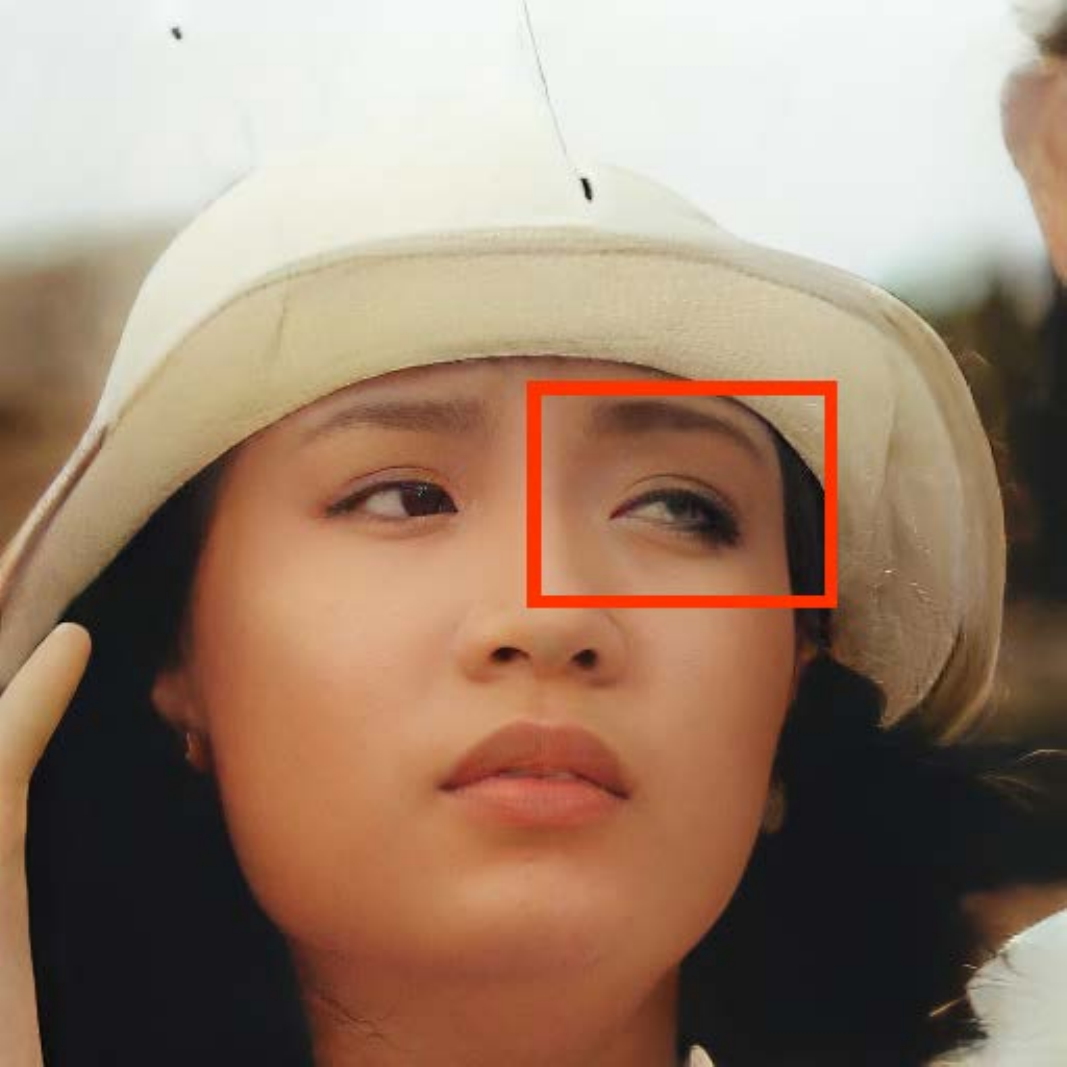}&
    \includegraphics[width=\swedm1]{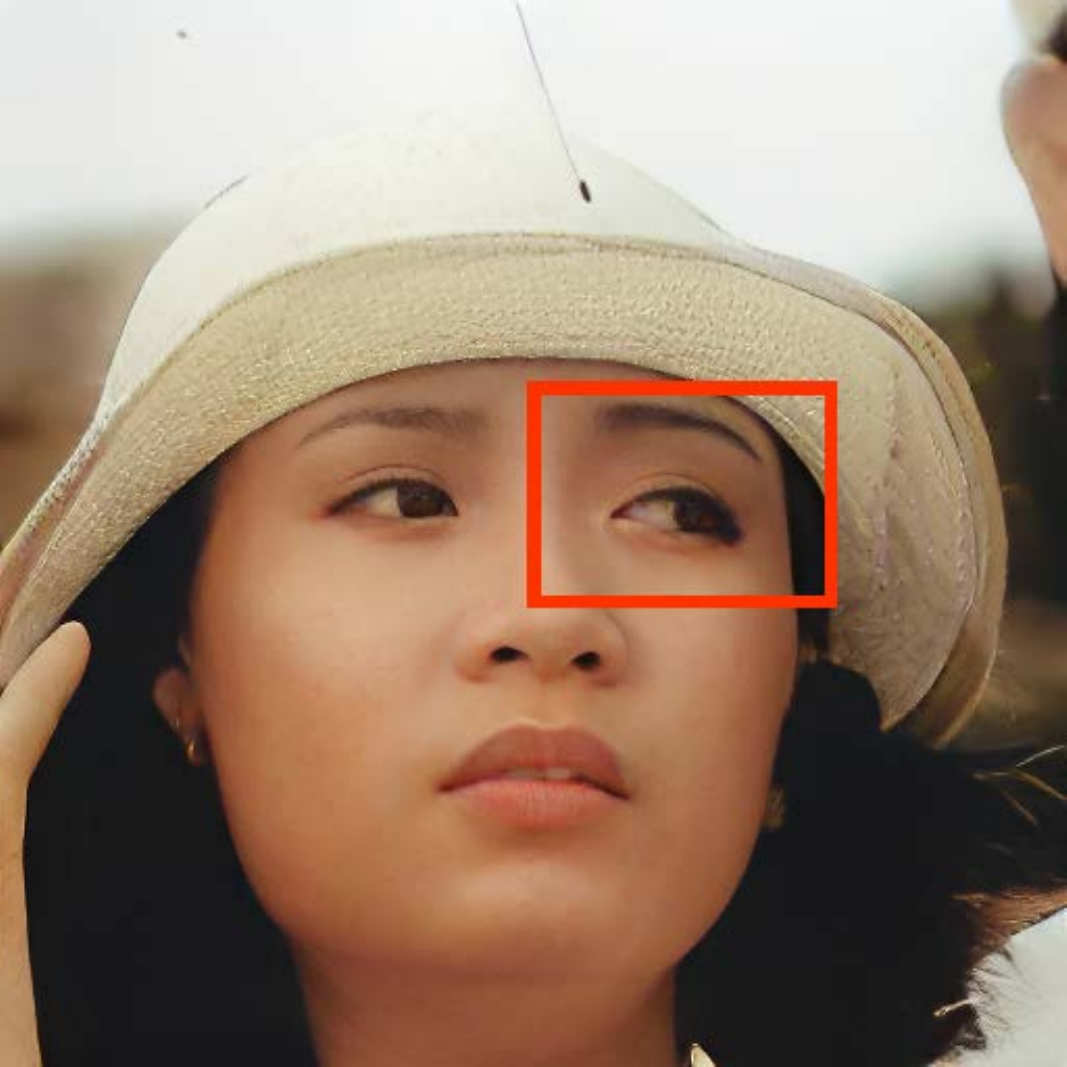}\\
    \includegraphics[width=\swedm1]{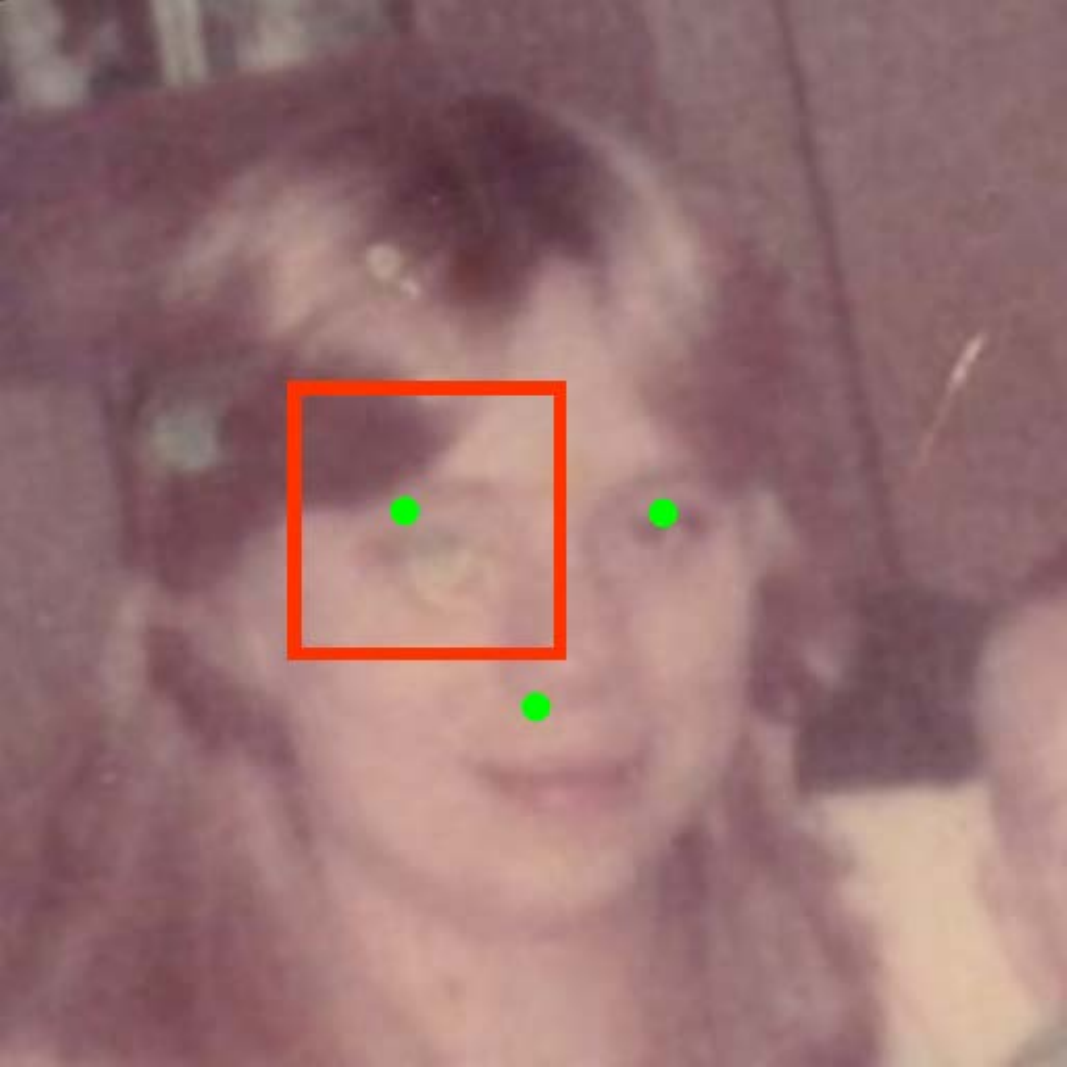}&
    \includegraphics[width=\swedm1]{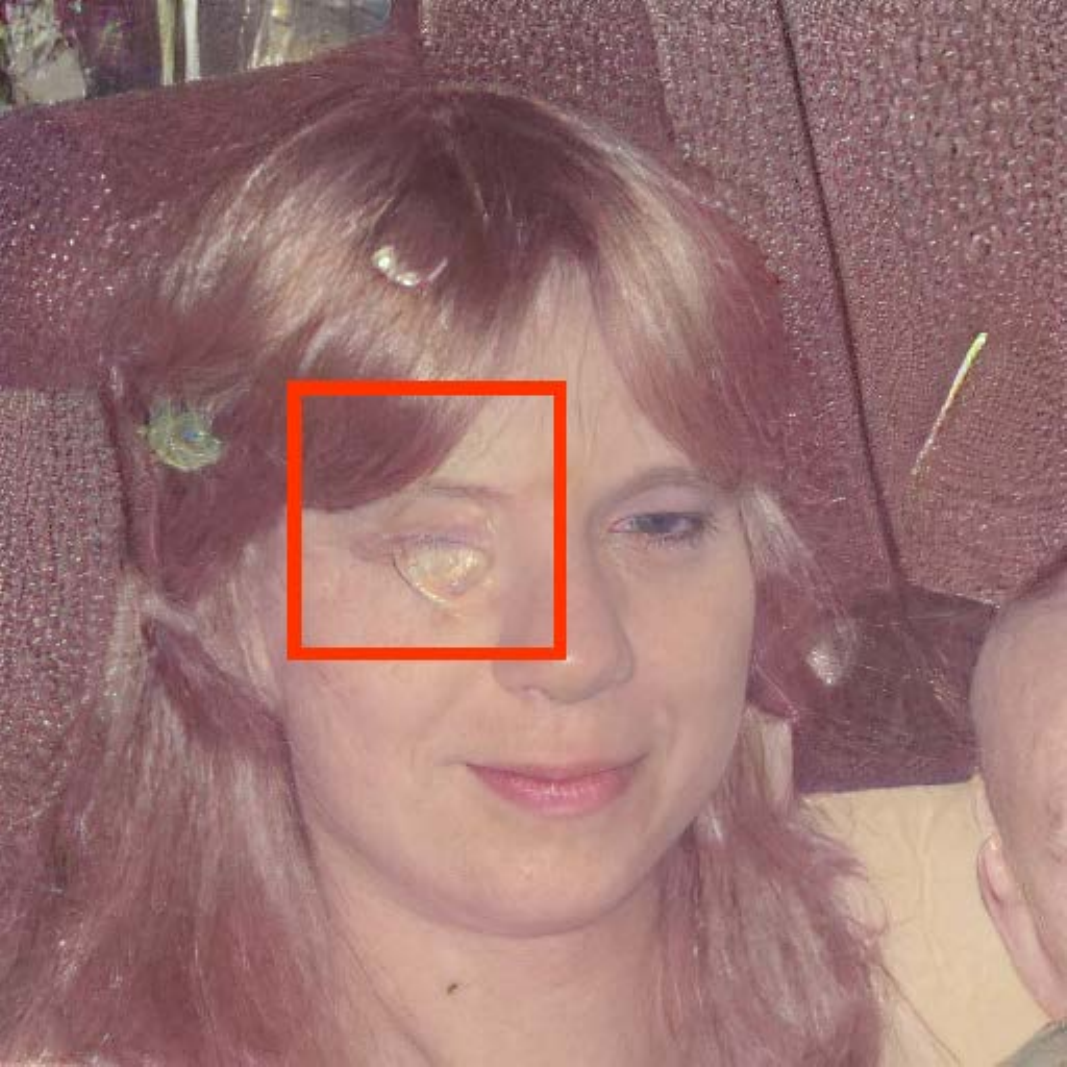}&
    \includegraphics[width=\swedm1]{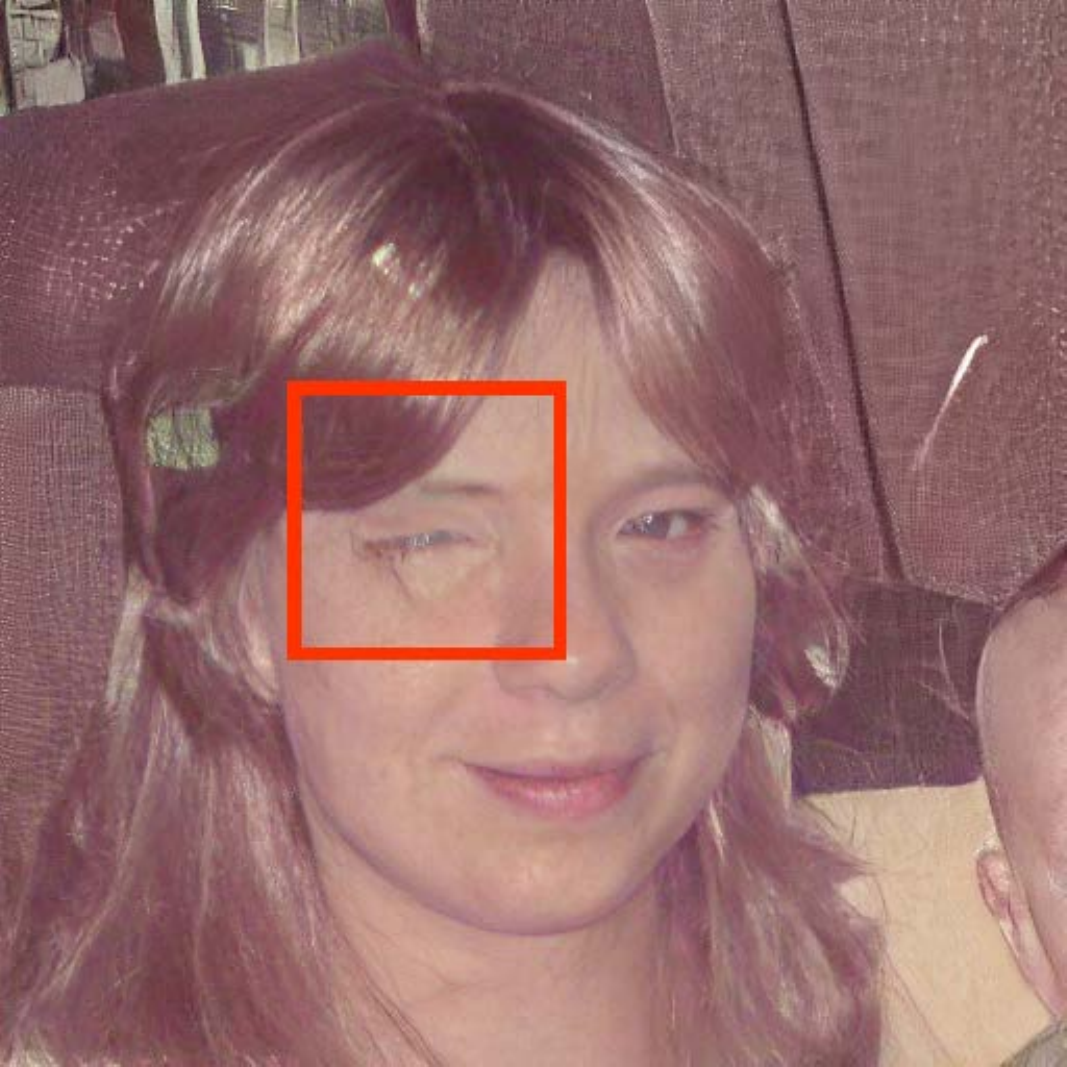}\\
    \includegraphics[width=\swedm1]{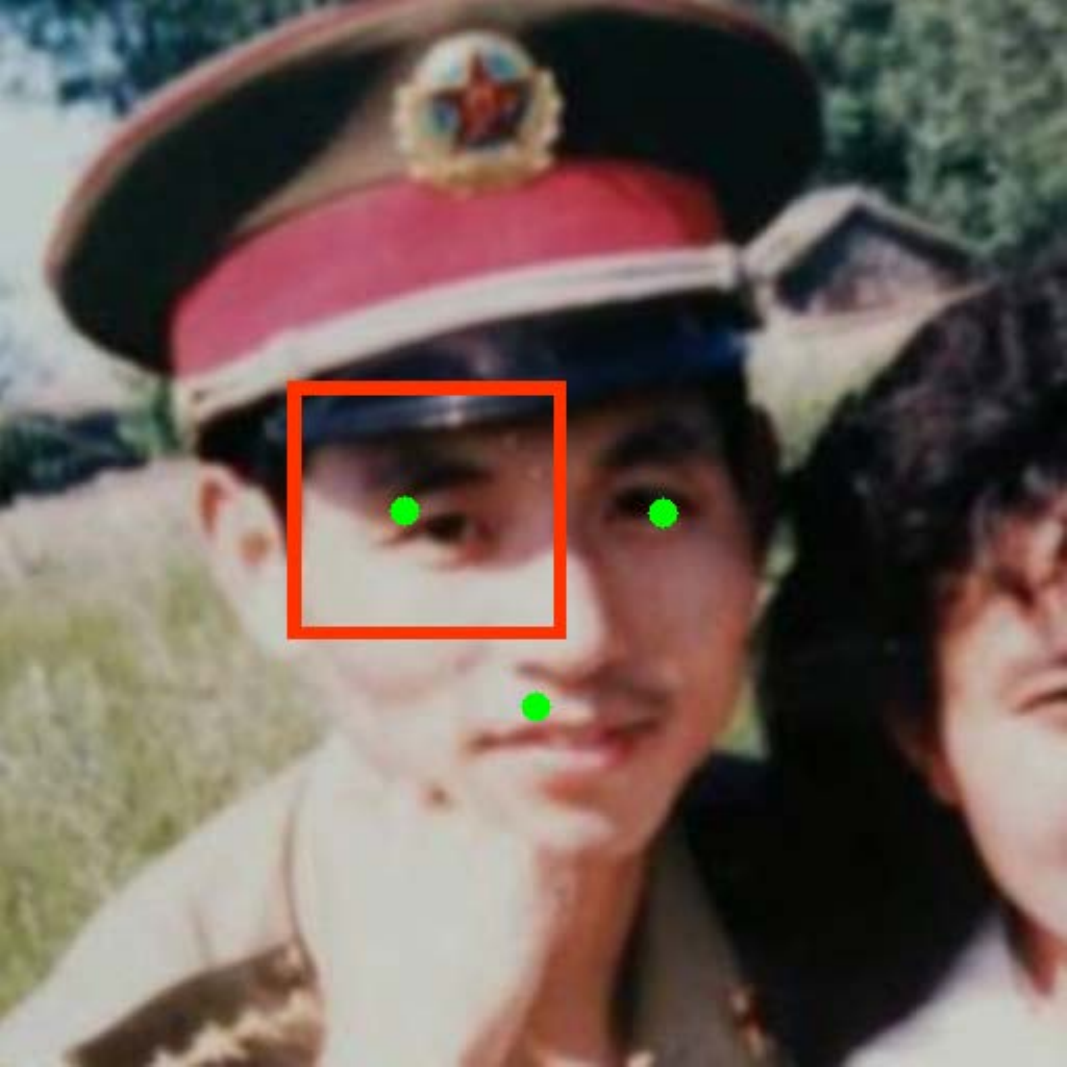}&
    \includegraphics[width=\swedm1]{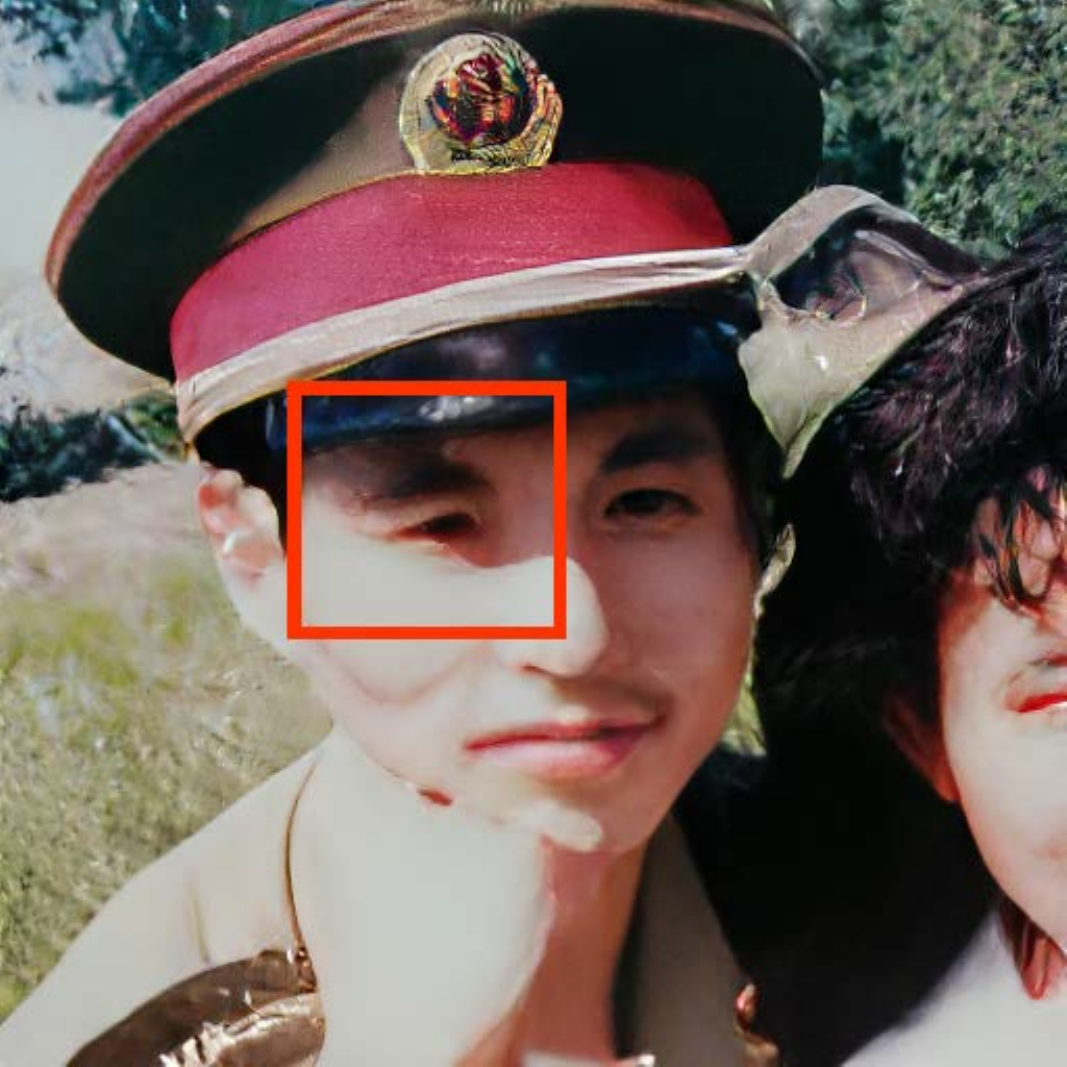}&
    \includegraphics[width=\swedm1]{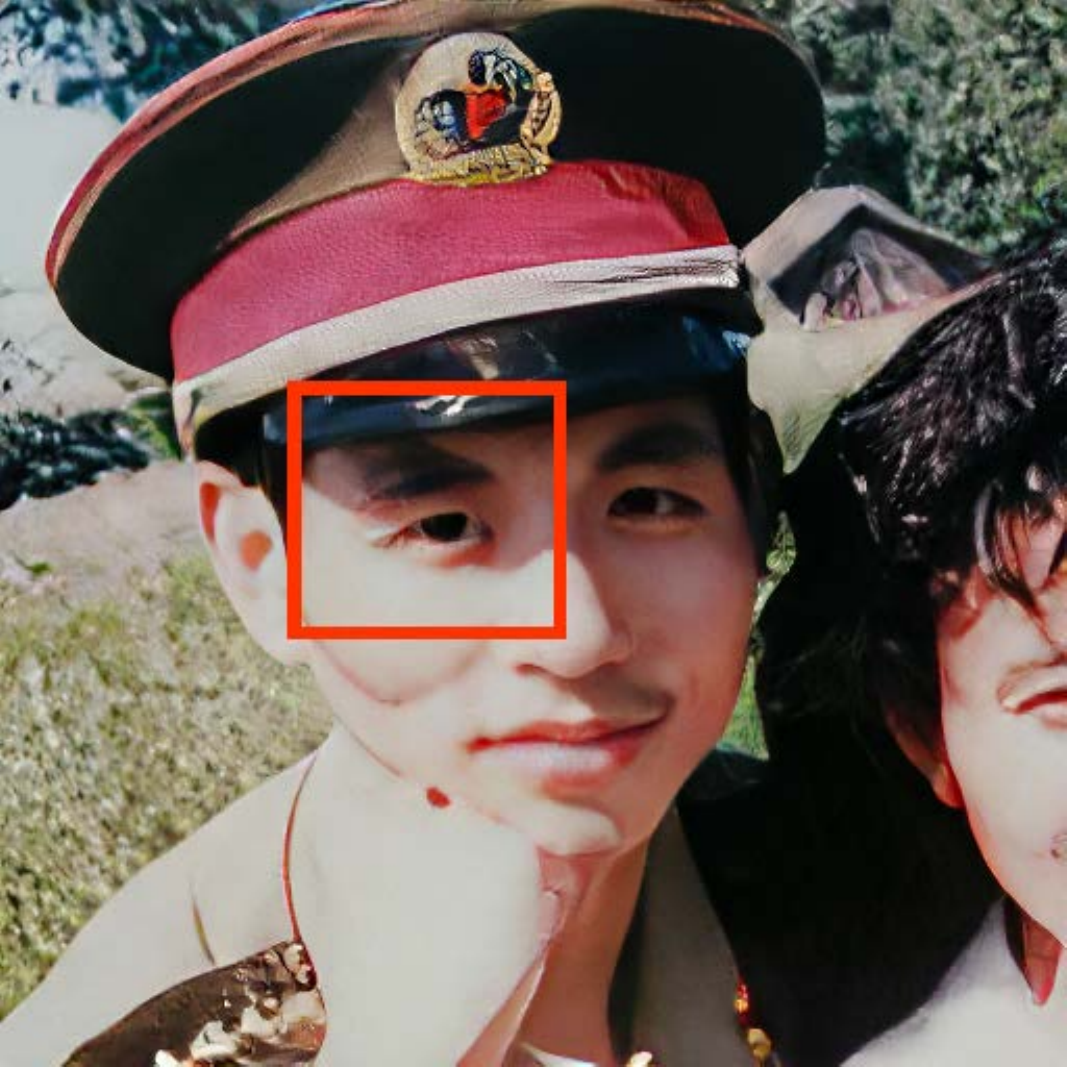}\\
    \includegraphics[width=\swedm1]{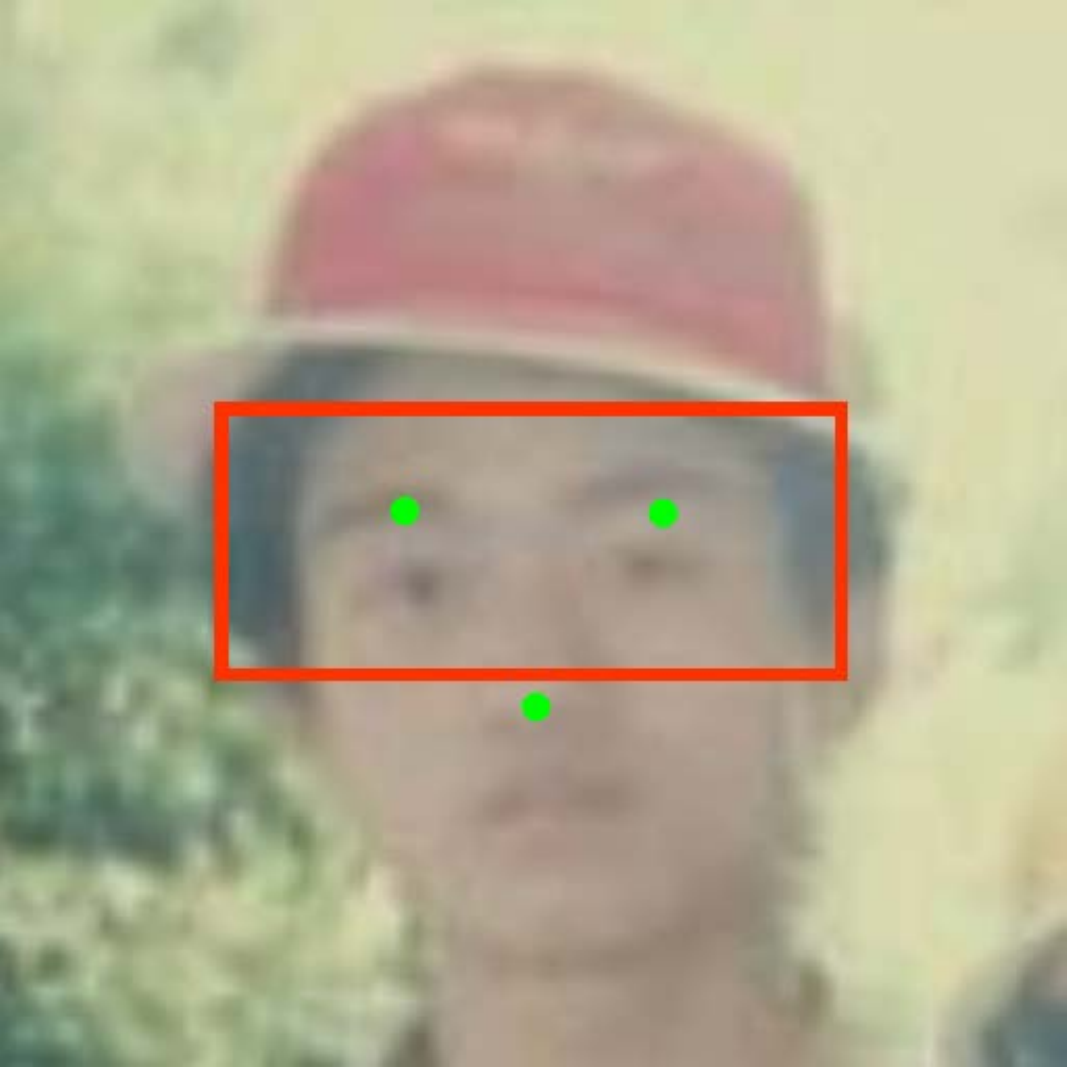}&
    \includegraphics[width=\swedm1]{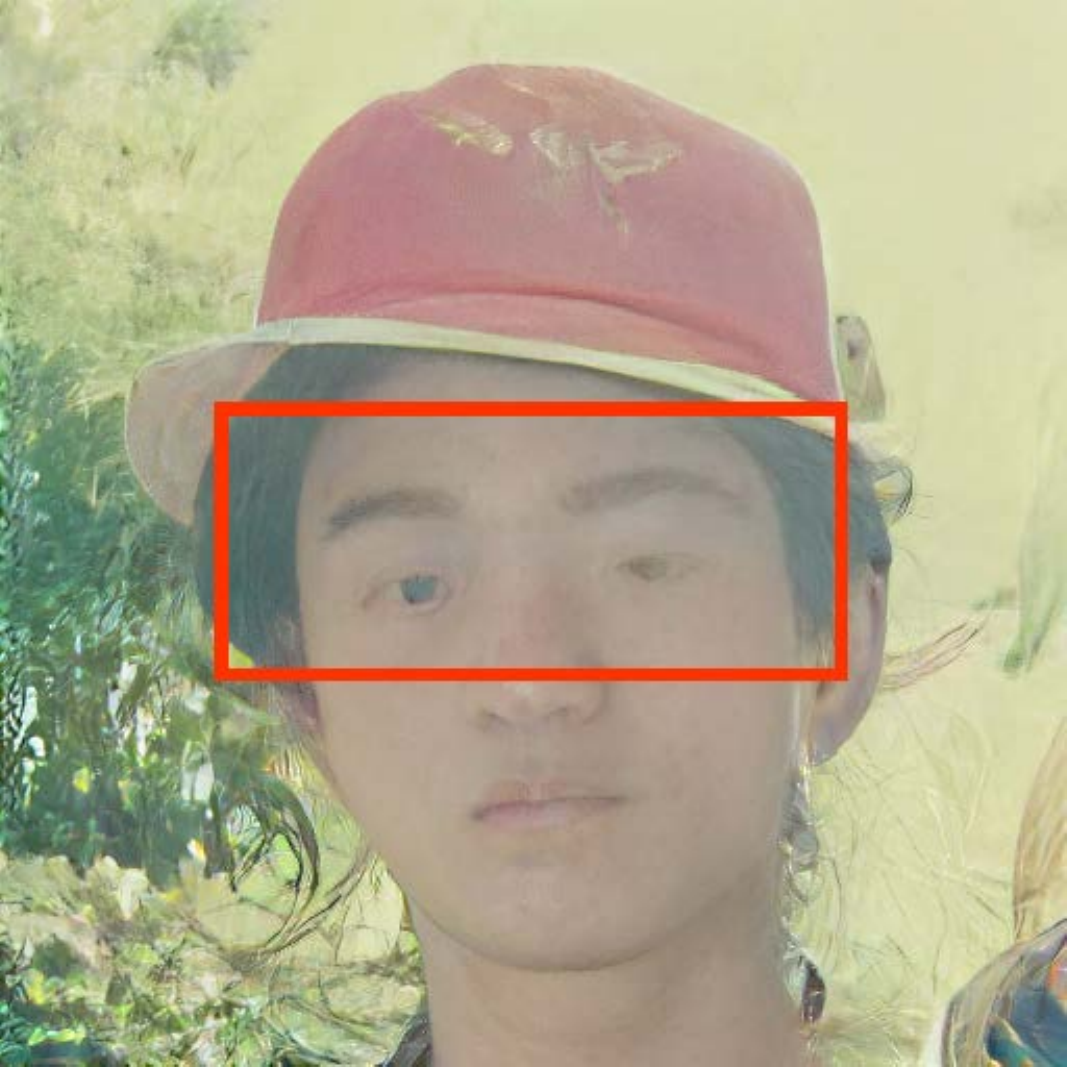}&
    \includegraphics[width=\swedm1]{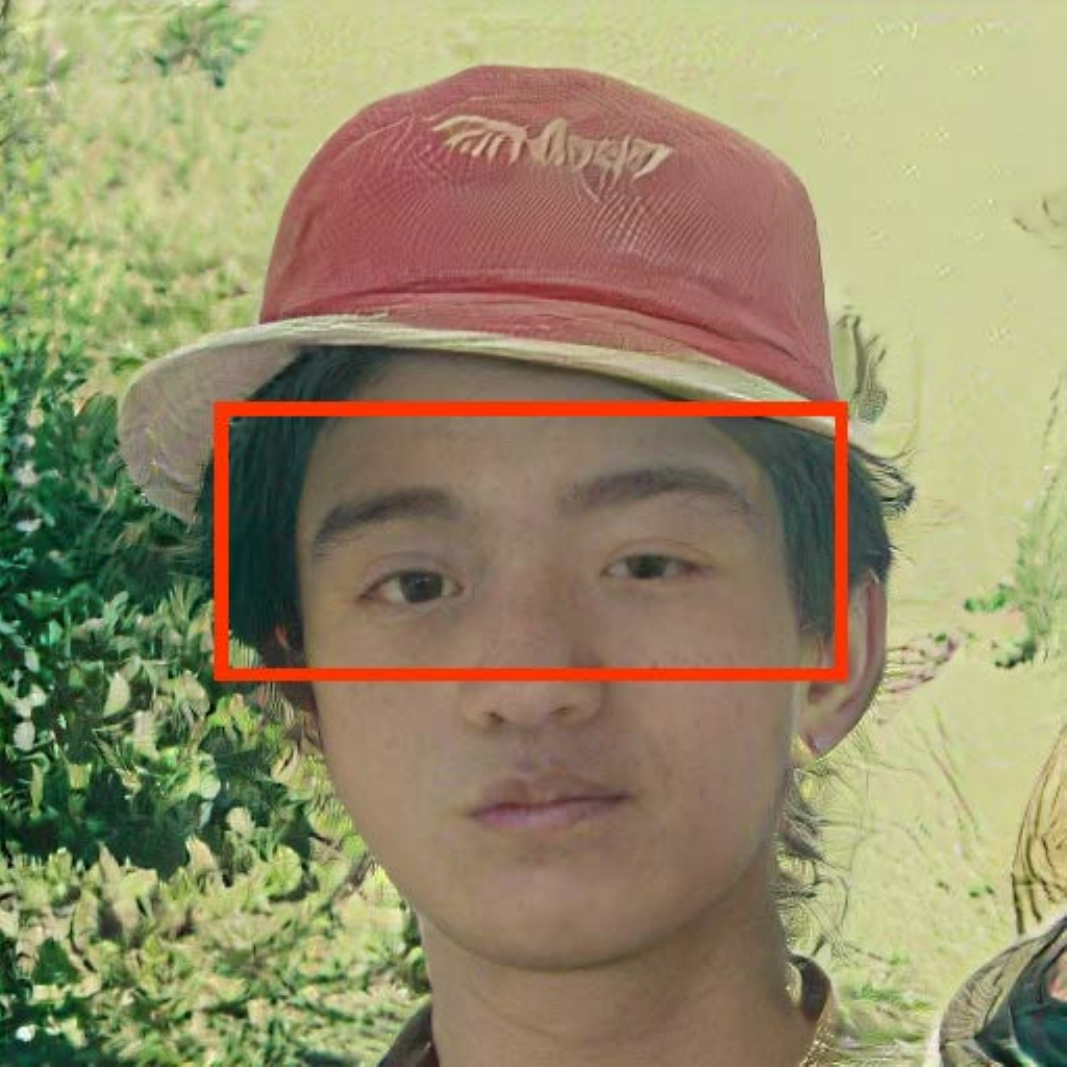}\\
    Input & Ours w/o EDM  & RestoreFormer++ 
\end{tabular}
\end{center}
\caption{
\textcolor{black}{
More qualitative results of our methods with or without EDM.
With the design of EDM, RestoreFormer++ can remove the haze covered on the faces, mitigate the artifacts raised by uneven degradation, alleviate the influence of bias introduced by face misalignment, and restore faces with a more natural appearance.
}}
\label{fig:edm_more}
\end{figure}

In this subsection, we analyze the effect of EDM on blind face restoration.
To make a more comprehensive comparison, except conducting experiments on our RestoreFormer++ with and without EDM, we also retrained PSFRGAN~\cite{chen2020generative} and GFP-GAN~\cite{wang2021towards} with EDM. 
The quantitative results in TABLE~\ref{tab:real_edm} show that methods with EDM perform better or comparable compared to their counterpart without EDM on three real-world datasets, especially on CelebChild-Test~\cite{wang2021towards} and WebPhoto-Test~\cite{wang2021towards}, whose degradations are more diverse and severe.
Since the degradations of LFW-Test~\cite{huang2008labeled} are relatively slight and regular, EDM has little effect on this dataset.
\textcolor{black}{
Besides, qualitative results show the notable effectiveness of EDM in real-world face image restoration, particularly towards face images with uneven degradation, haze, and aligned bias.
In the first sample in Fig.~\ref{fig:EDM}, the degradation in the red box is uneven (degradation in the center is more severe than the neighbouring regions), leading to a weird look of the restored right eye (blue area).
Compared to methods without EDM, their counterpart with EDM attains a clearer and more natural right eye.
The second sample in Fig.~\ref{fig:EDM} is covered with haze, and its left eyebrow is aligned to the left eye of the reference face. 
The restored results of methods without EDM are unclear, and their left eyes are not in the right position.
After fine-tuning with EDM, their results become clear, and their left eyes are restored to the right position with a more natural look.
More restored results of our proposed method with or without EDM are shown in Fig.~\ref{fig:edm_more}.
}

\begin{figure}[th]
\scriptsize
\begin{center}
\begin{tabular}{ccc}
    \includegraphics[width=\swgray]{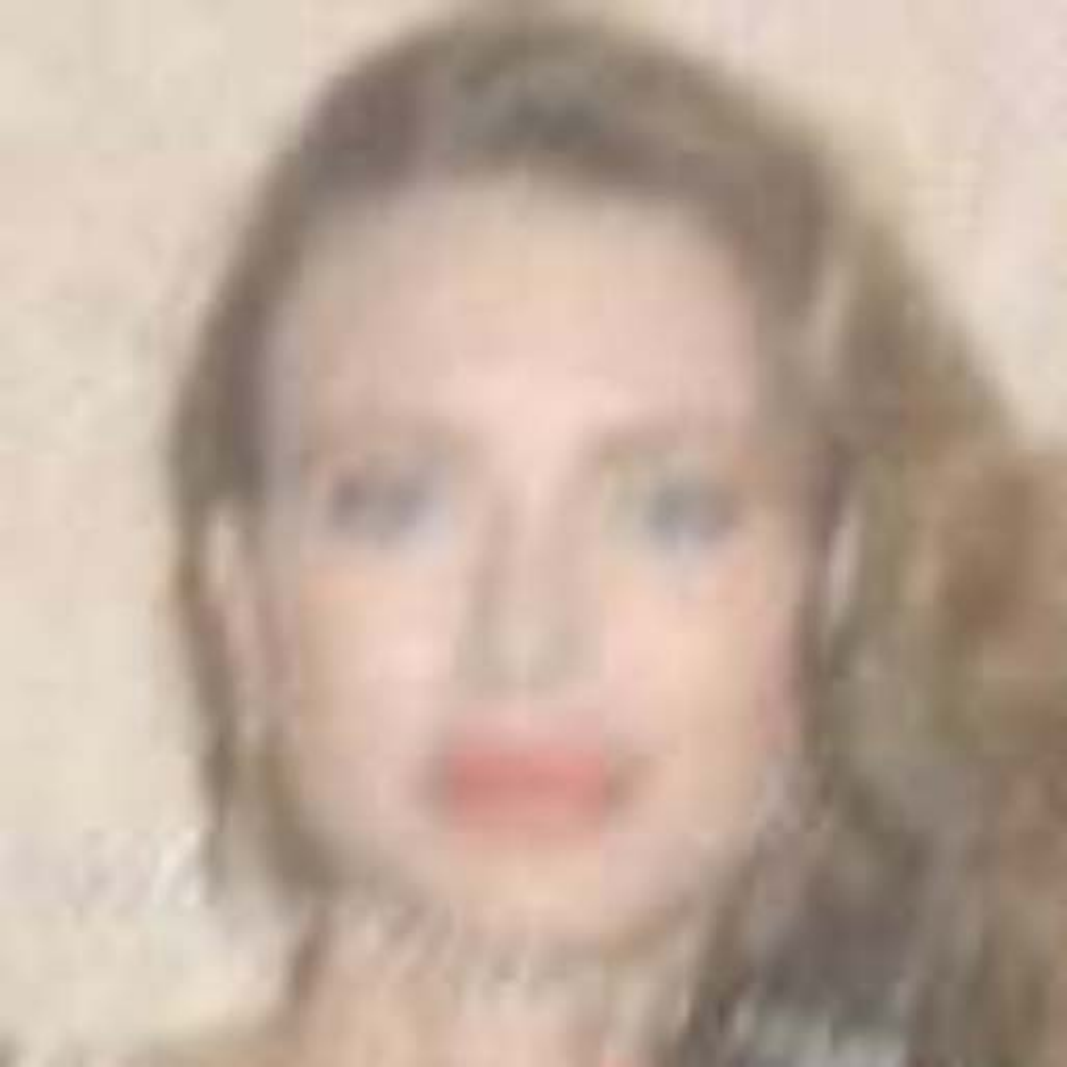}&
    \includegraphics[width=\swgray]{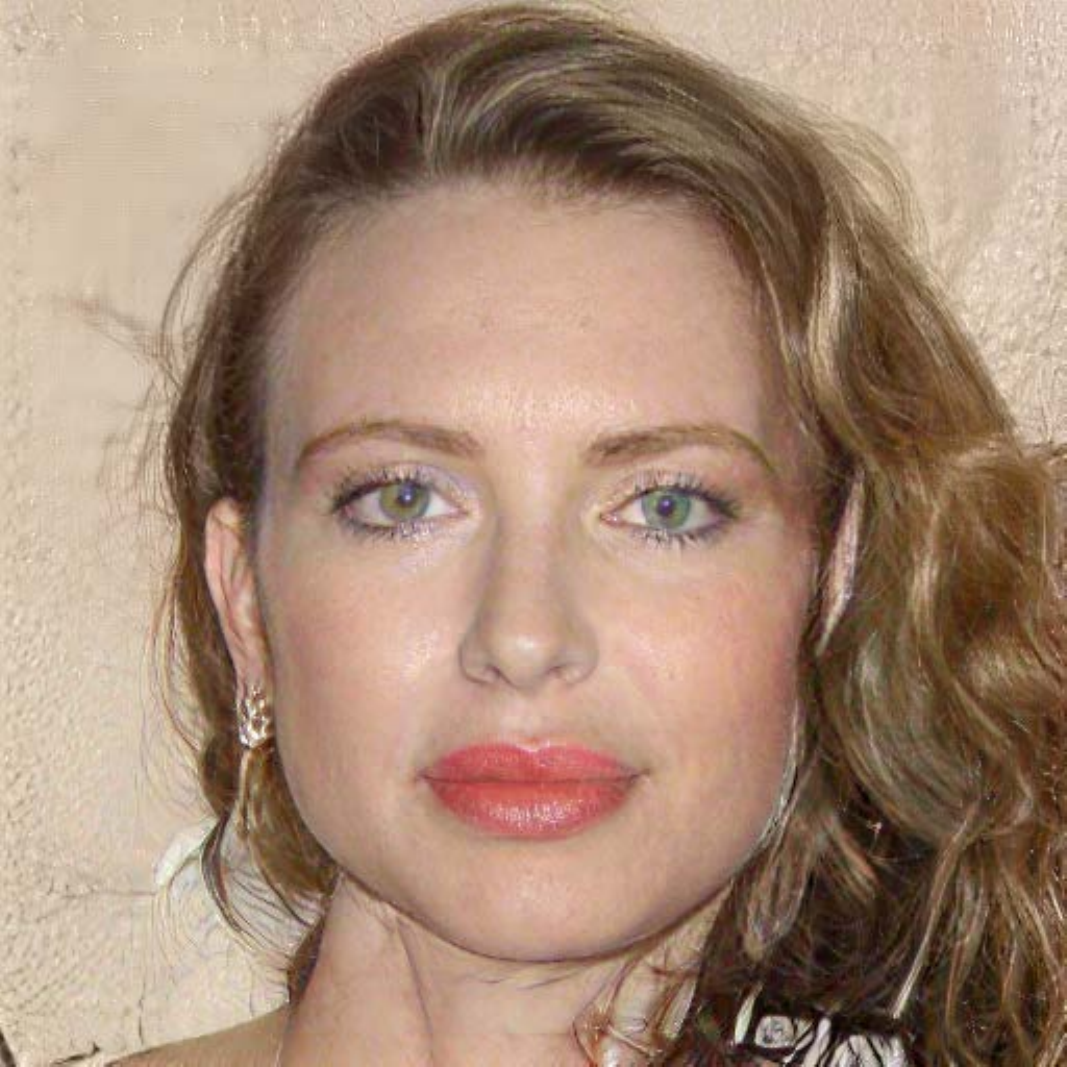}&
    \includegraphics[width=\swgray]{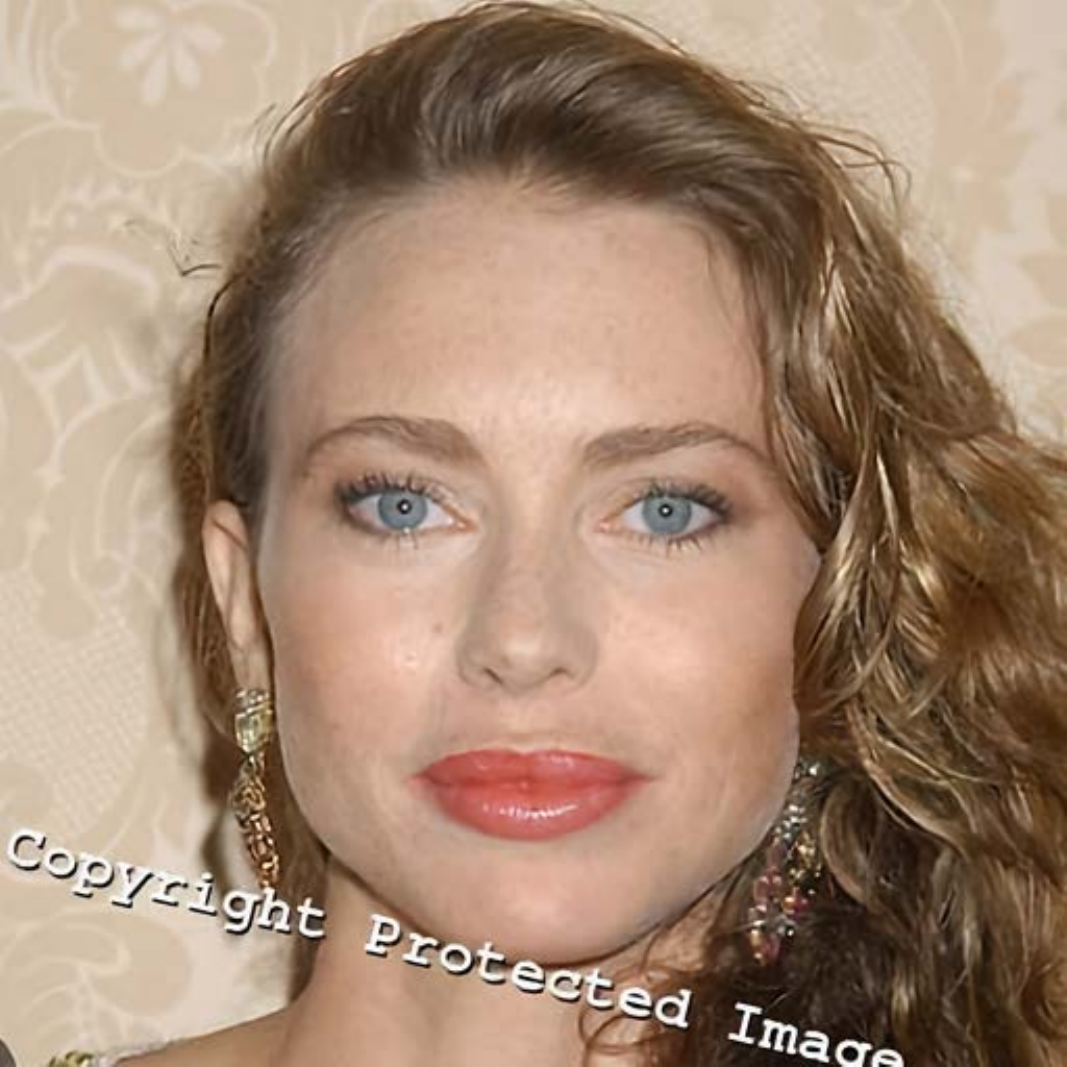} \\
    \includegraphics[width=\swgray]{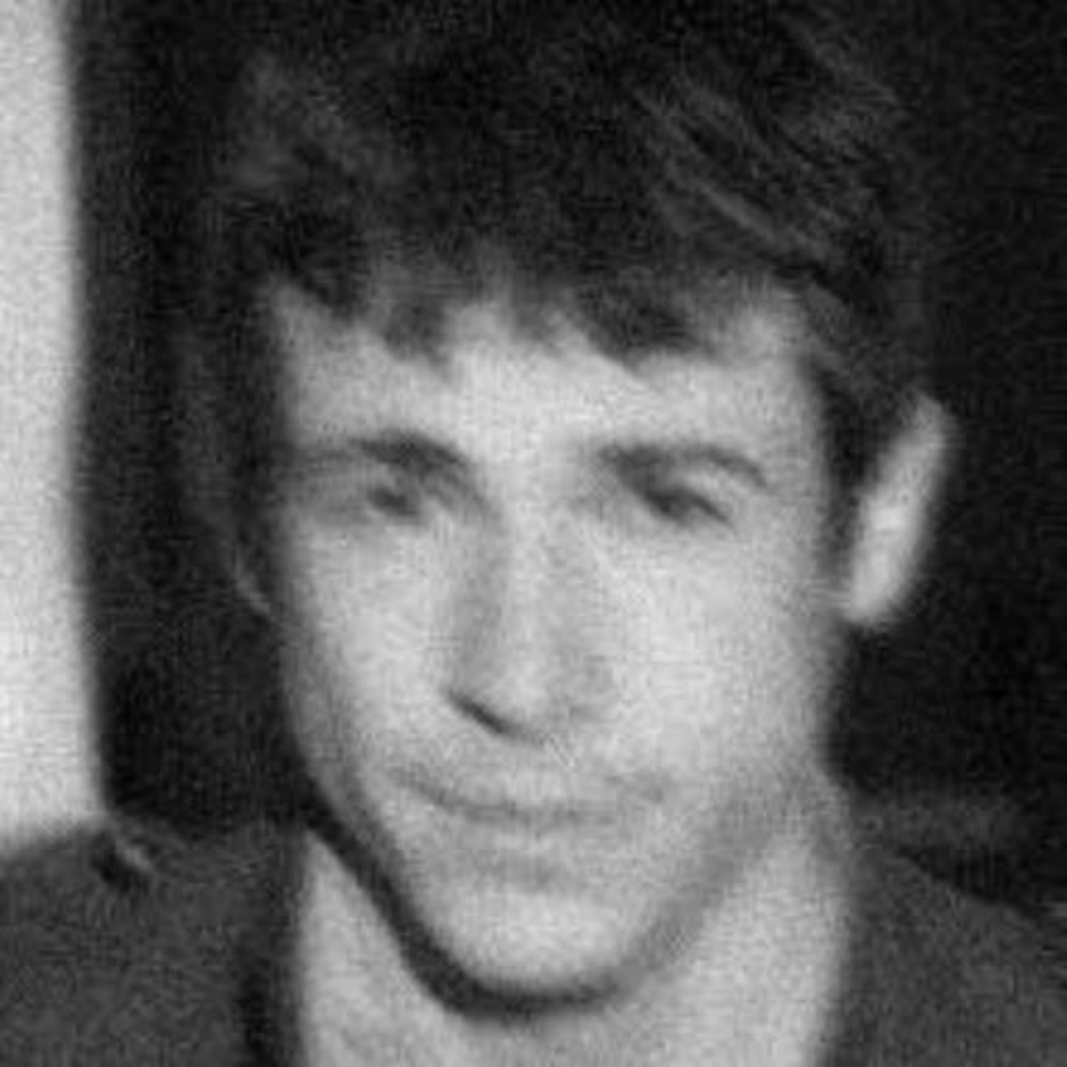}&
    \includegraphics[width=\swgray]{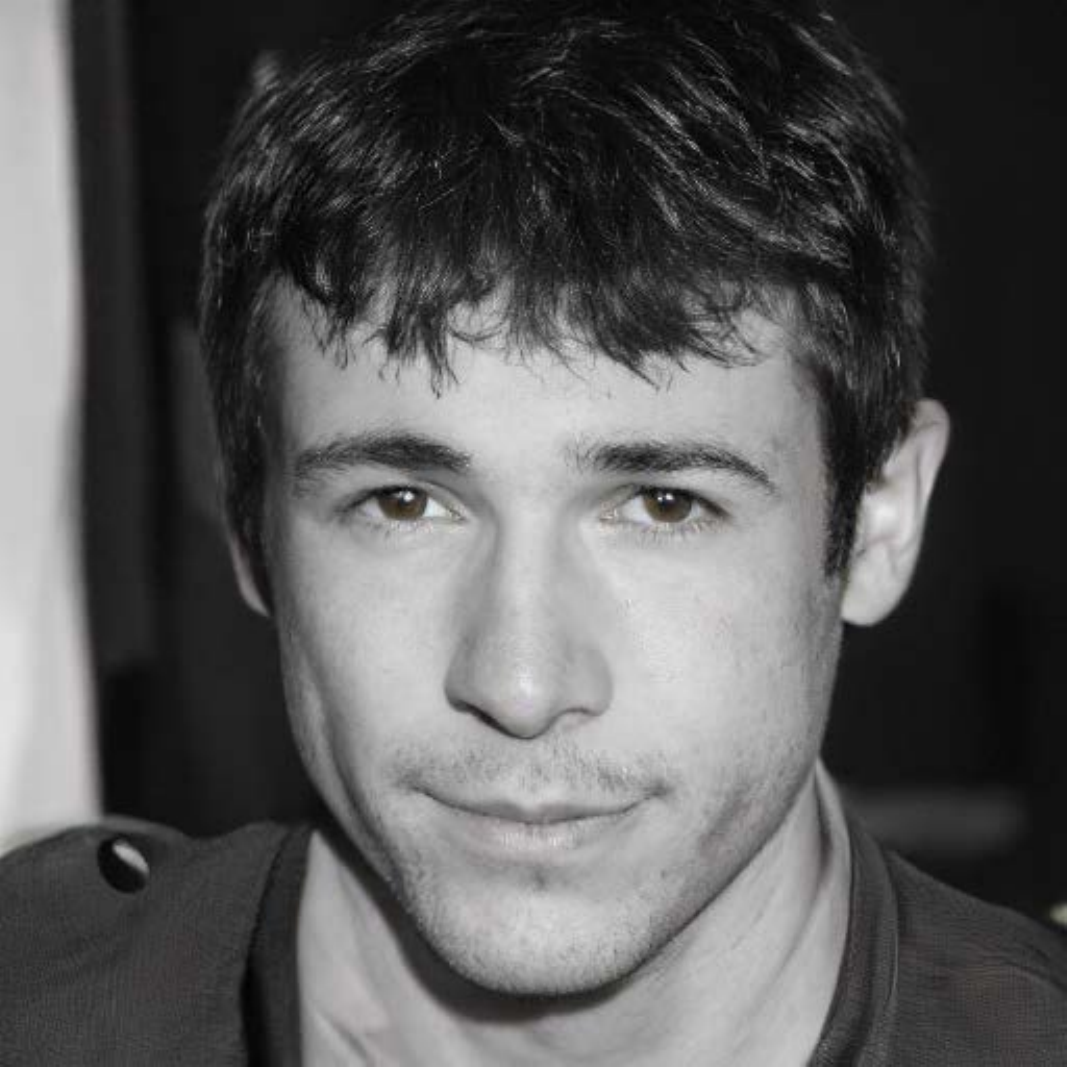}&
    \includegraphics[width=\swgray]{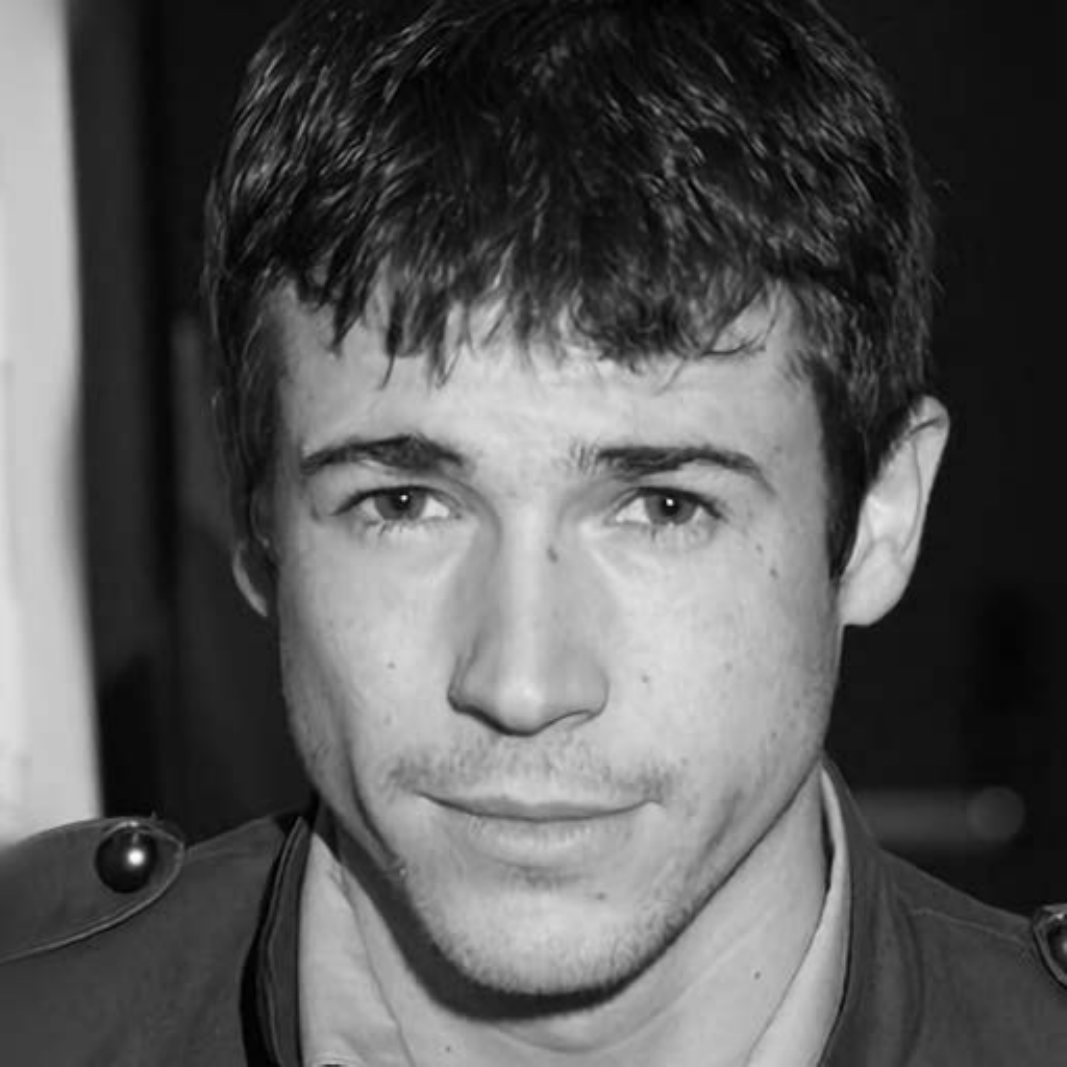} \\
    Input & RestoreFormer++ & GT \\
\end{tabular}
\end{center}
\caption{
\textcolor{black}{
Restored results of RestoreFormer++ on two synthetic degraded face images from CelebA-Test~\cite{huang2008labeled}.
The first sample shows that RestoreFormer++ aims to recover the original colors of the degraded face image. It does not change the colors of the restored results compared to the GT.
The second sample indicates that our RestoreFormer++ tends to keep the gray degraded face image gray after restoration.
}}
\label{fig:gray}
\end{figure}

\textcolor{black}{
\noindent\textbf{Discussion about color changes.}
Blind face restoration aims to remove the degradations in a face image and recover its high-quality facial structures. 
Its colors will be restored if the degraded face image contains colors. 
For example, pink color can be observed on the cheek and forehead of the first sample in Fig.~\ref{fig:results}. Methods including Wan~\textit{et al.}~\cite{wan2020bringing}, PULSE~\cite{menon2020pulse}, GPEN~\cite{yang2021gan}, VQFR~\cite{gu2022vqfr}, and our RestoreFormer~\cite{wang2022restoreformer} detect and recover this color. 
Since our priors matched from ROHQD are more accordant to blind face restoration and contextual information is considered while fusing the degraded face information and high-quality priors, the color restored with our RestoreFormer looks more harmonized and natural. 
Besides, since haze is common in real-world degraded face images, our EDM is proposed to endow RestoreFormer++ with the capability of haze removal. 
As shown in Fig.~\ref{fig:gray}, the capacity of haze removal of RestoreFormer++ aims to recover the original colors of the degraded face image. 
The colors of the restored results are close to their Ground Truth. 
The restored result of a gray degraded image is almost gray. 
With the help of EDM, the restored results of RestorFormer++ contain more facial details and look clearer. 
}

%

\begin{table}[h]
\renewcommand{\arraystretch}{1.3}
\caption{Quantitative results on CelebA-Test~\cite{huang2008labeled} of different loss settings.}
\label{tab:loss}
\centering
  \begin{tabular}{c|c|c|c|c|c|c|c} 
    \hline
    ~~$\mathcal{L}_{l1}$~~ & ~~$\mathcal{L}_{per}$~~ & ~~$\mathcal{L}_p$~~ & ~~$\mathcal{L}_{adv}$~~ & ~~$\mathcal{L}_{comp}$~~ & ~~$\mathcal{L}_{id}$~~ & ~~~FID$\downarrow$~~~ & ~~~IDD$\downarrow$~~~  \\
    \hline
    \checkmark&\checkmark&\checkmark&\checkmark&\checkmark&&45.14&0.6052\\
    \hline
    \checkmark&\checkmark&\checkmark&&&\checkmark&54.14&0.5378\\
    \hline
    \checkmark&\checkmark&\checkmark&\checkmark&&\checkmark&48.19&0.5384\\
    \hline
    \checkmark&\checkmark&&\checkmark&\checkmark&\checkmark&38.69&0.5873\\
    \hline
    \checkmark&&\checkmark&\checkmark&\checkmark&\checkmark&48.66&0.5619\\
    \hline
    \checkmark&\checkmark&\checkmark&\checkmark&\checkmark&\checkmark&38.41&0.5375\\
    \hline
  \end{tabular}
\end{table}

\subsubsection{Analysis of Losses.}
\textcolor{black}{During the training process}, RestoreFormer++ involves several loss functions, including L1 loss ($\mathcal{L}_{l1}$) and perceptual loss ( $\mathcal{L}_{per}$) for content consistence learning, degradation removal loss ($\mathcal{L}_p$) for reducing the prior searching error, adversarial loss on the whole image ($\mathcal{L}_{adv}$) and key components ($\mathcal{L}_{comp}$) for realness learning, and identity loss ($\mathcal{L}_{id}$) for identity learning.
To analyze the effect of each loss function, we remove one or two of them from the original RestoreFormer++.
%
\textcolor{black}{Since the training process without $\mathcal{L}_{l1}$ tends to collapse, we keep it in all the experiments.}
As shown in TABLE~\ref{tab:loss}, no matter removing which loss function, both scores in terms of FID and IDD increase, which means the realness and fidelity of their results become worse.
Specifically, $\mathcal{L}_{adv}$ and $\mathcal{L}_{comp}$ affect the realness of the final restored results more compared to their fidelity.
This phenomenon conforms to the characteristics of adversarial loss which aims at generating real images.
On the contrary, $\mathcal{L}_{id}$ and $\mathcal{L}_p$ affect the fidelity of the restored faces more than their realness.
$\mathcal{L}_{id}$ specifically aim at identity reserving.
$\mathcal{L}_p$ can reduce the bias introduced by noisy information by removing degradations in the encoding time.
$\mathcal{L}_{per}$ has similar effects on the realness and fidelity of the restored results for the reason that it can keep the semantic information while generating high-quality images~\cite{johnson2016perceptual}.


\begin{table}[h]
\renewcommand{\arraystretch}{1.3}
\caption{Analysis of efficiency in terms of running time, flops, and model size.}
\label{tab:efficiency}
\centering
  \begin{tabular}{c|c|c|c|c} 
    \hline
    Methods & ~~~Time/s~~~ & ~~~Flops/G~~~ & ~~~Size/M~~~ & ~~~$S$~~~ \\   
    \hline
    \hline
    DFDNet~\cite{li2020blind} & 2.218 & 602.732 & 240.117 & - \\
    PSFRGAN~\cite{chen2021progressive} & 0.2076 & 337.870 & 67.026 & - \\
    GPEN~\cite{yang2021gan} & 0.1193 & 168.279 & 71.005 & -\\
    GFP-GAN~\cite{wang2021towards} & 0.0083 & 54.734 & 76.564 & -\\
    VQFR~\cite{gu2022vqfr} & 0.7309 & 1071.492 & 71.829 & - \\
    \hline
    \textbf{RestoreFormer++} & 0.2229 & 343.071 & 72.680 & 1 \\
    \textbf{RestoreFormer++} & 0.2260 & 345.490 & 73.473 & 2 \\
    \textbf{RestoreFormer++} & 0.2643 & 374.493 & 74.265 & 3 \\
    \hline
  \end{tabular}
\end{table}

\subsection{Analysis of Efficiency}
To analyze the efficiency, we test the running time, Flops, and model size of several current methods and our RestoreFormer++ with different scale settings on a GeForce GTX 1060.
Comprehensively considering the efficiency in TABLE~\ref{tab:efficiency} and the effectiveness in TABLE~\ref{tab:celeba} and TABLE~\ref{tab:real_fid}, we can see that RestoreFormer++ attains high performance with more modest resource consumption.
Although GPEN~\cite{yang2021gan} and GFP-GAN~\cite{wang2021towards} can run \textcolor{black}{faster}, their performance is slightly inferior to the performance of RestoreFormer++.
Besides, although the performance of VQFR~\cite{gu2022vqfr} approaches to our RestoreFormer++ and their codebook shares the same idea as our ROHQD, it is time-consuming and needs more computations because its codebook is built at $32 \times 32$ resolution and the fusions between the degraded feature and priors involve resolutions from $32\times32$ to $512\times512$,  
On the contrary, our RestoreFormer++ can attain better performance with fusions only in several smaller scales ($16\times16$ and $32\times32$).
Since the performance of the three-scale setting is comparable to two scales setting but has a large increase in resource consumption, we adopt the two-scale setting in our RestoreFormer++.
It also validates that by modeling the contextual information in the face with MHCAs, feature interactions in smaller resolutions are enough for RestoreFormer++ to attain high-quality face images with both realness and fidelity.

\begin{figure}[!t]
    \centering
    \begin{tabular}{cccc}
        \includegraphics[width=\swlimitation]{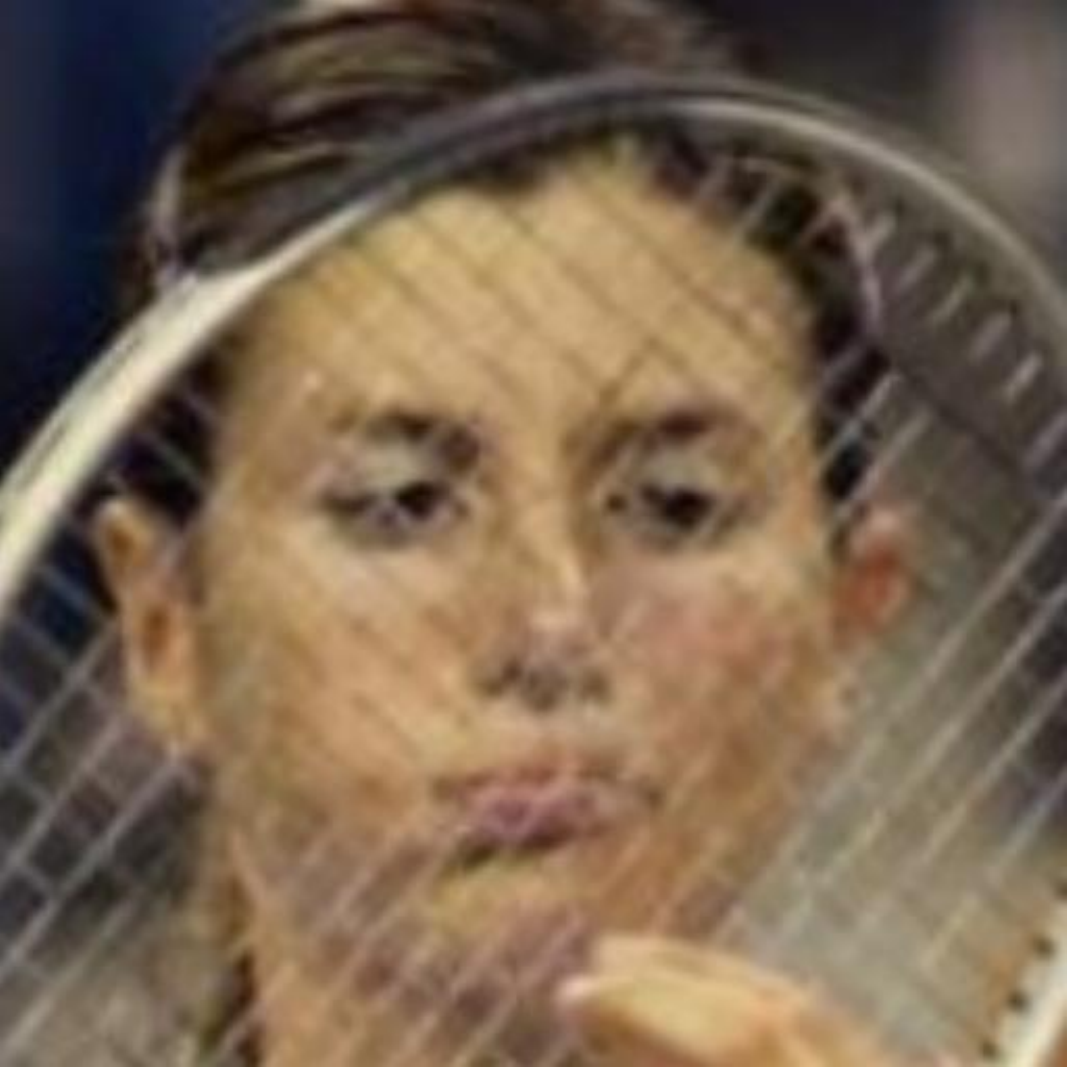} &
        \includegraphics[width=\swlimitation]{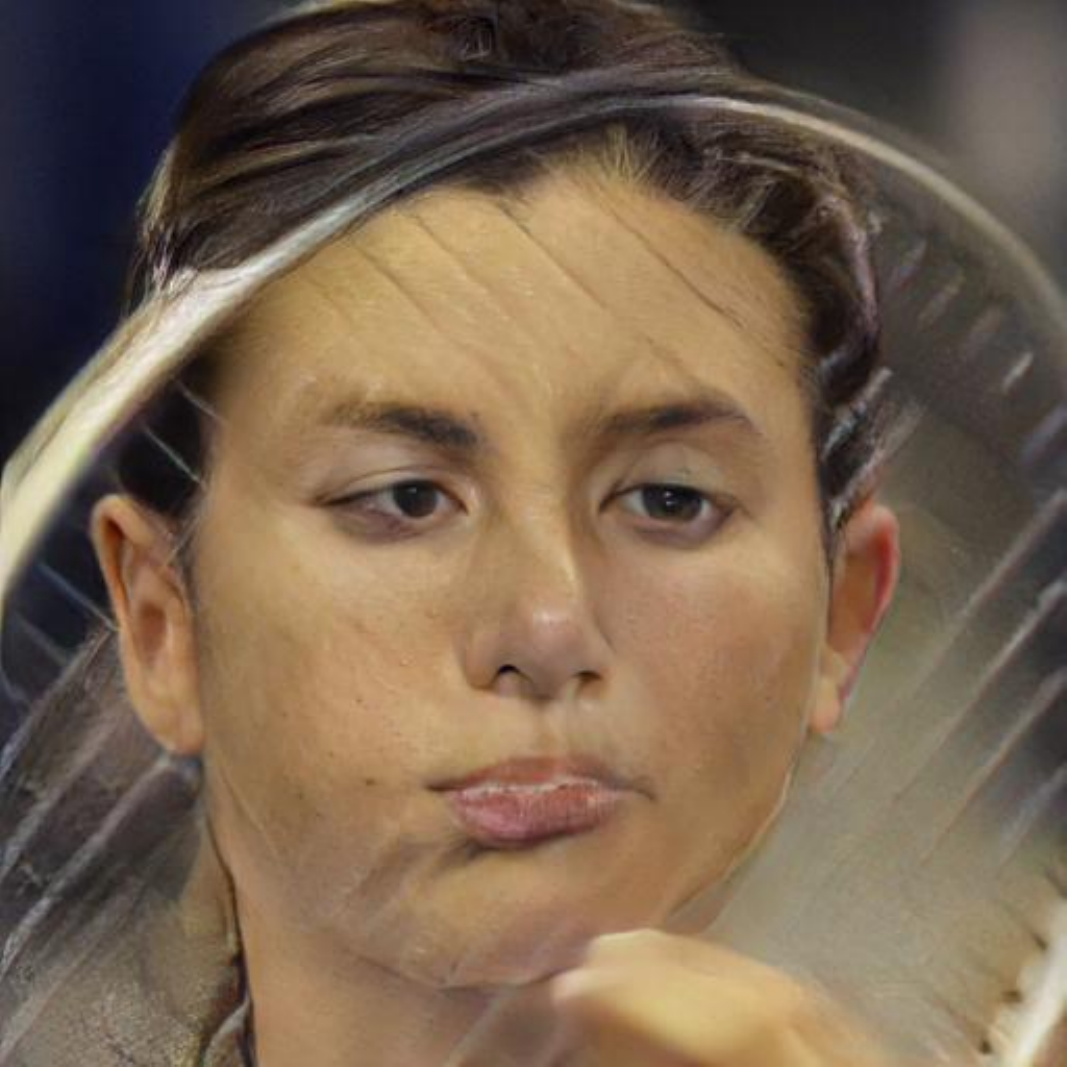} & 
        \includegraphics[width=\swlimitation]{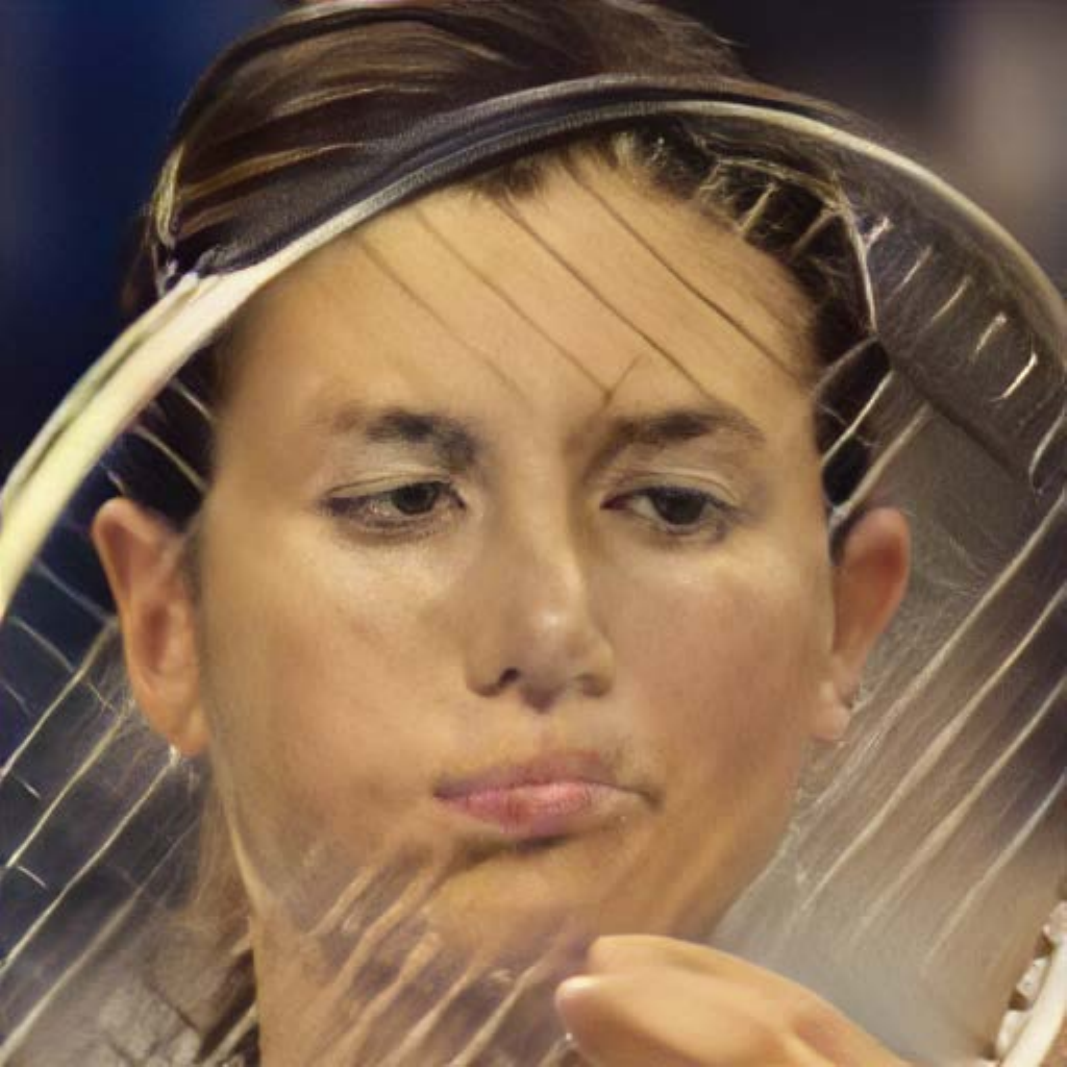} & 
        \includegraphics[width=\swlimitation]{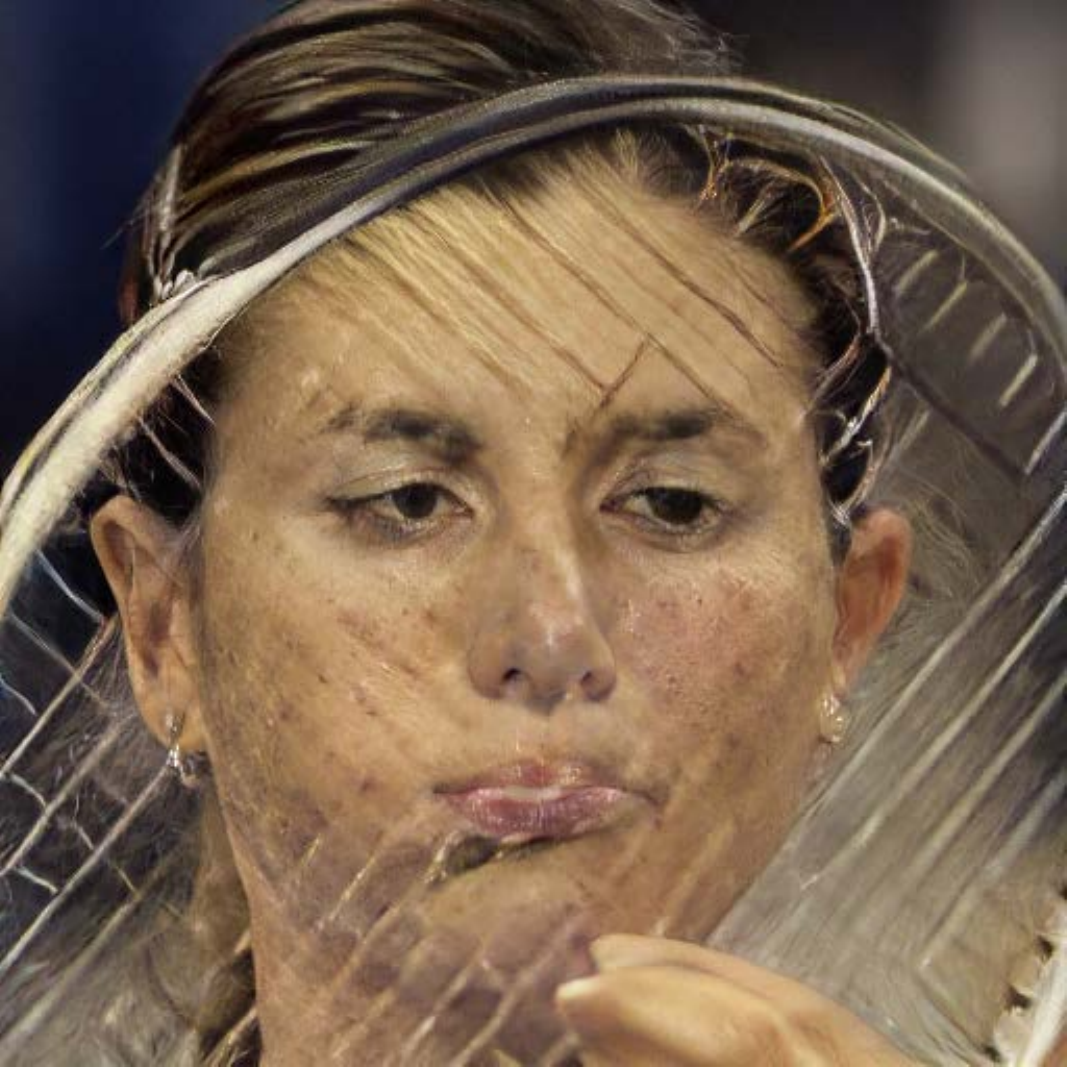}\\
        \includegraphics[width=\swlimitation]{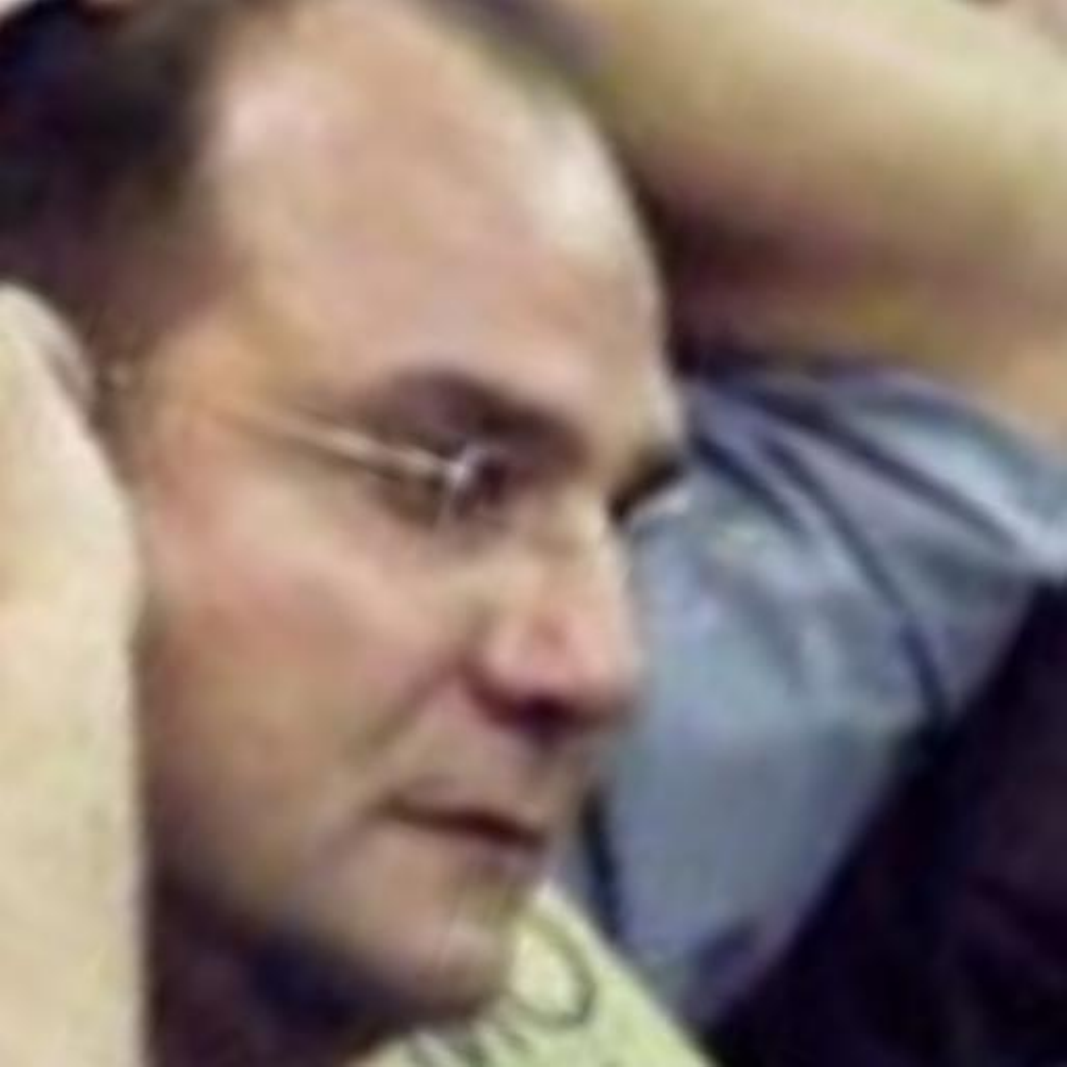} & 
        \includegraphics[width=\swlimitation]{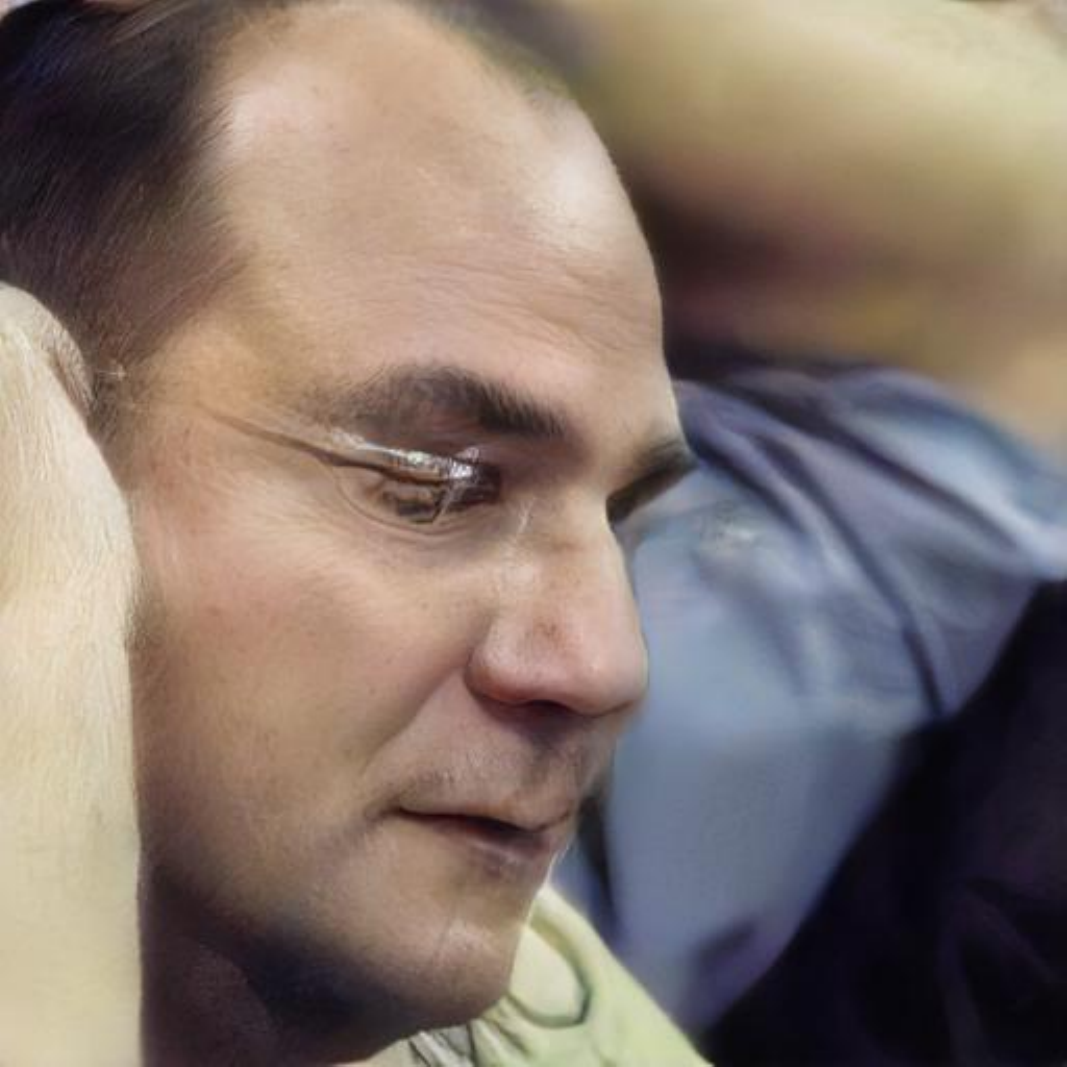} & 
        \includegraphics[width=\swlimitation]{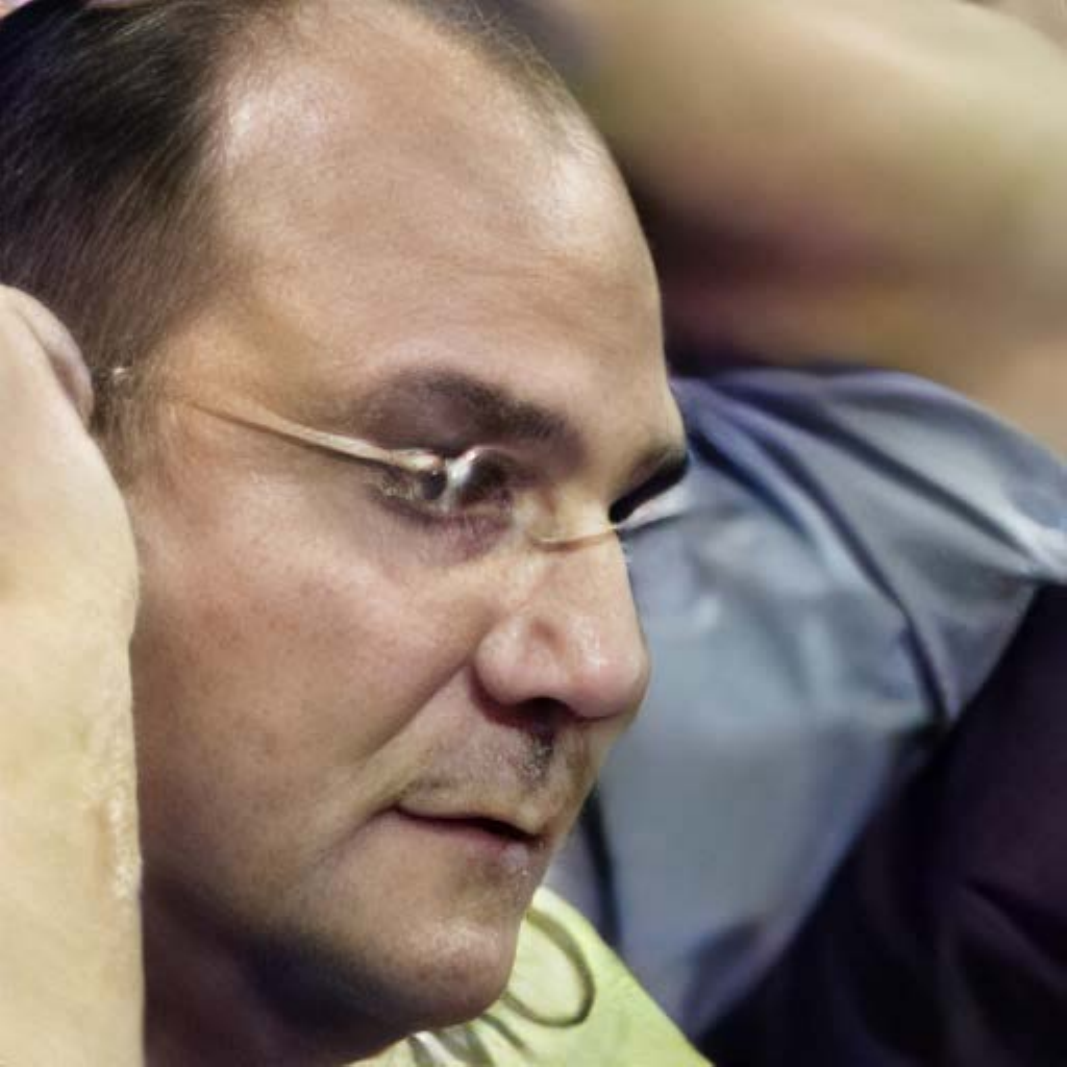} & 
        \includegraphics[width=\swlimitation]{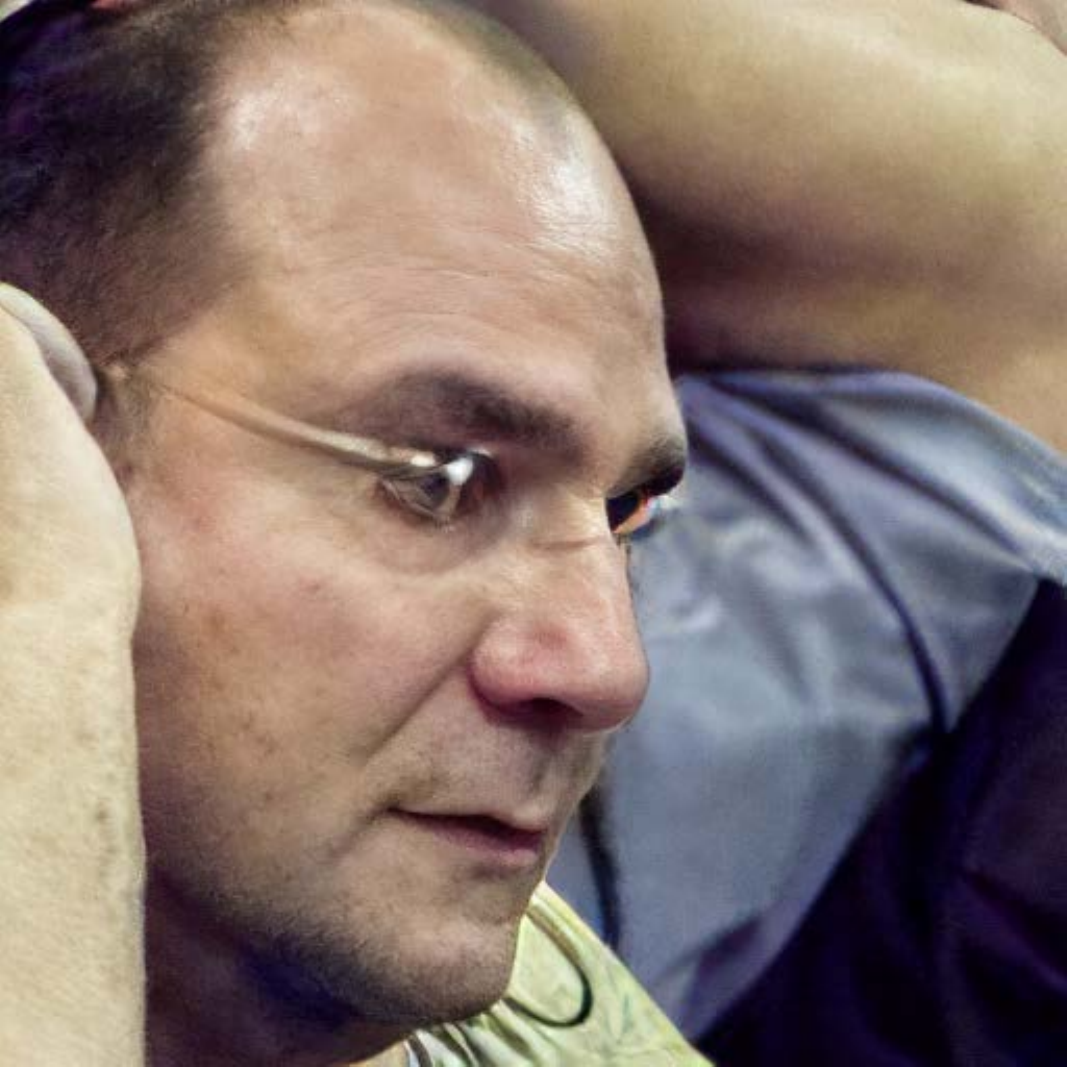}\\
        \scriptsize Input  & \scriptsize PSFRGAN~\cite{chen2021progressive} & \scriptsize GFP-GAN~\cite{wang2021towards} & \scriptsize \textbf{RestoreFormer++} \\
    \end{tabular}
    \caption{\textcolor{black}{RestoreFormer++ cannot handle faces with obstacles or large poses well.
    %
    Other methods have similar issues.}}
    \label{fig:limitation}
\end{figure}

\textcolor{black}{
\subsection{Limitations}
RestoreFormer++ cannot handle faces with obstacles or large poses well, which are also two issues for other methods. 
%
As shown in Fig.~\ref{fig:limitation}, the face in the first sample is covered with a tennis racket. 
PSFRGAN~\cite{chen2021progressive} and GFP-GAN~\cite{wang2021towards} tend to remove the obstacle.
Although our method can keep most of the tennis racket, it leads to artifacts.
In the second sample, all the approaches cannot restore complete glasses due to the large facial pose.
These limitations mainly result from the bias in the training data -- most of the high-quality face images in FFHQ~\cite{karras2019style} are near-frontal and without obstacles.
In the future, we will try to make an effort to mitigate these limitations from two aspects: 1) extending the diversity of FFHQ~\cite{karras2019style} with more facial poses and obstacles. 2) explicitly modeling the information of facial poses and obstacles, and then merging it into the face restoration model.
}

\section{Conclusion}
\label{sec:conclusion}
In this work, we propose a RestoreFormer++, which on the one hand introduces multi-head cross-attention mechanisms to model the fully-spatial interaction between the degraded face and its corresponding high-quality priors, and on the other hand, explores an extending degrading model to synthetic more realistic degraded face images for training.
The goal of RestoreFormer++ is to restore face images with higher realness and fidelity for both synthetic and real-world scenarios.
\textcolor{black}{
Specifically, the multi-head cross-attention mechanisms take the features of degraded faces as queries and the corresponding high-quality priors as the key-value pairs.
It models the contextual information based on both semantic and structural information in features at different resolutions.
Besides, the priors provided by our reconstruction-oriented high-quality dictionary are learned from plenty of high-quality face images with a face generation network cooperating with vector quantization.
They are more accordant to the face restoration task, enabling RestoreFormer++ to attain faces with better realness.
What is more, the extending degrading model contains more realistic degradations and considers face misalignment situations, allowing our RestoreFormer++ to further alleviate the synthetic-to-real-world gap and improve its robustness and generalization towards real-world scenarios.
Finally, we conduct extensive experiments on both synthetic and real-world datasets to demonstrate the superiority of the proposed method.
We also discuss the limitations of RestoreFormer++ and our future works for further improvement.
}

%



\ifCLASSOPTIONcompsoc
  \section*{Acknowledgments}
  This paper is partially supported by the National Key R\&D Program of China No.2022ZD0161000 and the General Research Fund of Hong Kong No.17200622. 
\else
  \section*{Acknowledgment}
\fi


\ifCLASSOPTIONcaptionsoff
  \newpage
\fi



\bibliographystyle{IEEEtran}
\bibliography{IEEEabrv,RestoreFormer}
%



%

\begin{IEEEbiography}[{\includegraphics[width=1in,height=1.25in,clip,keepaspectratio]{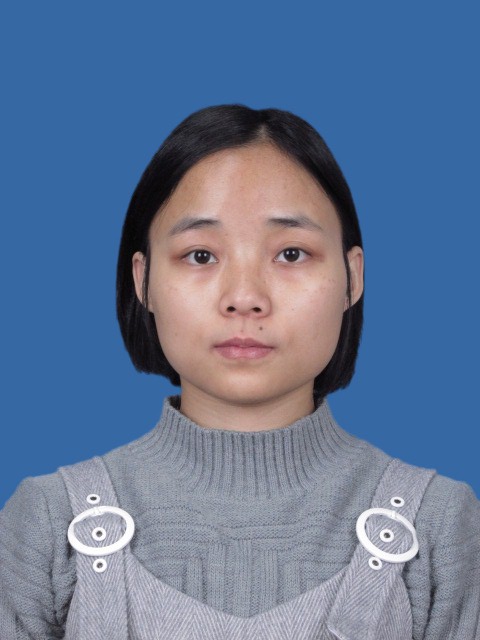}}]{Zhouxia Wang}
received the B.Eng. and M.Eng. degrees from the School of Data and Computer Science at Sun Yat-sen University, Guangzhou, China, in 2015 and 2018. She is currently a Ph.D. student at the University of Hong Kong (HKU), Hong Kong SAR, China. Her research interests include deep learning and low-level vision.
\end{IEEEbiography}

\begin{IEEEbiography}[{\includegraphics[width=1in,height=1.25in,clip,keepaspectratio]{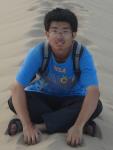}}]{Jiawei Zhang}
is a research scientist at SenseTime Research.
He received a PhD degree from City University of Hong Kong in 2018, a master degree from Institute of Acoustics, Chinese Academy of Sciences in 2014 and a bachelor degree from University of Science and Technology of China in 2011 receptively. His research interests include image deblurring, image super-resolution and related computer vision problems.
\end{IEEEbiography}

\begin{IEEEbiography}[{\includegraphics[width=1in,height=1.25in,clip,keepaspectratio]{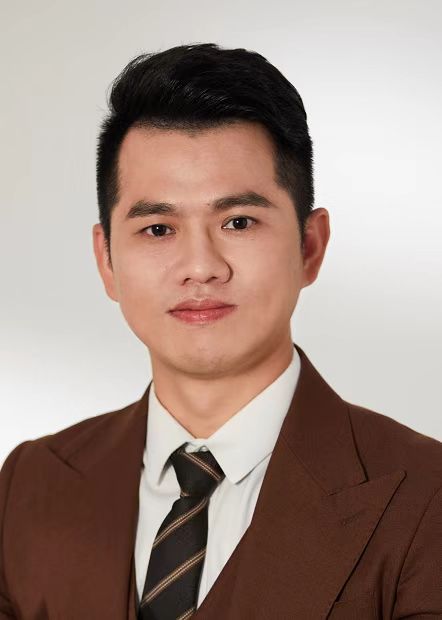}}]{Tianshui Chen}
received the Ph.D. degree in computer science from the School of Data and Computer Science at Sun Yat-sen University, Guangzhou, China, in 2018. From 2016 to 2017, he was a research assistant at Hong Kong Polytechnic University. He is currently an associated professor at Guangdong University of Technology. His current research interests include computer vision and machine learning. He has authored and coauthored more than 30 papers published in top-tier academic journals and conferences, including T-PAMI, T-NNLS, T-IP, T-MM, CVPR, ICCV, AAAI, IJCAI, ACM MM, etc. He has served as a reviewer for numerous academic journals and conferences. He was the recipient of the Best Paper Diamond Award at IEEE ICME 2017.
\end{IEEEbiography}

\begin{IEEEbiography}[{\includegraphics[width=1in,height=1.25in,clip,keepaspectratio]{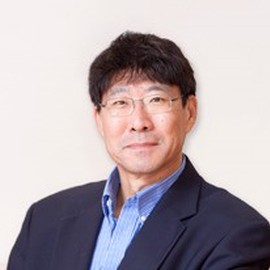}}]{Wenping Wang}
is now Professor of the Department of Computer Science \& Engineering at Texas A\&M University. He has been Chair Professor and Head (2012-2017) of the Department of Computer Science at the University of Hong Kong. His research interests are computer graphics and computer vision. He has over 300 technical publications in related fields. He serves or has served as journal associate editor of CAGD, CAG, TVCG, CGF, IEEE Computer Graphics and Applications, and IEEE Transactions on Computers. He has chaired a number of international conferences, including SPM 2006, SMI 2009, Pacific Graphics 2012, GD/SPM'13, SIGGRAPH Asia 2013, Geometry Summit 2019 and Geometry Summit 2023. He has been Founding Chairman of Asian Graphics Association (2016-2020).
\end{IEEEbiography}

\begin{IEEEbiography}[{\includegraphics[width=1in,height=1.25in,clip,keepaspectratio]{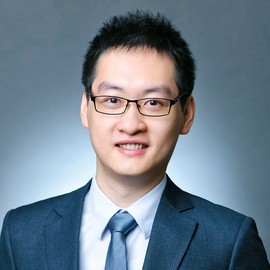}}]{Ping Luo}
is an Assistant Professor in the department of computer science, The University of Hong Kong (HKU). He received his PhD degree in 2014 from Information Engineering, the Chinese University of Hong Kong (CUHK), supervised by Prof. Xiaoou Tang and Prof. Xiaogang Wang. He was a Postdoctoral Fellow in CUHK from 2014 to 2016. He joined SenseTime Research as a Principal Research Scientist from 2017 to 2018. His research interests are machine learning and computer vision. He has published 100+ peer-reviewed articles in top-tier conferences and journals such as TPAMI, IJCV, ICML, ICLR, CVPR, and NIPS. His work has high impact with 18000+ citations according to Google Scholar. He has won a number of competitions and awards such as the first runner up in 2014 ImageNet ILSVRC Challenge, the first place in 2017 DAVIS Challenge on Video Object Segmentation, Gold medal in 2017 Youtube 8M Video Classification Challenge, the first place in 2018 Drivable Area Segmentation Challenge for Autonomous Driving, 2011 HK PhD Fellow Award, and 2013 Microsoft Research Fellow Award (ten PhDs in Asia).
\end{IEEEbiography}







\end{document}